\documentclass[letterpaper,twocolumn,10pt]{article}
\usepackage{usenix}
\usepackage{tikz}
\newcommand*\circled[1]{\tikz[baseline=(char.base)]{
            \node[shape=circle,draw,inner sep=0.2pt] (char) {#1};}}

\usepackage{tabularx}
\usepackage{array}
\usepackage{booktabs}
\usepackage{algorithmic}
\usepackage{epsfig,endnotes}
\usepackage[ruled, lined, linesnumbered,longend]{algorithm2e}
\usepackage{multirow}
\usepackage{url}
\newcommand{\Orca}{{\scshape Orca}\xspace}
\usepackage{subfig}
\usepackage{wrapfig}
\newcommand{\DELF}[1]{\iffalse #1 \fi}
\newcommand{\IdeatoRead}[1]{\iffalse #1 \fi}

\newcommand{\DELMayNeed}[1]{\iffalse #1 \fi}

\usepackage{amsthm}
\theoremstyle{definition}

\usepackage{color}
\usepackage{amsfonts}
\newtheorem{thm}{\textbf{Observation}}
\definecolor{light-gray}{gray}{0.95}
\usepackage{mathtools}      
\usepackage{cleveref}
\crefname{section}{§}{§§}
\Crefname{section}{§}{§§}
\crefformat{section}{§#2#1#3}

\newcommand{\sh}[1]{\textcolor{blue}{[??: #1]}}

\newcommand{\DEL}[1]{\iffalse #1 \fi}

\usepackage[normalem]{ulem}
\usepackage{mathtools}      

\newcommand\tsr[1]{{\color{red}#1}}

\newcommand\tfn[1]{{\color{black}#1}}
\newcommand\tsm[1]{{\color{violet}#1}}

\newcommand{\squishlist}{
\begin{list}{$\bullet$}
  { \setlength{\itemsep}{0pt}
     \setlength{\parsep}{0pt}
     \setlength{\topsep}{0pt}
     \setlength{\partopsep}{0pt}
     \setlength{\leftmargin}{0em}
     \setlength{\labelwidth}{0em}
     \setlength{\labelsep}{0.2em} } }

\newcommand{\squishlisttwo}{
\begin{list}{$\bullet$}
  { \setlength{\itemsep}{0pt}
     \setlength{\parsep}{0pt}
    \setlength{\topsep}{0pt}
    \setlength{\partopsep}{0pt}
    \setlength{\leftmargin}{2em}
    \setlength{\labelwidth}{1.5em}
    \setlength{\labelsep}{0.5em} } }

\newcommand{\squishend}{
  \end{list}  }

\newcommand{\sys}{\textsc{CacheOPT}\xspace}
	
\begin{document}




\title{Mitigating KV Cache Competition to Enhance User Experience in LLM Inference}
\author{
  Haiying Shen\\
  University of Virginia
  \and
  Tanmoy Sen\\
  University of Virginia
  \and
    Masahiro Tanaka\\
Microsoft
}
\maketitle

\vspace{-0.25in}


\pagestyle{plain}

\DEL{\tfn{Plan.
1. Modify up to the analysis section with fake figs. 07/22/2024 - 8 am

2. Modify the system design. 07/24/2024 - 8am

3. Modify the experimental evaluation. 07/24/2024 - midnight}}


\DEL{\begin{abstract}

In Large Language Model (LLM) serving, the KV-cache (KVC) bottleneck causes high tail Time-to-First-Token (TTFT) and Time-Between-Tokens (TBT), impairing user experience, particularly in time-sensitive applications. 
However, satisfying both TTFT and TBT service-level objectives (SLOs) is challenging. To address this, we propose a system, Mitigating KV Cache Competition (\sys), based on key insights from our measurements, incorporating novel components. First, it estimates a request's output length, which guarantees that its KVC demand is satisfied with a high specified probability. Second, it allocates the estimated KVC demand to a request, and reuses other requests' allocated KVC in order to avoid preemptions while reducing waiting time. Third, it proactively allocates KVC before instead of at the time a request exhausts its allocation and reserves KVC globally to prevent preemptions. Fourth, it chooses a request that has long SLO for TBT, long job remaining time and short preemption time to preempt. Fifth, it selects the shortest-latency strategy between swapping and recomputation for preemptions. Experimental results show that \sys achieves up to a 3.29$\times$ and 2.83$\times$ lower tail TBT and TTFT, and 47\% and 53\% higher TTFT and TBT SLO attainments than the state-of-the-art methods. 
\end{abstract}}

\DEL{if it can predict the output length with high confidence
integrates several strategies.
Instead of allocating a fixed block of KVC, \Sys employs demand-based KVC allocation, where the KVC allocated equals the predicted response length. Then, it dynamically adjusts the allocated KVC  through confidence score of the prediction
and then decides the padding based on the request arrival rate to reduce the waiting time
while avoiding request preemption.
\sys also prioritizes request eviction or preemption based on resource occupancy and projected completion time, and thus ensuring \tsr{low preemption latency  through efficient resource distribution}\sh{what's the ultimate goal?-done}. Additionally, \sys
chooses between either recomputation or swapping to reduce preemption time. }


\pagestyle{plain}

\DEL{L. Chen and H. Shen, Towards Resource-Efficient Cloud Systems: Avoiding Over-Provisioning in Demand-Prediction Based Resource Provisioning, Proc. of the 2016 IEEE International Conference on Big Data (IEEE BigData 2016), December 5-8, 2016, Washington D.C., USA
--you can refer to Section C padding, and then start writing a paper in overleaf. Share it with me now.
The idea is to predict do prediction of the output length, then based on the prediction confidence score, determine the padding. The sum of the padding is the buffer size.

Study the pro and cons of recomputation and swapping, and then based on the current condition, choose the best option.

Also, in swapping and recomputation, do not choose a request to swap based on FCFS. Choose a request occupy more space and will finish later. Add SLOs to applications as another factor to consider.

Before you sent me a list of different scheduling algorithms, such as least remaining time, etc. Now, in LLM, think about what additional factors we need to consider in selecting requests to form a batch.

Pls finish the draft with analysis section and exp. Section using dummy figs ASAP.
Let’s plan to complete this paper and submit it to ASPLOS for now, June }

\DEL{\tfn{Plan.
1. Modify up to the analysis section with fake figs. 07/22/2024 - 8 am

2. Modify the system design. 07/24/2024 - 8am

3. Modify the experimental evaluation. 07/24/2024 - midnight}}

\begin{abstract}


In Large Language Model (LLM) serving, the KV-cache (KVC) bottleneck causes high tail Time-to-First-Token (TTFT) and Time-Between-Tokens (TBT), impairing user experience, particularly in time-sensitive applications. However, satisfying both TTFT and TBT service-level objectives (SLOs) is challenging. To address this, we propose a system, named \sys for mitigating KV Cache competition, based on key insights from our measurements, incorporating novel components. First, it estimates a request's output length, bounding the deviation with a high specified probability, adjusted based on the request arrival rate. Second, it allocates the estimated KVC demand to a request, and reuses other requests' allocated KVC to avoid preemptions while reducing waiting time. Third, it proactively allocates KVC before instead of at the time a request exhausts its allocation and reserves KVC globally to prevent preemptions. Fourth, it chooses a request that has long TBT SLO, long job remaining time and short preemption time to preempt. Fifth, it selects the shortest-latency strategy between swapping and recomputation for preemptions. Experiments show that \sys achieves up to 3.29$\times$ and 2.83$\times$ lower tail TBT and tail TTFT, 47\% and 53\% higher TTFT and TBT SLO attainments, and supports up to 1.58$\times$ higher request arrival rate than the state-of-the-art methods. 
\end{abstract}

\vspace{-0.0in}
\section{Introduction}
\label{sec:intro}


The Large Language Model (LLM) models have found widespread applications across various domains, including automated customer support, chatting, and real-time translation. 
However, deploying LLMs in real-world applications often comes with high tail Time-to-First-Token (TTFT) or Time-Between-Tokens (TBT)~\cite{seldon2023llm,liu2024llumnix,Zhu2024SampleAttentionNA}
, which can impair user experience, especially in time-sensitive tasks. 
Reducing TTFT and TBT is crucial for ensuring seamless user interactions and satisfied user experience. 
It also enables the broader adoption of LLMs in different applications. 


Due to limited GPU memory and the large volume of KV-cache (KVC) values, KVC becomes a bottleneck, significantly impacting both TTFT and TBT. The first iteration-level scheduler \Orca~\cite{280922}  pre-allocates KVC for the maximum sequence length 
for each request. However, this approach wastes memory~\cite{jin2023s}, limits the batch size and hence GPU utilization (e.g., 0.4\%~\cite{jin2023s}) and throughput. To address this problem, vLLM~\cite{vllm} 
uses block-based KVC allocation approach, in which, a block, consisting of a fixed number of tokens, is allocated to a request each time. To form a batch in each iteration, the requests from the waiting queue are added to the batch until the whole KVC is allocated. While this approach significantly reduces KVC reserved waste, a running request may experience block allocation failure when it exhausts its allocation. In this case, the last arrived running request is selected to preempt based on the first-in-first-serve (FCFS) policy. 
Both approaches introduce delays, increasing TBT.


Another method allocates KVC to a request equal to the response length predicted by an LLM model~\cite{Zheng2023ResponseLP,Jin2023S3IG}. 
Compared to the block-based allocation approach, it reduces the chances of KVC allocation failures but still leads to certain reserved waste since allocated KVC is not used instantly and generates KVC allocation failures with underestimation.
\cite{Zheng2023ResponseLP} proposes to add a constant padding (e.g., 100 tokens) to the predicted value to avoid underprovisioning, 
indicates that overprovision is not as important as underprovision. 


\DEL{This problem can be solved by reusing allocated but unused KVC as proposed in~\cite{techreport} \sh{need to upload our paper to archieve-done --paper is uploaded, need to modify the verison}. }

\DEL{Deciding proper padding is required to overcome overprovision and underprovision. However, constant padding based on the predicted length leads to higher waiting time and lower execution time when the arrival rate is high, and vice-versa. Figure~\ref{fig:example} shows the impact of arrival rate on the waiting time for constant padding for consecutive two seconds. The blue box in the figure shows the constant added pad. If the pad is remained constant, the waiting queue keeps growing with higher arrival rate, as low amount of request can be allocated KVC, and thus leading higher overall waiting time.}

\DEL{
Consequently, such prediction-based KVC allocation suffers from either underprovision or overprovision of the KVC. Also, 
}



\DEL{\begin{figure}[t]
\centering
    \subfloat[Lower arrival rate.\vspace{-0.0in}\label{fig:ex1}]{{\includegraphics[width=0.48\linewidth,height=0.1\textheight]{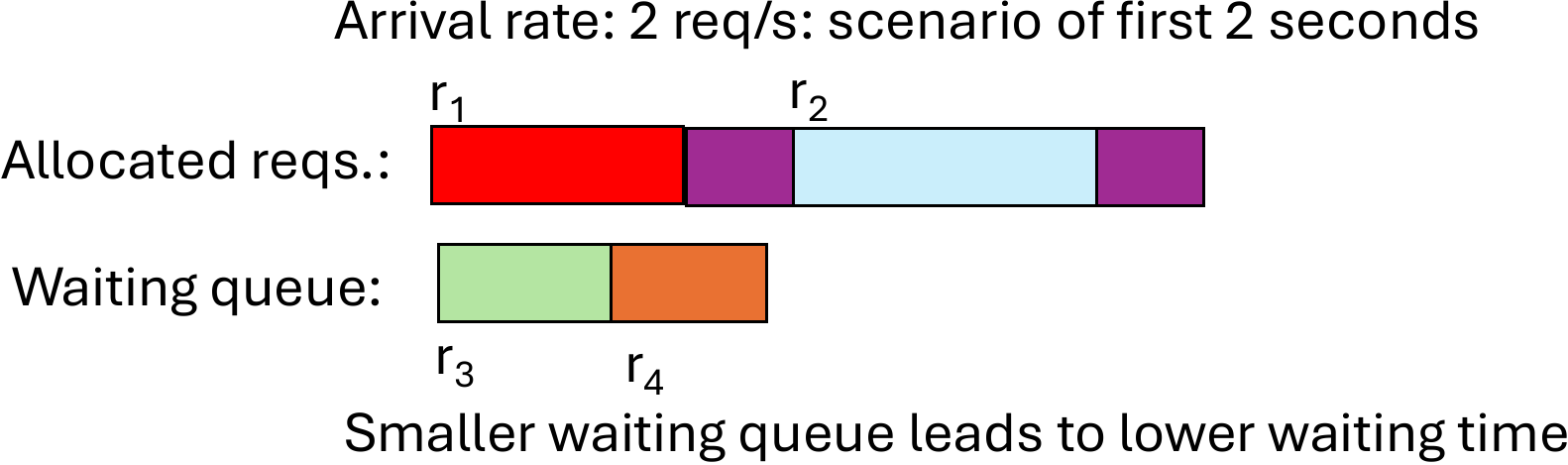} }}
    \hfill  \subfloat[Higher arrival rate.\vspace{-0.0in}\label{fig:ex-2}]{{\includegraphics[width=0.48\linewidth,height=0.1\textheight]{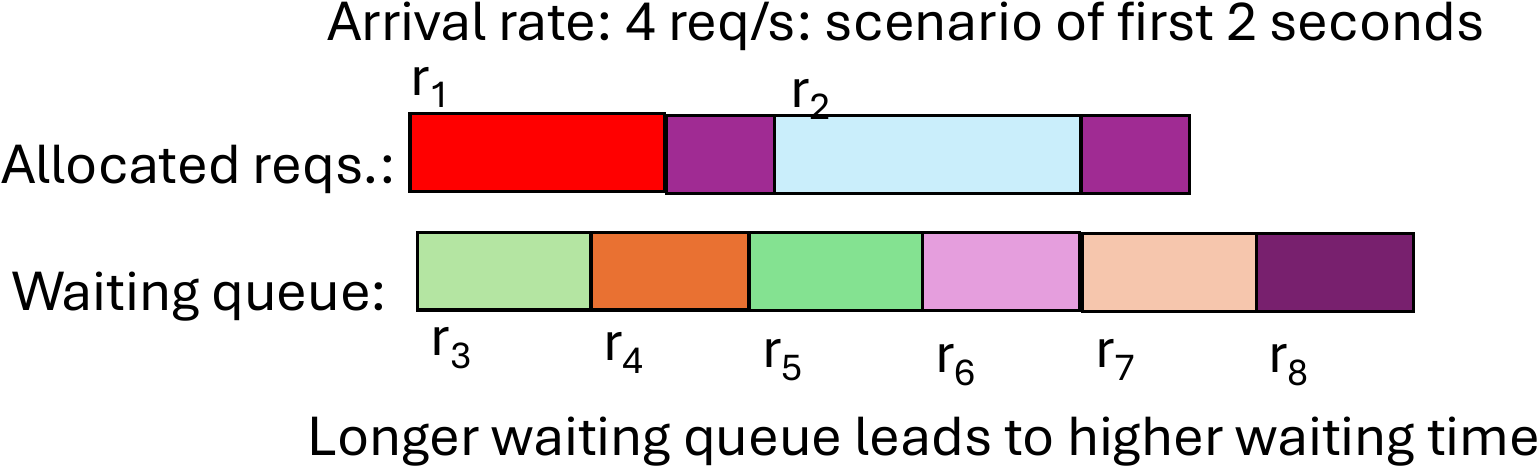} }}
    \hfill
    \vspace{-0.0in}
   \caption{\small{Examples of waiting time vs. arrival rate for constant padding. \vspace{-0.0in}}}%
    \label{fig:example}
\end{figure}
}

\DEL{\tsr{Deciding proper padding is required to overcome overprovision and underprovision. However, constant padding based on the predicted length leads to higher waiting time and lower execution time when the arrival rate is high, and vice-versa. Figure~\ref{fig:example} shows the impact of arrival rate on the waiting time for constant padding for consecutive two seconds. The blue box in the figure shows the constant added pad. If the pad is remained constant, the waiting queue keeps growing with higher arrival rate, as low amount of request can be allocated KVC, and thus leading higher overall waiting time.}
}


\DEL{As mentioned, when a KVC allocation failure occurs due to underprovisioning, a request is preempted through swapping or recomputation~\cite{vllm}. Swapping swaps the request’s KV values from GPU memory to CPU memory, bringing them back when KVC becomes available. Recomputation recomputes its KV values when the preempted sequence is rescheduled. Both approaches introduce significant delays for the preempted requests. 
}

\DEL{Swapping and recomputation generate different preemption times influenced by different factors (including the sequence length, PCIe bandwidth and computation capacity). The preemption time is defined as the time that the selected request is paused. The current LLM systems offer options to choose either one. However, we should choose the strategy that takes less time based on the current status of these factors.}

\DEL{Swapping 
is suitable when computational resources are scarce but demands high memory (PCIe) bandwidth and can introduce latency due to I/O overhead. 
Recomputation, on the other hand, eliminates the need for additional storage and avoids I/O delays, 
which can be beneficial when memory resources are limited or bandwidth is constrained, but it increases computational load and latency. 
Therefore, it is important to determine which approach (swapping or recomputation) generates the least latency for a preempted request.
}

\DEL{\sh{what is it?-done} }



\DEL{In this work, based on the demand-based KVC allocation, we aim to design a KVC allocation method and a preemption strategy to avoid KVC overflow and reduce LLM response latency. ?This work aims to efficiently batch the requests and adjust the predicted length of the batched request by considering a positive or negative \emph{cushion} to the predicted length based to reduce the amount of underprovision and overprovision, respectively, to reduce the latency. Then, to further recover from overprovisioning and underprovisioning, we decide the amount of padding based on the arrival rate using Hoeffding's inequality for balancing the execution and waiting time \tsr{as well as minimizing the memory violation severity, which is defines as the deficit of the allocated KVC and the KVC demand.} \sh{cushion is for accurate prediction based on confidence score but the padding part and its goal are missing here-done}

Beyond that, this work also provides an advanced strategy to evict or preempt the requests by ordering the requests based on the latency SLO, remaining processing time, and KVC occupancy compared to the typic FIFO or LRU strategy used in the state-of-the-art methods~\cite{vllm, Radix-attention}.
}


\DEL{{\color{purple}5th paragraph: To achieve the goal, We conducted experimental analysis and found:  describe the tradeoff caused by the arrival rate and padding \sh{not done-done, 2nd observation, trade-off between execution time and waiting time, arrival rate impacts it }, etc.}}

Different LLM applications and users often have varying SLOs for TTFT and TBT. For example, real-time translation applications prioritize ultra-low TTFT to deliver the first token as quickly as possible, enabling smoother conversation flow. In contrast, document summarization may tolerate higher TTFT in exchange for lower TBT. 
Moreover, users' reading speeds influence their tolerance for TBT. A fast reader using a summarization tool may prefer shorter TBT to maintain a seamless reading experience, while a slower reader may be less sensitive to longer TBT but require low TTFT for responsiveness. The diverse demands necessitate adaptable LLM systems capable of meeting varying TTFT and TBT SLOs.\looseness=-1

In this paper, we aim to mitigate KV cache competition while enhancing user
experience by satisfying the diverse TTFT and TBT SLOs through studying three problems:
\squishlist
    \item[(1)] How to determine the padding size to ensure that the KVC demand is satisfied with a high specified probability?
     \item[(2)] How to allocate KVC to a request initially and during token generation process? 
    \item[(3)] How to choose requests to preempt?
    \item[(4)] How to choose a preemption strategy between swapping
and recomputation to reduce the preemption time (defined as the time that a request
is halted)?
\squishend

To achieve this goal, we conducted an experiments based on real traces and made the following observations (Os):

\squishlist
    \item[(1)] 
Block-based KVC allocation increases preemptions and TBT, while prediction-based allocation raises TTFT. A new method is needed to balance both.  

     \item[(2)] Previous prediction-based methods often cause underprovisioning (increasing TBT) and overprovisioning (increasing TTFT). The padding size must be carefully determined, considering the request arrival rate, to balance TTFT and TBT.
    
    \DEL{\item With the increasing arrival rate, current block and ML-prediction-oriented KVC allocation suffers from increasing waiting time. A simplistic approach to adjust the predicted length based on confidence can help reduce the waiting time and the preemption time\sh{how can this method be applied to block-method?}.} 
    \DEL{\item Demand-based memory allocation
    approaches follow the same distribution pattern \sh{what is this "distribution pattern"?} for the predicted lengths for the different datasets, although the mean and standard deviation may vary\sh{what do you mean "the same" in spite of this?}. However, the distribution of the overprovisioned and underprovisioned tokens may differ for the datasets. \sh{what is the purpose to talk about these two observations?} \sh{are you sure what you wrote is correct? Please do not say or write anything that you actually do not know or are not certain, check online to be sure before write it down.}}

    \item[(3)] FCFS-based preemption results in more preemptions than approaches that account for remaining time and KVC usage.
    \item[(4)] 
For sequences exceeding the sweet spot length, swapping results in shorter times compared to recomputation.
\squishend

\DEL{{\color{purple}
\begin{enumerate}
    \item Demand-based or static ML-prediction-oriented KVC allocation suffers from high latency for increasing arrival rates. A dynamic adjustable length prediction scheme can reduce the likelihood of underprovisioning and overprovisioning, thereby minimizing latency.\sh{If aims to advocate "dynamic adjustable length", why discuss "increasing arrival rates". This should be separate.}
    \item Demand-based memory allocation
    approaches follow the same distribution pattern \sh{what is this "distribution pattern"?} for the predicted lengths for the different datasets, although the mean and standard deviation may vary\sh{what do you mean "the same" in spite of this?}. However, the distribution of the overprovisioned and underprovisioned tokens may differ for the datasets. \sh{what is the purpose of talking about these two observations?} \sh{are you sure what you wrote is correct? Please do not say or write anything that you actually do not know or are not certain, check online to be sure before write it down.}
    \item Many requests (48\%) are swapped out because of being underprovisioned for a low amount of tokens ($\leq$20 tokens). \sh{why is there 20 threshold?-done}
    \item \sh{first, you need to show the swapping and recomputation takes long time relative to iteration/response time.-done}While doing the preemption, eviction policies considering multiple factors, such as remaining processing time and KV-cache occupancy, can make more informed decisions, leading to improved overall performance compared to a random policy.\sh{what is the random policy? each method/term must be explained before using it.}
    \item A comprehensive approach to determine the preemption policy (swapping vs. recomputation), \sh{this observation is duplicated with the above} that considers various influencing factors can lead to better resource management. The factors include the size of the preempted data and memory (PCIe) bandwidth and real-time metrics like the input and output length of the requests, the number of running requests in the system, and GPU/memory usage.
\end{enumerate}}}



Leveraging these observations, we propose a system, that optimizes the cache operations for mitigating
KV cache competition (\sys). We name this \sys to reflect the system's focus on efficient cache usage. \sys consists of the following components.
\squishlist
    \item[(1)] \textbf{Confidence-based Padding.} We modify an LLM model to predict output length and deviation direction for adding or subtracting padding. Using Hoeffding's inequality theory~\cite{hoeffding1963,bentkus2004hoeffding}, we determine the padding needed to ensure a request's KVC demand is satisfied with a high specified
probability and adjust it based on arrival rate to balance TTFT and TBT. (O2)

    \item[(2)] \textbf{SLO-aware Batching and KVC Allocation.} We reuse allocated but unused KVC and select waiting and returned requests that must run to satisfy their TTFT and TBT SLOs. The remaining unallocated KVC is distributed among the selected requests to maximize throughput. Additionally, it proactively allocates KVC before instead of at the time a request exhausts its allocation and reserves KVC globally to prevent preemptions. (O1)

    \item[(3)] \textbf{Preemption Policy. } We order requests for preemption based on their latency SLO, remaining completion time and KVC occupancy in order to avoid SLO violation and preemptions, and reduce preemption time.
    (O3)

      \item[(4)] \textbf{Preemption Strategy Selection.} When the KVC size exceeds the observed sweet spot sequence length, we opt for swapping; otherwise, we choose recomputation. (O4)

\squishend

Experimental results show that \sys achieves up to a 3.29$\times$ and 2.83$\times$ lower tail TBT and TTFT and 47\% and 53\% higher TTFT and TBT SLO attainments than the state-of-the-art methods, respectively. We will distribute \sys's source code after the paper is accepted.

\section{Experimental Analysis}\vspace{-0.0in}
\label{sec:analysis}

\DEL{\begin{figure*}[t]
\centering
    \subfloat[Bucket 1.\vspace{-0.01in}\label{fig:fig1}]{{\includegraphics[width=0.24\linewidth,height=0.13\textheight]{Padding-FIgs/1.pdf} }}%
    \hfill
    \subfloat[Bucket 2.\vspace{-0.01in}\label{fig:fig2}]{{\includegraphics[width=0.24\linewidth,height=0.13\textheight]{Padding-FIgs/2.pdf} }}%
    \hfill
    \subfloat[Bucket 3.\vspace{-0.01in}\label{fig:fig3}]{{\includegraphics[width=0.24\linewidth,height=0.13\textheight]{Padding-FIgs/3.pdf} }}%
    \hfill
    \subfloat[Bucket 4.\vspace{-0.01in}\label{fig:fig4}]{{\includegraphics[width=0.24\linewidth,height=0.13\textheight]{Padding-FIgs/4.pdf} }}%
    \hfill

    \subfloat[Bucket 5.\vspace{-0.01in}\label{fig:fig5}]{{\includegraphics[width=0.24\linewidth,height=0.13\textheight]{Padding-FIgs/5.pdf} }}%
    \hfill
    \subfloat[Bucket 6.\vspace{-0.01in}\label{fig:fig6}]{{\includegraphics[width=0.24\linewidth,height=0.13\textheight]{Padding-FIgs/6.pdf} }}%
    \hfill
    \subfloat[Bucket 7.\vspace{-0.01in}\label{fig:fig7}]{{\includegraphics[width=0.24\linewidth,height=0.13\textheight]{Padding-FIgs/7.pdf} }}%
    \hfill
    \subfloat[Bucket 8.\vspace{-0.01in}\label{fig:fig8}]{{\includegraphics[width=0.24\linewidth,height=0.13\textheight]{Padding-FIgs/8.pdf} }}%
    \hfill
    \subfloat[Bucket 9.\vspace{-0.01in}\label{fig:fig9}]{{\includegraphics[width=0.24\linewidth,height=0.13\textheight]{Padding-FIgs/9.pdf} }}%
    \hfill
    \subfloat[Bucket 10.\vspace{-0.01in}\label{fig:fig10}]{{\includegraphics[width=0.24\linewidth,height=0.13\textheight]{Padding-FIgs/10.pdf} }}%
    \hfill
    \subfloat[Bucket 11.\vspace{-0.01in}\label{fig:fig11}]{{\includegraphics[width=0.24\linewidth,height=0.13\textheight]{Padding-FIgs/11.pdf} }}%
    \hfill
    \subfloat[Bucket 12.\vspace{-0.01in}\label{fig:fig12}]{{\includegraphics[width=0.24\linewidth,height=0.13\textheight]{Padding-FIgs/12.pdf} }}%
    \hfill

    \subfloat[Bucket 13.\vspace{-0.01in}\label{fig:fig13}]{{\includegraphics[width=0.24\linewidth,height=0.13\textheight]{Padding-FIgs/13.pdf} }}%
    \hfill
    \subfloat[Bucket 14.\vspace{-0.01in}\label{fig:fig14}]{{\includegraphics[width=0.24\linewidth,height=0.13\textheight]{Padding-FIgs/14.pdf} }}%
    \hfill
    \subfloat[Bucket 15.\vspace{-0.01in}\label{fig:fig15}]{{\includegraphics[width=0.24\linewidth,height=0.13\textheight]{Padding-FIgs/15.pdf} }}%
    \hfill
    \subfloat[Bucket 16.\vspace{-0.01in}\label{fig:fig16}]{{\includegraphics[width=0.24\linewidth,height=0.13\textheight]{Padding-FIgs/16.pdf} }}%
    \hfill

    \subfloat[Bucket 17.\vspace{-0.01in}\label{fig:fig17}]{{\includegraphics[width=0.24\linewidth,height=0.13\textheight]{Padding-FIgs/17.pdf} }}%
    \hfill
    \subfloat[Bucket 18.\vspace{-0.01in}\label{fig:fig18}]{{\includegraphics[width=0.24\linewidth,height=0.13\textheight]{Padding-FIgs/18.pdf} }}%
    \hfill
    \subfloat[Bucket 19.\vspace{-0.01in}\label{fig:fig19}]{{\includegraphics[width=0.24\linewidth,height=0.13\textheight]{Padding-FIgs/19.pdf} }}%
    \hfill
    \subfloat[Bucket 20.\vspace{-0.01in}\label{fig:fig20}]{{\includegraphics[width=0.24\linewidth,height=0.13\textheight]{Padding-FIgs/20.pdf} }}%
    \hfill

    \caption{\small{Probability distribution of difference of actual value and predicted value.\vspace{-0.15in}}}
    \label{fig:bucket-20}
\end{figure*}}


\DEL{\sh{need to add figs in the analysis section: token budget is target forward size
0.0: Y: CDF of iterations, X: Number of requests added to the batch
0.1: Y: CDF of requests, X: wait time (initial waiting time, not preemption time)-done}

\sh{0: Y: token budget used, Allocated KVC tokens --2 bars together, X: Each iteration 1, 2, 3, 4.....
1. Y: CDF of iterations, X: Unused token budget
2. Y: average \%, two bars for each X point: 1) token budget used, Allocated KVC tokens, X points: 1)Not reach token budget, 2) Reach token budget, 3) all results-done}

\sh{3. y: \% of iterations, X: Preemption time, -- not, if no preemption, then preemption time=0 -- showed you the figure, asked to remove.
4. Y: Num. of preemptions, X: each preemption policy -- you called it eviction policy
Use Splitwise paper dataset to set arrival rate
-done}
}

\DEL{\begin{thm}\label{1Obs:GPULimit}
When the token budget is used up but there still exist long-waiting requests, to reduce TTFT, we can add the requests to the batch if the increase iteration latency won't affect the iteration SLO of the requests in the batch.
\end{thm}


\begin{thm}\label{3PaddingValue}
The allocated KVC for each request (hence the request dequeueing speed) and the request arrival rate affect waiting time.
\end{thm}

\begin{thm}\label{4Policy}
Using FCFS to select a request to preempt may lead to more preemptions due to limited released KVC.
\end{thm}

\begin{thm}\label{5Strategy}
When the sequence length is long (greater than \sh{??k}), swapping generates shorter preemption time than recomputation.
\end{thm}
}

\subsection{Experiment Settings}\vspace{-0.0in}\label{sec:setting}
\noindent{\textbf{Machine settings.}} We used an AWS p4d.24xlarge instance, which features 8 NVIDIA A100 GPUs.
Each GPU has 80GB of memory. The GPUs are interconnected via a 600 GB/s NVSwitch. We executed the OPT-13B model~\cite{Zhang2022OPTOP} on a single GPU, 
and the OPT-175B model, partitioning the model across 8 GPUs with both model and tensor parallelism degree of 2 as in~\cite{280922}.

\DEL{\begin{figure*}[t]
\centering
    \subfloat[Bucket 2.\vspace{-0.01in}\label{fig:diff-fig1}]{{\includegraphics[width=0.24\linewidth,height=0.13\textheight]{Padding-FIgs/diff-2.pdf} }}%
    \hfill
    \subfloat[Bucket 3.\vspace{-0.01in}\label{fig:diff-fig2}]{{\includegraphics[width=0.24\linewidth,height=0.13\textheight]{Padding-FIgs/diff-bucket-3.pdf} }}%
    \hfill
    \subfloat[Bucket 18.\vspace{-0.01in}\label{fig:diff-fig3}]{{\includegraphics[width=0.24\linewidth,height=0.13\textheight]{Padding-FIgs/diff-18.pdf} }}%
    \hfill
    \subfloat[Bucket 20.\vspace{-0.01in}\label{fig:diff-fig4}]{{\includegraphics[width=0.24\linewidth,height=0.13\textheight]{Padding-FIgs/diff-20.pdf} }}%
    \hfill
    \caption{\small{Differnce of probability of training and inference for 4 buckets.\vspace{-0.15in}}}
    \label{fig:jct-4}
\end{figure*}}


\noindent{\textbf{Request settings.}} 
We used the Alpaca~\cite{alpaca}, ShareGPT~\cite{sharegpt}, and BookCorpus~\cite{soskkobayashi2018bookcorpus} traces. \DEL{Alpaca and ShareGPT contain 52K and 90K requests, respectively. BookCorpus includes 11K unpublished books featuring longer prompt lengths.}  
Table~\ref{tab:trace-table} shows their features and the mean request arrival rate setting. It
follows a Poisson distribution~\cite{280922,vllm}. The block size was set to 32. For BookCorpus~\cite{soskkobayashi2018bookcorpus}, we divided the prompts into 2048 tokens to meet the requirements of the LLM models. We used 3-hour trace for OPT-13B and 1-hour trace for OPT-175B.

\DEL{\begin{table}[]
\centering
\caption{Trace properties and experiment settings}\vspace{-0.15in}
\label{tab:trace-table}
\resizebox{\columnwidth}{!}{%
\begin{tabular}{|l|lll|lll|c|}
\hline
\multirow{2}{*}{Trace} &
  \multicolumn{3}{l|}{Input length} &
  \multicolumn{3}{l|}{Output length} &
  \multirow{2}{*}{Arrival rate} \\ \cline{2-7}
 &
  \multicolumn{1}{l|}{avg} &
  \multicolumn{1}{l|}{min} &
  max &
  \multicolumn{1}{l|}{avg} &
  \multicolumn{1}{l|}{min} &
  max &
   \\ \hline
Alpaca &
  \multicolumn{1}{l|}{19.31} &
  \multicolumn{1}{l|}{9} &
  2.47K &
  \multicolumn{1}{l|}{58.41} &
  \multicolumn{1}{l|}{13} &
  292 &
  32 \\ \hline
ShareGPT   & \multicolumn{1}{l|}{161.31}  & \multicolumn{1}{l|}{16} & 3.2K & \multicolumn{1}{l|}{337.99} & \multicolumn{1}{l|}{19} & 991  & 28  \\ \hline
BookCorpus & \multicolumn{1}{l|}{1952.11} & \multicolumn{1}{l|}{18} & 461K & \multicolumn{1}{l|}{681.2}  & \multicolumn{1}{l|}{32} & 1041 & 1.2 \\ \hline
\end{tabular}%
}
\end{table}}

\begin{table}[]
\centering
\caption{Trace properties and experiment settings.}
\label{tab:trace-table}
\vspace{-0.1in}
\resizebox{\columnwidth}{!}{%
\begin{tabular}{|l|lll|lll|l|c|}
\hline
\multirow{2}{*}{Trace} &
  \multicolumn{3}{l|}{Input length} &
  \multicolumn{3}{l|}{Output length} &
  \multirow{2}{*}{\begin{tabular}[c]{@{}l@{}}Number of \\ requests\end{tabular}} &
  \multirow{2}{*}{Mean arrival rate} \\ \cline{2-7}
           & \multicolumn{1}{l|}{avg}     & \multicolumn{1}{l|}{min} & max   & \multicolumn{1}{l|}{avg}    & \multicolumn{1}{l|}{min} & max  &     &     \\ \hline
Alpaca     & \multicolumn{1}{l|}{19.31}   & \multicolumn{1}{l|}{9}   & 2.47K & \multicolumn{1}{l|}{58.41}  & \multicolumn{1}{l|}{13}  & 292  & 52K & 32  \\ \hline
ShareGPT   & \multicolumn{1}{l|}{161.31}  & \multicolumn{1}{l|}{16}  & 3.2K  & \multicolumn{1}{l|}{337.99} & \multicolumn{1}{l|}{19}  & 991  & 90K & 28  \\ \hline
BookCorpus & \multicolumn{1}{l|}{1952.11} & \multicolumn{1}{l|}{18}  & 461K  & \multicolumn{1}{l|}{681.2}  & \multicolumn{1}{l|}{32}  & 1041 & 11K & 1.2 \\ \hline
\end{tabular}%
}
\end{table}

\begin{figure*}[t]
\centering
     \subfloat[Alpaca.\vspace{-0.01in}\label{fig:jct-alpaca-13}]{{\includegraphics[width=0.24\linewidth,height=0.13\textheight]{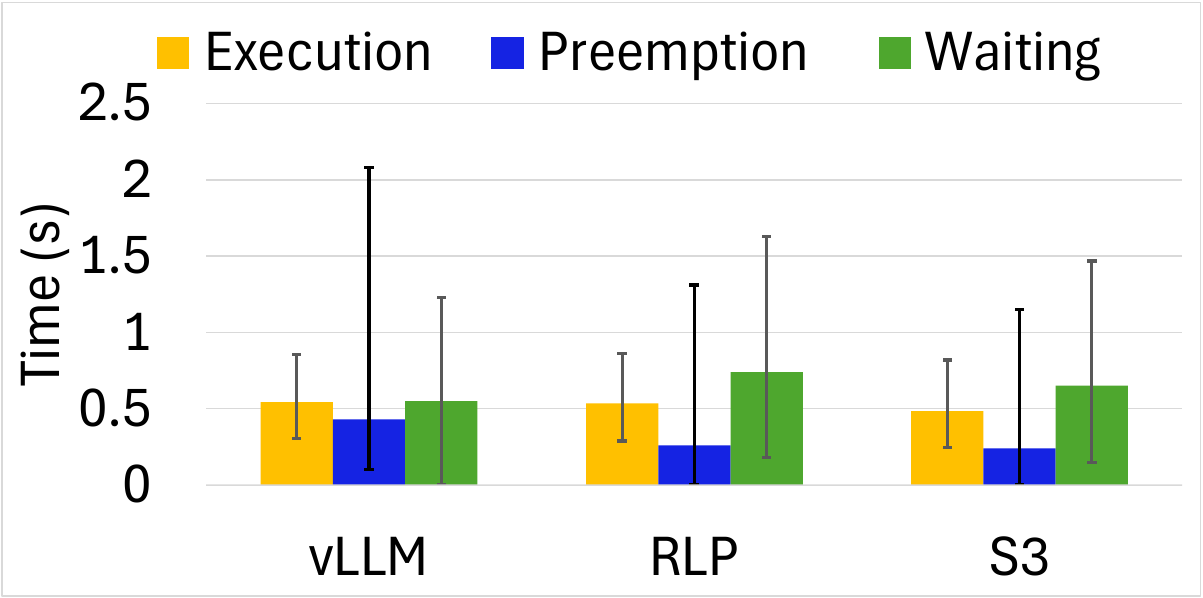} }}
    \hfill
    \subfloat[ShareGPT.\vspace{-0.01in}\label{fig:jct-share-13}]{{\includegraphics[width=0.24\linewidth,height=0.13\textheight]{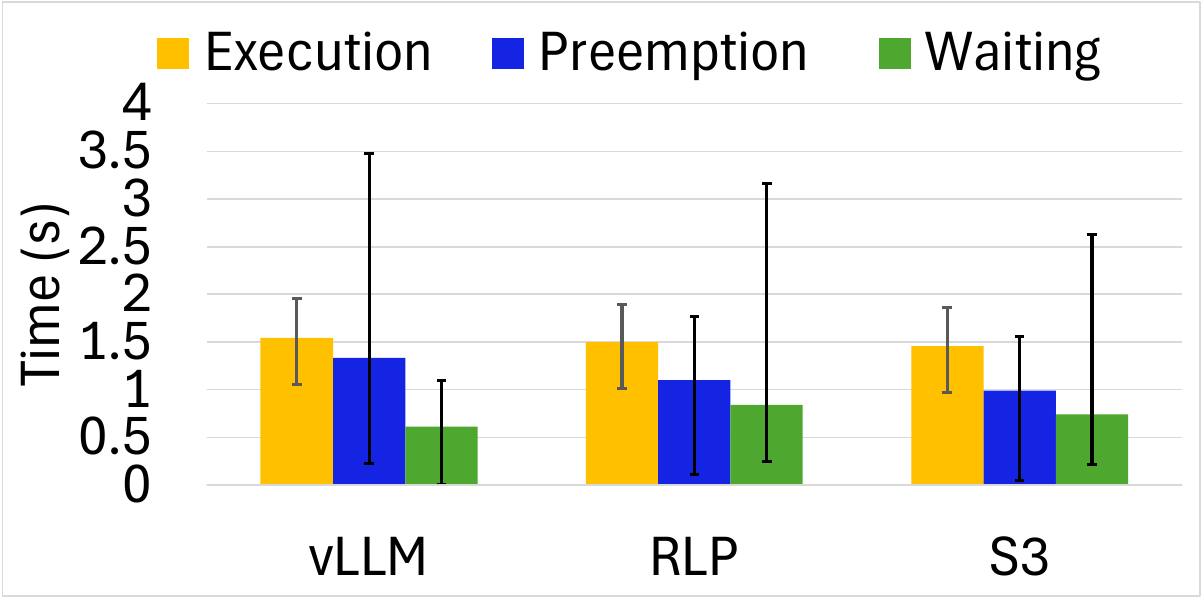} }}
    \hfill
    \subfloat[BookCorpus.\vspace{-0.01in}\label{fig:jct-bookcorpus-13}]{{\includegraphics[width=0.24\linewidth,height=0.13\textheight]{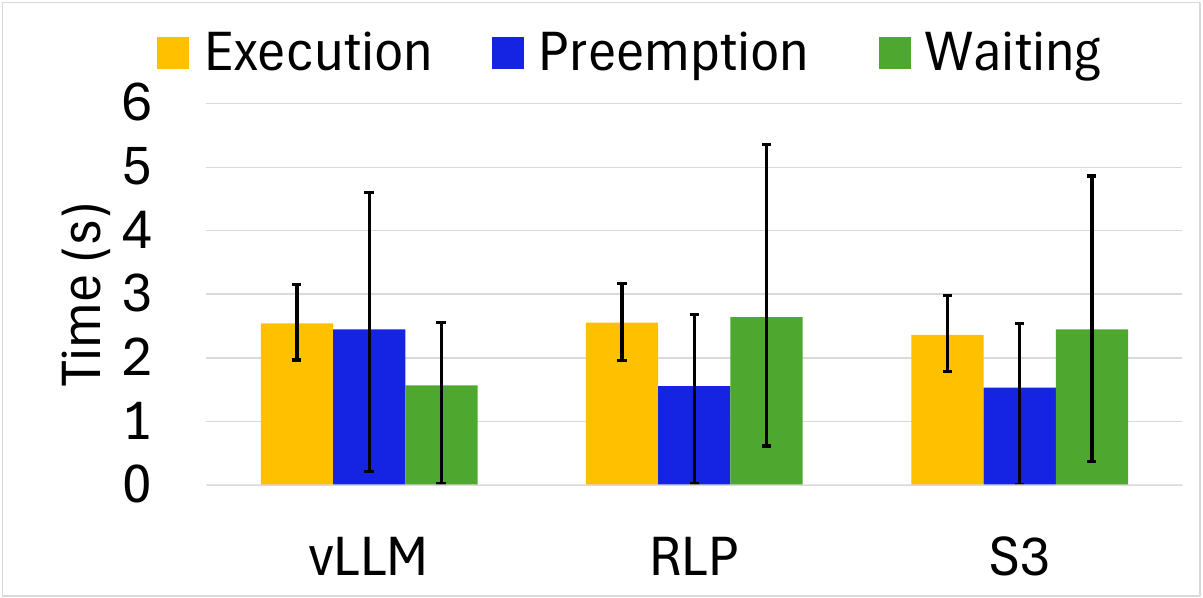} }}
    \hfill
    \subfloat[Number of preemptions.\vspace{-0.01in}\label{fig:preemptions}]{{\includegraphics[width=0.24\linewidth,height=0.13\textheight]{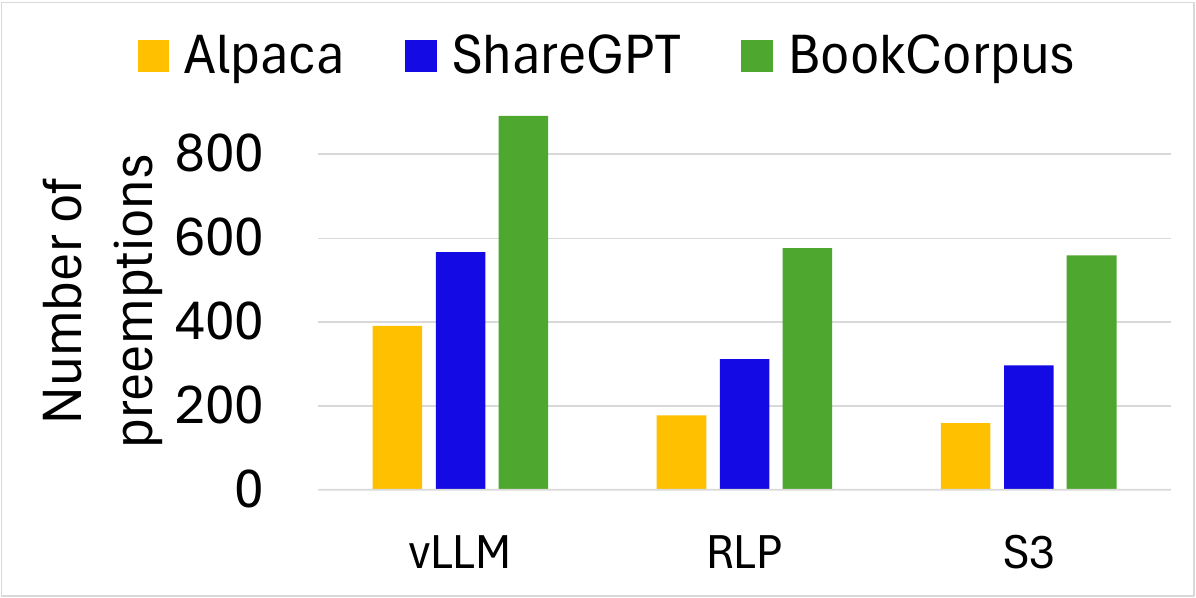} }}
    \hfill
    \vspace{-0.1in}
   \caption{{Measurements for different traces for OPT-13B. 
   }} 
    \label{fig:jct-time}
\end{figure*}

\begin{figure*}[t]
\centering
     \subfloat[Alpaca.\vspace{-0.01in}\label{fig:jct-alpaca-175}]{{\includegraphics[width=0.24\linewidth,height=0.13\textheight]{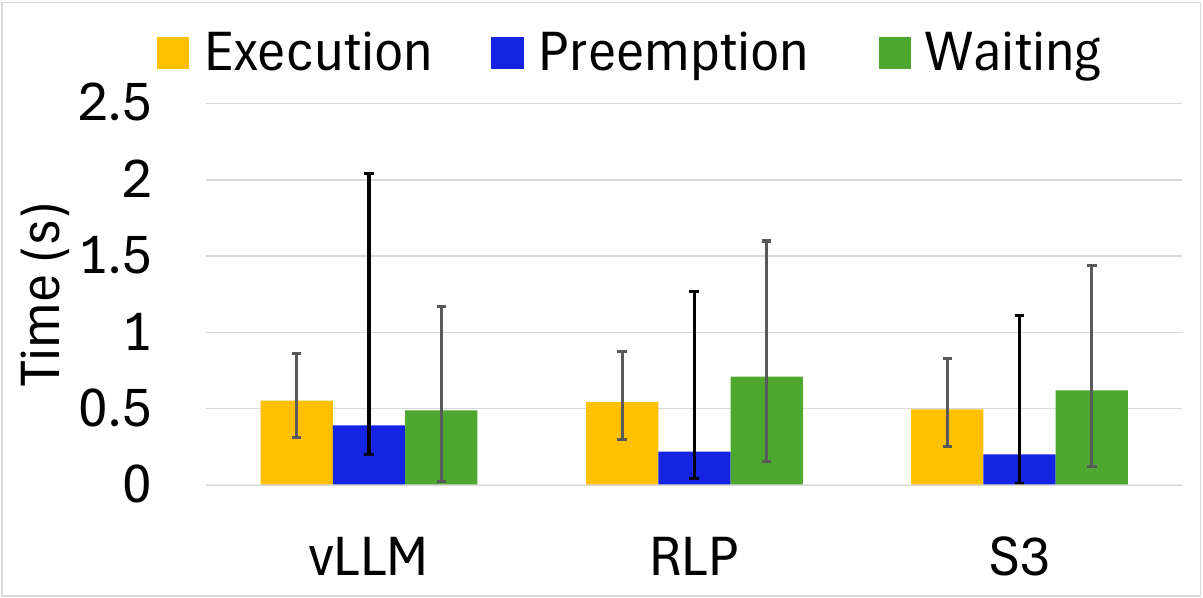} }}
    \hfill
    \subfloat[ShareGPT.\vspace{-0.01in}\label{fig:jct-sharegpt-175}]{{\includegraphics[width=0.24\linewidth,height=0.13\textheight]{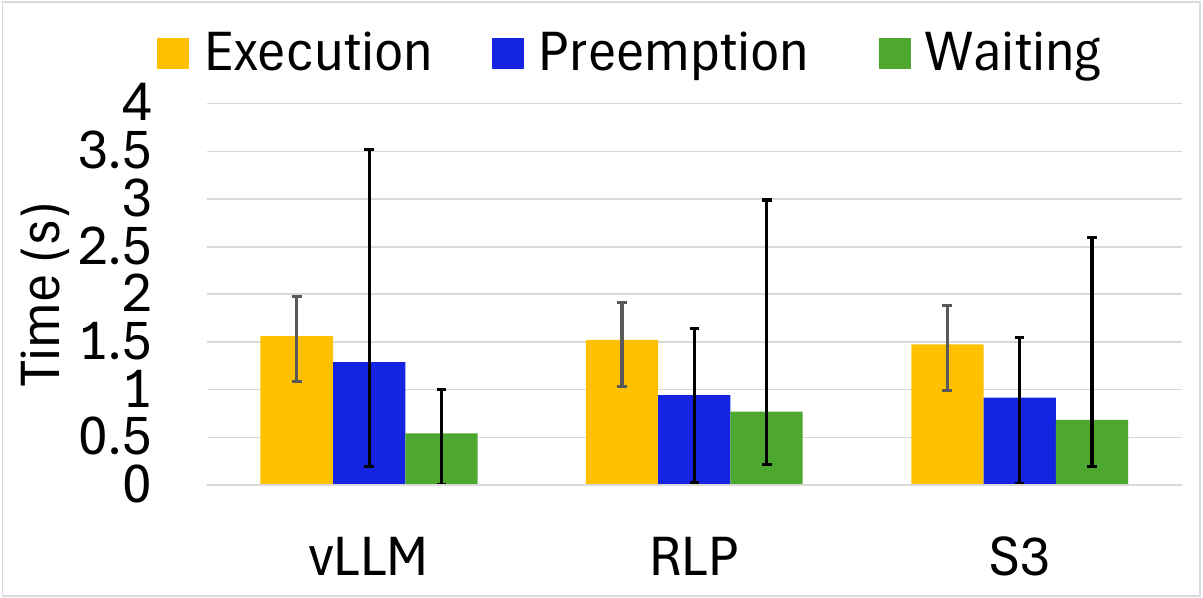} }}
    \hfill
    \subfloat[BookCorpus.\vspace{-0.01in}\label{fig:jct-bookcorpus-175}]{{\includegraphics[width=0.24\linewidth,height=0.13\textheight]{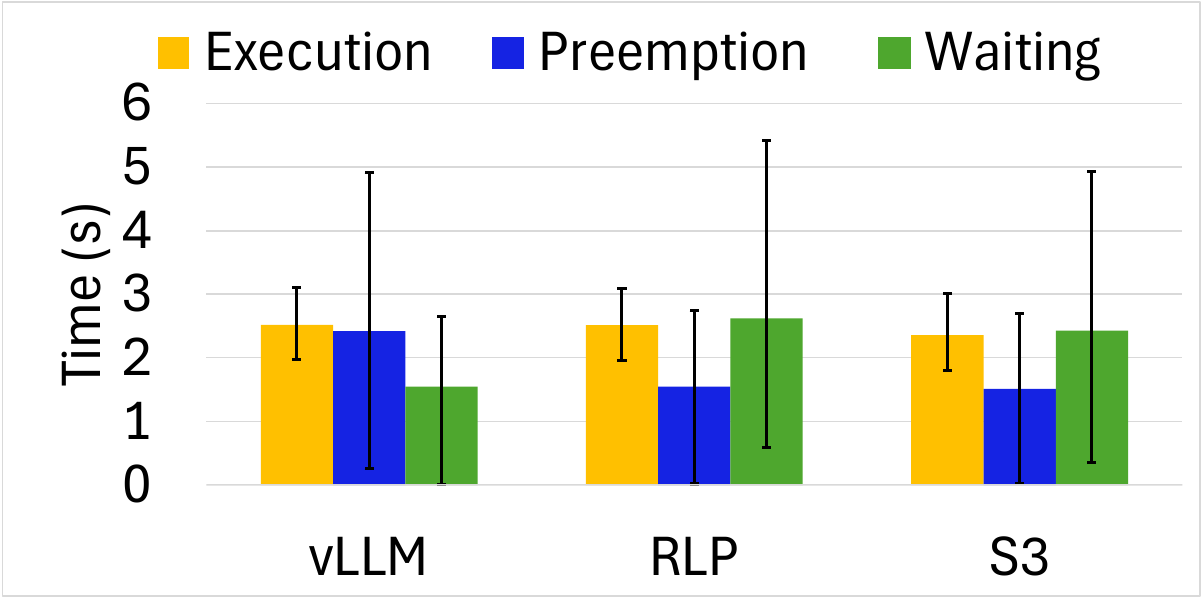} }}
    \hfill
    \subfloat[Number of prremptions.\vspace{-0.01in}\label{fig:preemptions}]{{\includegraphics[width=0.24\linewidth,height=0.13\textheight]{Padding-FIgs/number-of-preemptions.pdf} }}
    \hfill
     \vspace{-0.1in}
   \caption{{Measurements for different traces for OPT-175B. }}
    \label{fig:jct-time-175}
\end{figure*}


\noindent{\textbf{Schedulers.}} We conducted the experiment measurements for the following methods.  
1) Response Length Prediction (RLP)~\cite{Zheng2023ResponseLP} uses the LLM to predict the response length from a request and always adds a padding of 100 tokens. It tries to schedule the requests with similar lengths in the same batch. Any underprovisioned requests are preempted and later executed when there is sufficient KVC. 2) $S^3$~\cite{Jin2023S3IG} uses the LLM model to predict the response length bucket (with a 50-token increment) for each request and allocates the upper bound of the predicted bucket.If a request faces insufficiently allocated KVC, it is preempted, and its KVC demand is doubled for the next allocation. 3) vLLM~\cite{vllm} 
uses the block-based KVC allocation and FCFS-based preemption, and defers new requests until preempted ones are completed. 


To make these methods comparable, we used our own fine-tuned OPT-13B model for the response length prediction in RLP and $S^3$. \DEL{We used 60\%  requests for the fine-tuning, and the rest for the experiments out of 10K requests from each trace. }To avoid the interference on the LLM inference, we ran this model in another server. In this paper, \emph{predicted output length} refers to the output length from the LLM, and \emph{estimated output length} refers to the predicted output length adjusted with the padding. 

\subsection{Impact of KVC Allocation Methods on TTFT and TBT}

\begin{figure}[t]
\centering
\subfloat[OPT-13B.\label{fig:avg-added}]{\includegraphics[width=0.48\linewidth,height=0.13\textheight]{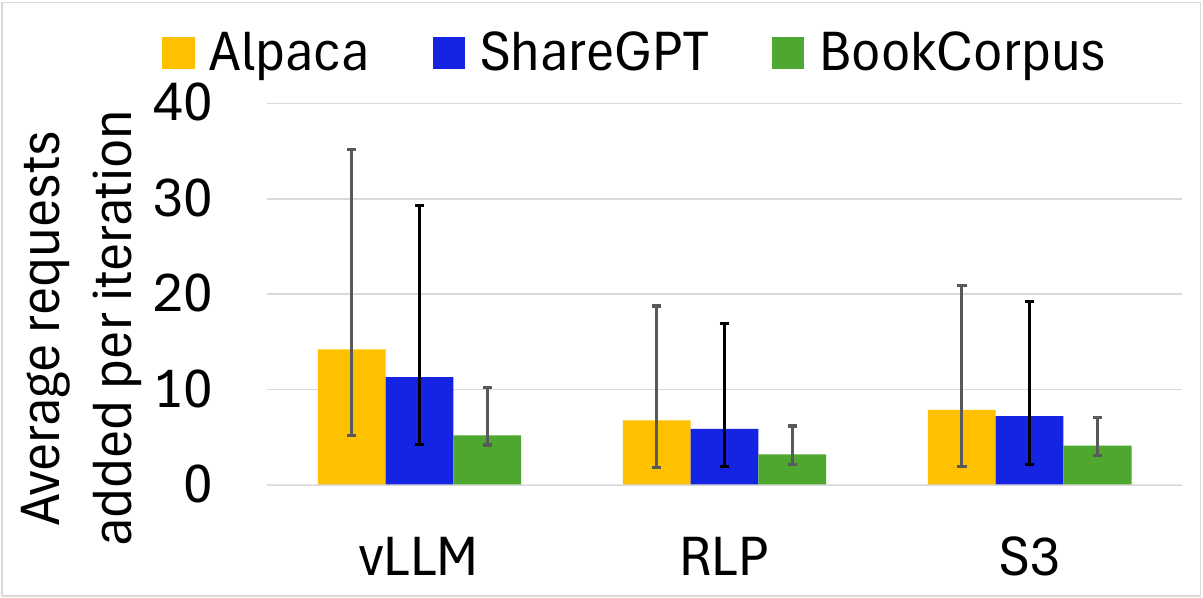}}\hfill
\subfloat[OPT-175B. \label{fig:average-added-175b}]{\includegraphics[width=0.48\linewidth,height=0.13\textheight]{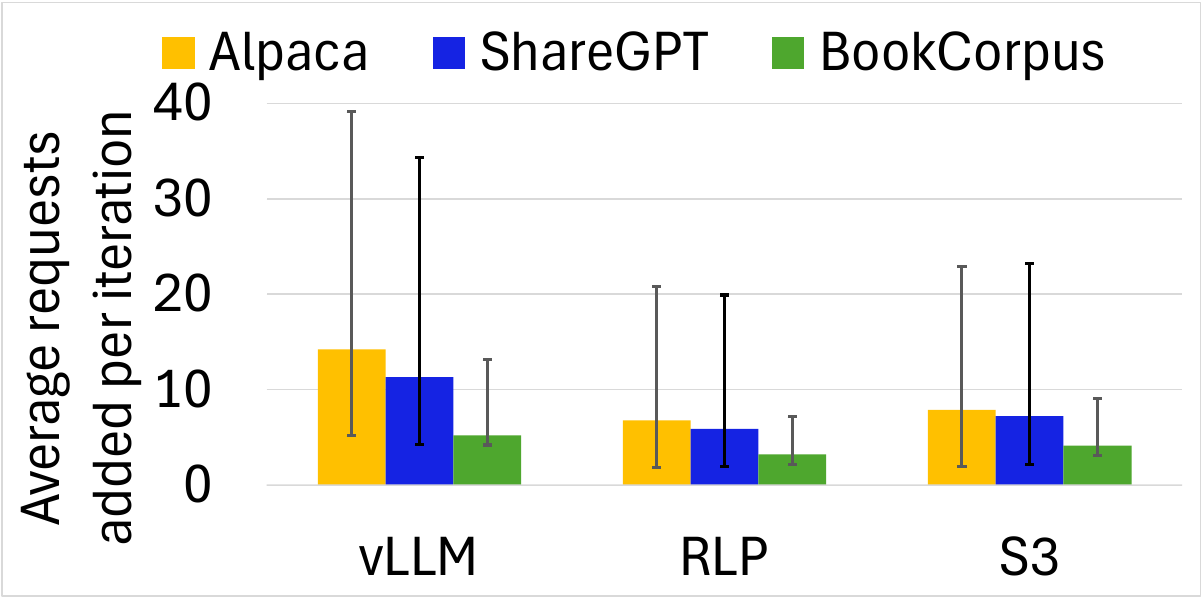}}
\hfill
\caption{\small Average requests added per iteration.}
\label{fig:request-added}
\end{figure}

Although~\cite{Zheng2023ResponseLP} indicates that overprovisioning is less critical than underprovisioning, this claim primarily applies to requests with overprovisioning. However, overprovisioning reduces the number of requests accommodated in a batch, increasing waiting time and TTFT. A request's \emph{waiting time} is the duration its prompt remains in the queue before execution begins; and its \emph{execution time} is the duration from when it is dispatched to the execution engine until completion, excluding preemption time. Figures~\ref{fig:jct-alpaca-13}-\ref{fig:jct-bookcorpus-13} and Figure~\ref{fig:jct-alpaca-175}-\ref{fig:jct-bookcorpus-175} 
show the average execution, preemption, and waiting times for OPT-13B and OPT-175B, with error bars representing the min and max values.
RLP and $S^3$ generate 29\%-49\% and 26\%-45 lower preemption time but 43\%-72\% and 27\%-65\% higher waiting time than vLLM. 
This is because that in vLLM, each request is only allocated with a block at a time, and it has to be preempted upon KVC allocation failure when it uses up its allocated block. RLP and $S^3$ allocate to a request the KVC equal to its estimated sequence length, so they reduce the number of requests accommodated in a batch and increase the TTFT but reduce the premption time, which occurs only at  underprovisioning. 
These findings are verified in Figure~\ref{fig:preemptions}, which shows the average number of preemptions, and  Figure~\ref{fig:request-added}, which shows the average number of requests added to the batch per iteration, with error bars representing the minimum and maximum values. $S^3$ has 7\% lower preemption time, 5\% lower waiting time, and adds 12-17\% more requests to the batch per iteration than RLP since $S^3$ adds 100-token padding while RLP's bucket size is 50. We calculated that the preemption time and the waiting time can take up to 71\% and 75\% of the execution time on average across the three datasets.\looseness=-1 

\begin{figure}[t]
\centering
 \DEL{\subfloat[Used token budgets and allocated KVC tokens.\vspace{-0.01in}\label{fig:tk1}]{{\includegraphics[width=0.48\linewidth,height=0.13\textheight]{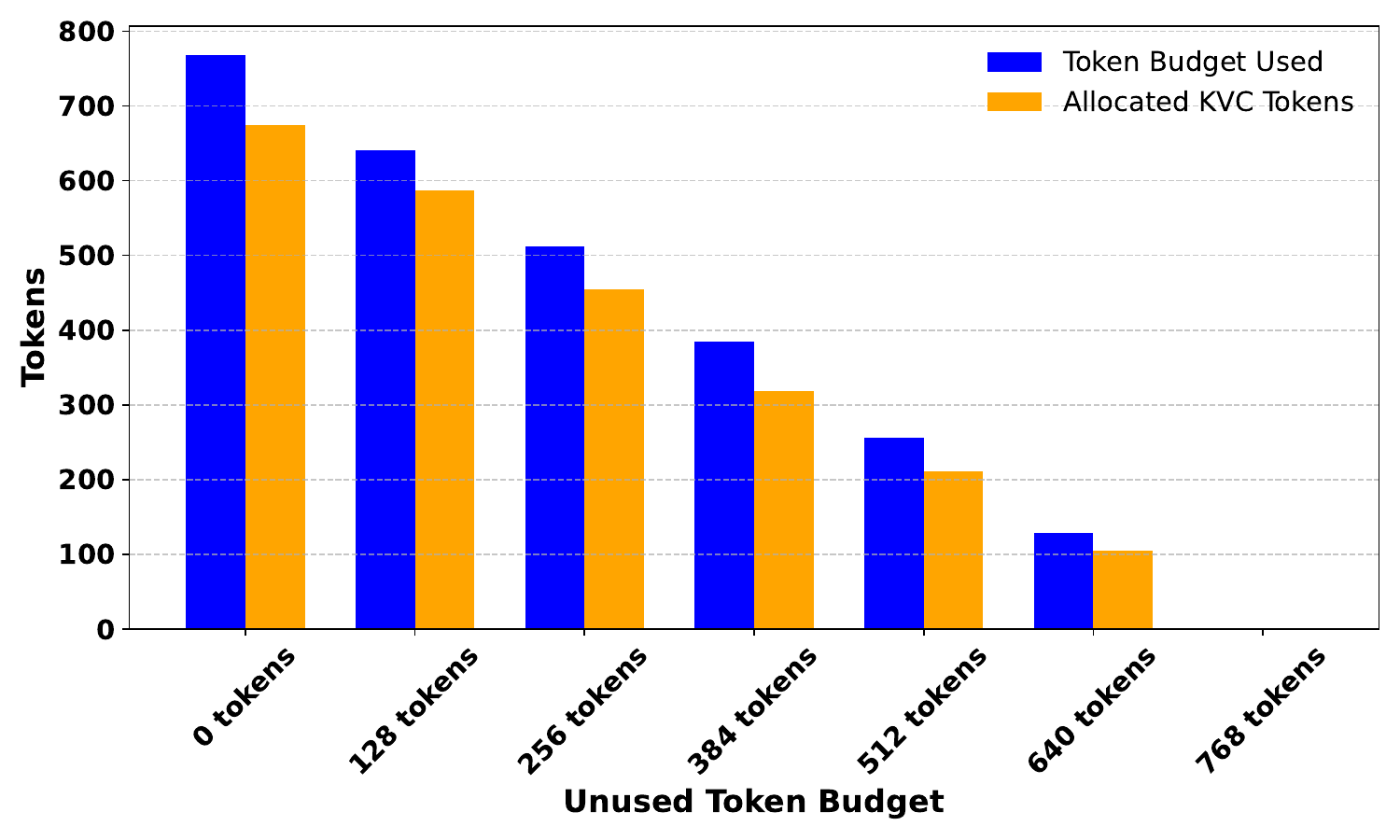} }}%
    \hfill}
    \subfloat[OPT-13B.\vspace{-0.01in}\label{fig:waiting-1}]{{\includegraphics[width=0.48\linewidth,height=0.13\textheight]{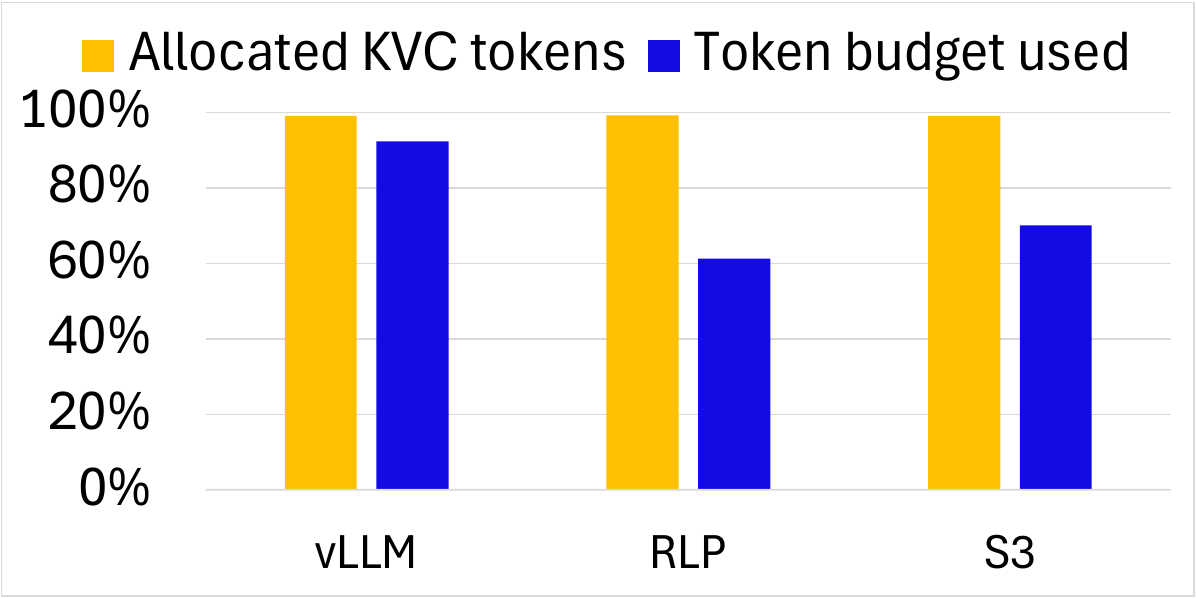} }}
    \hfill
    \subfloat[OPT-175B.\vspace{-0.01in}\label{fig:waiting-2}]{{\includegraphics[width=0.48\linewidth,height=0.13\textheight]{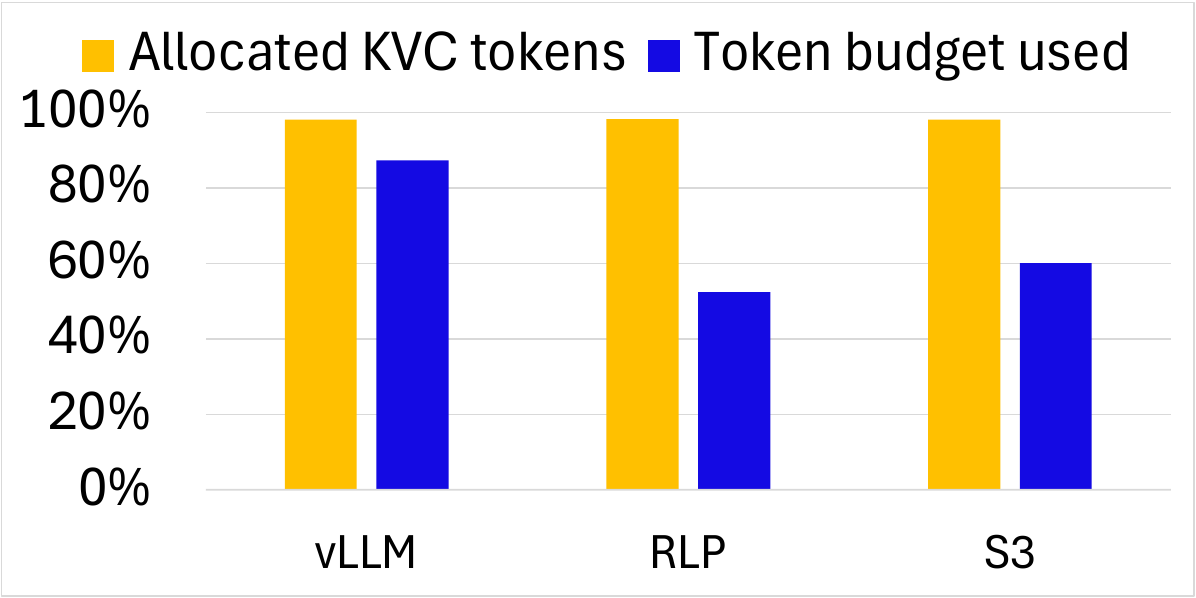} }}
    \hfill
\vspace{-0.1in}\caption{\small{Illustration of resource bottleneck.\vspace{-0.0in}}}
    \label{fig:waiting}
\end{figure}

\begin{figure}[t]
\centering
\subfloat[Overprovisioning.\label{fig:overprovisioned-13b}]{\includegraphics[width=0.48\linewidth,height=0.13\textheight]{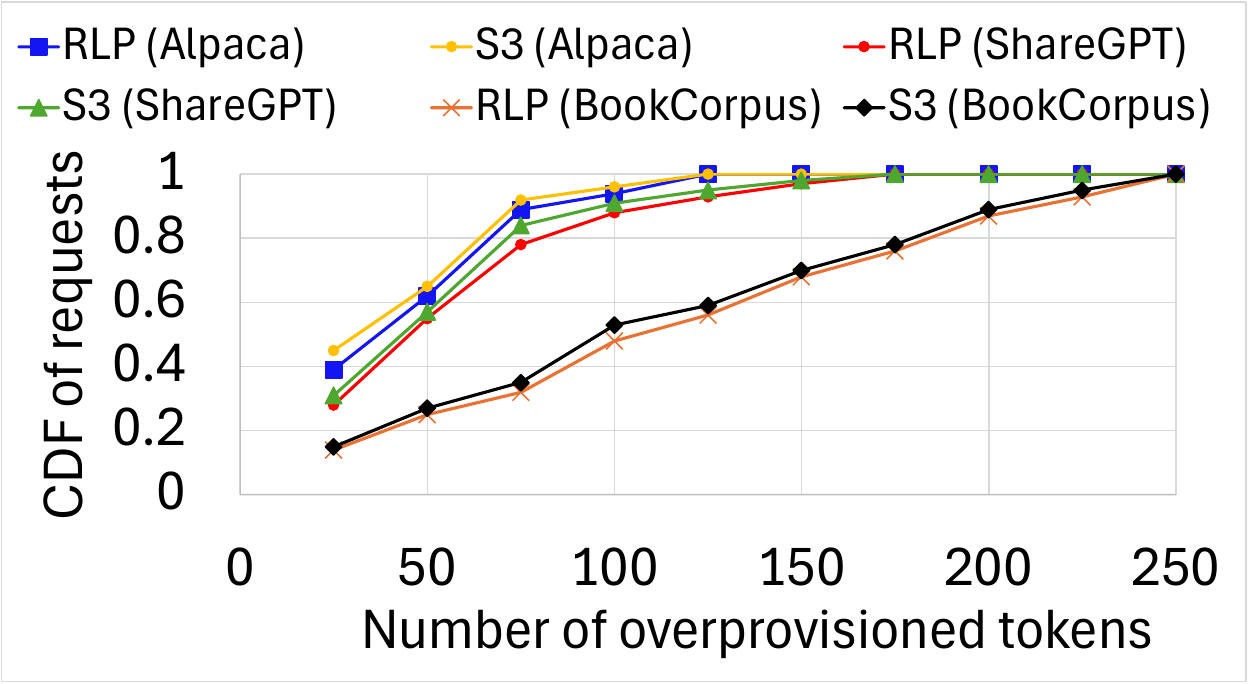}}\hfill
\subfloat[Underprovisioning. \label{fig:underprovisioned-13b}]{\includegraphics[width=0.48\linewidth,height=0.13\textheight]{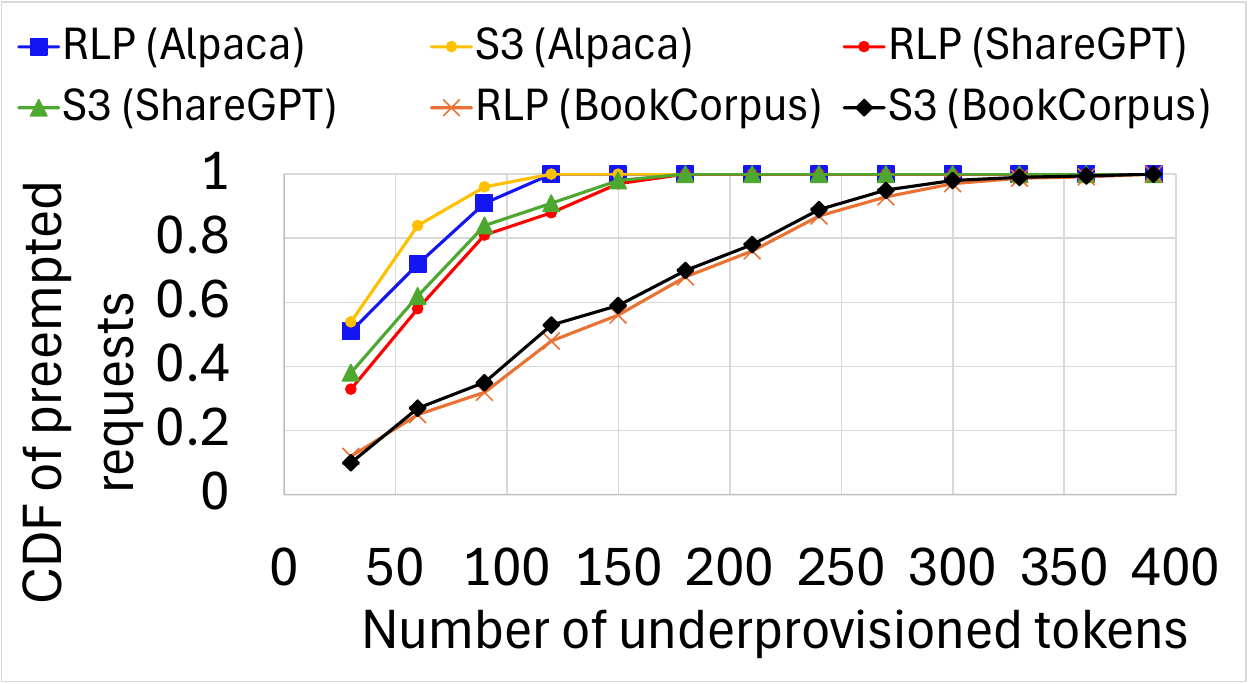}}
\hfill
\caption{\small CDF of requests vs. over/under provisioned tokens.}
\label{fig:cdf-over-under-13b}
\end{figure}

\begin{figure*}[t]
\centering
     \subfloat[Alpaca.\vspace{-0.01in}\label{fig:latency-alpaca-pad}]{{\includegraphics[width=0.32\linewidth,height=0.13\textheight]{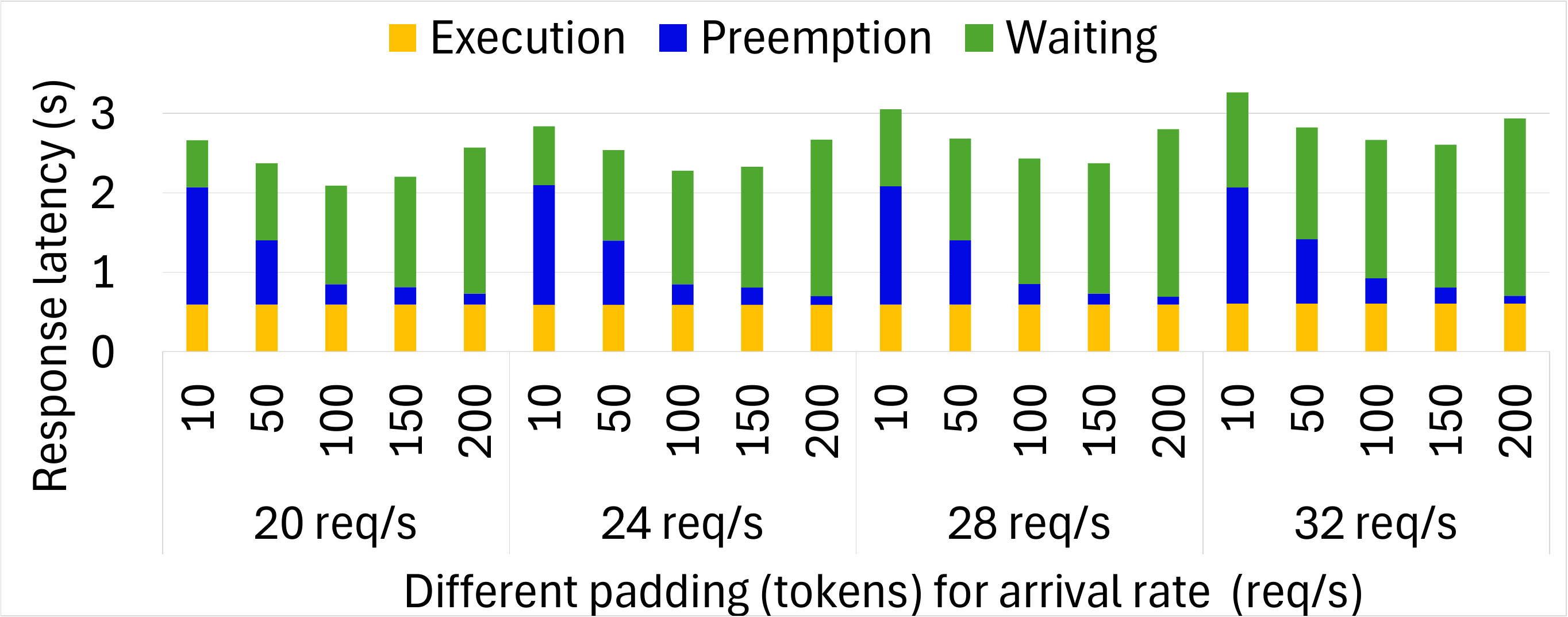} }}
    \hfill
    \subfloat[ShareGPT.\vspace{-0.01in}\label{fig:latency-sharegpt-pad}]{{\includegraphics[width=0.32\linewidth,height=0.13\textheight]{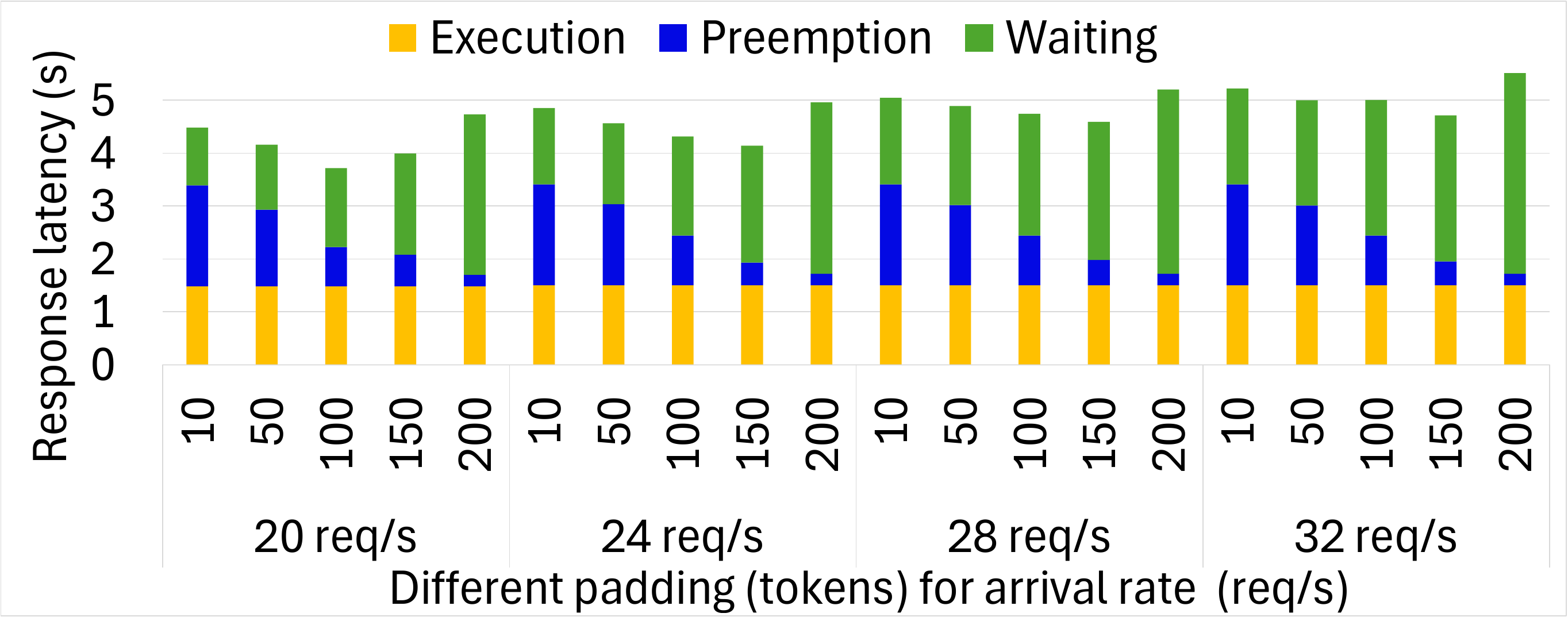}
    } }
    \hfill
    \subfloat[BookCorpus.\vspace{-0.01in}\label{fig:latency-bookcorpus-pad}]{{\includegraphics[width=0.32\linewidth,height=0.13\textheight]{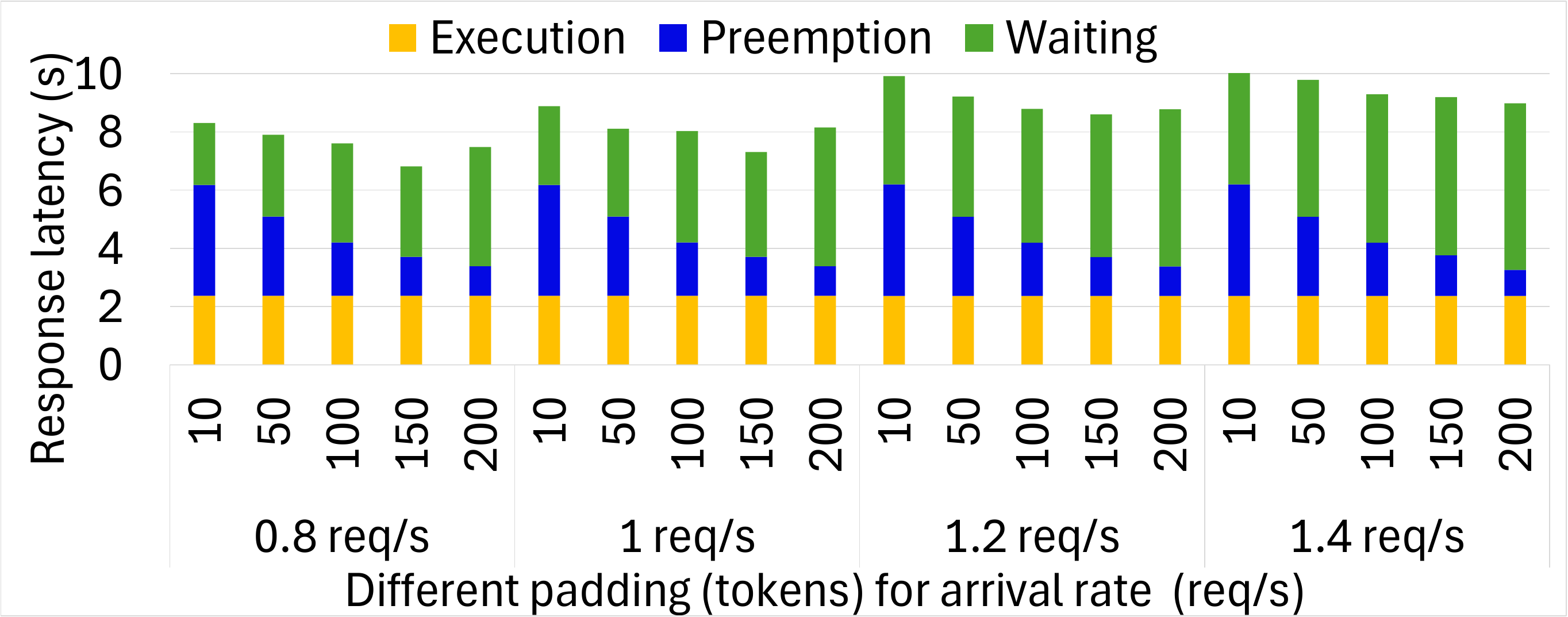} }}
    \hfill
     \vspace{-0.1in}
   \caption{Trade-off between preemption time and waiting time versus padding size for OPT-13B.}\vspace{-0.0in}%
    \label{fig:trade-off-2}
\end{figure*}

We find the \emph{token budget} for the number of tokens in a batch to maximize throughput as in~\cite{Agrawal2023SARATHIEL}. Figure~\ref{fig:waiting} shows the average token budget used and the average number of allocated KVC tokens at each iteration. We see that while all systems fully alloctae the KVC, vLLM more fully utilizes the token budget than RLP and $S^3$ by 37\% and 33\%. 

\begin{thm}\label{KVCLimit}
Due to KVC bottleneck, the block-based KVC allocation method increases preemptions and TBT, while the prediction-based KVC allocation method increases request waiting time and TTFT. A novel approach is required to limit both TTFT and TBT effectively.
\end{thm}

\subsection{Padding Size Determination and Impact}

Figures~\ref{fig:cdf-over-under-13b} shows the CDF of requests versus the number of over and under provisioned tokens for RLP and $S^3$. Overprovisioning and underprovisioning are consistently observed. Different requests exhibit varying degrees of these deviations, with longer prompts generally experiencing greater overprovisioning and underprovisioning amounts. 


Figure~\ref{fig:trade-off-2} shows the response latency decomposed to wating time, preemption time and execution time versus different padding size at different arrival ratefor RLP. At each arrival rate, we observe that as the padding size increases, the preemption time decreases, the waiting time increases, and the execution time remains stable. The total response latency exhibits a concave pattern, initially decreasing with smaller padding sizes and then increasing beyond a certain threshold. 
Also, the arrival rate increase leads to higher waiting time. The padding size that minimizes the response latency varies for different arrival rates. 

\DEL{Using these padding sizes, we measure the impact of request arrival rate on the response latency as shown in Figure~\ref{fig:trade-off-1}. 
We observe that the latency keeps increasing as the arrival rate increases because of the increasing waiting time. 
}

\DEL{\begin{figure}[t]
\centering
\subfloat[Overprovisioned.\label{fig:overprovisioned-175b}]{\includegraphics[width=0.48\linewidth,height=0.13\textheight]{Padding-FIgs/cdf-overprovisioned-all-175.pdf}}\hfill
\subfloat[Underprovisioned. \label{fig:underprovisioned-175b}]{\includegraphics[width=0.48\linewidth,height=0.13\textheight]{Padding-FIgs/cdf-underprovisioned-all-175b.pdf}}
\hfill
\caption{\small CDF of over and underprovisioned requests for OPT-175B.}
\label{fig:cdf-over-under-175b}
\end{figure}
}



\begin{thm}\label{PaddingDetermine}
Previous prediction-based methods often cause underprovisioning (increasing TBT) and overprovisioning (increasing
TTFT). The padding size must be carefully determined, conidering the request arrival rate, to balance TTFT and TBT.
\end{thm}

\begin{figure}[t]
\centering
     \subfloat[OPT-13B.\vspace{-0.01in}\label{fig:preempt-13}]{{\includegraphics[width=0.48\linewidth,height=0.13\textheight]{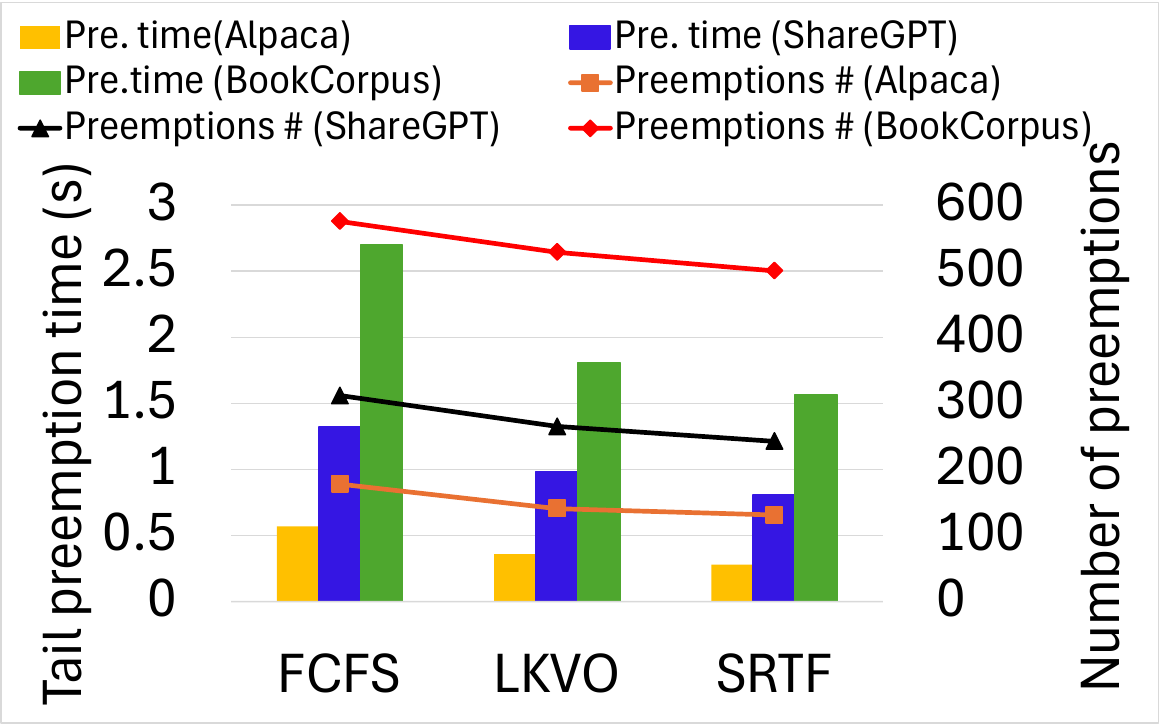} }}
     \subfloat[OPT-175B.\vspace{-0.01in}\label{fig:preempt-175}]{{\includegraphics[width=0.48\linewidth,height=0.13\textheight]{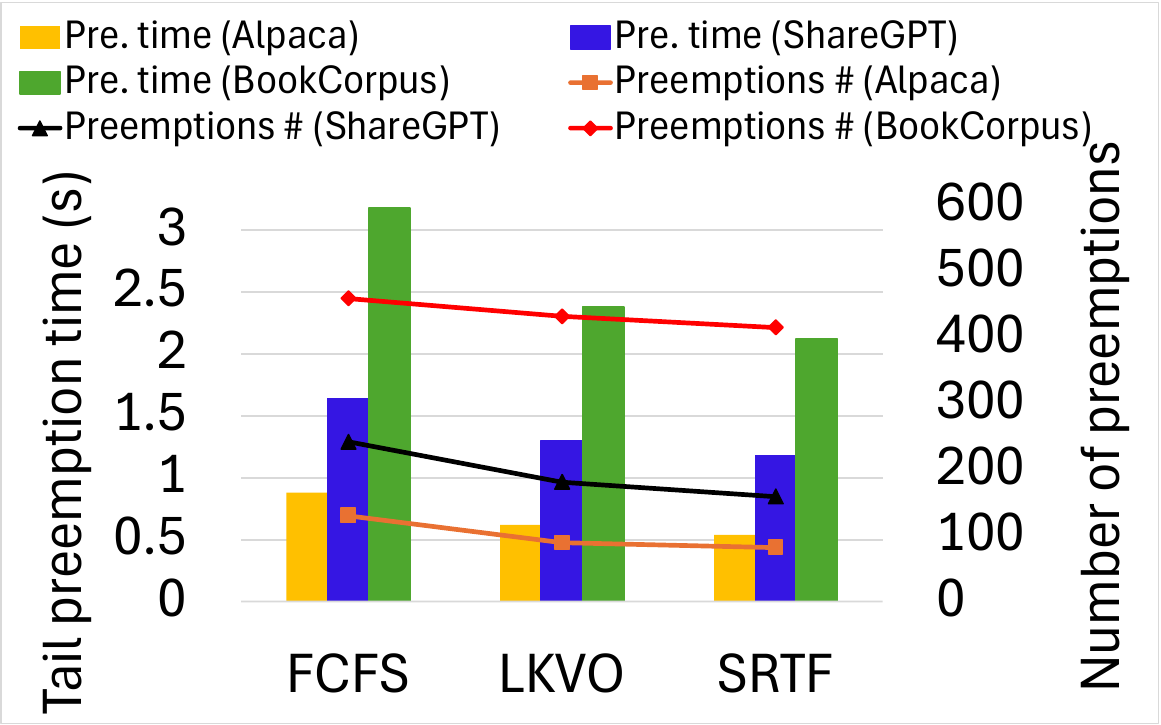} }}
     \DEL{\subfloat[ShareGPT.\vspace{-0.01in}\label{fig:preempt-sharegpt}]{{\includegraphics[width=0.32\linewidth,height=0.13\textheight]{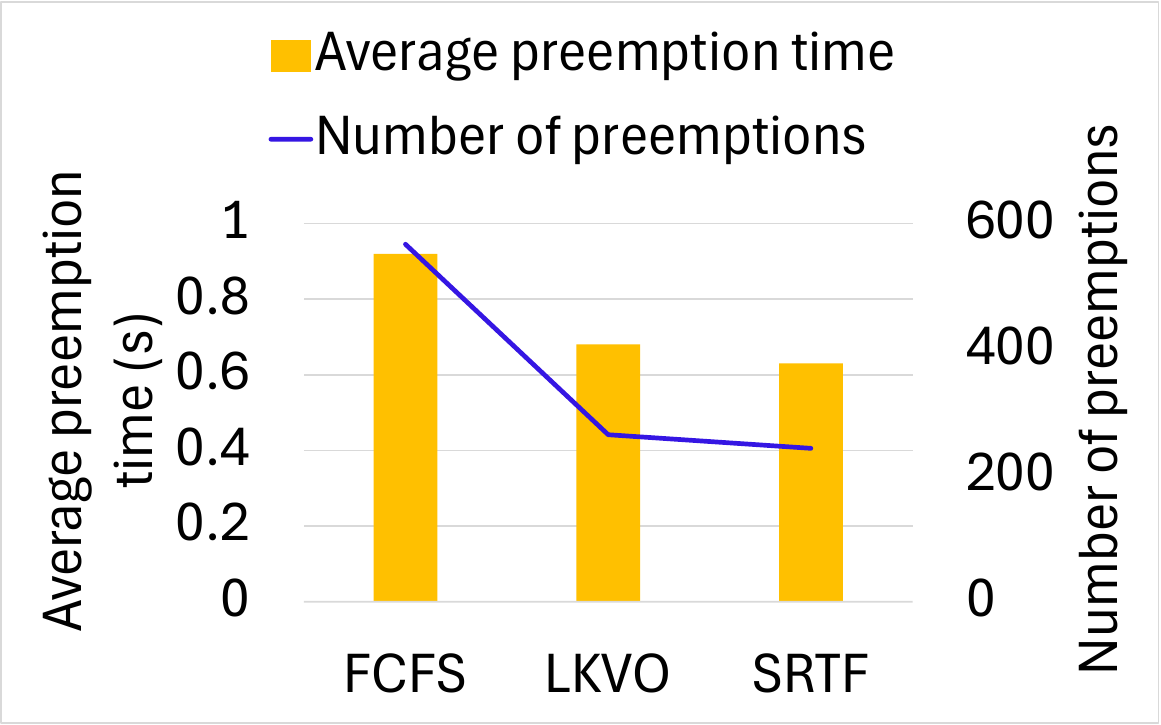} }}
     \subfloat[Bookcorpus.\vspace{-0.01in}\label{fig:preempt-bookcorpus}]{{\includegraphics[width=0.32\linewidth,height=0.13\textheight]{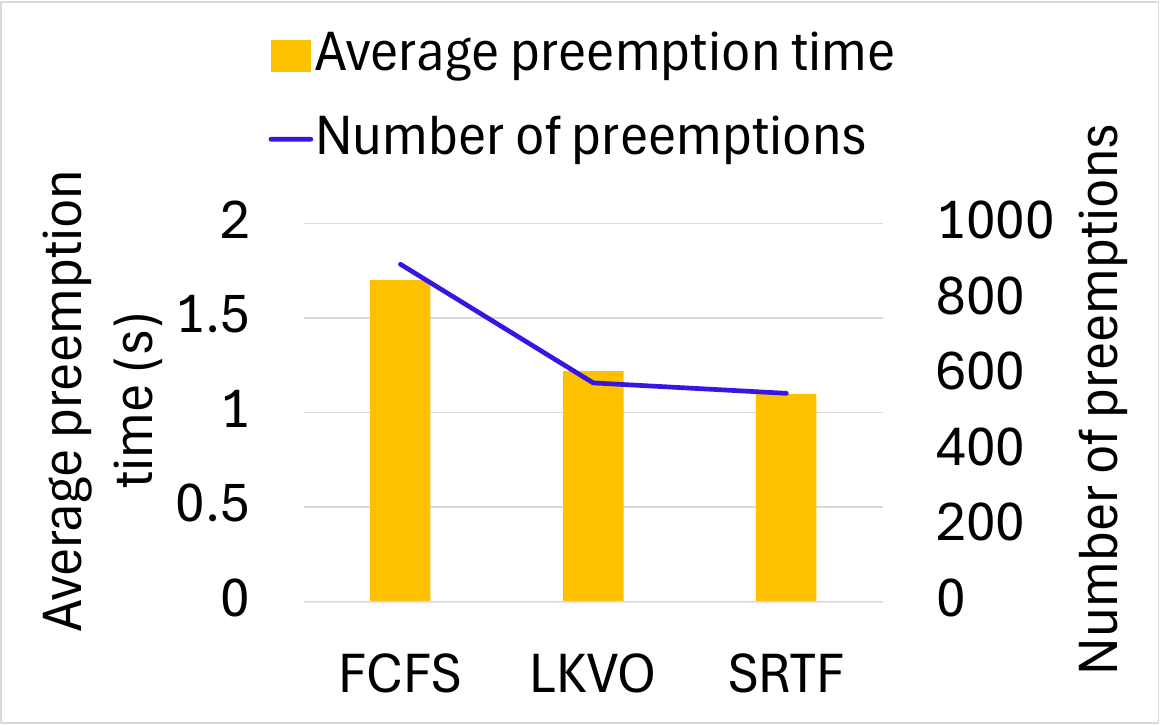} }}
     }
     \DEL{\subfloat[Total preemptions (Y: Num. of preemptions, X: each preemption policy).\vspace{-0.01in}\label{fig:-preemptionstotal}]{{\includegraphics[width=0.48\linewidth,height=0.13\textheight]{Padding-FIgs/total-preemptions.pdf} }}
    \hfill}
    \DEL{\subfloat[Iterations vs preemption time (y: \% of iterations, X: Preemption time, -- not, if no preemption, then preemption time=0).\vspace{-0.01in}\label{fig:iteration-preemption}]{{\includegraphics[width=0.32\linewidth,height=0.13\textheight]{Padding-FIgs/iterations-preemption-time.pdf} }}
    \hfill}
 \caption{\small{Performance of different preemption policies \DEL{\sh{it is impossible that the 2 figs are almost the same-done}\sh{did you see my comment in the exp. section? larger model should have less KVC competition since it uses 2 GPU memories-done}.}\vspace{-0.0in}}}%
    \label{fig:cdf-preemp-time-strategy}
\end{figure}

\DEL{\begin{figure*}[t]
\centering
     \subfloat[Alpaca.\vspace{-0.01in}\label{fig:preempt-alpaca-175}]{{\includegraphics[width=0.32\linewidth,height=0.13\textheight]{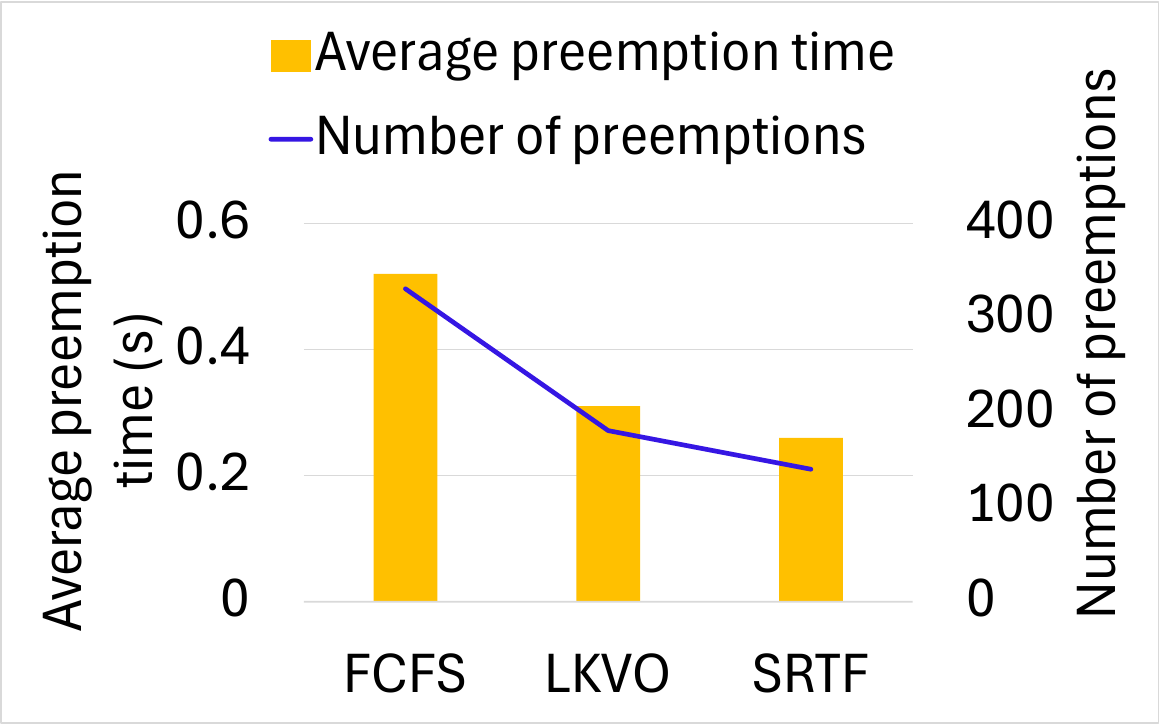} }}
     \subfloat[ShareGPT.\vspace{-0.01in}\label{fig:preempt-sharegpt-175}]{{\includegraphics[width=0.32\linewidth,height=0.13\textheight]{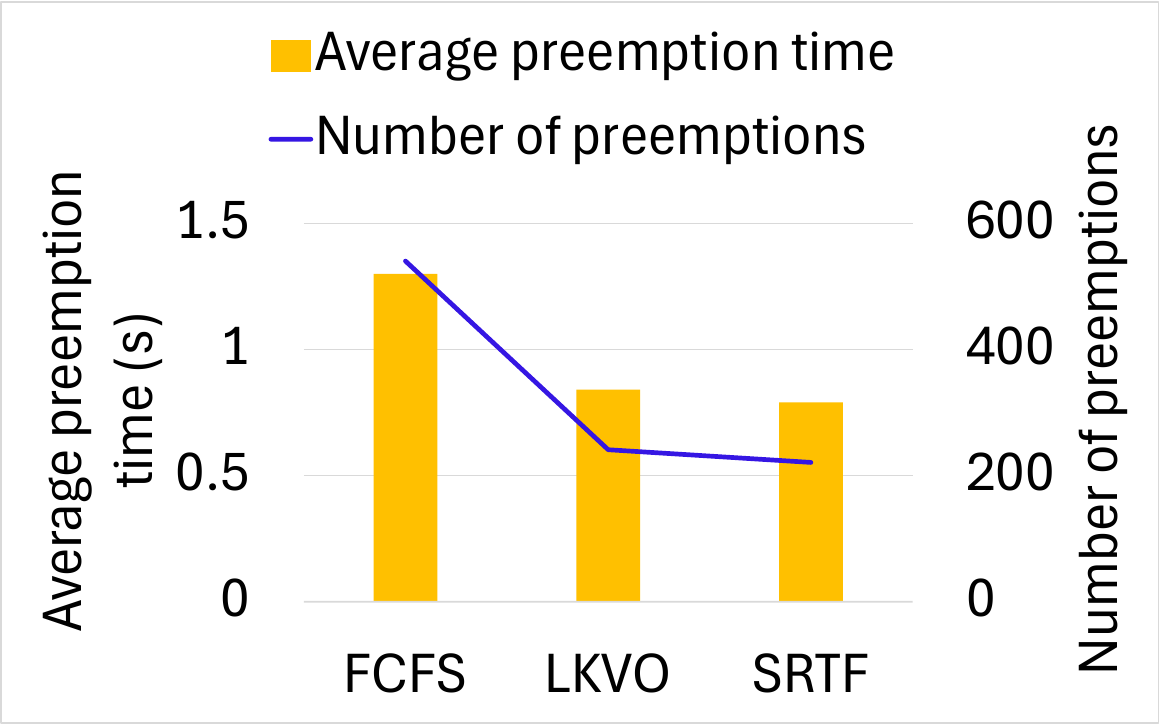} }}
     \subfloat[Bookcorpus.\vspace{-0.01in}\label{fig:preempt-bookcorpus-175}]{{\includegraphics[width=0.32\linewidth,height=0.13\textheight]{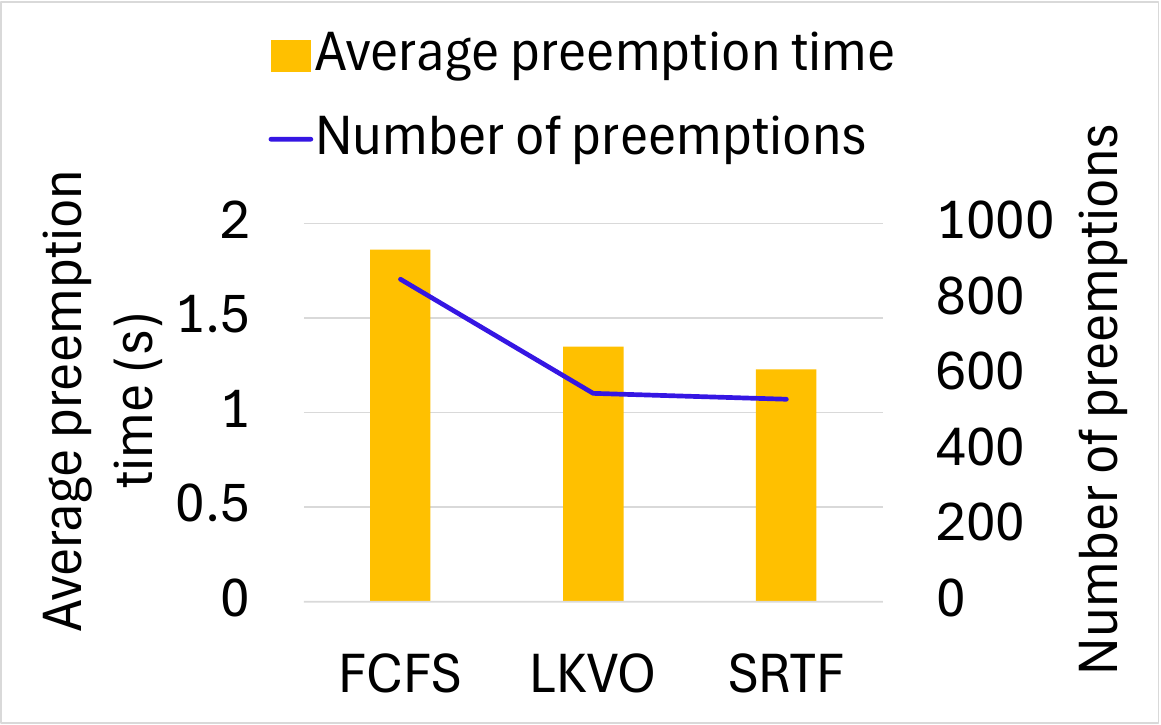} }}
     \DEL{\subfloat[Total preemptions (Y: Num. of preemptions, X: each preemption policy).\vspace{-0.01in}\label{fig:-preemptionstotal}]{{\includegraphics[width=0.48\linewidth,height=0.13\textheight]{Padding-FIgs/total-preemptions.pdf} }}
    \hfill}
    \DEL{\subfloat[Iterations vs preemption time (y: \% of iterations, X: Preemption time, -- not, if no preemption, then preemption time=0).\vspace{-0.01in}\label{fig:iteration-preemption}]{{\includegraphics[width=0.32\linewidth,height=0.13\textheight]{Padding-FIgs/iterations-preemption-time.pdf} }}
    \hfill}
 \caption{\small{Performance of different preemption policy for OPT-175B.\sh{combine the three figs to one fig}\vspace{-0.15in}}}%
    \label{fig:cdf-preemp-time-strategy-175}
\end{figure*}}

\subsection{Preemption Policy} \label{sec:policy}
\DEL{A preemption policy decides which request will be preempted. As we observe that the preemption time takes around 71\% 
of the execution time on average (Figures~\ref{fig:jct-time} and ~\ref{fig:jct-time-175}), we must design an efficient strategy to reduce the preemption time.}
In RLP, we used the following preemption policies that preempt the request: 1) with the lastest arrival time based on FCFS employed in vLLM, 2) with the longest remaining time based on the Shortest-Remaining-Time-First (SRTF), and 3) with the smallest occupied KVC based the Lowest-KVC-occupancy (LKVO). 
Figure~\ref{fig:cdf-preemp-time-strategy} 
shows the tail  preemption time and 
total number of preemptions, respectively.
FCFS exhibits 26\%-58\% and 39\%-1.03$\times$ higher tail preemption time, 9\%-31\% and 13\%-39\% \DEL{\sh{update data after you fix the similar problem-done}} more preemptions compared to LKVO and SRTF, respectively. SRTF helps reduce KVC competition by making running requests release their KVC sooner, thus lowering preemptions. LKVO minimizes the time spent on swapping or recomputation.

\DEL{\tsr{We also observe that OPT-175B has fewer number of preemptions because of less KVC contention for using 8GPUs. However, OPT-175B have higher tail preemption latency because it has additional time to synchronize the KVC across the GPUs in multiple machines.}}



\begin{thm}\label{4Policy}
FCFS leads to more frequent preemptions and increased preemption time compared to SRTF and LKVO. 
\end{thm}

\subsection{Preemption Strategy: Swapping or Recomputation?} 
\begin{figure}[t]
\centering
\DEL{   \subfloat[Preempted data size.\vspace{-0.01in}\label{fig:preempted-data-size}]{{\includegraphics[width=0.48\linewidth,height=0.13\textheight]{Padding-FIgs/preempted-data-size-2.pdf} }}
    \hfill}
\DEL{\subfloat[Available memory bandwidth.\vspace{-0.01in}\label{fig:exp-l3-s}]{{\includegraphics[width=0.24\linewidth,height=0.13\textheight]{Padding-FIgs/memory-bandwidth.pdf} }}
    \hfill
    \subfloat[Available GPU.\vspace{-0.01in}\label{fig:inp-length}]{{\includegraphics[width=0.24\linewidth,height=0.13\textheight]{Padding-FIgs/available-gpu.pdf}
    }}
    \hfill}
        \subfloat[100-100k sequence length.\vspace{-0.01in}\label{fig:seq-length}]{{\includegraphics[width=0.48\linewidth,height=0.13\textheight]{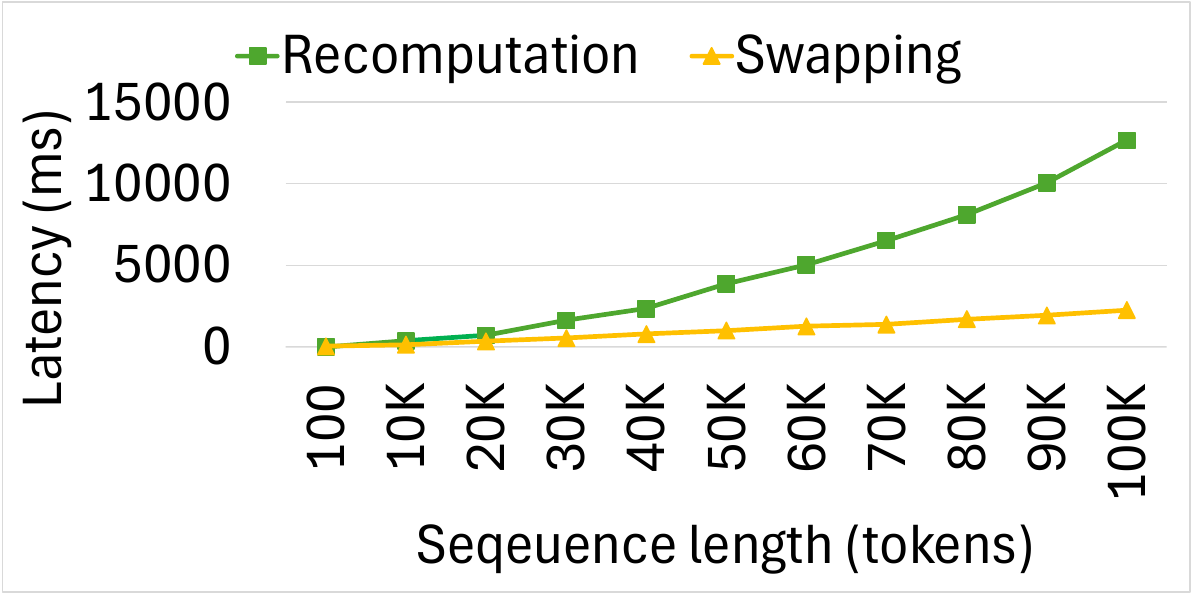} }}
    \hfill
    \subfloat[100-10k sequence length.\vspace{-0.01in}\label{fig:seq-length-2}]{{\includegraphics[width=0.48\linewidth,height=0.13\textheight]{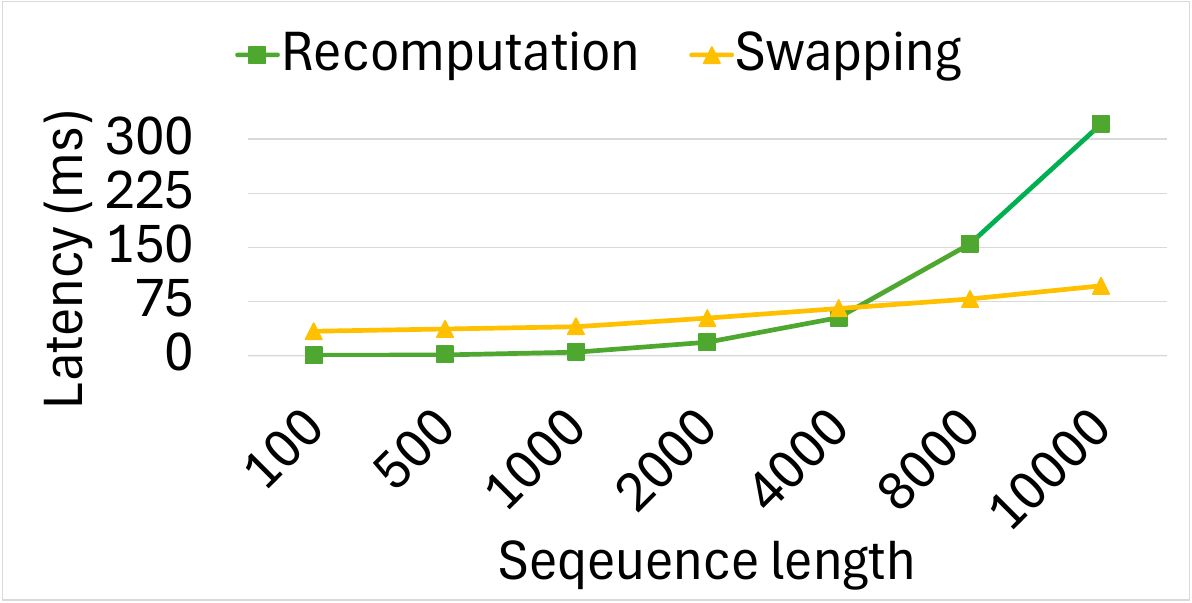} }}
    \hfill
    \hfill
     \vspace{-0.1in}
   \caption{\small{Latency of swapping and recomputation.\vspace{-0.0in}}}%
    \label{fig:policy-time}
\end{figure}

The time complexity of swapping is $O(s)$ and that of recomputation is $O(s^2)$, where $s$ is the sequence length. Figure~\ref{fig:policy-time} shows the latency for swapping including swapping in and out and recomputation for the varying sequence length. It confirms that swapping follows a linear growth, while recomputation follows a quadratic growth based on the sequence length.  \DEL{Figure~\ref{fig:seq-length} shows the recomputation and swapping latency for the sequence length from 100 to 100k with an increment of 10K at each step. For clear visibility, we show the results when the sequence length increases from 100 to 10k in Figure~\ref{fig:seq-length-2}.
}
We observe that swapping is faster when the sequence length exceeds 4000 tokens; otherwise, recomputation is faster. 
\DEL{Swapping is faster for long sequences due to its linear scaling with data size and the quadratic growth of recomputation latency.} However, vLLM and RLP use recomputation and $S^3$ uses swapping as the default preemption strategy irrespective of the sequence length, and users can choose a strategy at a time. 

\begin{thm}\label{5Strategy}

Relying on a single preemption strategy is inefficient. When the sequence length exceeds a threshold (e.g., 4k in our setting), swapping is faster than recomputation.
\end{thm}


\DEL{We first measure the KVC overprovisioning and underprovisioning in different schedulers. Figure~\ref{fig:ov-under-am
}\sh{} and Figure~\ref{fig:ov-under} \sh{draw figs with X=absolute number of tokens-done} plot the CDF of requests versus the number of overprovisioned and underprovisioned tokens, and their ratios over the output lengths. \tsr{For the vLLM, we get the final provisioned values, by calculating the difference between the response length and total allocated tokens.}  From the figure, we observe for vLLM, RLP, and $S^3$, 12\%, 18\%, and 14\% of the requests are overprovisioned by no less than 15\% the response length. Length-wise vLLM never gets overprovisioned by more than 32 tokens because of allocating block size of 32. 
Similarly, for vLLM, RLP, and $S^3$, 48\%, 46\%, and 46\% of the requests are underprovisioned by no less than 12\% of the response length. Underprovisioning in vLLM can occur at any iteration, whereas in RLP and $S^3$ it typically occurs during the final phase of token generation due to an underestimation of the response length\sh{how did you measure vLLM's final values in the figs?-done}.

\sh{didyou recover the above paragraph. If not, pls remove it.}}

\DEL{\subsection{Motivation for Padding Scheme}
\sh{Fig 1 and 2: blue color and black color are too similar, hard to distinguish. Make blue color blue. Double check all other figures.
-done}
\sh{Can you draw a fig to show the number of preemptions of these 3 methods? 3 bars.-done}

\sh{You did not show me Fig1 and 2, right? First, you need to have results for 3 datasets, not 1 dataset. Second, the results are not right. VLLM is completely different from other two methods and won't have similar results.}

We first measure the KVC overprovisioning and underprovisioning in different schedulers. Figure~\ref{fig:ov-under-am
}\sh{} and Figure~\ref{fig:ov-under} \sh{draw figs with X=absolute number of tokens-done} plot the CDF of requests versus the number of overprovisioned and underprovisioned tokens, and their ratios over the output lengths. \tsr{For the vLLM, we get the final provisioned values, by calculating the difference between the response length and total allocated tokens.}  From the figure, we observe for vLLM, RLP, and $S^3$, 12\%, 18\%, and 14\% of the requests are overprovisioned by no less than 15\% the response length. Length-wise vLLM never gets overprovisioned by more than 32 tokens because of allocating block size of 32. 
Similarly, for vLLM, RLP, and $S^3$, 48\%, 46\%, and 46\% of the requests are underprovisioned by no less than 12\% of the response length. Underprovisioning in vLLM can occur at any iteration, whereas in RLP and $S^3$ it typically occurs during the final phase of token generation due to an underestimation of the response length\sh{how did you measure vLLM's final values in the figs?-done}.
\tsr{So, we see these methods vary in their performance only by 1-2\%.}

\DEL{\sh{wrong results-already explained to you in our meeting on 11/21 zoom meeting-done, fixed}
\tsr{From the figure, we also observe that for \emph{SimplePad}, there are 30\% and 8\% requests that are overprovisioned and underprovisioned by 20 tokens or more\sh{X is ratio, how did you get 20?}. This is 18\%, 16\%, and 16\%\sh{list equations for these values} lower than the vLLM, RLP, and $S^3$, respectively, for underprovisioned requests. For the overprovisioned requests, compared to vLLM, RLP and $S^3$, the improvement is 10\%, 8\% and 6\%\sh{list equations for these values}, respectively.}}

\sh{when there is a preemption in vLLM, how did you calcuate "tokens needed to complete", every time is just needs one token KV cache-done, mentioned at the red beginning}
\sh{in vLLM, the overprovisioning cannt be more than 32 tokens-done, fixed, changed to \% of the response length, which is never greater than the 32 tokens}

\sh{also, need to discuss other X values-done}

\sh{do you really need to compare S3 and RLP? what's the purpose for our paper?-done, yes, its because of demand-based kvc allocation}\sh{pls give more details of your answer.-done, red above the section 2.2}

\DEL{\begin{equation}
    \Tilde{p_i} = p_i + (-1)^I \dfrac{c_i}{\lambda}
\end{equation}
Here, $I$ is an indicator variable that returns $1$ if the monitor observes overprovisioning in the majority of cases in requests with similar input length; for underprovisioning, the indicator variable returns $2$. This simple padding scheme tries to reduce the padding when the arrival rate is high because it can help pack more requests in the running queue to reduce the waiting time.}

\DEL{We add the performance of this method to Figures~\ref{fig:ov-under} to~\ref{fig:jct-time}. From the figure, we observe that for the \emph{SimplePad}, there are 30\% and 8\% requests that are overprovisioned and underprovisioned by 20 tokens or more. This is 18\%, 16\%, and 16\% lower than the vLLM, RLP, and $S^3$, respectively, for underprovisioned requests. }\DEL{For the overprovisioned requests, compared to vLLM, RLP and $S^3$, the improvement is 10\%, 8\% and 6\%, respectively. In terms of job completion time, \emph{SimplePad} has 2.5$\times$, 1.9$\times$, 1.85$\times$ lower than vLLM, RLP, and $S^3$, respectively, while the job completion time is \% higher than the \emph{Oracle}.}

\DEL{Next, we measure the impact of increasing arrival rate on the waiting time of these methods and plot it in Figure~\ref{fig:waiting-time}. From the Figure??, we observe that with the increasing arrival rate, the waiting time\sh{what is it? total or average? make sure in whole paper when you explain one metric, explain it clerly whether average, total or others} for the requests keeps increasing. The waiting time also increases for the \emph{Oracle}, but the change is smaller than the other methods.
For the increasing arrival rate, \emph{SimplePad} performs within 12\% of the \emph{Oracle}.}


\DEL{\begin{figure*}[t]
\centering
     \subfloat[Alpaca.\vspace{-0.01in}\label{fig:exp-13-s-c}]{{\includegraphics[width=0.32\linewidth,height=0.13\textheight]{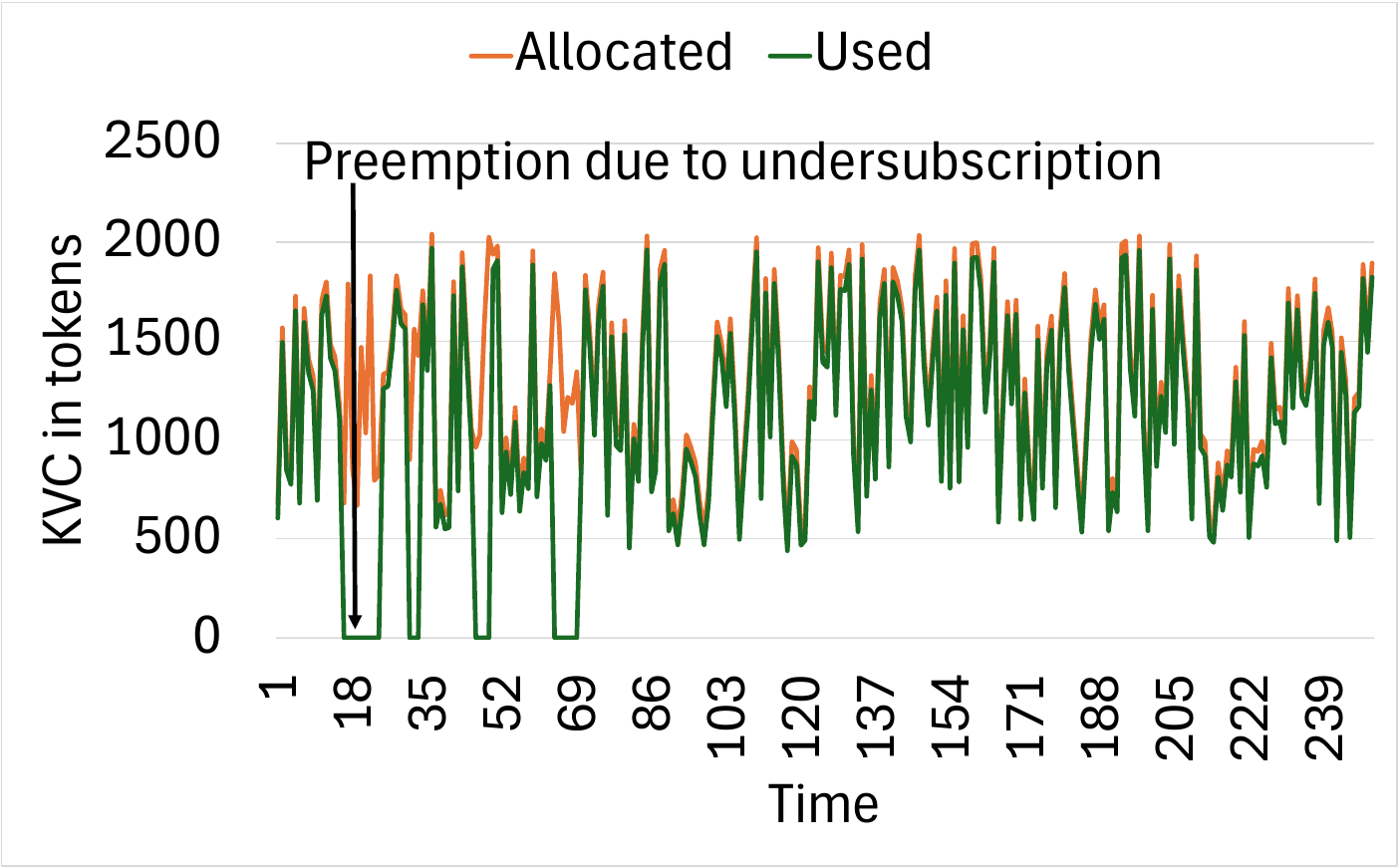} }}
    \hfill
    \subfloat[ShareGPT.\vspace{-0.01in}\label{fig:exp-175-s-c}]{{\includegraphics[width=0.32\linewidth,height=0.13\textheight]{Padding-FIgs/dynamic.pdf} }}
    \hfill
    \subfloat[BookCorpus.\vspace{-0.01in}\label{fig:exp-l3-s}]{{\includegraphics[width=0.32\linewidth,height=0.13\textheight]{Padding-FIgs/dynamic.pdf} }}
    \hfill
   \caption{\small{Utilization of different traces for dynamic padding.\vspace{-0.15in}}}%
    \label{fig:dynamic-exo}
\end{figure*}}

\begin{figure*}[t]
\centering
     \subfloat[Alpaca.\vspace{-0.01in}\label{fig:exp-13-s-c}]{{\includegraphics[width=0.32\linewidth,height=0.13\textheight]{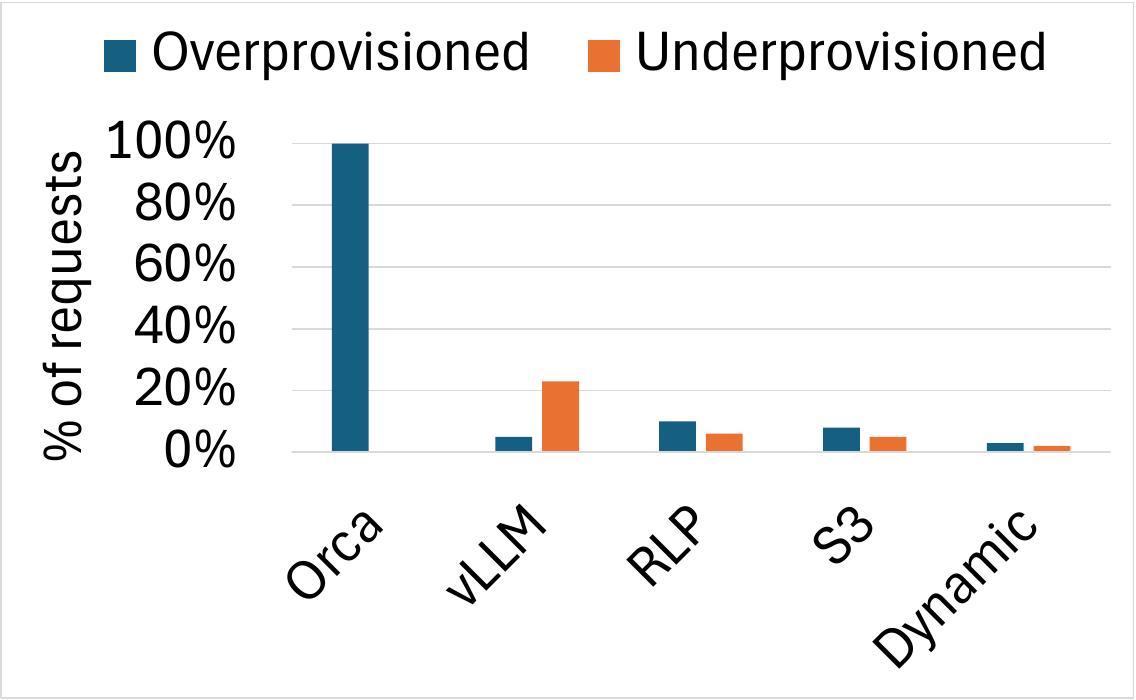} }}
    \hfill
    \subfloat[ShareGPT.\vspace{-0.01in}\label{fig:exp-175-s-c}]{{\includegraphics[width=0.32\linewidth,height=0.13\textheight]{Padding-FIgs/ov-underprovision.pdf} }}
    \hfill
    \subfloat[BookCorpus.\vspace{-0.01in}\label{fig:exp-l3-s}]{{\includegraphics[width=0.32\linewidth,height=0.13\textheight]{Padding-FIgs/ov-underprovision.pdf} }}
    \hfill
   \caption{\small{Over and Underprovisioned request percentage for different traces.\vspace{-0.15in}}}%
    \label{fig:ov-under}
\end{figure*}

\begin{figure}[t]
\centering
     \subfloat[Overprovisioned.\vspace{-0.01in}\label{fig:exp-13-s-c}]{{\includegraphics[width=0.48\linewidth,height=0.13\textheight]{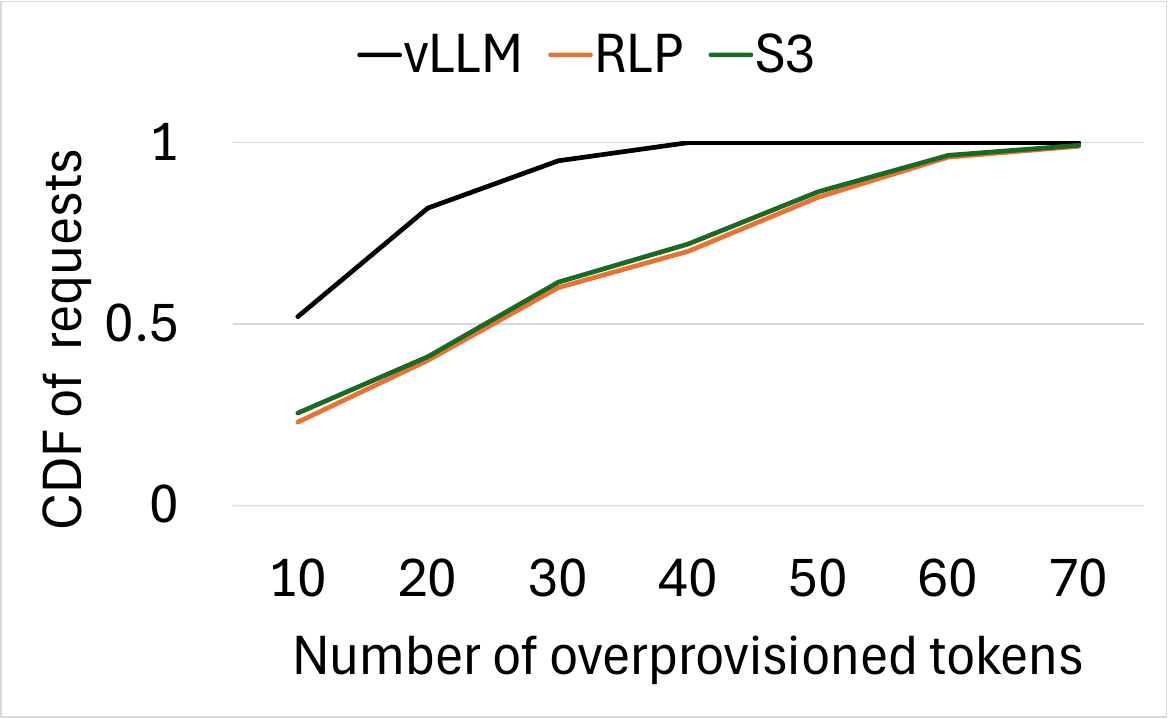} }}
    \hfill
    \hfill
    \subfloat[Underprovisioned\vspace{-0.01in}\label{fig:swapped}]{{\includegraphics[width=0.48\linewidth,height=0.13\textheight]{Padding-FIgs/underprovisioned-tokens.pdf} }}
    \hfill
 \caption{\small{CDF of over and underprovisioned requests for Alpaca. \vspace{-0.15in}}}%
    \label{fig:ov-under-amount}
\end{figure}

\DEL{\begin{figure}[t]
\centering
     \subfloat[Overprovisioned.\vspace{-0.01in}\label{fig:exp-13-s-c}]{{\includegraphics[width=0.48\linewidth,height=0.13\textheight]{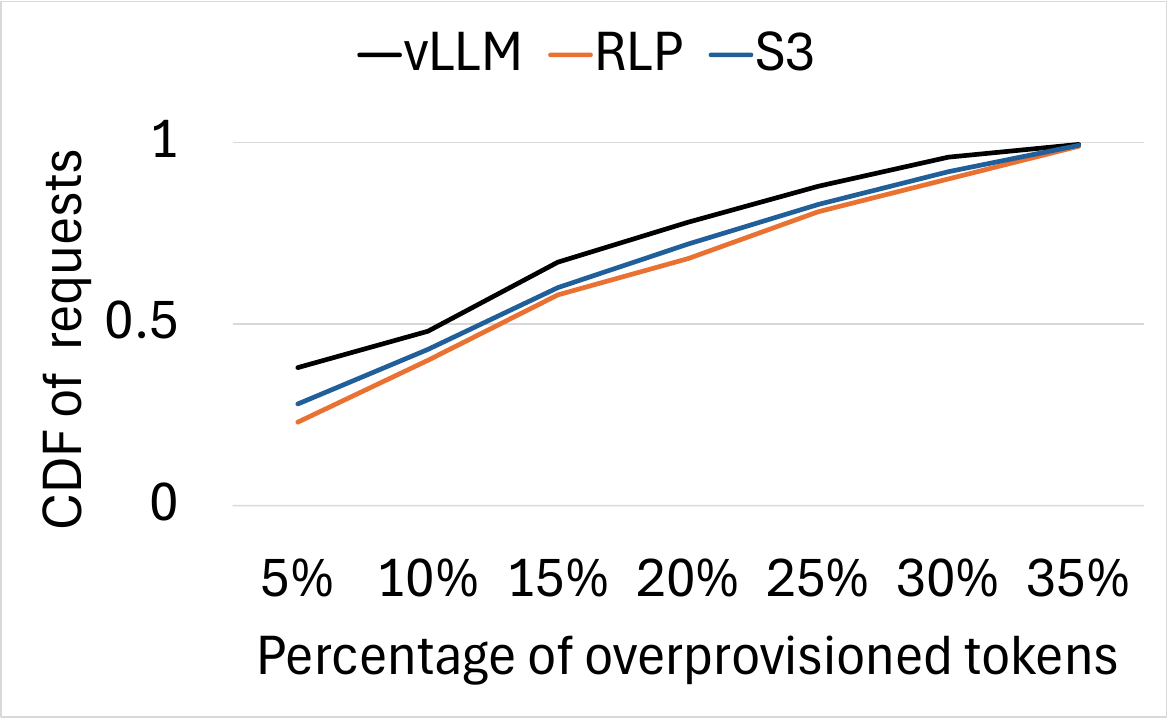} }}
    \hfill
    \hfill
    \subfloat[Underprovisioned\vspace{-0.01in}\label{fig:swapped}]{{\includegraphics[width=0.48\linewidth,height=0.13\textheight]{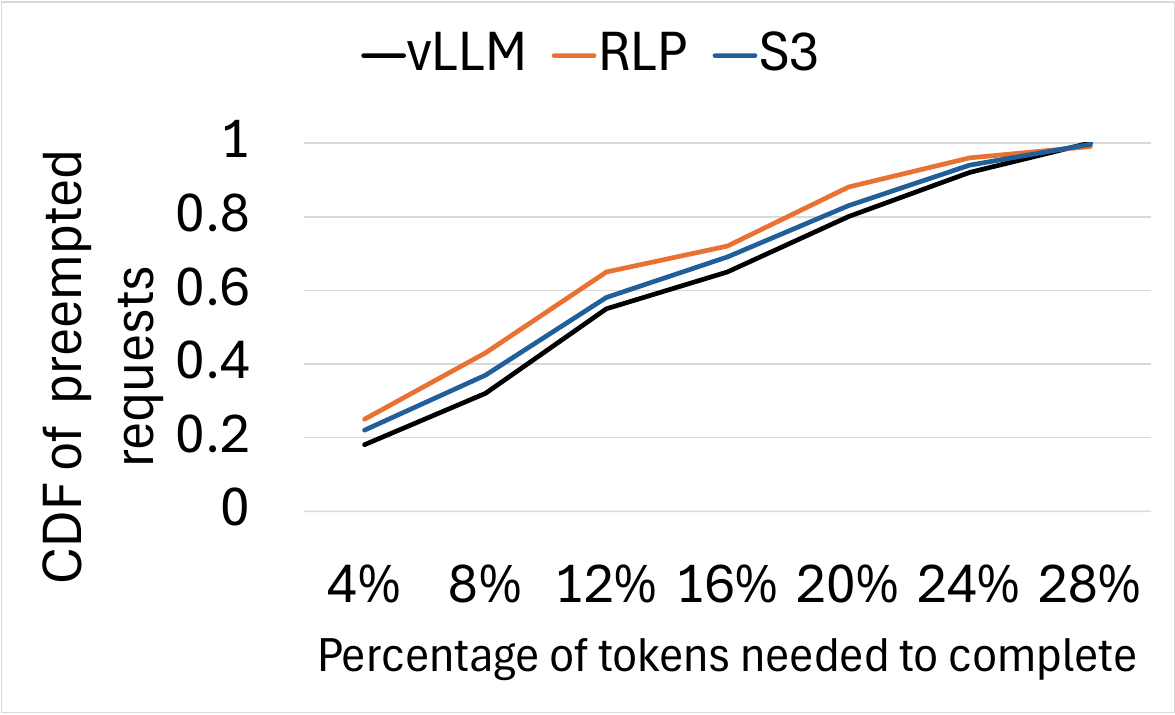} }}
    \hfill
 \caption{\small{CDF of over and underprovisioned requests for Alpaca. \sh{which dataset? make sure you indicate dataset for each fig-done}.\vspace{-0.15in}}}%
    \label{fig:ov-under}
\end{figure}}

\begin{figure}[t]
\centering
     \subfloat[Alpaca.\vspace{-0.01in}\label{fig:ov-alpaca-count}]{{\includegraphics[width=0.32\linewidth,height=0.13\textheight]{Padding-FIgs/over-alpaca-count.pdf} }}
    \hfill
    \subfloat[ShareGPT.\vspace{-0.01in}\label{fig:ov-sharegpt-count}]{{\includegraphics[width=0.32\linewidth,height=0.13\textheight]{Padding-FIgs/over-sharegpt-count.pdf} }}
    \hfill
    \subfloat[BookCorpus.\vspace{-0.01in}\label{fig:ov-bookcorpus-amount}]{{\includegraphics[width=0.32\linewidth,height=0.13\textheight]{Padding-FIgs/over-bookcorpus-count.pdf} }}
    \hfill
    \DEL{\subfloat[Underprovisioned\vspace{-0.01in}\label{fig:swapped}]{{\includegraphics[width=0.48\linewidth,height=0.13\textheight]{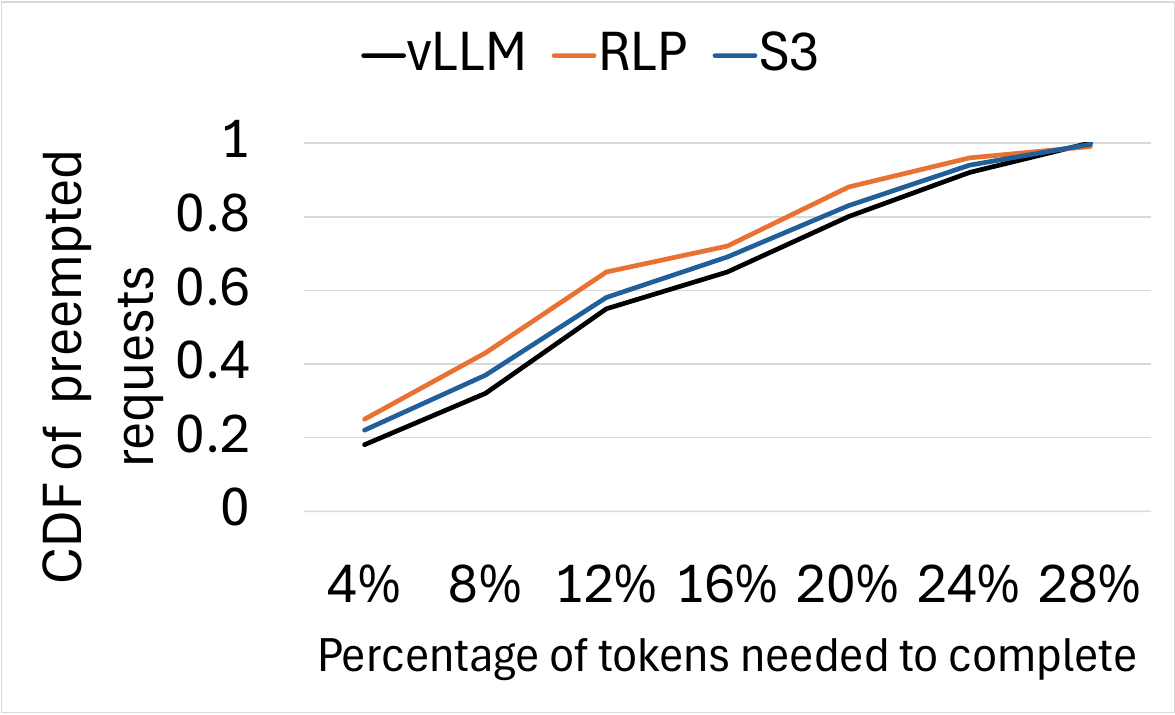} }}
    \hfill}
 \caption{\small{CDF of overprovisioned  requests. \sh{which dataset? make sure you indicate dataset for each fig-done}.\vspace{-0.15in}}}%
    \label{fig:ov-under-ov-amount}
\end{figure}

\begin{figure}[t]
\centering
     \subfloat[Alpaca.\vspace{-0.01in}\label{fig:un-alpaca-count}]{{\includegraphics[width=0.32\linewidth,height=0.13\textheight]{Padding-FIgs/under-alpaca-count.pdf} }}
    \hfill
    \subfloat[ShareGPT.\vspace{-0.01in}\label{fig:un-sharegpt-count}]{{\includegraphics[width=0.32\linewidth,height=0.13\textheight]{Padding-FIgs/under-sharegpt-count.pdf} }}
    \hfill
    \subfloat[BookCorpus.\vspace{-0.01in}\label{fig:un-bookcorpus-amount}]{{\includegraphics[width=0.32\linewidth,height=0.13\textheight]{Padding-FIgs/under-bookcorpus-amount.pdf} }}
    \hfill
    \DEL{\subfloat[Underprovisioned\vspace{-0.01in}\label{fig:swapped}]{{\includegraphics[width=0.48\linewidth,height=0.13\textheight]{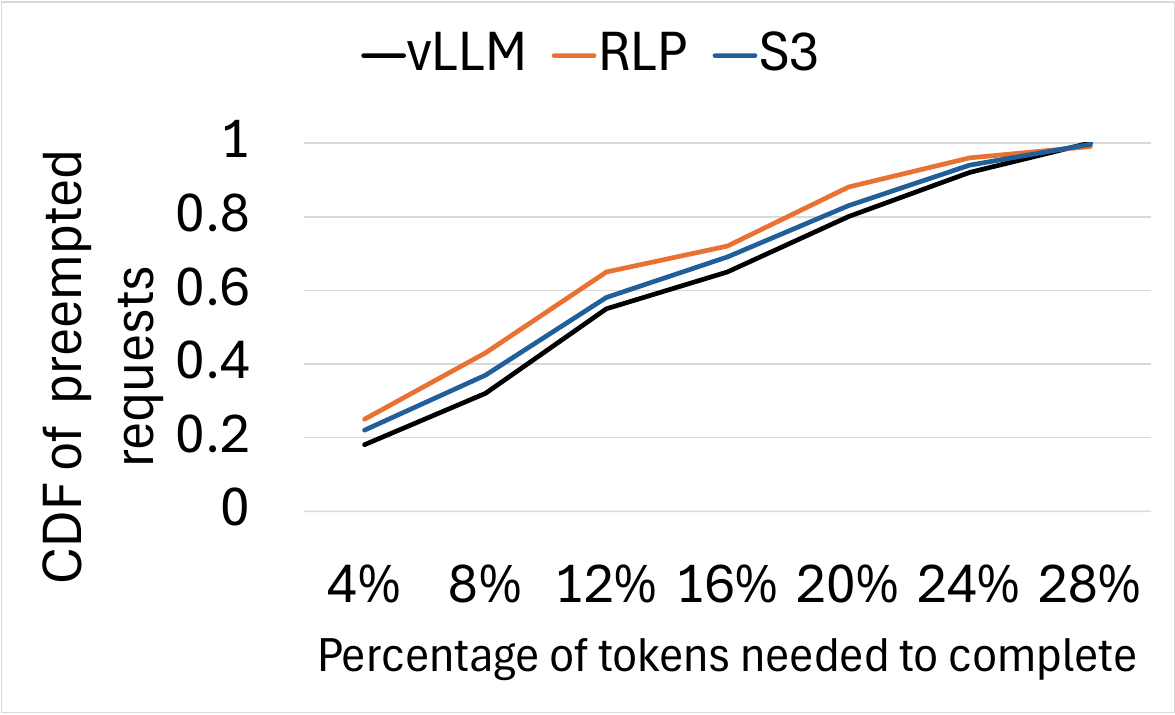} }}
    \hfill}
 \caption{\small{CDF of underprovisioned request  requests. \sh{which dataset? make sure you indicate dataset for each fig-done}.\vspace{-0.15in}}}%
    \label{fig:ov-under-un-amount}
\end{figure}

\begin{figure}[t]
\centering
     \subfloat[Alpaca.\vspace{-0.01in}\label{fig:ov-alpaca}]{{\includegraphics[width=0.32\linewidth,height=0.13\textheight]{Padding-FIgs/over-alpaca.pdf} }}
    \hfill
    \subfloat[ShareGPT.\vspace{-0.01in}\label{fig:ov-sharegpt}]{{\includegraphics[width=0.32\linewidth,height=0.13\textheight]{Padding-FIgs/over-sharegpt.pdf} }}
    \hfill
    \subfloat[BookCorpus.\vspace{-0.01in}\label{fig:ov-bookcorpus}]{{\includegraphics[width=0.32\linewidth,height=0.13\textheight]{Padding-FIgs/over-bookcorpus.pdf} }}
    \hfill
    \DEL{\subfloat[Underprovisioned\vspace{-0.01in}\label{fig:swapped}]{{\includegraphics[width=0.48\linewidth,height=0.13\textheight]{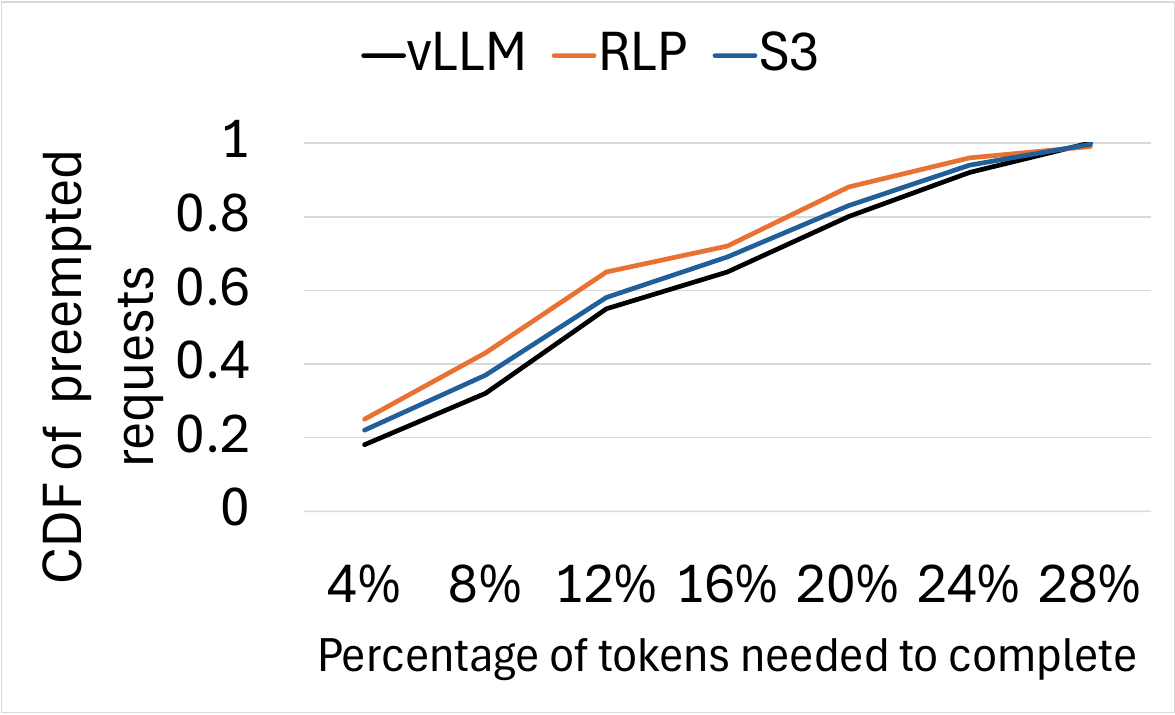} }}
    \hfill}
 \caption{\small{CDF of overprovisioned  requests. \sh{which dataset? make sure you indicate dataset for each fig-done}.\vspace{-0.15in}}}%
    \label{fig:ov-under-ov}
\end{figure}

\begin{figure}[t]
\centering
     \subfloat[Alpaca.\vspace{-0.01in}\label{fig:un-alpaca}]{{\includegraphics[width=0.32\linewidth,height=0.13\textheight]{Padding-FIgs/under-alpaca.pdf} }}
    \hfill
    \subfloat[ShareGPT.\vspace{-0.01in}\label{fig:un-sharegpt}]{{\includegraphics[width=0.32\linewidth,height=0.13\textheight]{Padding-FIgs/under-sharegpt.pdf} }}
    \hfill
    \subfloat[BookCorpus.\vspace{-0.01in}\label{fig:un-bookcorpus}]{{\includegraphics[width=0.32\linewidth,height=0.13\textheight]{Padding-FIgs/under-bookcorpus.pdf} }}
    \hfill
 \caption{\small{CDF of underprovisioned  requests. \sh{which dataset? make sure you indicate dataset for each fig-done}.\vspace{-0.15in}}}%
    \label{fig:ov-under-under}
\end{figure}
}





\DEL{\begin{thm}\label{padding}

\end{thm}}

\DEL{\begin{thm}
Workloads with low arrival rates incur high KV-cache waste for the static-padding-based approaches compared to the high arrival scenario. A padding can increase the KV-cache utilization by \% for all workloads.
\end{thm}}

\DEL{\begin{thm}
The probability values of the confidence score for the requests with similar response lengths follow a similar distribution. As a result, the confidence score can be considered while adjusting the amount of padding.
\end{thm}}

\DEL{\begin{figure}
    \centering
    \includegraphics[width=0.5\linewidth,height=0.13\textheight]{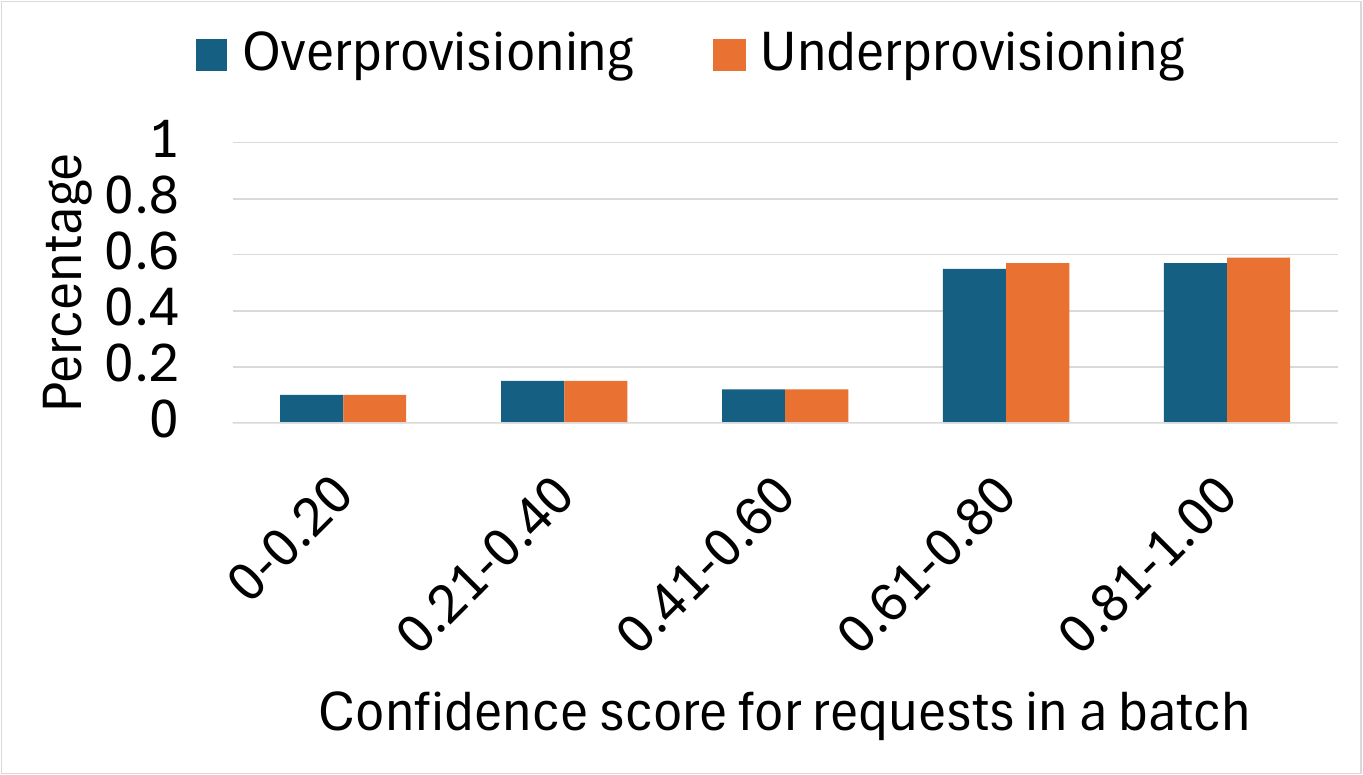}
    \caption{(Fake) Existence of overprovisioned/underprovisioned requests in the same batch.}
    \label{fig:confidence}
\end{figure}

\begin{thm}
Requests within the same batch with different confidence scores suffer from underprovision and overprovision. (Figure~\ref{fig:confidence})
\end{thm}}

\DEL{\begin{thm}
A small amount of fixed-size buffer can reduce the latency for underprovisioned requests that have a shorter remaining time.
\end{thm}}

\DEL{\begin{figure*}[t]
\centering
    \subfloat[Request types.\vspace{-0.01in}\label{fig:req-type}]{{\includegraphics[width=0.32\linewidth,height=0.13\textheight]{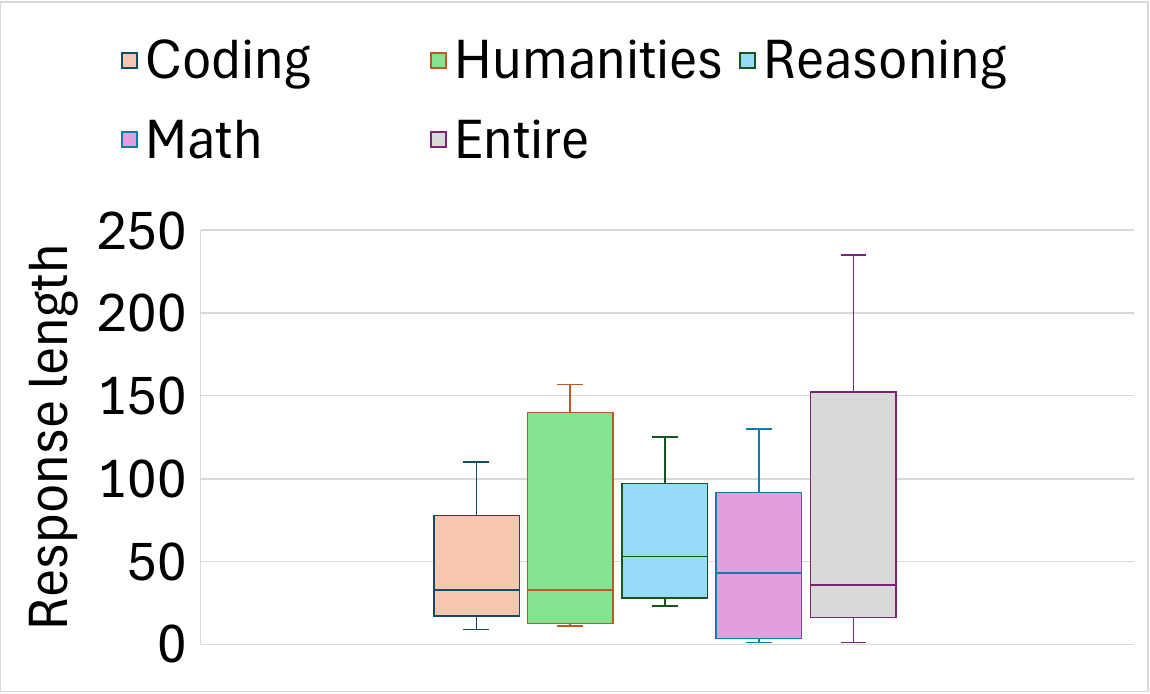} }}
    \hfill
     \subfloat[ Coding.\vspace{-0.01in}\label{fig:req-coding}]{{\includegraphics[width=0.32\linewidth,height=0.13\textheight]{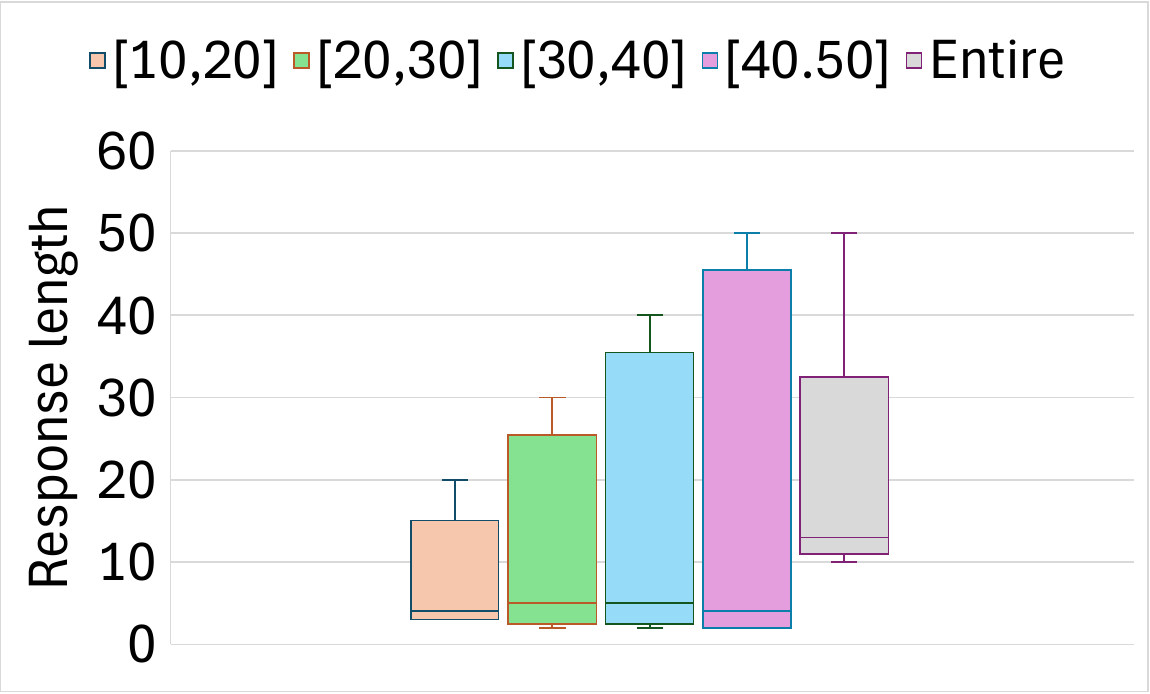} }}
    \hfill
    \subfloat[ Humanities.\vspace{-0.01in}\label{fig:req-hum}]{{\includegraphics[width=0.32\linewidth,height=0.13\textheight]{Padding-FIgs/category-distri.pdf} }}
    \hfill
    \subfloat[ Reasoning.\vspace{-0.01in}\label{fig:req-reason}]{{\includegraphics[width=0.32\linewidth,height=0.13\textheight]{Padding-FIgs/category-distri.pdf} }}
    \hfill  \subfloat[Math.\vspace{-0.01in}\label{fig:exp-l3-s}]{{\includegraphics[width=0.32\linewidth,height=0.13\textheight]{Padding-FIgs/category-distri.pdf} }}
    \subfloat[ Entire.\vspace{-0.01in}\label{fig:req-entire}]{{\includegraphics[width=0.32\linewidth,height=0.13\textheight]{Padding-FIgs/category-distri.pdf} }}
    \hfill
   \caption{\small{Distribution of length according to type and input length.\vspace{-0.15in}}}%
    \label{fig:distribution}
\end{figure*}}

\DEL{This mixed approach performs 12\%, 13\%, and 14\%, 22\% better than the FCFS, SRTF, LKVO, and Random, respectively.}

\DEL{\begin{thm}\label{ob3}
While doing the preemption, eviction policies considering multiple factors, such as\sh{do not use "such as", need to include all factors} remaining processing time and KV-cache occupancy, can make more informed decisions, leading to improved overall performance compared to a random policy. (Figure~\ref{fig:cdf-preemp-time-strategy})
\end{thm}}

\DEL{\subsection{Preemption Strategy: Swapping or Recomputation?} There are two primary strategies to perform the preemption in the KV-cache: 1) Swapping, and 2) Recomputation. 
The time complexility of swapping is $O(n)$ and that of recomputation is $O(n^2)$, where $n$ is the sequence length. Figure~\ref{fig:seq-length} shows the latency for swapping and recomputation for the varying sequence length. It shows that swapping follows a linear growth, while recomputation follows a quadratic growth based on the sequence length. We further observe that swapping performs better when the sequence length exceeds 1500, while recomputation performs better for the lower sequence length than 1500. 
At lower sequence lengths, the number of operations required to recompute the intermediate results (e.g., attention values) is small\sh{Masahiro, is this sentence correct?} so the time required for the computations is significantly less than the time needed for data transfers in swapping. Swapping is more efficient for long sequences due to its linear scaling with data size and the quadratic growth of recomputation costs for larger sequence lengths.

Figure~\ref{fig:policy-time}\sh{in the figues, change Y name to exactly the latency name. what latency?-done} shows the \tsr{average latency for the preempted requests for performing swapping and recomputaion to return to running state.} \sh{isn't it just premption time?-done, latency to return to running state} for the FCFS eviction policy. 
\sh{you need to describe how you did the experiment, how did you get the different values of X, how did you get the Y value.-done} \sh{descriptoin below is not clear. go over with me orally.} \tsr{For this experiment, while varying the preempted data size, we keep track of the preempted data size for the A100 GPU. \DEL{On the other hand, while measuring for the memory (PCIe) bandwidth, we queue data transfers in a stream and use \emph{cudaStreamSynchronize  to add pauses or control the pace at which transfers are complete.} Finally, when we measure the available GPU (\%), we keep track of the available GPU  over time while performing the eviction and cluster (5 clusters) them to plot the average latency for each cluster. The decision to choose between swapping and recomputation based on preempted data size, available GPU (\%), and available memory (PCIe) bandwidth is rooted in minimizing resource costs and latency.}
\DEL{Preempted data size matters because larger data incurs higher I/O costs when swapped, making recomputation potentially more efficient for big data sizes, while smaller data may be swapped quickly.} \DEL{Available GPU memory dictates whether storing data temporarily is feasible without risking memory overload, thus favoring swapping if memory is sufficient and recomputation if it’s constrained. Finally, available memory bandwidth affects data transfer speeds, making swapping preferable when bandwidth is high, but pushing towards recomputation in low-bandwidth scenarios to avoid delays.}}

For different preempted data size\sh{at the beginning, you need to explain how recomputation is executed, use what to compute what. Then, here, you can explain why the data size does not affect the recomputation latency--I feel it affects-done}, we observe that the recomputation latency increases while varying the data size, the recomputation for the prompt processing keep increasing \sh{this reason cannot get a conclusion before it-done}. The swapping time also increases with the increasing data size. However, the increase is comparatively lower than recomputation because of the high memory (PCIe) bandwidth of A100 GPU between CPU and GPU. \tsr{The impact of high bandwidth is visible for larger size because of high data transfer speed. On the other hand, the latency is so small that for smaller preempted data size, the difference becomes negligible.} \sh{why this reason leads to the above result? -done . you need to avoid being farfetching or handwaving-done} . However, the recomputation is more efficient for smaller preempted data size, because it can be recomputed quickly for the available GPU. Thus, recomputation is more efficient when the data
size is small while swapping is more efficient when the data size is large \sh{transfer more data does not increase latency?-done, increases}. \sh{this should be "available memory bandwidth"-done}.}

\DEL{\tsr{The available memory (PCIe) bandwidth has no impact on the latency for return to the running state in case of the recomputation because it is mostly dependent on the available GPU. However, the latency keeps decreasing as the available memory (PCIe) bandwidth keeps increasing due to the higher transfer capability. In contrast, the available GPU (\%) has no impact on the latency due to the swapping, however, the recomputation time keeps decreasing for the available GPU (\%).}}

\DEL{In addition to the preempted data size and bandwidth as mentioned in~\cite{vllm}, we also consider several other factors that may influence the recomputation and the swapping. The factors include the GPU and memory usage,  input/output length of the requests, and number of requests in the running and waiting queue. We plot these factors from Figure~\ref{fig:preempted-data-size} to \ref{fig:out-length-2}. From the figure, we observe these factors show varying impact on the recomputation and the swapping time. Mostly, as the values of these factors increase, the latency of swapping drops, while the latency of recomputation either remains constant or increases\sh{explain why for each factor. do not artificially create reasons. need to think deeply and really use exp. to verify. Be truthful.}. Only the waiting requests as in Figure~\ref{fig:out-length-2} does not have any impact.\sh{then no need to show it} } 


\DEL{\begin{thm}\label{4Policy}
Using FCFS to select a request to preempt may lead to more preemptions due to limited released KVC.
\end{thm}}

\DEL{\begin{thm}\label{ob4}
A comprehensive approach to determining the preemption policy (swapping vs. recomputation) that considers various influencing factors can lead to lower latency. The factors include the size of the preempted data, available memory (PCIe) bandwidth and available GPU (\%). (Figure~\ref{fig:policy-time})
\end{thm}}


\section{System Design}
\subsection{Solution Overview}

\begin{figure}
    \centering
\includegraphics[width=0.65\columnwidth]{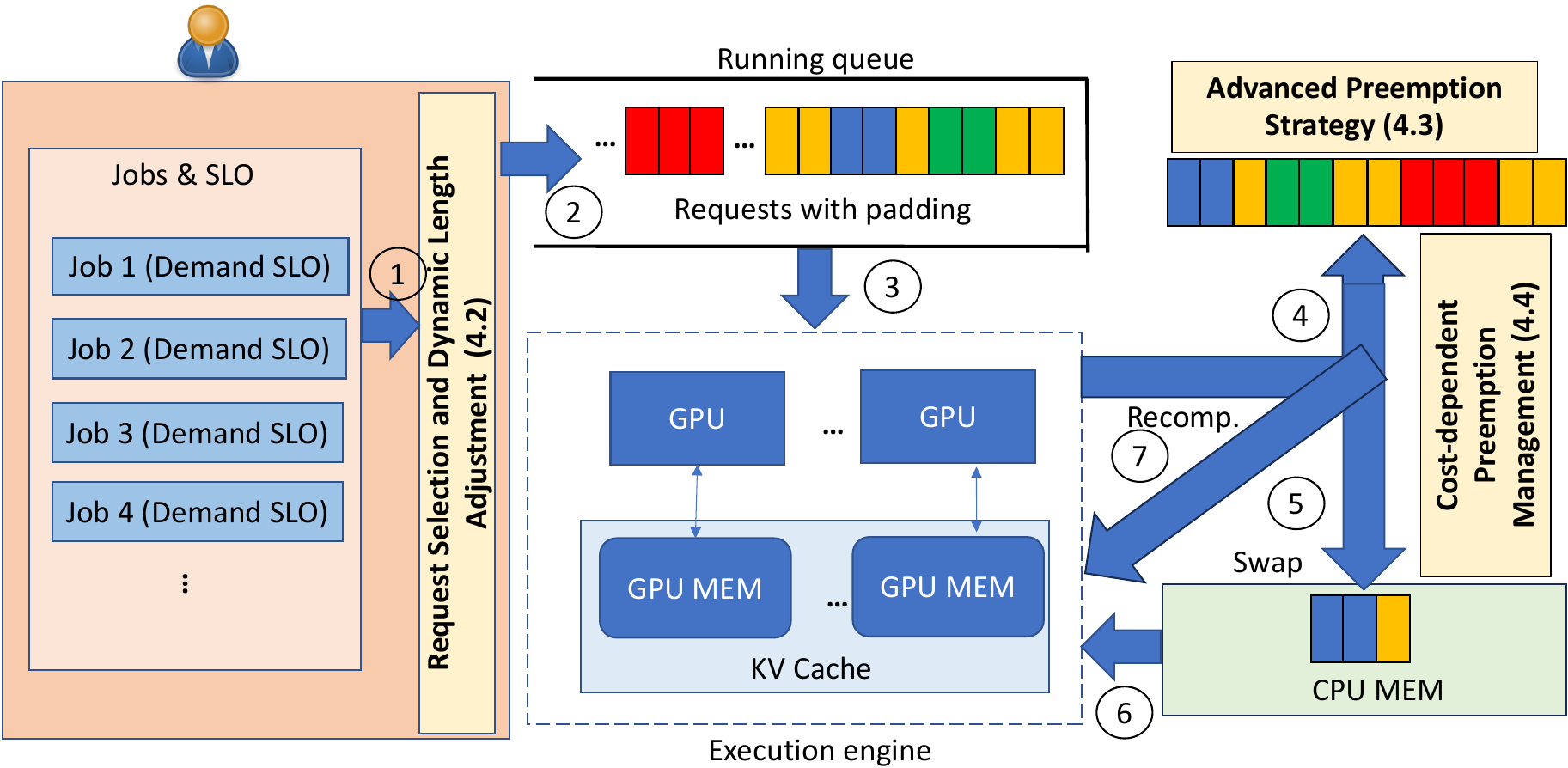}
    \caption{Architecture of \sys.}
    \label{fig:component1}\vspace{-0.0in}
\end{figure}

Based on our observations, we propose \sys. \sys consists of the following components as shown in Figure~\ref{fig:component1}.
\squishlist
\item[(1)] \emph{Confidence-based padding} guided by O\ref{PaddingDetermine} (\ref{sec:padding}). 
\item[(2)] \emph{SLO-aware batching and KVC allocation} per O\ref{KVCLimit} (\ref{sec:allocation}).\looseness=-1
\item[(3)] \emph{Preemption policy} guided by O\ref{4Policy} (\ref{sec:preemption}).
\item[(4)] \emph{Preemption strategy selection} guided by O\ref{5Strategy} (\ref{sec:strategy}).
\squishend

\DEL{

\squishlist
\item[(1)] \textbf{Confidence-based padding (\ref{sec:padding})}. Guided by Observation~\ref{3PaddingValue}, 
\sys modifies and fine-tunes an LLM model to predict output length along with the confidence score and the prediction deviation direction (overprediction or underprediction). It then dynamically determines the padding based on the prediction confidence using the Hoeffding's inequality theory~\cite{hoeffding1963,bentkus2004hoeffding} to bound the probability that the estimated value deviates from its real value.

\item[(2)] \textbf{SLO-aware batching and KVC allocation (\ref{sec:allocation}).} Motivated by Observations~\ref{1Obs:KVCLimit} and~\ref{PaddingDetermine}, \sys strategically selects waiting requests and allocates KVC to meet their TTFT and TBT SLOs. It reuses unutilized KVC, proactively allocates KVC before requests exhaust their allocated KVC, and reserves KVC for global sharing when requests require additional KVC.

\item[(3)] \textbf{Preemption policy (\ref{sec:preemption}).} Guided by Observation~\ref{4Policy}, \sys selects requests to be preempted by jointly considering the SLO, SRTF and LKVO metrics in order to reduce the preemption time. 



\item[(4)] \textbf{Preemption strategy selection (\ref{sec:strategy}).} Guided by Observation~\ref{5Strategy}, \sys avoids latency estimation for swapping and recomputation, instead opting to decide based on whether the sequence length exceeds the threshold. 
\squishend
}


In Figure~\ref{fig:component1}, users' requests are initially entered to the waiting queue. \emph{Confidence-based padding} (\circled{1}) is executed to estimate the output length of each request when it waits in the queue. After each iteration, \emph{SLO-aware batching and KVC allocation} (\circled{2}) selects waiting requests to form a batch and allocates KVC to each batched request. Next, the batch is forwarded to the execution engine to be executed. When a request experiences under-provisioning, the \emph{preemption policy}  (\circled{3}) is executed to choose the requests to be preempted. The preempted requests are entered to the waiting queue and ordered. 
For a request to be preempted, the \emph{preemption strategy selection} (\circled{4}) is executed to reduce the preemption time of this request and other running requests. 


\subsection{Confidence-based Padding}
\label{sec:padding}

Given the autoregressive nature of LLMs, accurately predicting the response length of a request poses a challenge. RLP achieves an accuracy of 67\%~\cite{Zheng2023ResponseLP}, while $S^3$ reaches 79\%~\cite{Jin2023S3IG}, resulting in overprovisioning and underprovisioning and increasing TTFT and TBT (O\ref{PaddingDetermine}). 
To address this issue, we propose a confidence-based padding method, leveraging Hoeffding's inequality to bound overprovisioning and underprovisioning with high probability or confidence.

\DEL{\sys uses the OPT-13B model as in~\cite{Zheng2023ResponseLP}) running in another server to predict the response length. 
In this fine-tuning process, the LLM model receives each prompt and the Root Mean Squared Error (RMSE) of the running requests\sh{what are the running requests?} as an input and produces a predicted response length and a confidence score as the outputs\sh{in fine-tuning, you enter the ground truth for the outputs? how did you get confidence score ground truth?}.
These outputs\sh{what are these outputs? this is not inference process, it is fine tuning} are then compared against the actual response length to determine if the prediction was accurate. Then, we calculate the RMSE between predicted and actual response lengths. By analyzing\sh{how analyze? Fine tuning process is already the process to tune the parameters} both the RMSE and the confidence scores generated by the LLM, the model adjusts the model parameters to minimize prediction errors by reducing RMSE, and thus refining the model's ability to predict response lengths for various prompts accurately.

In the prediction stage, the LLM only receives the prompt as input \sh{in the above, you had RMSE as input, but now, it is not an input anymore?},  it directly outputs two values: (1) a predicted response length, estimating the number of tokens required to satisfy the prompt, and (2) a confidence score indicating the model’s certainty about its length prediction (3) whether the predicted length is assumed to be overprovisioned or underprovisioned. \sh{these 3 outputs are the same as the 3 outputs in fine tuning, right?}
These predictions are then evaluated against the actual response length obtained from the inference process, allowing us to assess accuracy by calculating the RMSE between the predicted and actual lengths \sh{how do you use this RMSE?}. }



\begin{figure}
    \centering  \includegraphics[width=0.65\columnwidth,height= 0.11\textheight]{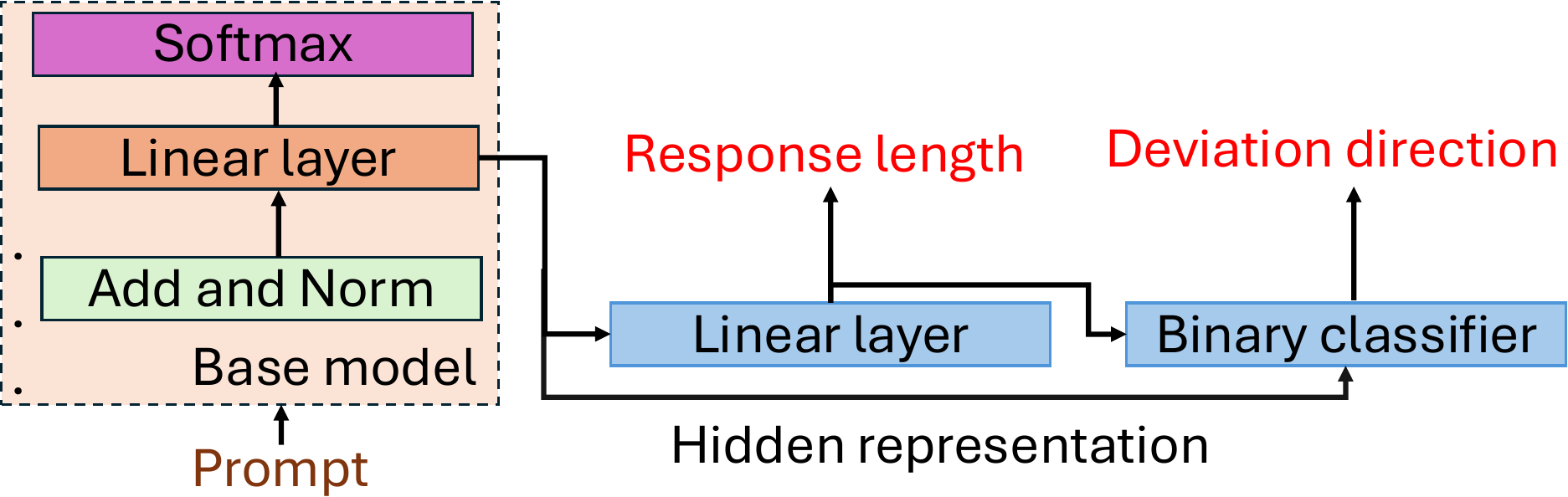}
    \caption{Our model for output length prediction. }
    \label{fig:base-model}
\end{figure}

\DEL{\tsr{We use DistilBERT as the base model and extend its architecture to predict three outputs: response length, the confidence or reliability of the length prediction, and whether the resource allocation results in overprovisioning or underprovisioning. To achieve this, the original model is enhanced by adding three task-specific heads directly to its pooled output representation.

The first head predicts the response length using a regression layer that outputs a single scalar value. The second head predicts the confidence score, which is a continuous value between 0 and 1, indicating how reliable the length prediction is. This is implemented using a linear regression layer. The third head predicts the provisioning status, indicating whether the system is overprovisioned (1) or underprovisioned (0), using another fully connected layer with a sigmoid activation function to produce a binary output.

Each head operates on the pooled representation of the input prompt from BERT’s encoder. The pooled output representation in BERT refers to the hidden state of the final layer of the encoder. This representation is designed to capture a global summary of the input sequence and is commonly used for classification or regression tasks. The outputs of these added layers are configured with appropriate loss functions to optimize their respective tasks. Specifically, the response length is optimized using Root Mean Squared Error (RMSE), which is configured during the fine-tuning process, as the regression layer itself does not produce the RMSE directly. Similarly, the confidence head uses RMSE for training, which is also defined externally . The provisioning status head uses Binary Cross-Entropy (BCE) loss, which is applied during training and not inherently part of the sigmoid layer.

The total training loss combines these three losses as follows:
\begin{equation}
    \mathbf{L}_{\text{total}} = \beta \cdot \mathbf{L}_{\text{length}} + \delta \cdot \mathbf{L}_{\text{confidence}} + \gamma \cdot \mathbf{L}_{\text{provisioning}},
\end{equation}
where \( \beta, \delta, \gamma \) are weights that control the relative contributions of each task to the overall training process.

The model requires labeled data for training. The ground truth for response length is directly derived from the actual output length of the prompt. The confidence score is calculated using the formula:
\begin{equation}
    y_{\text{confidence}} = 1 - \frac{\lvert \text{predicted\_length} - \text{actual\_length} \rvert}{\text{actual\_length}},
\end{equation}
where higher errors result in lower confidence. This is derived from the historical predicted lengths of similar prompts and their actual length. The provisioning status is labeled to determine if the predicted length is significantly higher (overprovisioned) or lower (underprovisioned) than the actual length:
\begin{equation}
    y_{\text{provisioning}} = 
    \begin{cases} 
    1, & \text{if } (\text{predicted\_length} > \text{actual\_length}), \\
    0, & \text{if } (\text{actual\_length} < \text{predicted\_length})
    \end{cases}
\end{equation}

During inference, the model processes a prompt $i$ as input and generates three outputs: the predicted response length $\hat{X}_i$, the confidence score $c_i$, and the provisioning status $ps_i$. These outputs enable the system to dynamically allocate resources based on the predicted length, adjust provisioning strategies using the confidence score, and identify whether the current provisioning state is over or under-resourced.

Using DistilBERT as the base model ensures that the input prompt is processed bidirectionally, allowing the model to generate a holistic representation of the input text. This bidirectional encoding makes DistilBERT particularly well-suited for tasks like response length prediction, where understanding the entire prompt context is critical for accurate predictions.}

-----------------------------------------------------------------------------------------
Alternative description as asked on deec3rd morning:}

\noindent \textbf{Output length prediction.} Previous methods relying on LLMs to predict output length lack the ability to provide high confidence to bound the overprediction or underprediction, or to identify the deviation direction. Our method addresses these issues. 
As in~\cite{Zheng2023ResponseLP}, we use OPT-13B as the base model and extend its architecture to predict response length, and deviation direction. As shown in Figure~\ref{fig:base-model}, we add two layers (in blue) after the last linear layer in the base model. This layer outputs the hidden representation of the input prompt, 
which is the input of our added linear layer. 
The added binary classifier outputs the deviation direction (0/1). \looseness=-1


\DEL{We specify 64 bins for the output length: [1, 32], [33, 64], $\cdots$, [2016, 2048]. 
In the linear layer ${WX}+b$, the weight matrix ${W}$ has dimensions $64 \times 2048$, the input ${X}$ has dimensions $2048 \times 1$, and the bias ${b}$ is of size $64 \times 1$. This layer produces an output tensor of size $64 \times 1$, where each value corresponds to a specific bin. The softmax layer outputs the probabilities for all bins. 
We choose the bin with the highest probability. We use the probability as the confidence score, and use the center of the bin as the predicted output length.}

\DEL{The \textbf{confidence score} is derived as the probability assigned to the predicted bin, providing a natural interpretation of the model's certainty. For this purpose, we pass the output of the linear classifer through a added softmax layer, which gives us the confidence score for the predicted bin.}



The added linear layer predicts the output length. The binary classifier is a fully connected layer. Its input includes the hidden representation of the last linear layer of the base model and that of our added linear layer.


\DEL{We derive the label using the following equation:
\begin{equation}
    y_{\text{dd}} = 
    \begin{cases} 
    1, & \text{if } (\text{predicted\_length} > \text{actual\_length}), \\
    0, &  otherwise.
    \end{cases}
\end{equation} }

The input to the model is the prompt itself. The fine-tuning is conducted in two steps. First, we fine-tune the model without the binary classifier using the ground-truth of the output lengths. Then, using the fine-turned predictor, we predict the output lengths and collect the data of deviation direction. Second, we use this collected data to find-tune the whole model including the binary classifier.   

\DEL{
For the response length classification,  we choose the Root Mean Square Error (RMSE) as the loss function. 
For the binary classifier, we choose the  binary cross-entropy as the loss function. 
}


\DEL{The model requires labeled data for training. For response length prediction, the actual response length is mapped to its corresponding bin during preprocessing. For the provisioning status layer, we use the predicted length from the previously trained linear layer and original length as input.
The provisioning status is labeled--explain how you get this training data-done

to determine if the predicted length is significantly higher (overprovisioned) or lower (underprovisioned) than the actual length. This is derived from the historical predicted lengths of similar prompts and their actual length. So, we derive the label using the following equation :
\begin{equation}
    y_{\text{provisioning}} = 
    \begin{cases} 
    1, & \text{if } (\text{predicted\_length} > \text{actual\_length}), \\
    0, & \text{if } (\text{actual\_length} < \text{predicted\_length})
    \end{cases}
\end{equation}
}



\DEL{This approach makes the model efficient, scalable, and practical for real-world multi-task scenarios. By leveraging the power of LoRA, fine-tuning OPT-13B becomes computationally feasible even for large-scale tasks.}

\DEL{\tsm{To estimate the confidence score, we perform residual analysis on the training dataset during fine-tuning. Residuals $r_i$ were defined as the difference between the predicted and actual lengths, and a Gaussian distribution was fit to these residuals.
For this purpose during fine-tuning, the residuals \(r_i = \hat{X}_i - p_i\) (where \(\hat{X}_i\) is the predicted response length and \(p_i\) is the true length) are analyzed to calibrate the model’s uncertainty. The residuals were modeled as a normal distribution:
\[
r_i \sim \mathbf{N}(\mu_r, \sigma_r^2)
\]
where \(\mu_r\) is the mean residual, representing the model’s average bias, and \(\sigma_r^2\) is the residual variance, capturing the variability in prediction errors. These statistics are stored as calibration parameters.

During inference, the confidence score for a predicted response length \(\hat{y}\) is calculated using the residual distribution. Assuming the residual for a new prompt follows the same distribution, the confidence is estimated as:
\[
\text{Confidence} = 1 - \frac{|r_{\text{approx}}|}{k \cdot \sigma_r}
\]
where \(r_{\text{approx}} = \hat{y} - \mu_r\) is the approximated residual, and \(k\) is a scaling factor that controls the sensitivity of confidence scores. Confidence is normalized to lie within \([0, 1]\) to ensure it is interpretable and bounded.

This approach enables the model to estimate confidence scores during inference without requiring the true response length. By leveraging the residual statistics observed during training, the confidence score reflects the model's reliability for a given prediction, providing a practical measure of uncertainty.}}






\DEL{\noindent{\textbf{What the model did?}}
The fine-tuning process trained the model to minimize the RMSE between its predictions and the actual response lengths while simultaneously learning to associate a confidence score with each prediction and determine the direction of provisioning (over or under). The ground truth response length directly influenced the model's loss function \tsr{(i.e., the RMSE)}, driving parameter updates to reduce errors and improve accuracy in both response length prediction and confidence score estimation \tsr{using the Adam optimizer during the fine-tuning process.}
}


\DEL{In this fine-tuning process, the LLM model receives each prompt and the Root Mean Squared Error (RMSE) of the \tsr{input} requests \sh{which input requests? the whole process is not clear.} as an input and produces a predicted response length and a confidence score as the outputs. \tsr{Note that, the ground truth data only contains the actual length. The fine-tuning like the original training has adjusts the parameters of the model to reduce the RMSE to generate the fine-tune model, for which the LLM model generates the intermediate outputs of the predicted length and the confidence score.}}

\DEL{These \tsr{intermediate outputs} \sh{what are these outputs? this is not inference process, it is fine tuning-done, these are intermediate during the training}  are then compared against the actual response length to determine if the prediction was accurate. Then, we calculate the RMSE between predicted and actual response lengths. By \tsr{calculating the} \sh{how analyze? Fine tuning process is already the process to tune the parameters-done} \sh{is this done in the inference step, if not, delete it}both the RMSE and the confidence scores generated by the LLM, the model adjusts the model parameters to minimize prediction errors by reducing RMSE, and thus refining the model's ability to predict response lengths for various prompts accurately.}

\DEL{In the prediction stage, the fine-tuned LLM receives the same instruction and prompt as input, as mentioned above, and outputs the predicted response length. It also determines whether the response is overprovisioned or underprovisioned and provides a confidence level for that prediction.} 



\DEL{and then add two padding sizes to increase the KVC demand satisfaction ratio.
The first padding size is determined by the confidence score $c$ of the prediction, so the real value has $c$ probability to be within the adjusted predicted value. Based on this adjusted predicted value, another padding determined offline is added to
ensure that the expected probability that the allocated KVC is
sufficient to meet the demand ($\overline{Pr}$) is greater than $1-\epsilon$, where $\epsilon$ is a small probability and the total KVC space is minimized. 
}


\DEL{From the analysis, we realized that underprovisioned requests have a higher impact on the running requests because of the high preemption latency. Meanwhile, the overprovisioned requests show a higher waiting time for the requests in the waiting queue. Therefore, while allocating the resources, \sys initially opts for allocating more to the running requests to finish them early. 
The amount of resources allocated is dependent on the confidence of the response length predictor model.

Let us consider the predicted length for a request $i$ by the predicted length is $l_i$ and the confidence of the model on its prediction is $c_i$, where $c_i$ is in the range of $[0,1]$.  Then the total allocated KV-cache memory, $M_i$ for that request is:
\begin{equation}
    M_i = l_i +  \lfloor l_i \times \dfrac{1}{c_i} \rfloor
\end{equation}where the padding amount is : $\lfloor l_i \times \dfrac{1}{c_i} \rfloor$.

This ensures the more confident the predictor is on its predicted response length, the less padding is added to the predicted length.

Finally, the total KVC memory, $M$ for all the requests in a current batch is combined and allocated. That is,
\begin{equation}
    M = \sum_{\forall_i \in B} M_i
\end{equation}where $B$ is the current batch.
And, the total buffer for all the requests in the current batch would be the summation of the padding for all requests in the batch:
\begin{equation}
    Bf = \sum_{\forall_i \in B} \lfloor l_i \times \dfrac{1}{c_i} \rfloor
\end{equation}
}

\DEL{Setting the padding to a high value can lead to using the confidence $c_i$ to be overprovisioned\sh{rephrase}, which avoids under-provisioning but may result in resource waste. The variation of prediction errors depends on cloud system load levels \cite{7840604}\sh{what is it? never simply copy sentences from other papers.}.} 


\noindent \textbf{Padding determination.} For a given predicted output length, if its actual output lengths follow a specific probability distribution, the padding for the predicted output length can be determined based on this distribution to achieve a certain confidence. However,  the probability distributions of the 9k fine-tuning requests and 6k inference requests does not follow a certain distribution. Therefore, this approach is not viable. 

We then leverage the Hoeffding's inequality theory.
It is used to bound the deviation with a specified confidence. 
Let $X_1, X_2, ..., X_n$ be independent random variables, where $a_i \leq X_i \leq b_i$ \text{ for } $i = 1, 2, ..., n.$
Let us define $S = X_1 + X_2 + ... + X_n$ and use $E[S]$ to denote the expected value of the sum.
Then, for any limit $t > 0$: \vspace{-0.15in}
\begin{equation}\label{Eq:positive}
\small
 P(S - E[S] \geq t) \leq \exp \left( \frac{-2t^2}{\sum_{i=1}^{n} (b_i - a_i)^2} \right) \vspace{-0.1in}
\end{equation}

\DEL{\begin{equation*}\label{Eq:negative}
P(S - E[S] \leq -t) \leq \exp \left( \frac{-2t^2}{\sum_{i=1}^{n} (b_i - a_i)^2} \right)
\end{equation*}}

\DEL{Combining both sides: \sh{make sure each notation is defined-done}
\begin{equation}\label{eq:hoef}
\small
 P(|S - E[S]| \geq t) \leq 2 \cdot \exp \left( \frac{-2t^2}{\sum_{j=1}^{n} (b_j - a_j)^2} \right)
\end{equation}, where $E[S]$ is the expected value of the sum.}

We use $X_i$ as the actual output length of request $i$, $\hat{X}_i$  as its predicted length,  $t_i$ is the maximum allowed deviation, and $(b_i-a_i)$ defines the range of request length for request $i$ in the predicted values. Then, 
Equation~\eqref{Eq:positive} becomes:\vspace{-0.1in}
\begin{equation}\label{length-hoef}
\small
    \DEL{P(|X_i - \hat{X}_i| \geq t) \leq 2 \cdot \exp \left( \frac{-2t^2}{(b_i-a_i)^2} \right)}
    P((X_i - \hat{X}_i) \geq t_i) \leq  \exp \left( \frac{-2t_i^2}{(b_i-a_i)^2} \right) \vspace{-0.1in}
\end{equation} Let $c_i$ be the specified confidence, i.e., $P((X_i - \hat{X}_i) \geq t_i = 1-c_i$. By solving the equation, we get: \vspace{-0.1in}
\begin{equation}\label{eq:t}
\small
\DEL{t = \sqrt{\frac{(b_i-a_i)^2}{2} \cdot \ln \left(  \frac{2} {(1 - c_i) }  \right)}}
t_i = \sqrt{ - \frac{(b_i-a_i)^2}{2} \cdot \ln (1 - c_i) }\vspace{-0.1in}
\end{equation} We use this $t_i$ as the padding for underprediction. From Equation~\eqref{eq:t}, higher $c_i$ leads to more padding and vice versa. Padding $t_i$ for overprediction is calculated similarly. 
Based on the deviation direction, the padding is added or subtracted from the predicted output length. 


\DEL{When the confidence score is high, after the padding $t$ is added or subtracted from the predicted output length to adjust it to the estimated output length, the probability that the actual output length is within the estimated output length is high; otherwise, the probability is low, which leads to high overprovisioning or underprovisioning. 
}

Based on O\ref{PaddingDetermine}, a higher request arrival rate results in longer waiting times, necessitating a lower padding size hence lower confidence, and vice versa. Thus, we propose dynamically adjusting $c_i$ based on the request arrival rate \( \lambda \): \vspace{-0.1in}\begin{equation}\label{equ:c}
\small
c_i = \frac{\alpha}{1 + \beta \lambda},\vspace{-0.05in}
\end{equation}  where   \( 0 < \alpha \leq 1 \) controls the maximum possible confidence, and \( \beta>0 \) regulates the sensitivity of the confidence to the arrival rate changes. 
At low arrival rates (\( \lambda \to 0 \)), the denominator \( 1 + \beta \lambda \) approaches \( 1 \), resulting in \( c_i \approx \alpha \). As the arrival rate \( \lambda \) increases, the denominator grows, causing \( c_i \) to decrease. This reflects reduced confidence in predictions when KVC is limited for the requests. 
The parameters here are empirically determined. 



\DEL{\tsr{We get this equation by replacing the Equation~\eqref{eq:hoef} with n=1. When determining the padding for a single request, we are concerned with the deviation of that specific request's actual length from its predicted length.  Therefore, we are effectively considering a sample size of 1 (i.e., n=1).}

\tsr{We use the absolute value $|X_i - \hat{X}_i|$ because we care about the magnitude of the deviation (also avoid any negative value), regardless of whether the actual length is longer or shorter than predicted. This sets a bound. The predicted length is not way off on either side. Moreover, fixing one side leads to a tighter bound and can lead to higher variations. Also, we adjust for overprovision and underprovision later.}
}



\DEL{We use a simple inverse relation where higher arrival rate $\lambda$ leads to a lower confidence score $c_i$, ensuring less padding and vice-versa. Our formulated equation is as follows: 
\begin{equation}
    c = \dfrac{k}{\lambda}
\end{equation}, where $k$ is a contant scaling factor to ensure $c$ remains within the valid range $[0,1]$ and $\lambda$ is the mean arrival rate. To prevent  $c$ from exceeding $1.0$ or dropping below a reasonable threshold, we normalize  $\lambda$ using the expected range of arrival rates. 
We calculate the $\lambda_{norm} = \dfrac{\lambda}{\lambda_{mean}}$. Here, $\lambda_{mean}$ is the average of observed mean arrival rate by the system. That is, if we set different arrival rates $\lambda_1$, $\lambda_2$, $\cdots.. \lambda_n$ for a system, $\lamda_{mean} = \dfrac{\lambda_1+\lambda_2+ \cdots+\lambda_n}{n}$ So, the equation of $c$ becomes: 
\begin{equation}
    c = \dfrac{k}{\lambda_{norm}} = \dfrac{k\times \lambda_{mean}} {\lambda}
\end{equation} 
However, in any case, if $c$ goes beyond 1, we clip to the value of 1.}
 

\DEL{*  It can also replaced $(b-a)$ as we are considering only one.}

\DEL{\begin{figure}
    \centering  \includegraphics[width=0.6\columnwidth,height= 0.11\textheight]{Padding-FIgs/requests-confidence-score.pdf}
    \caption{Confidence score-wise request distribution.\sh{need to redraw}}
    \label{fig:conf-score}
\end{figure}}

\DEL{\textbf{Goal: Bounding the Deviation}

$\text{Ensure } X_i \leq \hat{X}_i + t \text{ with high probability.}$

\textbf{Original Hoeffding's Inequality~\cite{hoeffding1963}:}

$\text{Let } X_1, X_2, ..., X_n \text{ be independent random variables, where } a_i \leq X_i \leq b_i \text{ for } i = 1, 2, ..., n.

\text{Let } S = X_1 + X_2 + ... + X_n.

\text{Then, for any } t > 0:$

\begin{equation*}
   P(S - E[S] \geq t) \leq \exp \left( \frac{-2t^2}{\sum_{i=1}^{n} (b_i - a_i)^2} \right)
\end{equation*}

P(S - E[S] \leq -t) \leq \exp \left( \frac{-2t^2}{\sum_{i=1}^{n} (b_i - a_i)^2} \right)

\text{Or, combining both sides:}

P(|S - E[S]| \geq t) \leq 2 \cdot \exp \left( \frac{-2t^2}{\sum_{i=1}^{n} (b_i - a_i)^2} \right)

So, Hoeffding's inequality is all about bounding the probability of large deviations. It says how much is the actual sum deviates from the expected sum.

\textbf{Modifying the Hoefding's Probability:}

\tsr{So, Hoeffding inequality helps us answer the question: "How likely is it that the actual value ($X_i$) is far away from the expected value ($\hat{X}_i$)?}

\tsr{Following, is same as the $P(y-hat{y})$ mentioned by the ChatGPT.}

$P(|X_i - \hat{X}_i| \geq t) \leq 2 \cdot \exp \left( \frac{-2t^2}{(b_i-a_i)^2} \right)$

\text{where:}

* $X_i: \text{actual length of request } i$

* $\hat{X}_i: \text{predicted length of request } i$

* $t: \text{maximum allowed deviation}$

\tsr{* We replace the hoeffding's inequality with n=1. When determining the padding for a single request, we are concerned with the deviation of that specific request's actual length from its predicted length.  Therefore, we are effectively considering a sample size of 1 (i.e., n=1).}

* $(b_i-a_i): \text{range of request length for i}$. It can also replaced $(b-a)$ as we are considering only one.

\tsr{We use the absolute value $|X_i - \hat{X}_i|$ because we care about the magnitude of the deviation (also avoid any negative value), regardless of whether the actual length is longer or shorter than predicted. This sets a bound. The predicted length is not way off on either side. Moreover, fixing one side leads to a tighter bound and can lead to higher variations. Also, We adjust for overprovisiond and underprivision later.}

\textbf{Confidence Level}

$(1 - c_i) \cdot \gamma: \text{target probability of exceeding cushioned estimate}$

\text{where:}

* $c_i: \text{confidence score of prediction } \hat{X}_i$

* $\gamma$: \text{overall tolerance for exceeding cushioned estimate}

\textbf{Setting } t

\textbf{Target Probability:}
$2 \cdot \exp \left( \frac{-2t^2}{(b_i-a_i)^2} \right) = (1 - c_i) 

\textbf{Solve for } t:
\tsr{$t = \sqrt{ \frac{(b_i-a_i)^2}{2} \cdot \ln \left(  \frac{2} {(1 - c_i) }  \right) }$}
\textbf{Interpretation}
* $\text{Higher } c_i \text{ or lower } \gamma \Rightarrow \text{larger } t \Rightarrow \text{more cushion}.$ This makes sense because we're being more cautious.

\textbf{Example}

$c_i = 0.8,  \gamma = 0.05, \ (b_i-a_i) = 1000 \Rightarrow t \approx 158.11$

\therefore \text{Add } \approx 158 \text{ tokens of cushion.}}

\subsection{SLO-aware Batching and KVC Allocation}\label{sec:allocation}

\DEL{Design guidance: 1. To avoid preemptions,
\textbf{Guidance 1.1)} in initial KVC allocation, we need to allocate KVC to a request as close to its demand as possible while limiting waiting time and maximizing throughput. 
(Method1: Arrival rate based KVC allocation)

\textbf{Guidance 2)} To reduce adverse influence from preemptions, when we cut allocated KVC for requests in batching or in selecting a request to preempt, we need to choose requests that have less impact from preemptions. The requests include those that have higher delay tolerance or longer response lengths (as they will have more opportunities to receive KVC later).

\textbf{Guidance 1.2)} in run time, we need to ensure that a request can receive KVC upon request proactively (reservation and pre-allocation) instead of waiting for KVC opportunistically. (Method 2: Proactive KVC allocation and global reservation)
}



\DEL{Both KVC and GPU compute resources are needed for executing a request. Previous work aims to fully utilize either the GPU compute resource~\cite{vllm} or the KVC~\cite{Agrawal2023SARATHIEL} and select the waiting requests in the FCFS manner. 
Instead, our goal in this paper is to satisfy the SLOs for TTFT and TBT to improve user experience. For this purpose, even though token budget is used up, if the waiting time is still long due to fast arrival rate, we still add requests to the batch to increase the dequeueing speed if this does not violate the SLOs of the requests in the batch. Based on Observation~\ref{1Obs:KVCLimit}, we aim to fully utilize the allocated KVC to mitigate the KVC bottleneck.
}

\DEL{The unallocated KVC determines the number of tokens that can be included in a batch for processing, thereby impacting throughput. Allocating one block per request in a batch may cause KVC allocation failures during token generation for some requests, increasing TBTs for evicted requests. Conversely, if we allocate a request's total KVC demand, the risk of allocation failures decreases but it limits the unallocated KVC hence the number of requests per batch, increasing TTFT. Therefore, optimizing the allocated KVC for each request is essential.}



This component tackles the challenge outlined in O\ref{KVCLimit} by incorporating (1) an embedding method to reuse allocated but unused KVC (\ref{sec:embedding}), (2) a request selection and KVC allocation strategy (\ref{sec:selection}), and (3) preemption-avoidance mechanisms (\ref{sec:global}). Key notations are summarized in Table~\ref{tab:notations}.


\begin{table}[ht]
\centering
\caption{Summary of notations.}
\vspace{-0.15in}
\footnotesize
\resizebox{\columnwidth}{!}{\begin{tabular}{|c|l|}
\hline
\textbf{Notation} & \textbf{Description} \\ \hline

$\text{SLO}_{ttft}$  & TTFT SLO. \\ \hline
$\text{SLO}_{tbt}$ &  TBT SLO. \\ \hline



$\mathbf{N}_w$ & Waiting requests that must execute next to meet $\text{SLO}_{ttft}$ \\ \hline
$\mathbf{N}_r$ & Returned requests that used up allocated KVC but must execute next to meet $\text{SLO}_{tbt}$ \\ \hline
$\mathbf{N}'_w$ & Selected waiting requests that contribute to exhausting the token budget \\ \hline
$\mathbf{N}'_r$ & Returned requests that used up allocated KVC but not needing immediate execution \\ \hline
$\text{T}^{max}_\text{I}$ & Maximum iteration latency historically observed \\ \hline
$s^p_i$ & Prompt length of request $i$ \\ \hline
$s^o_i$ & Estimated output length of request $i$ \\ \hline
$M_i$ & Total KVC demand for request $i$: $M_i = s^p_i + s^o_i$ \\ \hline
$A'_{kvc}$ & Unallocated KVC available for the current iteration \\ \hline
$B$ & Fixed block size for KVC allocation \\ \hline
$D_{kvc}$ & Total KVC demanded for critical requests:
            $D_{kvc} = \sum_i^{|\mathbf{N}_w|}(s^p_i + B) + \sum_i |\mathbf{N}_r| B$ \\ \hline
$a_i$ & Allocated KVC for a running request $i$ \\ \hline
$u_i$ & Currently used/occupied KVC for a running request $i$ \\ \hline
$\gamma$ & Factor used to determine KVC allocation cuts based on SLO priority\\ \hline
$E_{kvc}$ & Excess demanded KVC: $E_{kvc} = \sum M_i - A'_{kvc}$ \\ \hline
\end{tabular}}
\label{tab:notations}
\end{table}

\begin{figure}
\centering
\includegraphics[width=0.5\columnwidth]{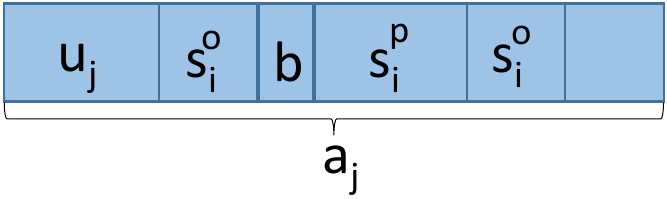}
\vspace{-0.0in}
\caption{Emdedding method to reuse allocated but unused KVC.}\vspace{-0.0in}
\label{fig:emdedding}
\end{figure}

\subsubsection{Embedding Method} \label{sec:embedding} As shown in Figure~\ref{fig:emdedding}, a running request $r_j$'s allocated KVC is denoted by $a_j$ and its currently used KVC is denoted by $u_j$. If another request $r_i$ uses $r_j$'s allocated but unused KVC starting from the location marked by the red line, when both $r_i$ and $r_j$ run $s^o_i$ iterations, $r_i$ completes and releases its KVC and $r_j$ has reached the point $b$ tokens before $r_i$'s KVC space. $b$ functions as a buffer to handle the inaccurate estimation and it is set to a small number (e.g., 8 tokens). Therefore, if $a_j-(u_j+s^o_i)-(s^p_i+s^o_i)\geq b$, it means that we can allocate $r_i$ in the allocated KVC of $r_j$ starting from the red line $s_0=a_j - (u_j + s^o_i) - b$. To choose a request to embed request $r_i$, we choose the running request $r_j$ that has the minimum remaining allocated KVC 
in order to reduce reserved waste. 
Due to inaccurate estimation of $s^o_i$, if $r_i$ does not complete by the time when $r_j$ needs the KVC allocated to $r_i$, then $r_i$ must be preempted. The value of $b$ is empirically determined; a larger $b$ leads to higher reserved waste while a smaller $b$ may cause the preemption of $r_i$.



\subsubsection{Request Selection and KVC Allocation} \label{sec:selection}

The requests in a returned batch after an iteration are called \emph{returned requests}. In forming a new batch and allocating KVC, we make sure that (1) the waiting requests' $\text{SLO}_{ttft}$ and the returned requests' $\text{SLO}_{tbt}$ must be satisfied, and (2) we allocate the KVC to a request as close to its demand as possible to avoid preemptions to reduce TBTs while reducing reserved waste to reduce both TTFTs. 

The waiting requests are ordered by the ascending order of their Remaining Time (RT) to their $\text{SLO}_{ttft}$ and the $\text{SLO}_{tbt}$ for preempted requests, denoted by $\text{RT}_{ttft}$ and $\text{RT}_{tbt}$. The requests are sequentially selected to add to the batch. We use $\text{T}^{max}_\text{I}$ to denote the maximum iteration latency historically. Then, the waiting requests that have $(\text{RT}_{ttft}- \text{T}^{max}_\text{I})<\epsilon$, where $\epsilon$ is a small number, must be batched in order to satisfy their $\text{SLO}_{ttft}$. We use $\mathbf{N}_w$ to denote the group of such waiting requests, and $\mathbf{N}_r$  the group of returned requests that have used up their allocated KVC and must execute in the next iteration to satisfy their $\text{SLO}_{tbt}$, that is, $\text{RT}_{tbt}- \text{T}^{max}_\text{I}<\epsilon$. The requests in $\mathbf{N}_w$ and $\mathbf{N}_r$ are called \emph{critical requests}, which need to allocate KVC first. 
Below, we present how to allocte KVC to critical requests and then to other returned requests and requests selected to exhaust the token budget (named as non-critical requests).

\textbf{(1) Allocate KVC to critical requests.} We define a small block (e.g., 8 blocks), denoted by $B$. The basic KVC requirement of a critical waiting request equals to the sum of its prompt length and a block size: $(s^p_i + B)$, and that of a critical returned request equals to $B$. The total basic KVC requirement of critical requests equals to $D_{kvc}=\sum_i^{|\mathbf{N}_w|} (s^p_i+B)+\sum_i|\mathbf{N}_r| B$. We sort the critical requests in descending order of their basic requirements and allocate each request using the embedding method first. Then, we allocate the remaining critical requests to unallocated KVC. 
If the current unallocated KVC $A'_{kvc}<D_{kvc}$, preemptions are executed to make $A'_{kvc}=D_{kvc}$. Which requests to choose for preemptions is introduced in \ref{sec:preemption}. When $A'_{kvc}\geq D_{kvc}$, we allocate the basic required KVC to each remaining critical request. 
Then, if there remains unallocated KVC, we allocate it to all critical requests and other non-critical requests 
The details are presented below.


\DEL{the number of non-critical requests should equal to the token budget, denoted by $B_T$~\cite{deepspeed-fastgen}.
Therefore, in addition to  $\mathbf{N}'_r$, other requests ($\mathbf{N}'_w$) are sequentially selected from the waiting requests to use up the token budget.
}


\DEL{arrival rate is $\lambda$ and the throughput is $\mu$. The demand $\sum M_i$ should be updated to $\sum M_i \cdot \frac{\lambda}{\mu}$. The reduction $\sum M_i-\sum M_i \cdot \frac{\lambda}{\mu}$ is amortized among the requests based on the ratio of $\frac{SLO}{\sum SLO_i} s^p_i$.
}

\textbf{(2) Allocating remaining KVC for the next iteration.} We use $\mathbf{N}'_r$ to denote the group of returned requests that are not critical. 
In addition to critical requests, $\mathbf{N}'_r$, we sequentially select requests from the waiting requests to use up the token budget (denoted by $\mathbf{N}'_w$) to minimize waiting time and maximize throughput. 
All non-critical requests ($\mathbf{N}'_w$ and $\mathbf{N}'_r$) and critical requests ($\mathbf{N}_w$ and $\mathbf{N}_r$) will join the KVC allocation. We use $n$ to denote the total number of these requests. The next question is how to distribute $A'_{kvc}$ among these requests. 
In this step, we allocate KVC to each request to match its total demand ($M_i = S_i^p + S_i^o$) as closely as possible while accommodating the $n$ requests to the batch.



For non-critical requests, we use the embedding method first. If it successfully allocates a request, the request's $M_i$ becomes 0 . Then, we allocate $A'_{kvc}$ to the remaining requests for the following three cases: 

\squishlist


\item When $\sum_{i=1}^n M_i = A'_{kvc}$, the requests are added to the batch and are guaranteed to receive the KVC they demand during execution.

\item When $\sum_{i=1}^n M_i < A'_{kvc}$, adding more requests can more fully allocate the KVC, increase dequeuing speed and increase throughput slightly, but may also increase iteration time. The iteration time, estimated based on profiled data, increases almost linearly with the number of tokens in a batch~\cite{Agrawal2023SARATHIEL}. Therefore, more requests are sequentially added from the queue until $\sum_{i=1}^n M_i = A'_{kvc}$ or the SLO of any request in the batch is violated. 



\DEL{After the first method, these embedded requests' $M_i$ becomes 0. 
Then, we update the total demanded KVC, $\sum M_i$. 
}

\item When $\sum_{i=1}^n M_i> A'_{kvc}$, we amortize the excess demands among the requests, and allow a request to use the released KVC from another request once the latter exhausts its allocated KVC. 
A request with a looser SLO and with a longer predicted output length  should have a higher KVC demand cut since it is less delay-sensitive and also have more opportunities to receive KVC. Therefore, we incorporate these two factors in the amortization process. The weight for each request is computed as:
$w_i = \frac{\text{RT}_i}{\sum_{i} \text{RT}_i} \times \frac{s_i^p}{\sum_{i} s_i^p}$. Subsequently, the KVC allocated to each request is determined by: $A_i = A'_{kvc} \times \frac{w_i}{\sum_{i=1}^n w_i}$.




After the above amortization, a request $r_i$ may not receive its demanded KVC fully. To address this, we find a running request $r_j$ that will release its KVC shortely before $r_i$ exhausts its allocated KVC, and the released KVC is no less than $r_i$'s additional demand. Then, $r_i$ will use $r_j$'s released KVC.



\squishend




\subsubsection{Proactive KVC Allocation and Global Reservation} 
\label{sec:global}

A request may not receive its KVC demand $M_i$ (an unfulfilled request) and then if it exhausts its allocated KVC without any request completing, a preemption occurs. The proposed proactive KVC allocation and global reservation aim to avoid the preemptions. In the previous systems, completed requests' released KVC is used to accommodate requests from the waiting queue. To avoid preemptions, after allocating the KVC to the critical requests, we include unfulfilled running requests, predicted to complete within $m$ iterations (where $m$ is a small number), into the non-critical request group for the remaining KVC allocation.

\DEL{we use the remaining unallocated KVC also for unfulfilled running requests, which
are predicted to complete within $m$iterations, where $m$ is a small number, 
Spcifically, we include these requests with the newly added requests to allocate $A'_{kvc}$ or allocated but unused KVC as described above. 
}

Due to the autoregressive nature of LLMs and the imprecision in output length estimation, a request may not receive the necessary KVC when needed. Assigning additional padding to each request can lead to wasted reserved memory that may go unused. To address this, we reserve a shared KVC space for all requests, allowing them to use the space in case of KVC allocation failure. 

\DEL{For this purpose, we rely on the additive-increase/multiplicative-decrease (AIMD) algorithm. It is a feedback control algorithm best known for its use in TCP congestion control. AIMD combines linear growth of the congestion window when there is no congestion with an exponential reduction when congestion is detected.  Suppose that $\lambda$ is the mean
 arrival rate of requests and $\mu$ is the service rate capacity of the
 server (i.e., how many requests are processed in a unit time).

\begin{equation}\label{eq:all}
w(t+1) =
\begin{cases}
    w(t)+a, & \text{if } \lambda/\mu \leq threshold \\
   w(t)\cdot b, & \text{if } \lambda/\mu>threshold
\end{cases}
\end{equation}

As $b$ is typically set to 1/2, we also set $b=1/2$ here. $a$ is set to a certain percentage of the adjusted predicted output length.
Initially, each request received the KVC equal to the adjusted predicted output length $p$. When $\lambda/\mu>threshold$, each request only receives $p\cdot b$ KVC. Otherwise, each request receives $p+a$ KVC.
}

\DEL{\begin{algorithm}[t]
\footnotesize
    \SetAlgoLined
    \LinesNumbered
    \SetCommentSty{small}
    \SetKwInOut{Input}{Input}
    \SetKwInOut{Output}{Output}

    \Input{
    Queueing requests, returned batch and unallocated KVC $A'_{kvc}$}\\
    \Output{Formed batch and allocated KVC to the requests in the batch}
$|\mathbf{N}_w|$=num. of queueing requests with $\text{RT}_{ttft}-\max T_I<b$\\
$|\mathbf{N}_w|$=num. of requests to reach target forward size $B_T$\\
Pick up $\max{|\mathbf{N}_w|, n_c}$ requests $\rightarrow$ ${G}$\\
\For{each request $r_i$$\in G$}{
predict its output length\\
\If{confidence<0.9}{
Allocate one block to it\\
$G-r_i \rightarrow G$}
\Else{$s^o_i$=predicted value+padding\\
$M_i=s^o_i+s^p_i$\\}
}

\If{$\sum_{i=1}^n M_i= A'_{kvc}$}
{Allocate $M_i$ to reach $r_i$}
else \If{$\sum_{i=1}^n M_i< A'_{kvc}$}
{Sequentially pick up requests, decide its $s^o_i$ and $M_i$ until $\sum_{i=1}^n M_i= A'_{kvc}$\\}
else {
/*Find hosting running request for each request in G */\\
Sort requests $\in G$ in descending of $M_i=s^o_i+s^p_i$\\
Sort running requests $G_r$ in descending of $a_j-u_j$\\
\For{each request $r_i$$\in G$}{
\If{find running $r_j$ to host it: $(a_j-u_j)-S_i^a-M_i>b$}{
{Allocate $r_i$'s KVC to $r_j$'s allocated but unused KVC\\
$M_i=0\\
}}
}
$E_{kvc}=\sum M_i-A'_{kvc}$\\
\If{$E_{kvc}=0$}{Assign $M_i$ to request $r_i$ $\in G$ to AVC}
\Else{\If{$E_{kvc}<0$}{add more requests from the queuing to make $E_{kvc}=0$}
\Else{
Find $\alpha$ to make $E_{kvc}=\sum_i[\alpha \cdot \frac{SLO}{\sum SLO_i} s^p_i]$\\
each request's cut demand equals to  $\alpha \cdot \frac{SLO}{\sum SLO_i} s^p_i$\\
Allocate the KVC to the request\\

\If {Successfully find a running request $r_j$, that will release its KVC sometime before $r_i$ used up its allocated KVC $a_i$, i.e., $s^o_j-u_j-(a_i-s^p_i)=0$, and the released KVC is no less than $r_i$'s demand, i.e., $s^p_j+s^o_j=s^o_i-(a_i-s^p_i)$} {Pre-allocte the KVC to the request} \\
\Else{find a request that satisfy $s^p_j+s^o_j\leq s^o_i-(a_i-s^p_i)$ with $\max\{s^p_j+s^o_j-[s^o_i-(a_i-s^p_i)]\}$ to allocte to the requst}

}

}

}
/* Allocate KVC for each request in G */\\

    \caption{Forming a batch and allocating KVC.}
    \end{algorithm}}

\subsection{Preemption Policy}\label{sec:preemption}
The preemption policy determines the order for requests to be preempted to reduce both the number and duration of preemptions.
Based on O\ref{4Policy}, 
\sys considers three factors: 1) TBT SLO, 2) SRTF, and 3) LKVO. The TBT SLO of a request needs to be satisfied so we choose a loose-SLO request to be preempted. The reasons for considering SRTF and LKVO are explained in \ref{sec:policy}. The remaining processing time is measured by the remaining predicted output tokens. For each factor, we set up certain magnitude ranges and order requests accordingly. For example, 0.05-0.2s, 0.2-0.5s, and 0.5-2s for the TBT SLO, and 0-128, 128-256, 256-384, 384-512, ... in tokens for the remaining processing time and occupied KV-cache.
The requests first are ordered based on the descending order of TBT SLOs. Then, for the requests with similar TBT SLOs, they are ordered in the descending order of remaining processing time. Next, for the requests with similar remaining processing time, they are ordered in the ascending order of the KVC occupancy. Finally, \sys preempts the first request.

\DEL{We then consider the remaining processing time because a request with a longer remaining processing time could be delayed longer and have more opportunities to receive released KVC. Moreover, a request with a smaller KV-cache occupancy should be preempted first in order to reduce swapping or recomputation time and TBT.}


\DEL{\textbf{Sorting requests.} The requests are sorted based on the adverse impact from preemptions and predicted output length; requests with lower impact and longer output lengths should cut KVC first. The requests should not be greatly impacted by preemptions and have long response length (i.e., cause high reserve memory waste and will have more opportunities to receive released KVC) are selected first to cut allocated KVC or preempt. Thus, the higher delay tolerance and the longer response length, the higher cut on the allocated KVC. We first choose throughput-oriented jobs}

\DEL{First, we choose the offline throughput-oriented jobs in the descending order of the remaining JCT, and then choose loose TBT-SLO jobs in the descrinding order of the TBT-SLO. To jointly consider the response length, we propose to cut the allocated KVC by x\% of the response length and the minimum KVC allocated to such a job is one block. After each offline throughput-oriented job receives one block, if still $\sum_{i=1}^n M_i< A'_{kvc}$, we need to cut the allocated KVC on loose TBT-SLO jobs. A higher TBT-SLO should have a higher cut. Assume we need to cut $S$-token allocated KVC. Then, each request $r_i$ cuts $S\cdot \frac{TBT-SLO_i}{\sum TBT-SLO}$, where $TBT-SLO_i$ is $r_i$'s TBT-SLO(?think details later).

If reducing the allocated KVC of the newly added requests in the batch until each of the requests only is allocated with one block still cannot satisfy $\sum_{i=1}^n M_i \leq A'_{kvc}$, then we must reduce the allocated KVC of running requests. (?need to think how to cut)}




\subsection{Preemption Strategy Selection}\label{sec:strategy}
\DEL{\cite{vllm} indicates that whether we choose swapping or recomputation depends on the available memory bandwidth and available GPU compute resource. Since both resouces are dedicated to the LLM inference, their available amounts should not vary greatly over time. Therefore, }

As per O\ref{5Strategy}, if the sequence length is greater than a sweet spot, swapping is more time-efficient than recomputation. 
The key is to find this sweet spot. For this purpose, we can profiling and regression models for estimation.  

\DEL{As a result, we determine the recompuation latency and swapping latency directly from the sequence length via profiling. Then, for a predicted length by the LLM, we choose the preemption strategy based on the regression outputs. }

Profiling is conducted using the LLM system with the specific settings including the type of GPUs, the number of GPUs, the tensor parallelism (TP) and model parallelism (MP) degrees and the model. \DEL{For example, we use one A100 GPU for the OPT-13B model with TP=1 and MP=1, and 8 A100 GPU for the OPT-175B with TP=2 and MP=2. We build two regression models for a setting.} 
For a given sequence length \( S \), total memory bandwidth \( M_b \) and total GPU capacity \( G \), of the target hardware (e.g., A100), we measure the recomputation time for a range of sequence lengths, and use the data to train a polynomial regressor: 
\vspace{-0.12in}
\[
L_r(S) = \alpha_r \cdot S^{\beta_r} + \kappa_r \cdot S + \epsilon_r,\vspace{-0.12in}
\] where $\alpha_r$, $\beta_r$, $\kappa_r$ and $\epsilon_r$ are parameters determined during training. 
We choose the polynomial regressor because reecomputation latency typically involves complex operations like matrix multiplications and other non-linear computations that grow in polynomial complexity with the size of the input~\cite{narayanan2021efficient}. 

\DEL{NVIDIA Nsight Systems to determine recomputation and swapping latencies. This ensures realistic latency measurements across various sequence lengths on the target hardware, such A100 GPUs in our scenario
\noindent\textbf{Recomputation profiling.}
Recomputation latency is measured by simulating the recomputation of intermediate states, such as attention scores and MLP activations. For a given sequence length \( S \), total memory bandwidth, \( M_b \) and total GPU capacity, \( G \) of the target hardware (A100):
\begin{enumerate}
    \item Synthetic data matching the input and output dimensions of the LLM model is generated.
    \item A forward pass is executed to recompute the necessary states.
    \item The GPU time is measured using NVIDIA Nsight Systems, capturing the total latency for recomputation.
\end{enumerate}
This process is repeated for a range of sequence lengths for the total  \( M_b \) and \( G \) of A100 and , and the results are used to fit a polynomial regressor: 
\[
L_r(S) = \alpha_r \cdot S^{1.8}.
\]  where $\alpha_r$ is a parameter determined during the regression from the profiled data.

The regressor outputs the recomputation latency $L_r$.

\noindent\textbf{Swapping profiling.}
Swapping latency is measured by transferring data between GPU and CPU memory, simulating the eviction and restoration of key-value cache. For a given sequence length \( S \), with the same \( M_b \) and \( G \) as above:
\begin{enumerate}
    \item A data block of size proportional to \( S \) is allocated in GPU memory.
    \item The block is transferred to CPU memory and back to GPU memory.
    \item The total time for the transfer is recorded using NVIDIA Nsight Systems, capturing PCIe  overhead.
\end{enumerate}}

Similarly, we profile swapping times across various sequence lengths and use the results to train a linear regressor.:\vspace{-0.15in}
\[
L_s(S, M_b, G) = \gamma_s \cdot S + \delta_s,\vspace{-0.1in}
\]  where $\gamma_s$ and $\delta_s$ are determined during training. 
We choose the linear regressor because swapping latency linearly increases based on the data size~\cite{megatron}. We calculate the sweet spot $S_s$, defined as the sequence length satisfying $L_r(S) = L_s(S)$.




\DEL{For a given sequence length \( S \), the strategy is:
\[
\text{Choose: }
\begin{cases}
\text{Recomputation, if } S < S_s, \\
\text{Swapping, if } S > S_s,\\
\text{random choose one, if } S = S_s.
\end{cases}
\]
}

\DEL{when $S = S_s$, we choose recomputation. Swapping can be pipelined with the batch inference computation, and then the next token of the swapped request will be output in the next iteration along with other requests, and the iteration time for the whole batch is the time to output the token. In recomputation, the sequence will be combined with other input sequences for processing, and after the whole batch is processed, the next token is output. Since we pay more attention to reducing the TBT of the preempted requests than the throughput in this paper, so we choose the recomputation.}


\DEL{\section*{3. Example Calculation}

\subsection*{Regressor Parameters}
Assume the following regressor parameters based on profiling data:
\[
L_r(S) = 0.0015 \cdot S^{1.8}, \quad L_s(S) = 0.02 \cdot S + 200.
\]

\subsection*{For \( S = 1000 \):}
\[
L_r(1000) = 0.0015 \cdot 1000^{1.8} \approx 114.79 \, \text{ms},
\]
\[
L_s(1000) = 0.02 \cdot 1000 + 200 = 220 \, \text{ms}.
\]
\textbf{Decision:} \( L_r(1000) < L_s(1000) \), so choose \textbf{Recomputation}.

\subsection*{For \( S = 20,000 \):}
\[
L_r(20000) = 0.0015 \cdot 20000^{1.8} \approx 9800 \, \text{ms},
\]
\[
L_s(20000) = 0.02 \cdot 20000 + 200 = 600 \, \text{ms}.
\]
\textbf{Decision:} \( L_r(20000) > L_s(20000) \), so choose \textbf{Swapping}.}

\DEL{This strategy uses profiling data collected via NVIDIA Nsight Systems to generate regressors for recomputation and swapping latencies. The decision metric ensures optimal performance by dynamically selecting the method with lower latency based on the sequence length. Profiling ensures that the decision-making process is based on realistic hardware performance, optimizing latency for varying sequence lengths.}

\DEL{Between the preemption mechanisms, swapping and recomputation have different impact based on {\color{red} preempted datasize, available memory (PCIe) and available GPU }\sh{this is not the only factor-done}. Considering this difference in impact of both of these preemption schemes, we choose the preemption scheme considering some key objectives as follows.
\begin{itemize}
    \item Minimize computation overhead and latency.
    \item Efficiently utilize memory and computational resources.
\end{itemize}

\DEL{In order to meet these objectives \sys considers several key factors.
\begin{itemize}
    \item  \textbf{Current System Load:} \sys includes memory usage, CPU/GPU utilization, memory (PCIe) bandwidth, the number of active requests, GPU and memory usage.
    \item \textbf{Request Characteristics:} \sys considers request size (input/output length).
\end{itemize}}

@@@observation 2 is not good. cant compare with oracle
@@@ combine observation 3 with observation 1.

Upon considering these mentioned characteristics, \sys follows a greedy rule-based heuristics to make the preemption decision. In the swapping procedure, \sys selects a set
of requests to evict and transfer their KV-cache to the CPU.
Once it preempts a request and evicts its KVC, \sys
stops accepting new requests until all preempted sequences
are completed. Once a request is completed, its KVC are freed
from memory, and the KVC of a preempted sequence are
brought back in to continue the processing of that sequence. The request to be brought back is chosen as it fits in the freed memory.

In the recomputation case, \sys recomputes the KVC when the preempted sequences are rescheduled. However, the tokens generated at decoding before preemption can be concatenated with the original user prompt as a new prompt while doing the recomputation.

\sys estimates the unit pausing latency per token for both the swapping and recomputation based on the preempted data size, memory (PCIe) bandwidth and the available GPU capacity while evicting a request. During the eviction, \sys chooses the policy that provides lower unit pausing latency. }

\DEL{\sys continuously tracks memory usage, CPU/GPU utilization, and the number of active requests. From the tracking, it gather statistics on average recomputation time and swapping time. Now, if memory usage is high, prioritizing swapping over recomputation is necessary to free up space.
On the other hand, If CPU/GPU utilization is high, recomputation needs to be preferred to avoid additional I/O overhead.}



\DEL{Keeping all these heuristics in mind, \sys calculates the cost of recomputation and swapping for each heuristic. \sys uses the historical data and real-time performance metrics of GPU utilization, and number of running requests, and their corresponding SLO to estimate the recomputation time using a simple random forest regression model. A similar regression model is used to estimate the swapping time for each request. This regression for swapping time considers the number of requests, their SLOs, size of the KV-cace of each request and the current delay of disk I/O to estimate the swapping time. Upon calculating both of these costs, we follow the above mentioned rules to determine the preemption policy of a request. The order of the factors during preemption is the current system load, request characteristics and the prediction confidence. However, it is quite easy to formulate a weight-based mechanism to consider all these to find out the preemption policy of a request.} 

\DEL{\begin{algorithm}
\label{preemption}
\caption{Preemption Policy}
\begin{algorithmic}[1]
\REQUIRE request, system\_load, prediction\_confidence
\STATE recomputation\_cost $\leftarrow$ estimate\_recomputation\_cost(request)
\STATE swapping\_cost $\leftarrow$ estimate\_swapping\_cost(request)

\IF{system\_load['memory\_usage'] > MEMORY\_THRESHOLD}
    \IF{swapping\_cost < recomputation\_cost}
        \RETURN 'swap'
    \ELSE
        \RETURN 'recompute'
    \ENDIF
\ENDIF

\IF{system\_load['cpu\_utilization'] > CPU\_THRESHOLD \OR system\_load['gpu\_utilization'] > GPU\_THRESHOLD}
    \IF{recomputation\_cost < swapping\_cost}
        \RETURN 'recompute'
    \ELSE
        \RETURN 'swap'
    \ENDIF
\ENDIF

\IF{prediction\_confidence < CONFIDENCE\_THRESHOLD}
    \IF{recomputation\_cost < swapping\_cost}
        \RETURN 'recompute'
    \ELSE
        \RETURN 'swap'
    \ENDIF
\ENDIF

\IF{request['priority'] = 'high' \AND request['SLO'] < SLO\_THRESHOLD}
    \IF{recomputation\_cost < swapping\_cost}
        \RETURN 'recompute'
    \ELSE
        \RETURN 'swap'
    \ENDIF
\ENDIF

\IF{request['remaining\_time'] > TIME\_THRESHOLD}
    \IF{swapping\_cost < recomputation\_cost}
        \RETURN 'swap'
    \ELSE
        \RETURN 'recompute'
    \ENDIF
\ENDIF

\RETURN 'swap' \textbf{if} swapping\_cost < recomputation\_cost \textbf{else} 'recompute'

\end{algorithmic}
\end{algorithm}}

\section{Implementation} We implemented \sys based on the vLLM source code~\cite{vllm}, comprising about 6K lines of Python code. To integrate confidence-based padding, we modified the \emph{scheduler.py} file. Specifically, we added a new function, \emph{compute\_confidence\_padding} and extended the \emph{allocate\_kvc} function in vLLM to incorporate this padding method. 

We modified \emph{kvcache.py} to include our \emph{embedding method}. We added a function named \emph{form\_batch} in the \emph{scheduler.py}. 
We further added functionalities in the \emph{allocate\_kvc} function to allocate KVC. New function \emph{reuse\_unused\_kvc} was integrated into the \emph{TokenBlockManager} class in vLLM to reallocate unused KVC. We also added function  \emph{proactive\_allocation} in \emph{scheduler.py} to proactively allocate KVC. 

We added another function \emph{global\_reservation} that sets up a reserved portion of the KVC. To track reserved KVC, in the \emph{TokenBlockManager} class in \emph{kvcache.py}, we added a function named \emph{track\_reserved\_blocks} to manage the tracking, allocation, and release of reserved blocks. This function also ensures that the reserved blocks are separate from the general allocation pool, preventing conflicts during allocation. Function \emph{allocate\_reserved\_kvc} is added in \emph{scheduler.py} to  assign reserved KV-cache blocks to requests.

We replaced the default FCFS preemption policy in the \emph{evict\_request} function in vLLM with our preemption policy. 
Besides, we added a function, \emph{preempt\_request\_strategy}, which decides the preemption strategy. 


\section{Performance Evaluation}
\label{sec:evaluation-results}


\noindent{\textbf{Experiment Settings.}}
Unless otherwise specified, the experiment settings are the same as in \ref{sec:analysis}. \DEL{We used the same models with the same AWS machine configuration mentioned in~\ref{sec:setting}, respectively, following strategies in~\cite{vllm}. In the experiment,} \DEL{We added the Llama-3-8B and Llama-3-70B models. Llama-3-8B ran on one GPU and Llama-3-70B ran on 4 GPUs within one machine.} 
For setting the TTFT and TBT SLO, we measured the average TTFT and TBT across all compared methods and multiply them by a SLO scale. The SLO scale was randomly selected from [0.5, 2.5]. Furthermore, to make the TTFT SLO reasonable for a long prompt that needs chunking~\cite{deepspeed-fastgen}, we increased its TTFT SLO by multiplying it with a scaling factor equal to the number of chunks. 
We used 3-hour trace for OPT-13B and used 1-hour trace for OPT-175B. By default, we set the block size (B) to 8 tokens, the number of preallocation iterations $m$ to 2, the number of reserved blocks to 8, the confidence to 90\%, $\alpha=8$, and $\beta=100$, respectively.   


\noindent{\textbf{Output length predictor.}}
We used 
9K requests and 6k for inference for testing the performance. \DEL{Out of those requests, 67\% of the data is used for fine-tuning and 33\% for testing.} The fine-tuning process took around 9 hours. 

\DEL{Instead of combining the response length prediction task with the request processing on the same LLM as in~\cite{Zheng2023ResponseLP}, we used a separate LLM running on another server (with four A100 GPUs) for response length prediction to avoid interference on the request processing. For more details of the response length predictor, please refer to~\cite{Zheng2023ResponseLP}.
}

\DEL{The memory management process\sh{what is it?} was built using the \emph{append\_token} function scheme in~\cite{vllm}\sh{what does it do?}.-done} 

\DEL{\tsr{vLLM does support dynamic KVC updates or optimized memory access for padding and embedding mechanisms.} We adapted the \emph{vLLM} kernel to optimize read and write operations for KVC. These changes include adjustments to thread block configurations to improve memory access patterns and support for dynamic KVC block updates during runtime. The changes in the kernel align with the padding and embedding mechanisms, and thus ensure efficient execution without significant overhead. Here, a thread block is a group of threads that execute concurrently and can share data through memory. \sh{how did you adapt?-done}.  \sh{what vLLM cannot do, what you did to enable vLLM to do these are not clear-done, deleted, vLLM, kernel does it, nothing new, we just set a value}}

\DEL{A GPU thread block was assigned to read/write the memory to ensure parallel memory access. Allocation for the predicted response length with buffering for the requests was handled using a separate thread block
\sh{what is "thread block"? I feel writing here is very hectic and random. For each of our methods, you need to explain what you modifed or added.-done, deleted unnecessary things}}

\noindent{\textbf{Compared Methods.}}
We compared \sys with vLLM, RLP,  $S^3$, Sarathi-Serve (Sarathi for simplicity). Due to space limit, unless otherwise specified, we plot the average of the results across three datasets for each model. 

\DEL{\tsr{We also tested the variants of \sys. \sys/C (i.e., without \emph{Confidence-based padding}), which replaces the confidence-based padding with the static padding. Then, we measured the \sys/E, that does the eviction in the First-Come-First-Serve (FCFS) manner. Finally, we tested \sys/P, where we replaced the preemption strategy with a random selection of recomputation and swapping, respectively.}  We also tested \sys with full knowledge of the response lengths and denoted it by \emph{Oracle}.}

\DEL{\begin{figure*}[t]
\centering
     \subfloat[ShareGPT.\vspace{-0.01in}\label{fig:exp-sharegpt}]{{\includegraphics[width=0.32\linewidth,height=0.13\textheight]{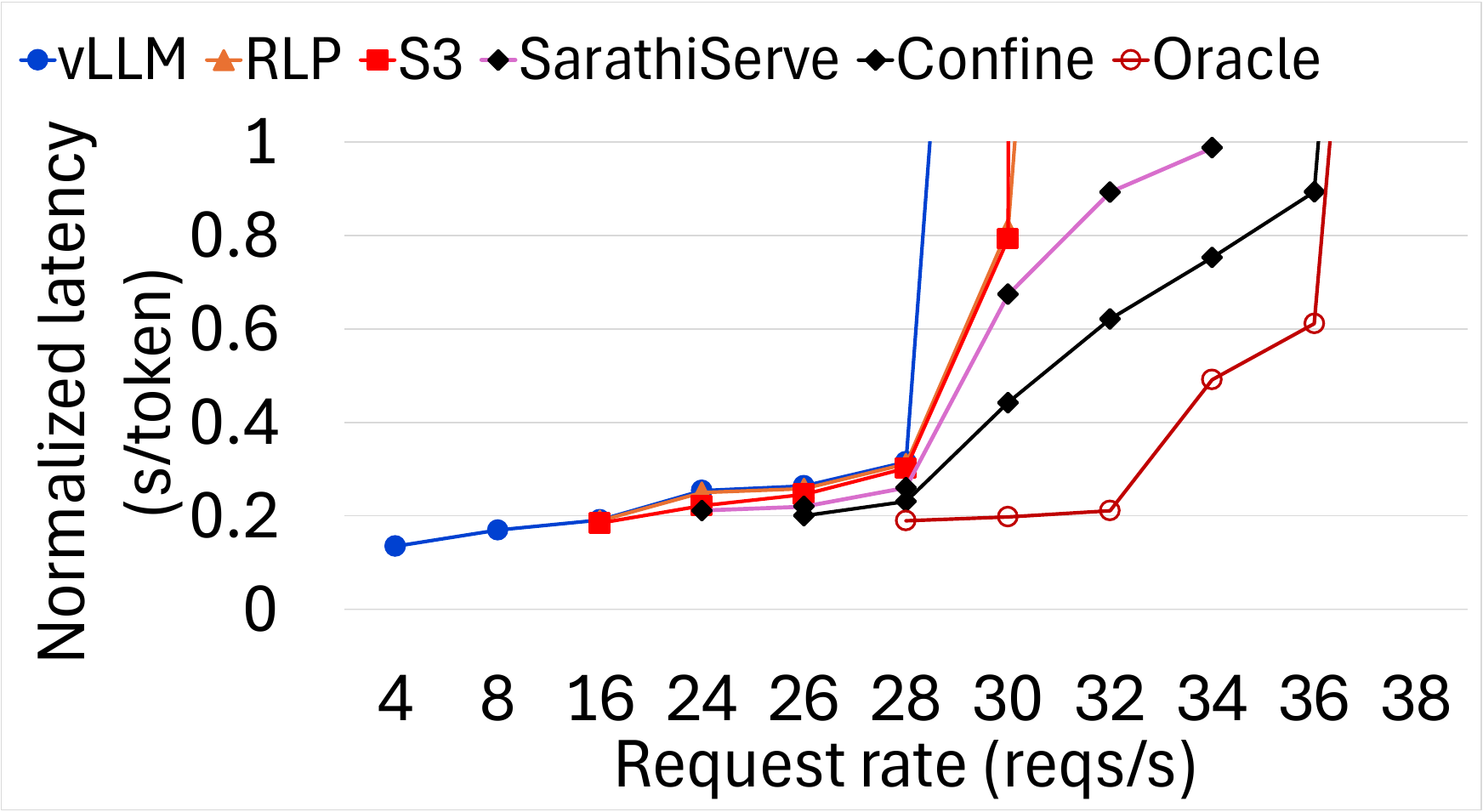} }}
    \hfill
    \subfloat[BookCorpus.\vspace{-0.01in}\label{fig:exp-book}]{{\includegraphics[width=0.32\linewidth,height=0.13\textheight]{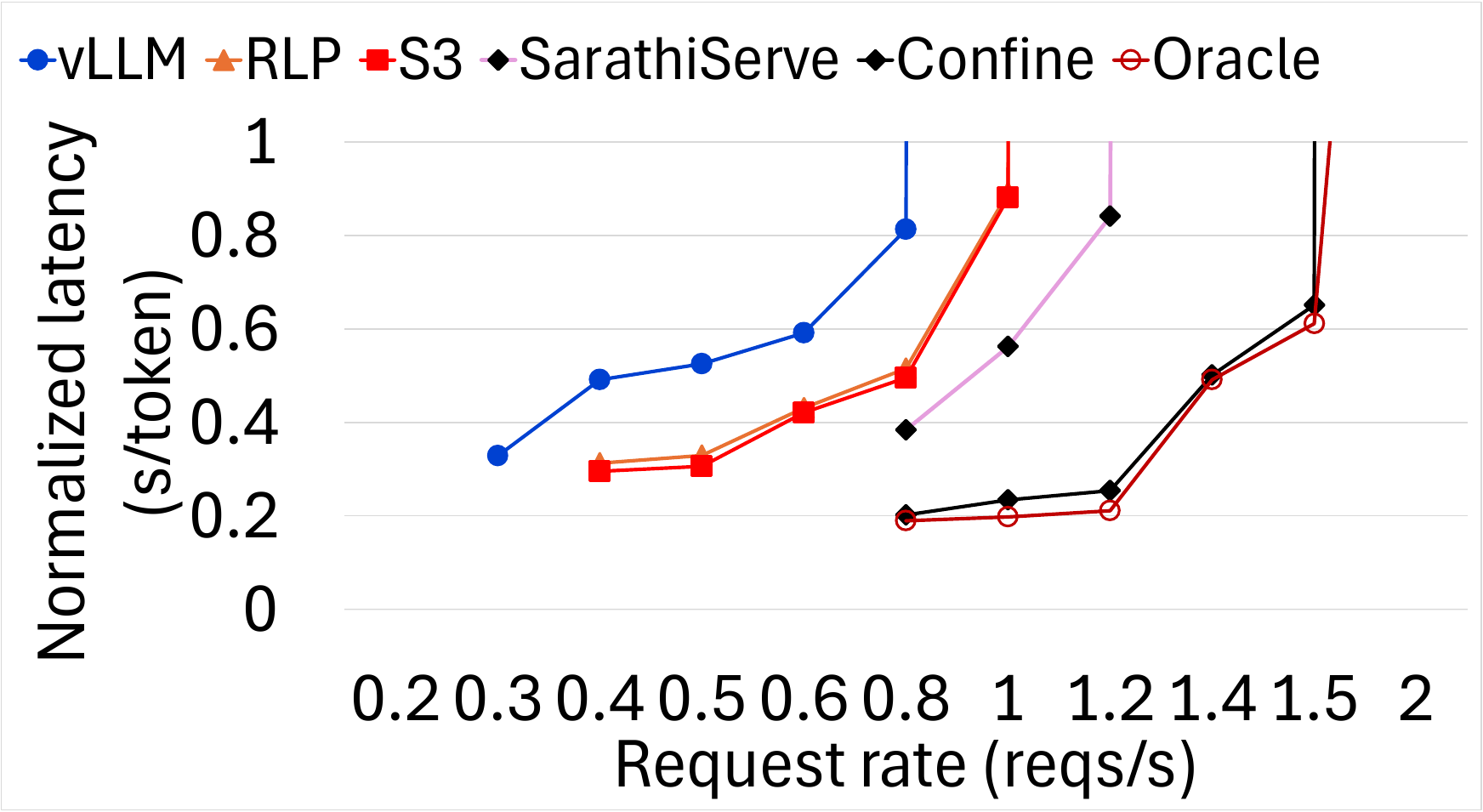} }}
    \hfill
    \subfloat[Alpaca.\vspace{-0.01in}\label{fig:exp-alpaca}]{{\includegraphics[width=0.32\linewidth,height=0.13\textheight]{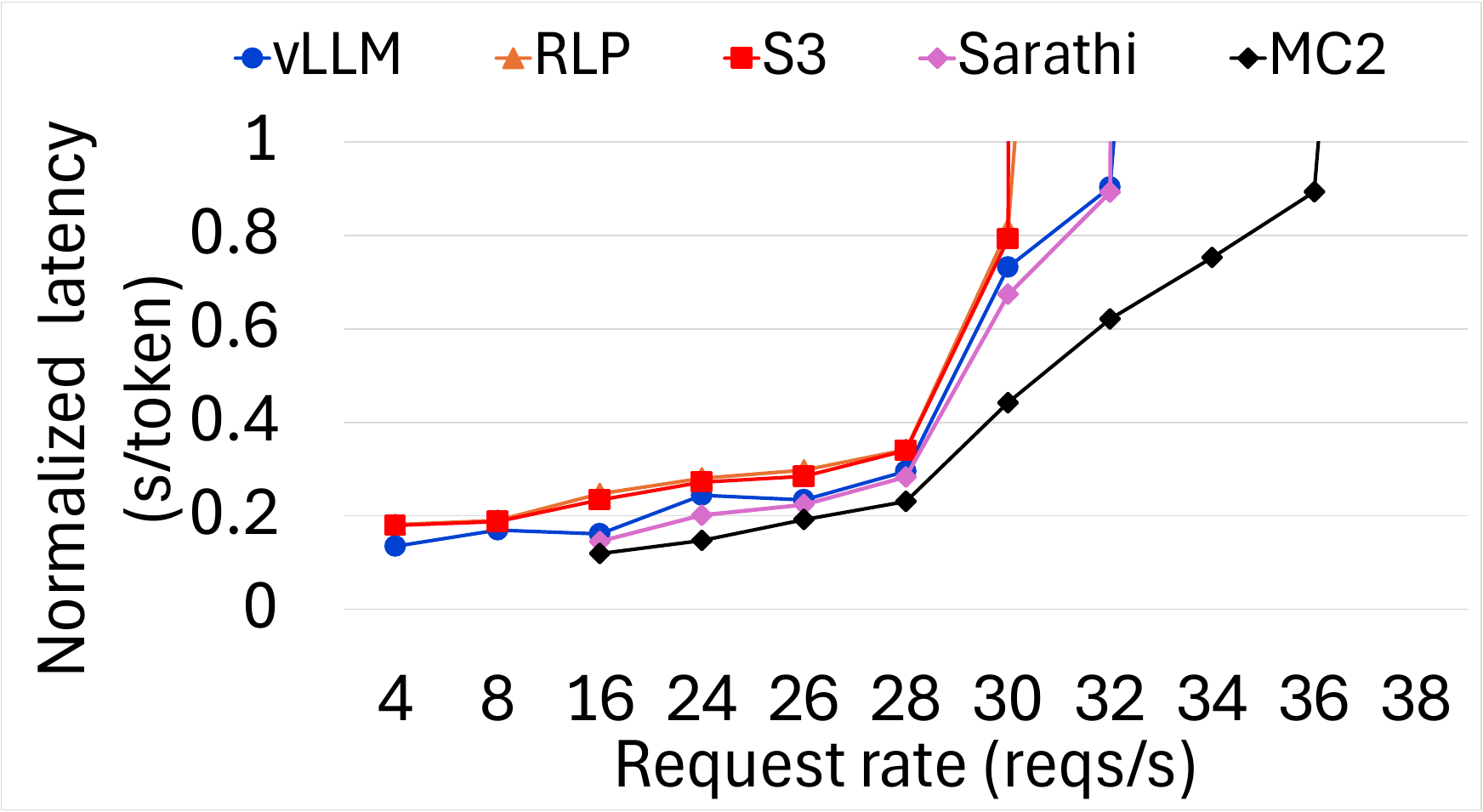} }}
    \hfill
   \caption{\small{End-to-end latency performance comparison with varied request rates on the three traces.\vspace{-0.0in}}}%
    \label{fig:overall}
\end{figure*}}

\begin{figure*}[t]
\centering
     \subfloat[Tail TTFT\vspace{-0.01in}\label{fig:exp-tfft}]{{\includegraphics[width=0.24\linewidth,height=0.13\textheight]{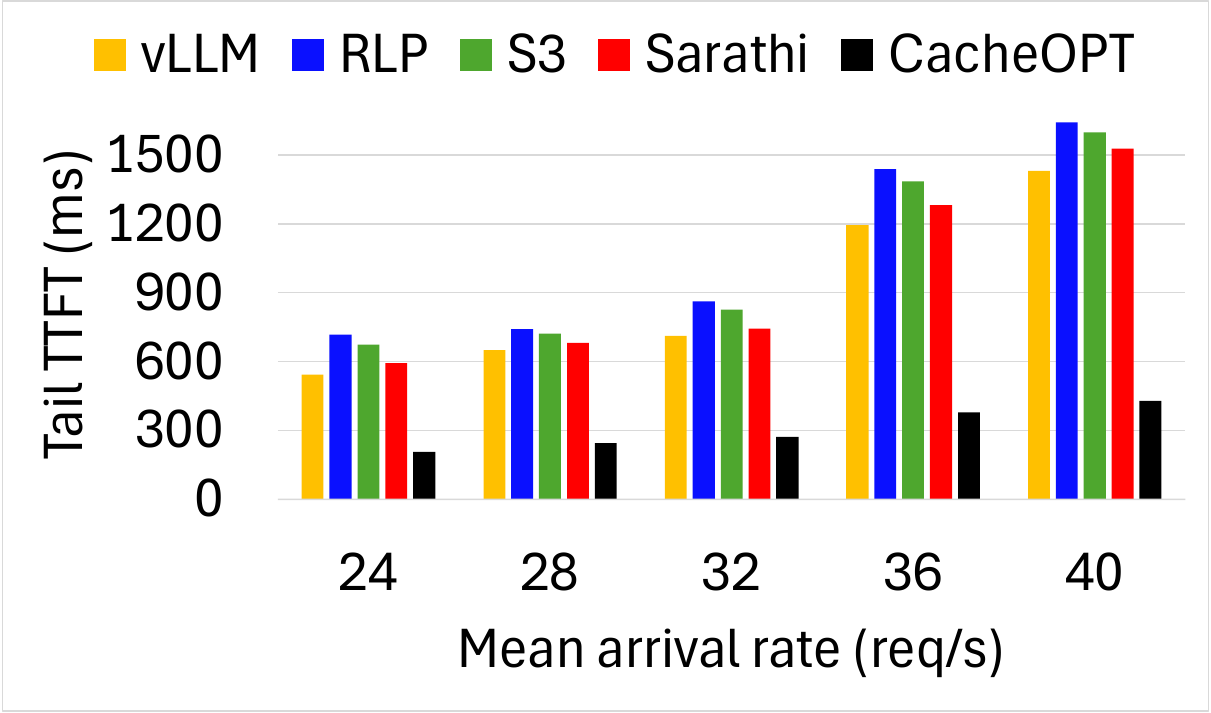} }}
    \hfill
    \subfloat[TTFT SLO attainment\vspace{-0.01in}\label{fig:exp-ttft-slo}]{{\includegraphics[width=0.24\linewidth,height=0.13\textheight]{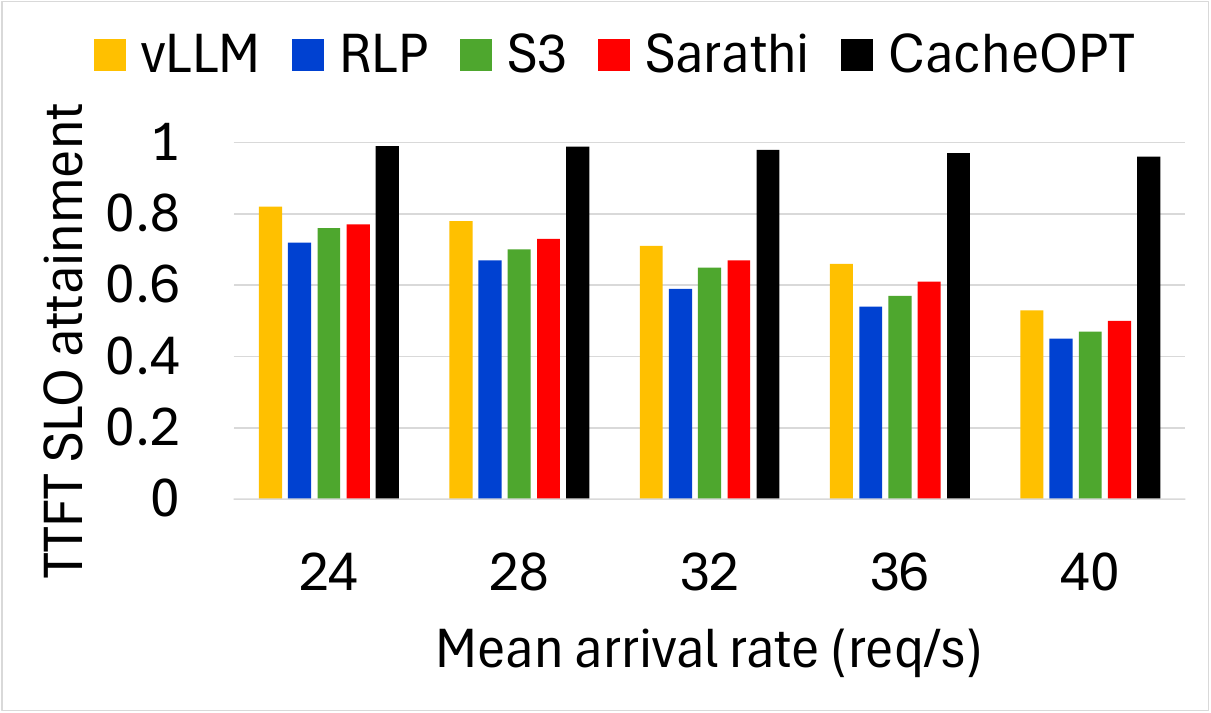}}} 
    \hfill
    \subfloat[
    Tail TBT\vspace{-0.01in}\label{fig:exp-tbt}]{{\includegraphics[width=0.24\linewidth,height=0.13\textheight]{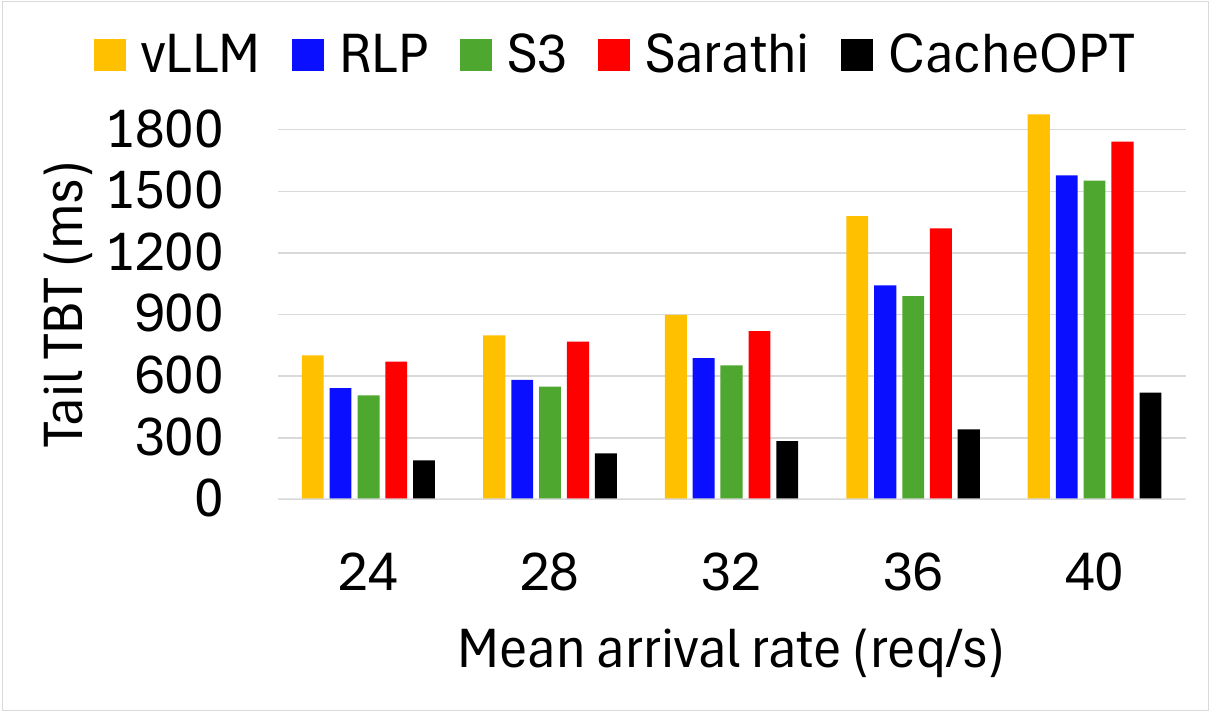} }}
    \hfill
    \subfloat[TBT SLO attainment\vspace{-0.01in}\label{fig:exp-tbt-slo}]{{\includegraphics[width=0.24\linewidth,height=0.13\textheight]{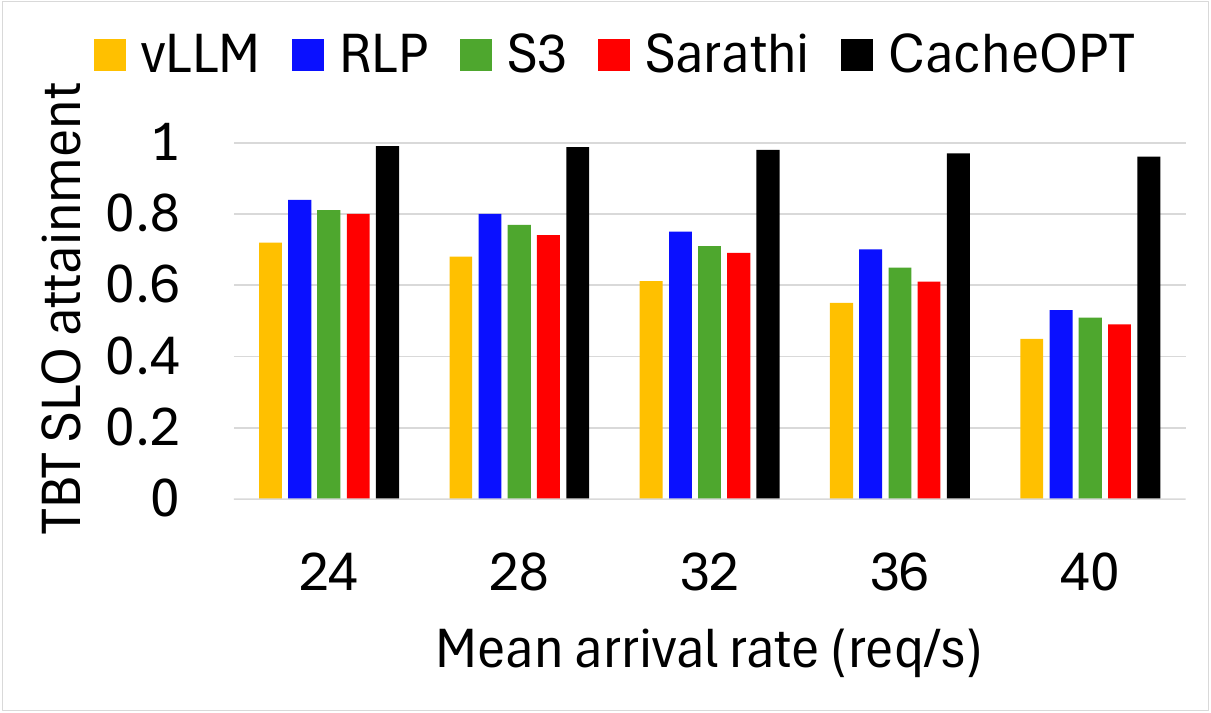} }}
    \hfill
    \DEL{\subfloat[Tail TTFT\vspace{-0.01in}\label{fig:exp-tfft}]{{\includegraphics[width=0.24\linewidth,height=0.13\textheight]{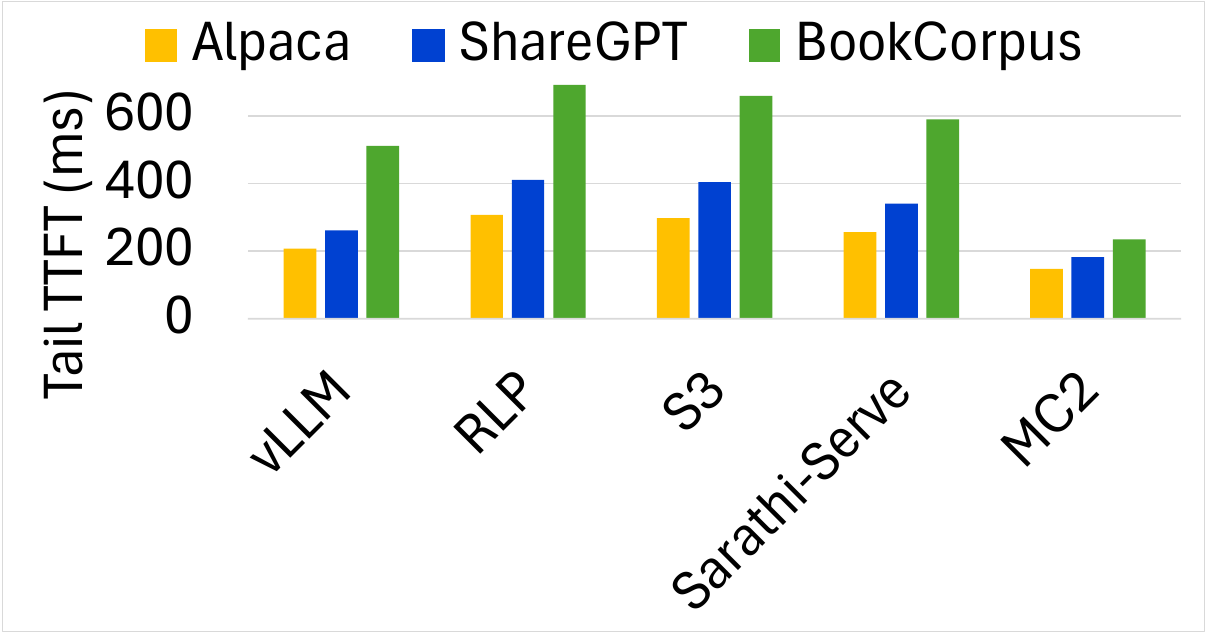} }}
    \hfill
    \subfloat[TTFT SLO attainment\vspace{-0.01in}\label{fig:exp-ttft-slo}]{{\includegraphics[width=0.24\linewidth,height=0.13\textheight]{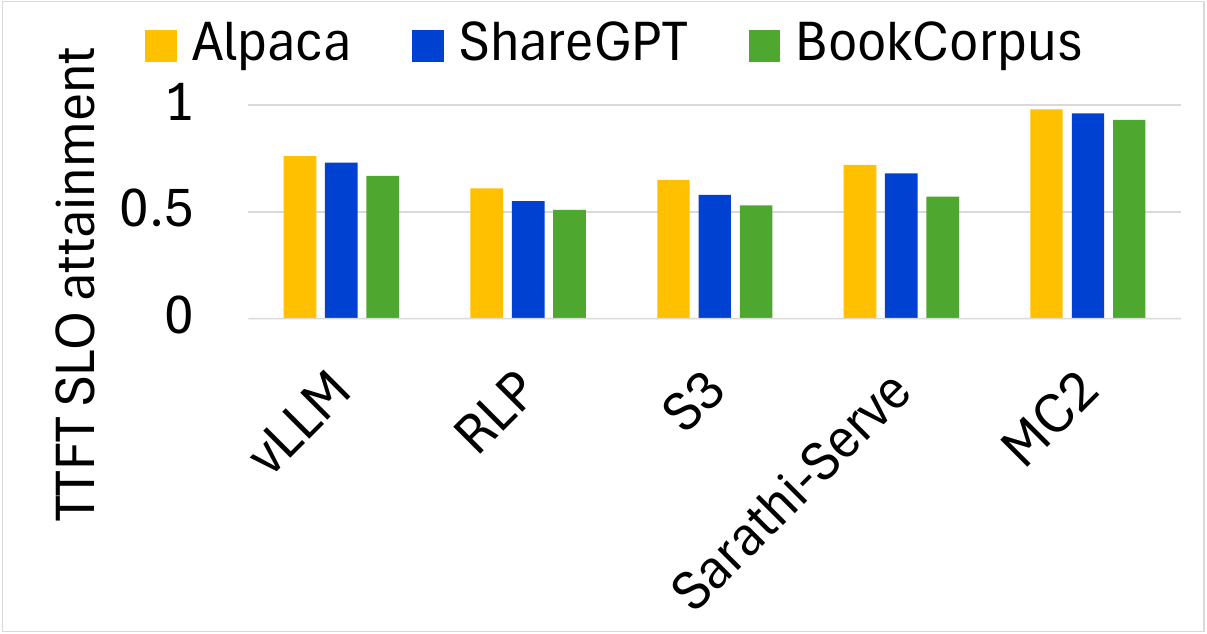} }}
    \hfill
    \subfloat[ Tail TBT\vspace{-0.01in}\label{fig:exp-tbt}]{{\includegraphics[width=0.32\linewidth,height=0.13\textheight]{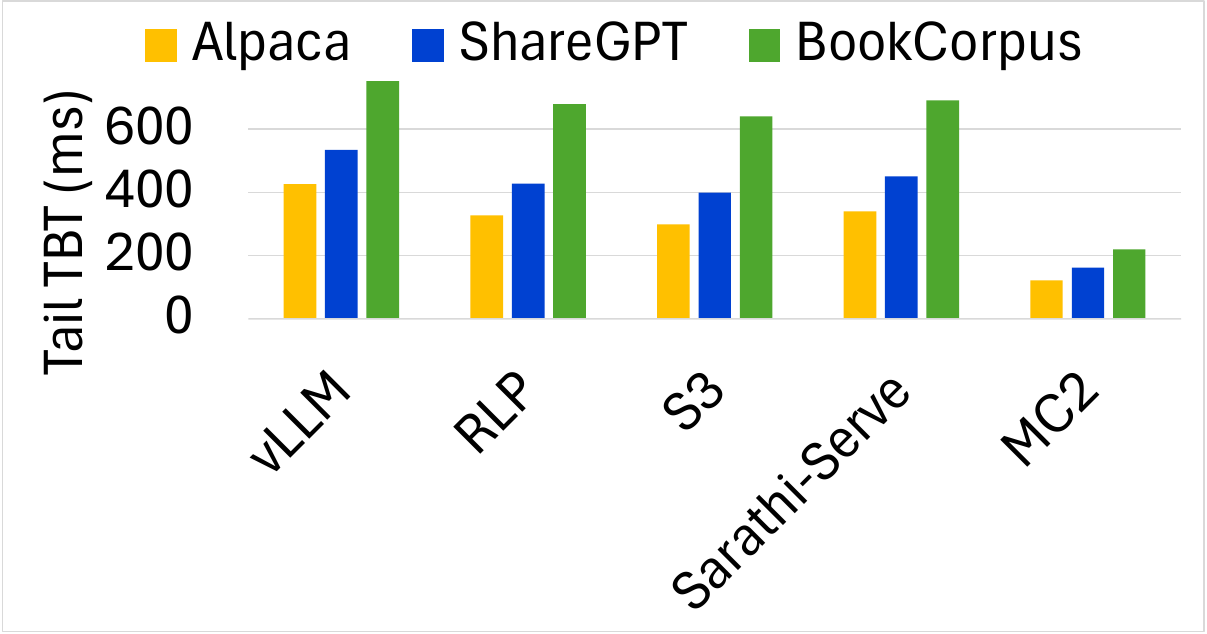} }}
    \hfill
    \subfloat[TBT SLO attainment\vspace{-0.01in}\label{fig:exp-tbt-slo}]{{\includegraphics[width=0.32\linewidth,height=0.13\textheight]{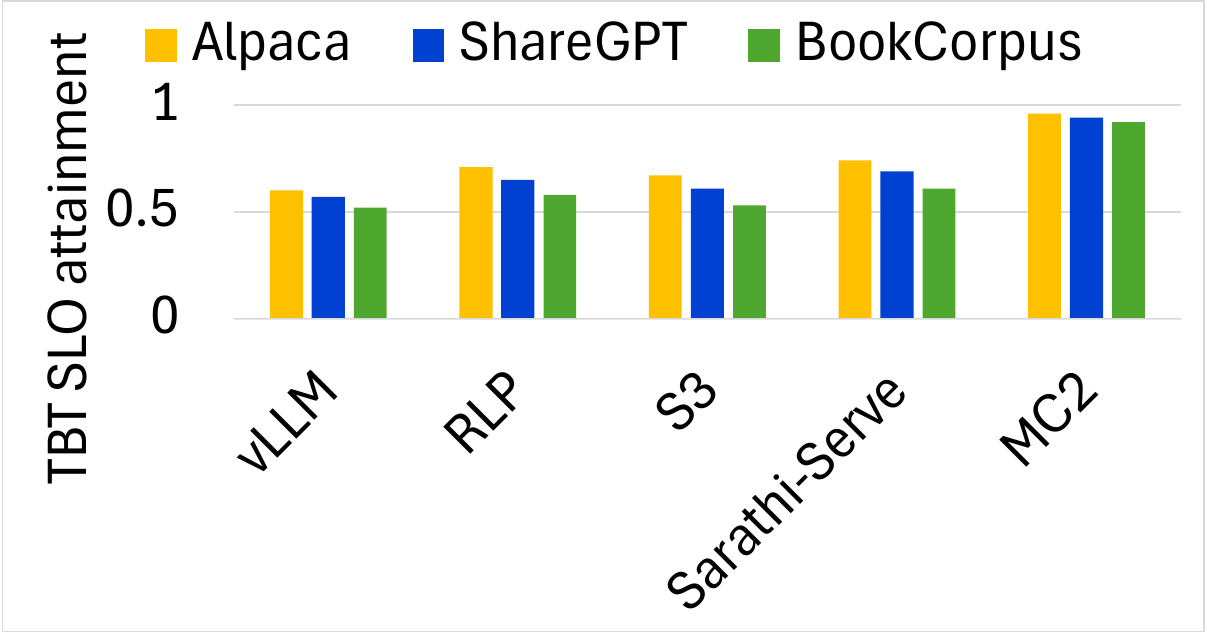} }}
    \hfill}
    \DEL{\subfloat[Overprovisioning rate.\vspace{-0.01in}\label{fig:exp-ov}]{{\includegraphics[width=0.32\linewidth,height=0.13\textheight]{Padding-FIgs/overprovisioning-rate-13b-2.pdf} }}
    \hfill
    \subfloat[Underprovisioning rate.\vspace{-0.01in}\label{fig:exp-under}]{{\includegraphics[width=0.32\linewidth,height=0.13\textheight]{Padding-FIgs/Underprovisioning-rate-13b-color.pdf} }}
    \hfill}
    \subfloat[Alpaca.\vspace{-0.01in}\label{fig:alpaca-13}]{{\includegraphics[width=0.32\linewidth,height=0.13\textheight]{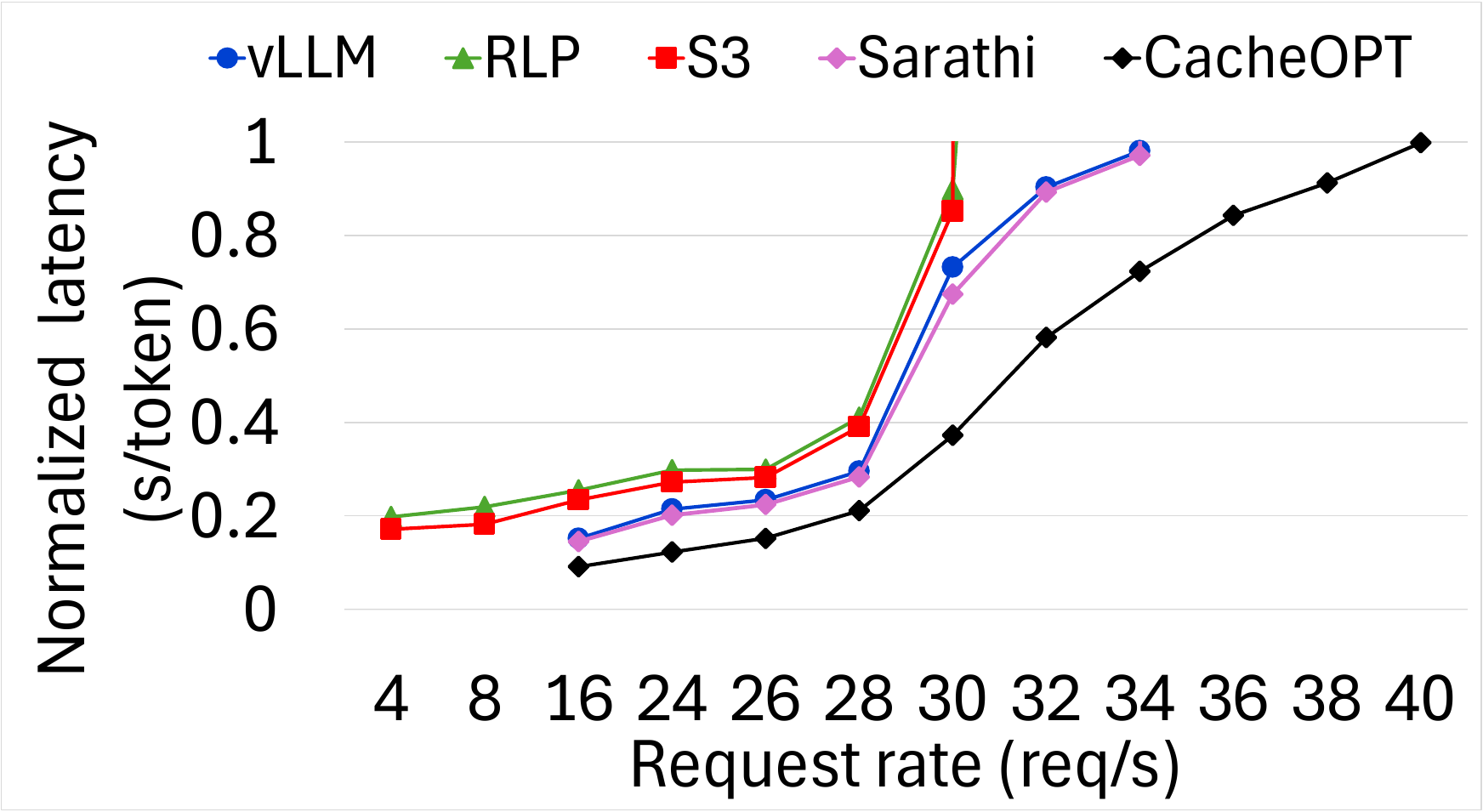} }}
    \hfill
    \subfloat[ShareGPT\vspace{-0.01in}\label{fig:sha-13}]{{\includegraphics[width=0.32\linewidth,height=0.13\textheight]{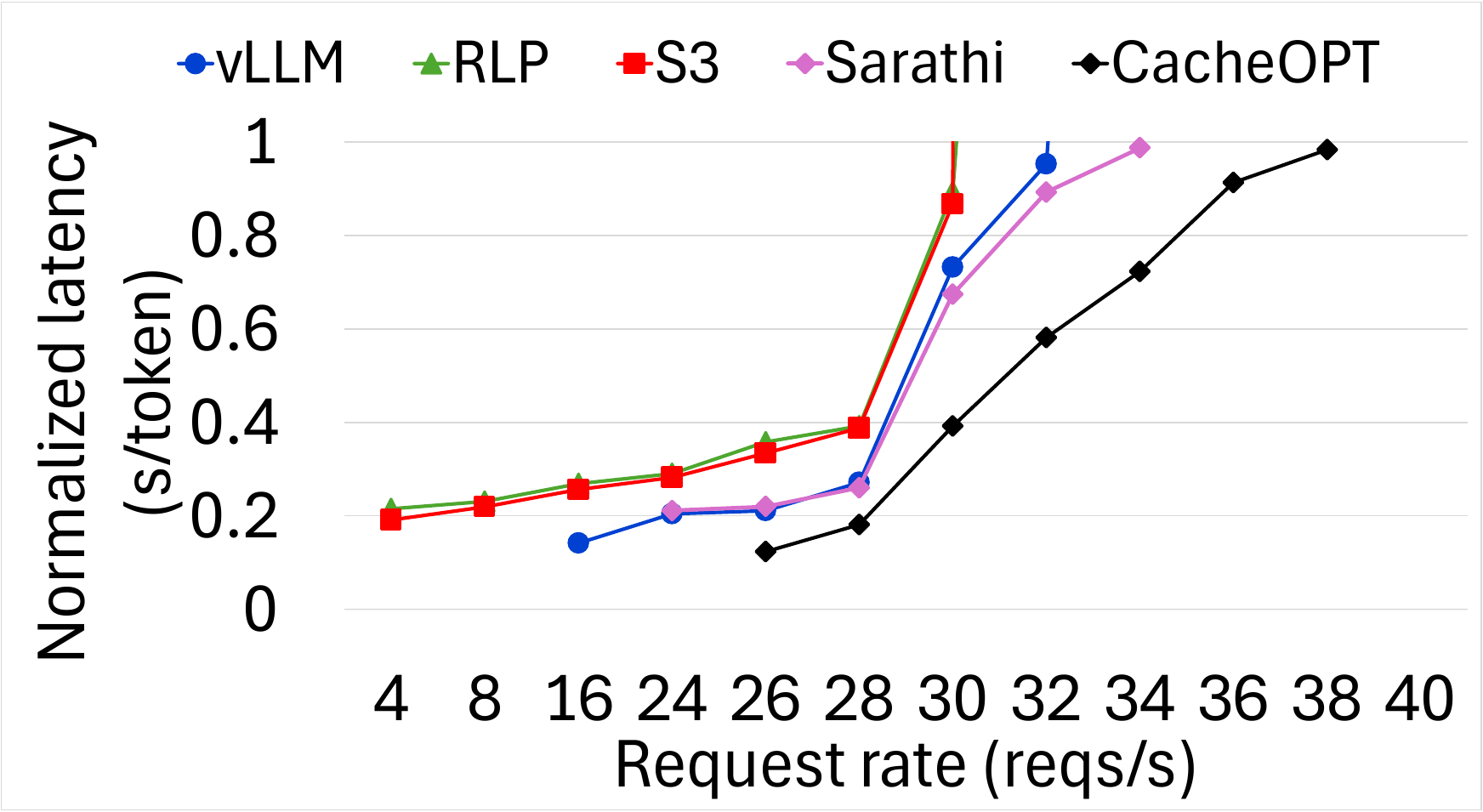} }}
    \hfill
    \subfloat[Bookcorpus\vspace{-0.01in}\label{fig:book-13-e}]{{\includegraphics[width=0.32\linewidth,height=0.13\textheight]{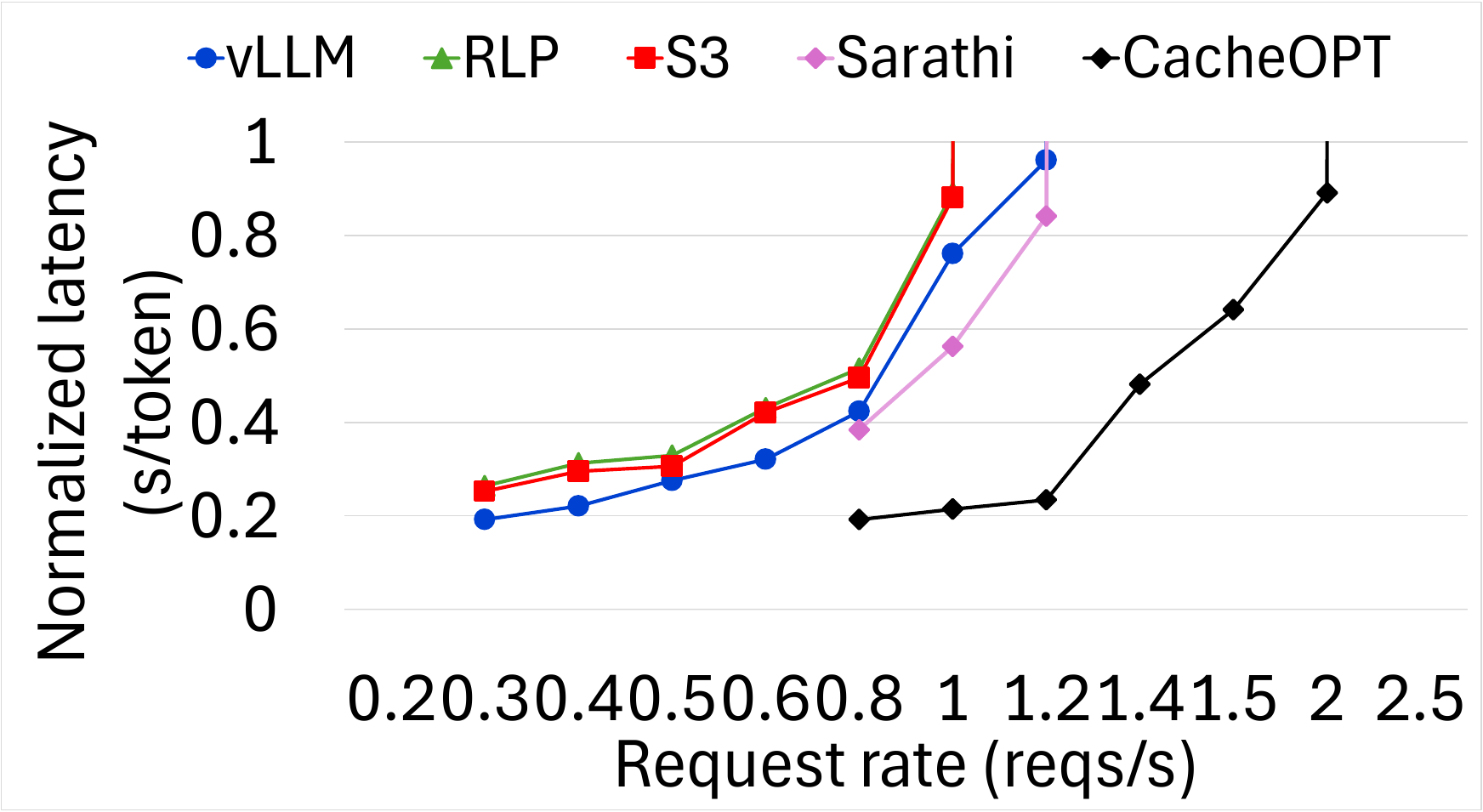} }}
    \hfill
    \DEL{\subfloat[TTFT for different arrival rates for OPT-13B.\vspace{-0.01in}\label{fig:exp-arr-13}]{{\includegraphics[width=0.32\linewidth,height=0.13\textheight]{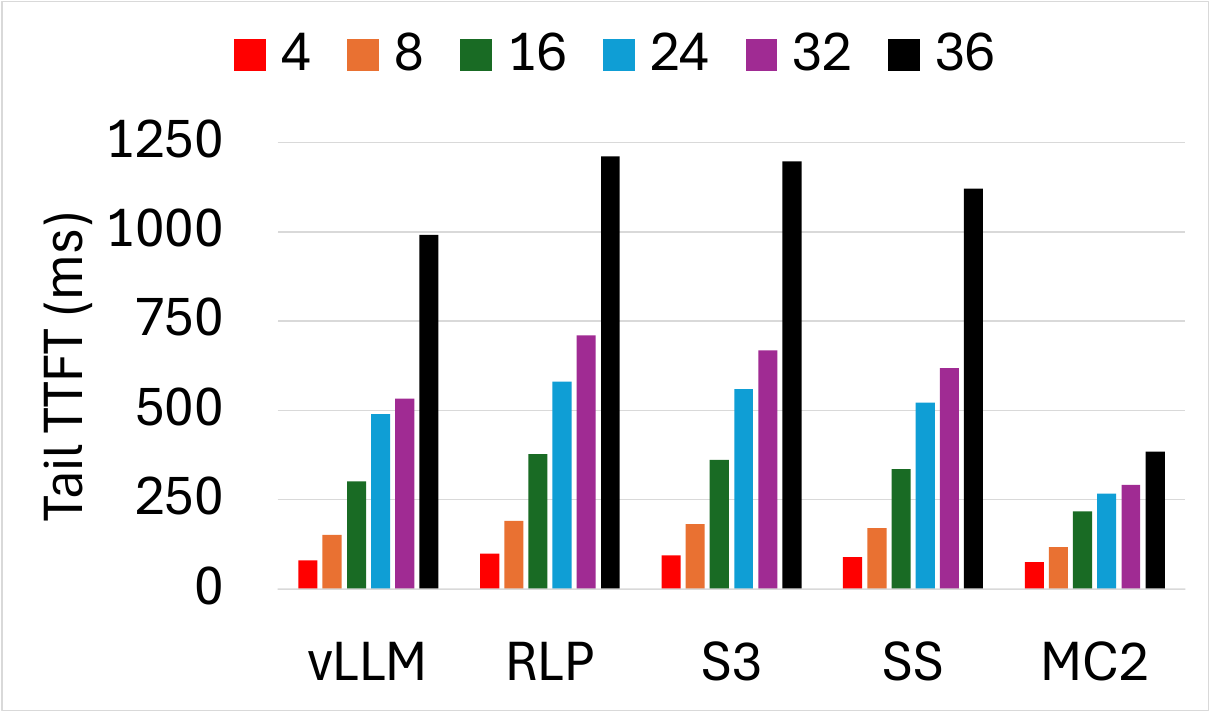} }}
    \hfill
    \subfloat[TBT for different arrival rates for OPT-13B.\vspace{-0.01in}\label{fig:exp-arr-tbt-13}]{{\includegraphics[width=0.32\linewidth,height=0.13\textheight]{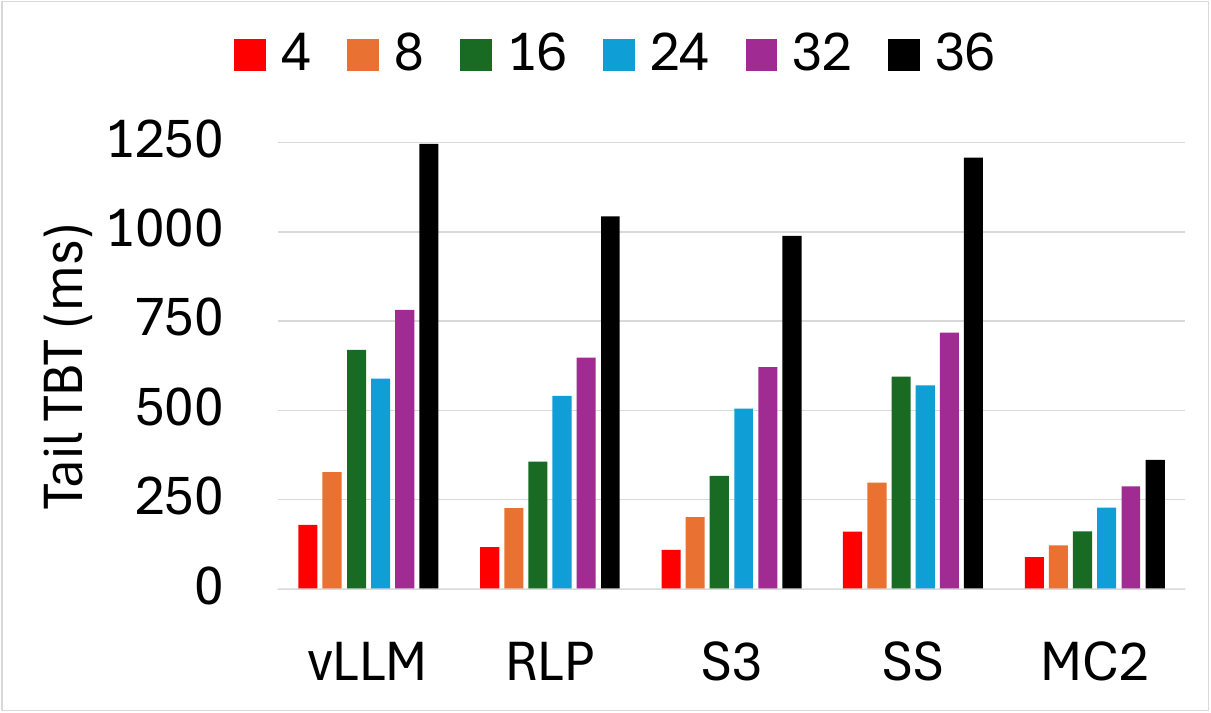} }}
    \hfill}
   \caption{\small{End-to-end latency performance for OPT-13B. \vspace{-0.0in}}}%
    \label{fig:overall-13b}
\end{figure*}

\begin{figure*}[t]
\centering
     \subfloat[Tail TTFT\vspace{-0.01in}\label{fig:exp-tfft-175}]{{\includegraphics[width=0.24\linewidth,height=0.13\textheight]{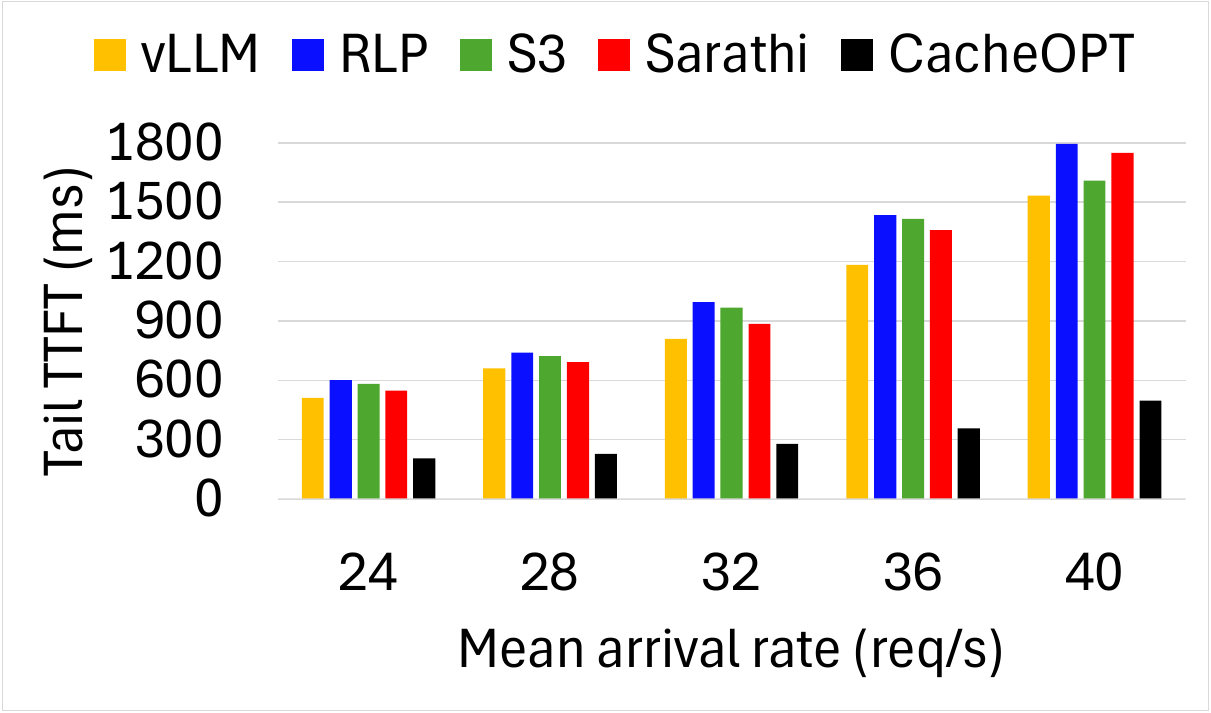} }}
    \hfill
    \subfloat[TTFT SLO attainment\vspace{-0.01in}\label{fig:exp-ttft-slo-175}]{{\includegraphics[width=0.24\linewidth,height=0.13\textheight]{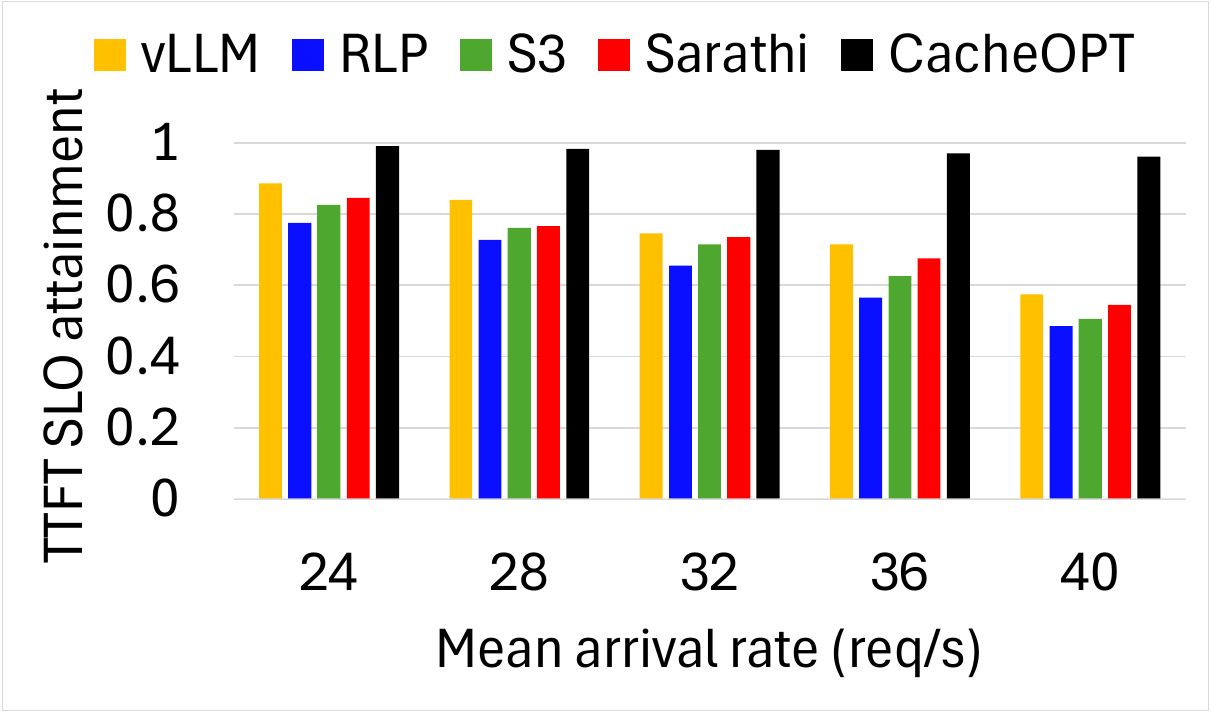} }}
    \hfill
    \subfloat[Tail TBT\vspace{-0.01in}\label{fig:exp-tbt-175}]{{\includegraphics[width=0.24\linewidth,height=0.13\textheight]{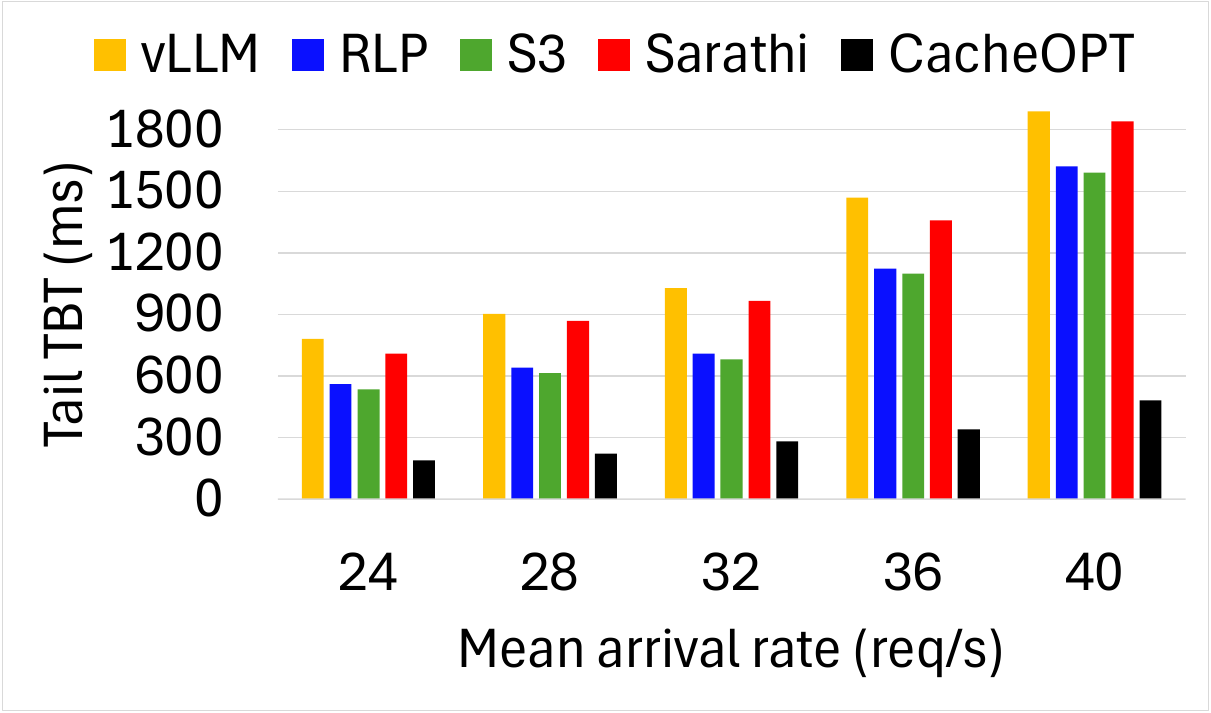} }}
    \hfill
    \subfloat[TBT SLO attainment\vspace{-0.01in}\label{fig:exp-tbt-slo-175}]{{\includegraphics[width=0.24\linewidth,height=0.13\textheight]{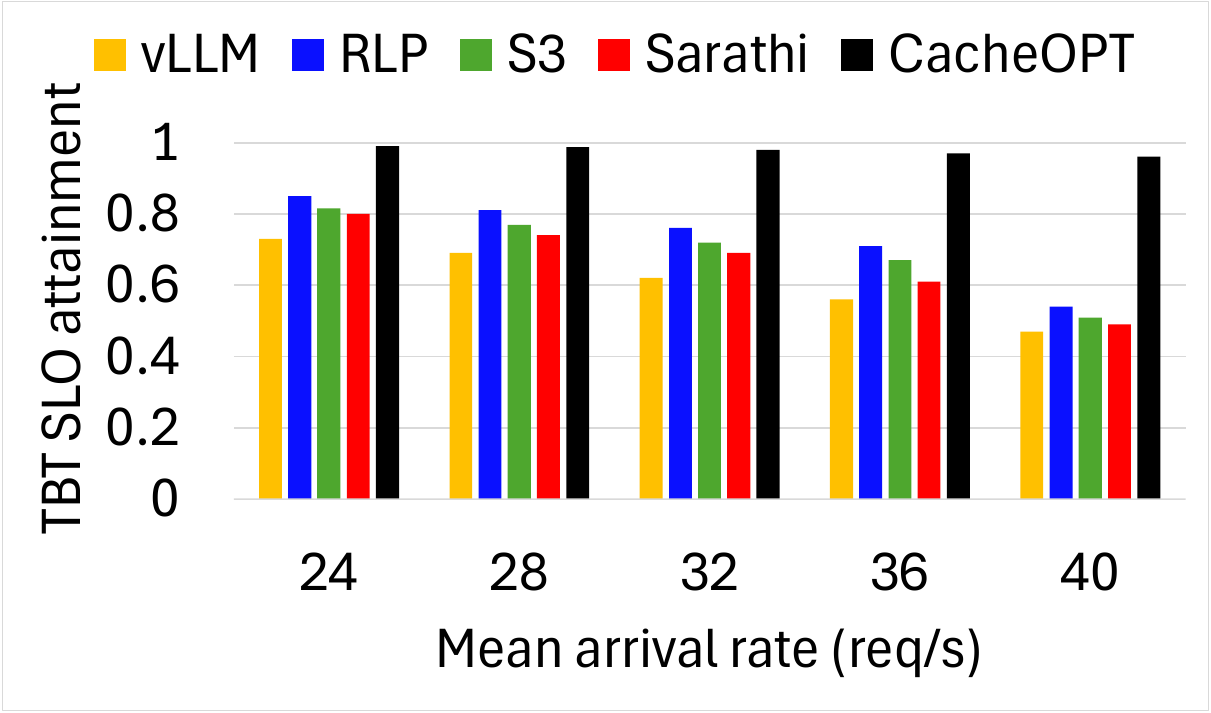} }}
    \hfill
\subfloat[Alpaca\vspace{-0.01in}\label{fig:alpaca-175}]{{\includegraphics[width=0.32\linewidth,height=0.13\textheight]{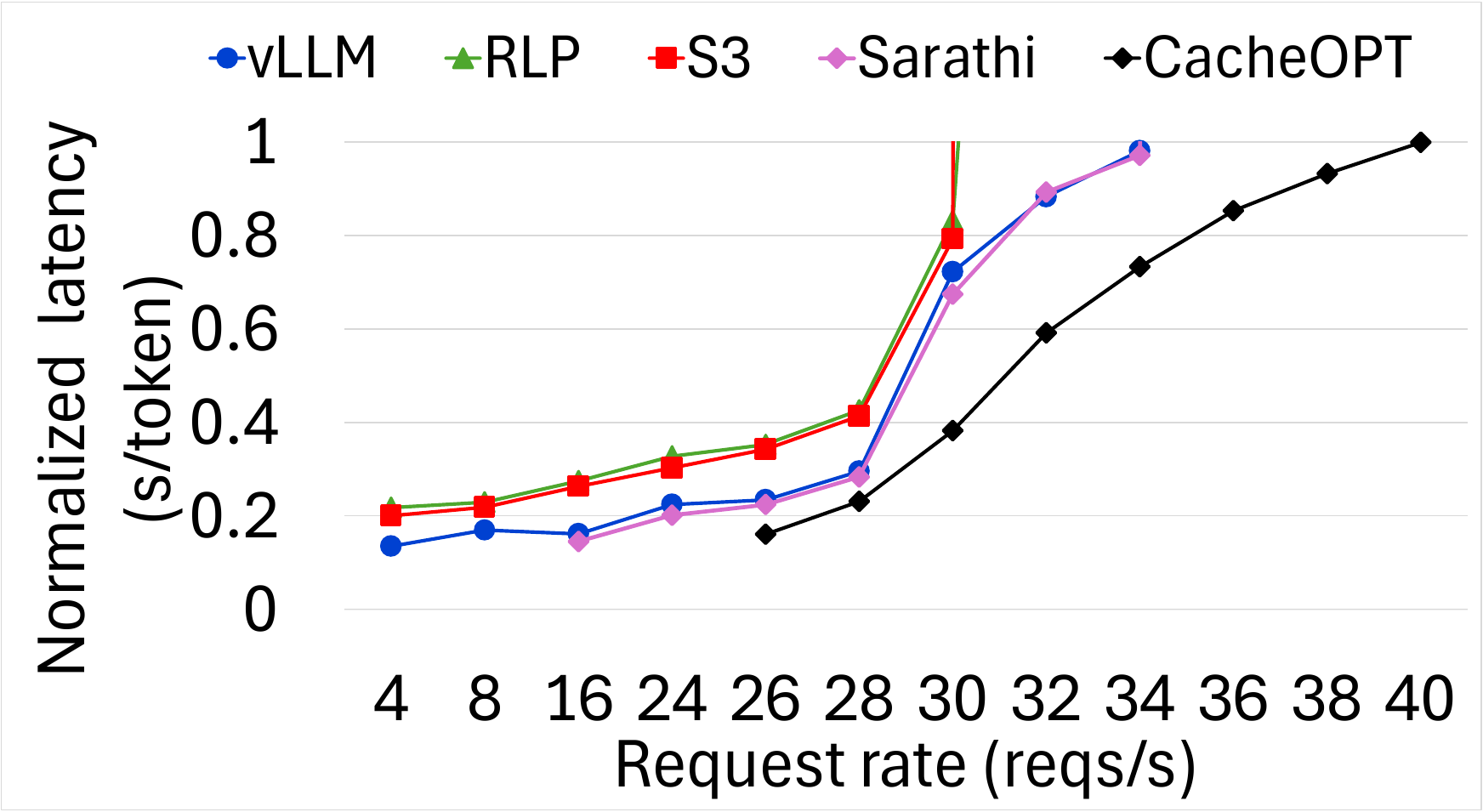} }}
    \hfill
\subfloat[ShareGPT\vspace{-0.01in}\label{fig:sha-175-e}]{{\includegraphics[width=0.32\linewidth,height=0.13\textheight]{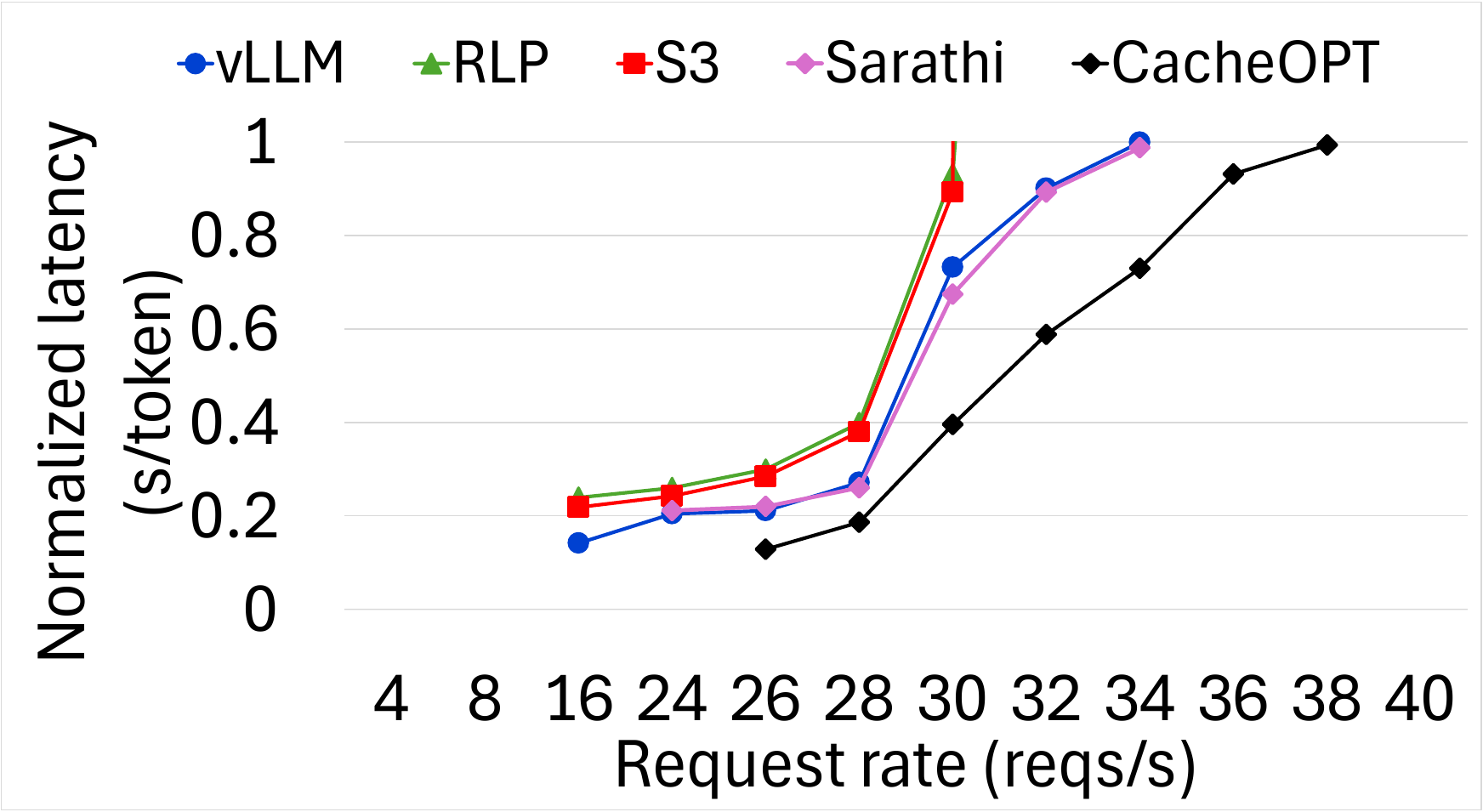} }}
    \hfill
\subfloat[Bookcorpus\vspace{-0.01in}\label{fig:book-175-e}]{{\includegraphics[width=0.32\linewidth,height=0.13\textheight]{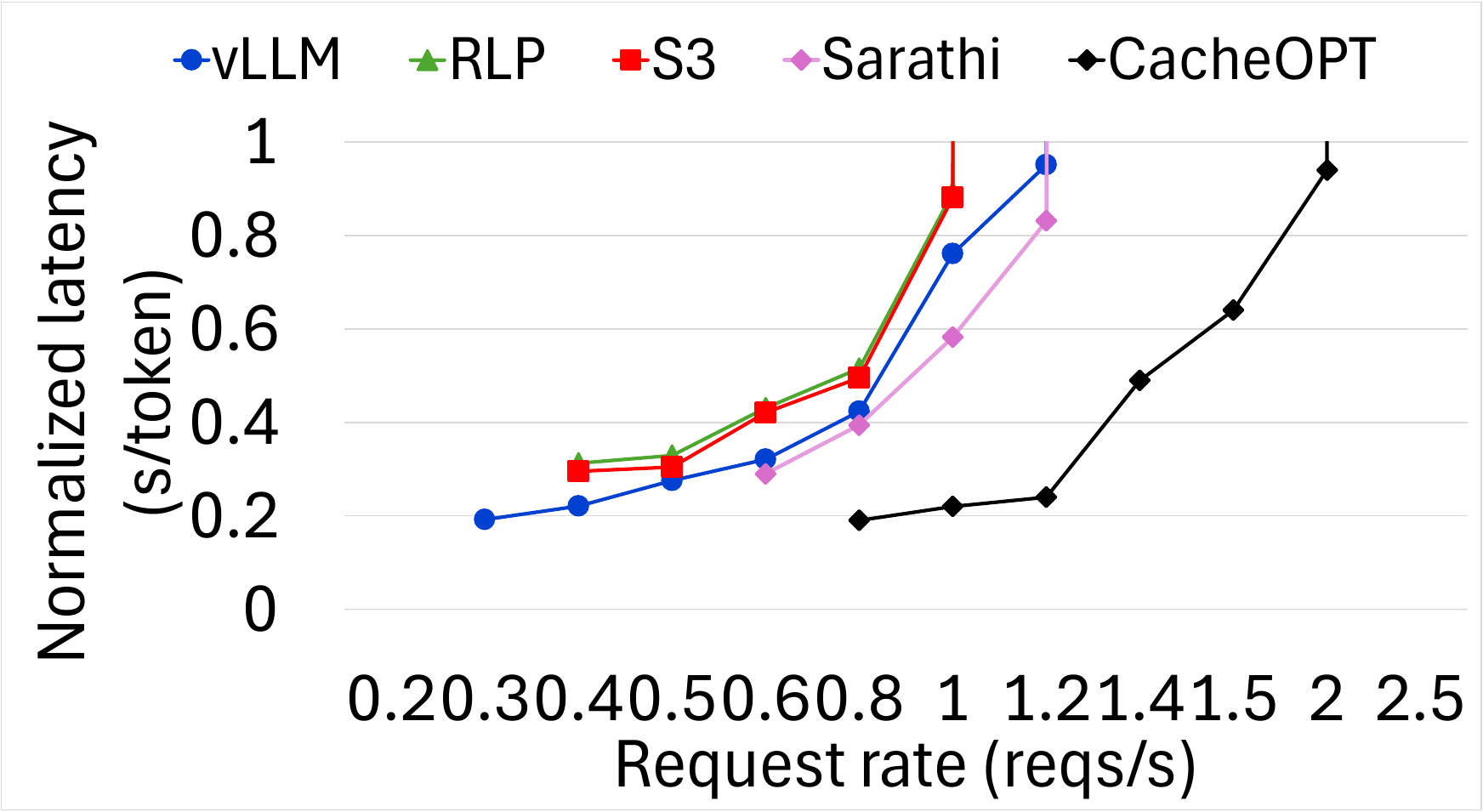} }}
    \hfill
   \caption{\small{End-to-end latency performance for OPT-175B.\vspace{-0.0in}}}%
    \label{fig:overall-175b}
\end{figure*}

\DEL{\begin{figure*}[t]
\centering
     \subfloat[Tail TTFT\vspace{-0.01in}\label{fig:ttft-llama-8b}]{{\includegraphics[width=0.24\linewidth,height=0.13\textheight]{Padding-FIgs/ttft-llama-8b.pdf} }}
    \hfill
    \DEL{\subfloat[TTFT.\vspace{-0.01in}\label{fig:exp-tfft-175}]{{\includegraphics[width=0.32\linewidth,height=0.13\textheight]{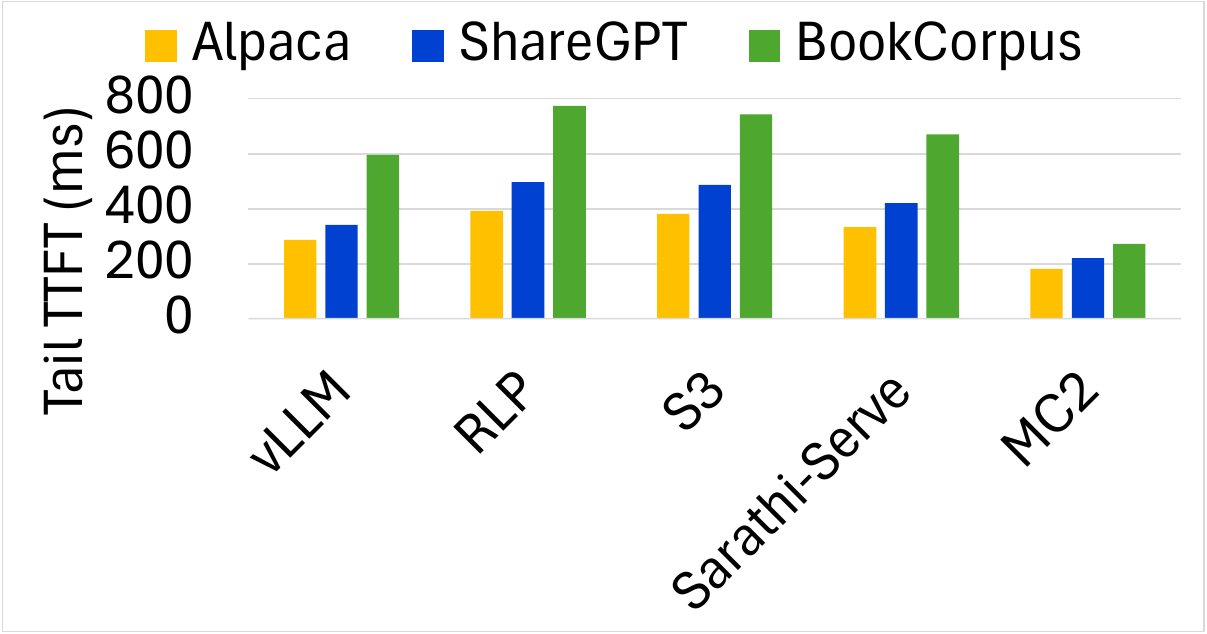} }}
    \hfill}
    \subfloat[TTFT SLO attainment\vspace{-0.01in}\label{fig:ttft-slo-llama-8b}]{{\includegraphics[width=0.24\linewidth,height=0.13\textheight]{Padding-FIgs/ttft-slo-llama-8b.pdf} }}
    \hfill
    \subfloat[Tail TBT\vspace{-0.01in}\label{fig:tbt-llama-8b}]{{\includegraphics[width=0.24\linewidth,height=0.13\textheight]{Padding-FIgs/tbt-llama-8b.pdf} }}
    \hfill
    \DEL{\subfloat[TBT.\vspace{-0.01in}\label{fig:exp-tbt-175}]{{\includegraphics[width=0.32\linewidth,height=0.13\textheight]{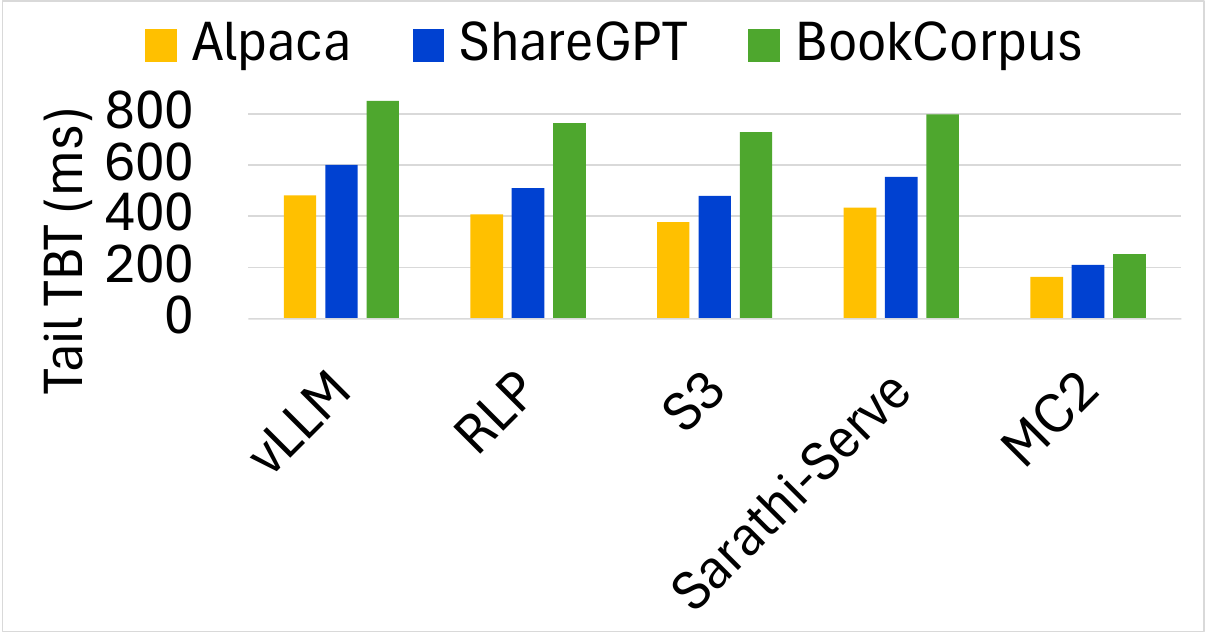} }}
    \hfill}
    \subfloat[TBT SLO attainment\vspace{-0.01in}\label{fig:tbt-slo-8b}]{{\includegraphics[width=0.24\linewidth,height=0.13\textheight]{Padding-FIgs/tbt-slo-attainment-llama-8b.pdf} }}
    \hfill
    \DEL{\subfloat[Overprovisioning rate.\vspace{-0.01in}\label{fig:overprovision-llama-8b}]{{\includegraphics[width=0.32\linewidth,height=0.13\textheight]{Padding-FIgs/overprovisioning-rate-llama-8b.pdf} }}
    \hfill
    \subfloat[Underprovisioning rate.\vspace{-0.01in}\label{fig:under-llama-8b}]{{\includegraphics[width=0.32\linewidth,height=0.13\textheight]{Padding-FIgs/underprovisioning-rate-llama-8b.pdf} }}
    \hfill}
    \subfloat[Alpaca\vspace{-0.01in}\label{fig:alpaca-llama-8b}]{{\includegraphics[width=0.32\linewidth,height=0.13\textheight]{Padding-FIgs/padding-alpaca-llama-8b.pdf} }}
    \hfill
    \subfloat[ShareGPT\vspace{-0.01in}\label{fig:sha-llama-8b}]{{\includegraphics[width=0.32\linewidth,height=0.13\textheight]{Padding-FIgs/padding-sharegpt-llama-8b.pdf} }}
    \hfill
    \subfloat[Bookcorpus\vspace{-0.01in}\label{fig:book-llama-8b}]{{\includegraphics[width=0.32\linewidth,height=0.13\textheight]{Padding-FIgs/padding-bookcorpus-llama-8b.pdf} }}
    \hfill
    \DEL{\subfloat[TTFT for different arrival rates for OPT-175B.\vspace{-0.01in}\label{fig:exp-arr-175}]{{\includegraphics[width=0.32\linewidth,height=0.13\textheight]{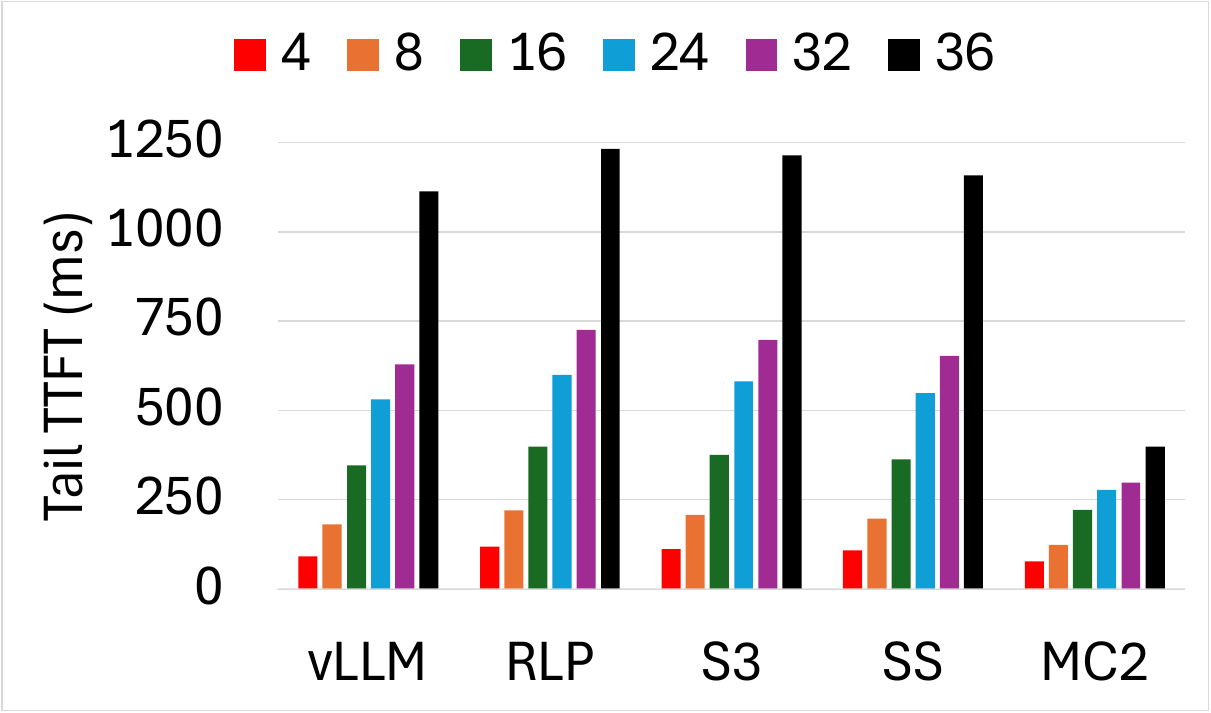} }}
    \hfill
    \subfloat[TBT for different arrival rates for OPT-175B.\vspace{-0.01in}\label{fig:exp-arr-tbt-175}]{{\includegraphics[width=0.32\linewidth,height=0.13\textheight]{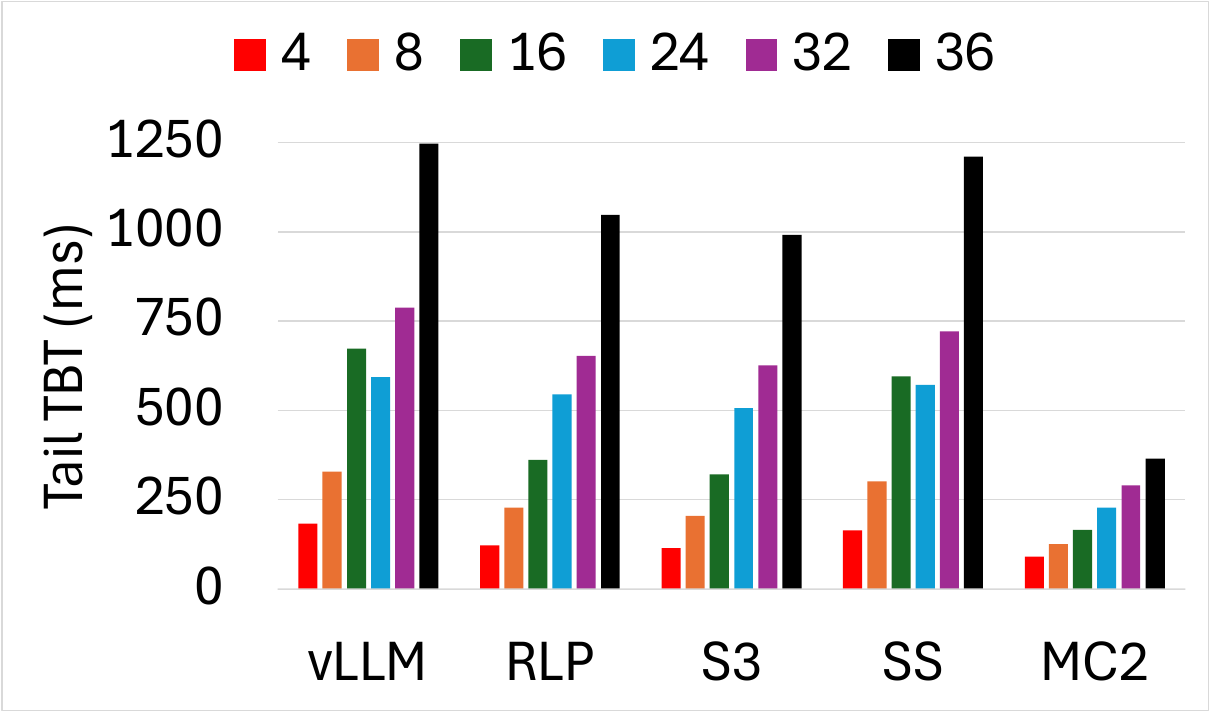} }}
    \hfill}
   \caption{\small{End-to-end latency performance for Llama-3-8B.\vspace{-0.0in}}}%
    \label{fig:overall-llama-8b}
\end{figure*}}

\DEL{\begin{figure*}[t]
\centering
     \subfloat[Tail TTFT\vspace{-0.01in}\label{fig:ttft-llama-70b}]{{\includegraphics[width=0.24\linewidth,height=0.13\textheight]{Padding-FIgs/ttft-llama-70b.pdf}}}
    \hfill
    \DEL{\subfloat[TTFT.\vspace{-0.01in}\label{fig:exp-tfft-175}]{{\includegraphics[width=0.24\linewidth,height=0.13\textheight]{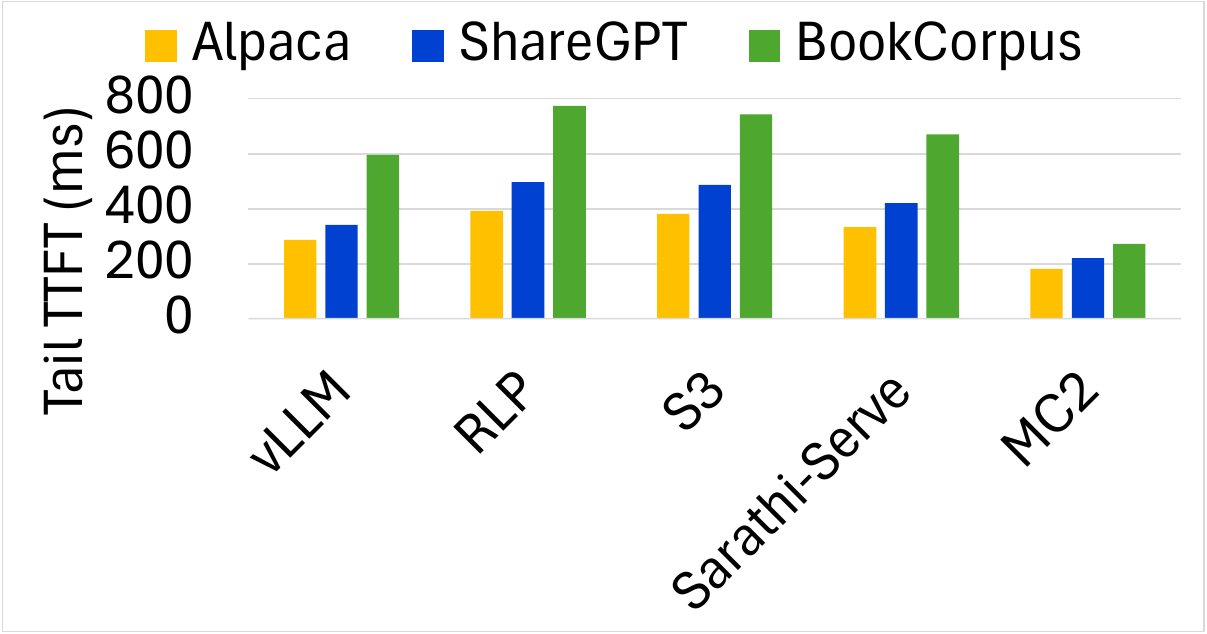} }}
    \hfill}
    \subfloat[TTFT SLO attainment\vspace{-0.01in}\label{fig:ttft-slo-llama-70b}]{{\includegraphics[width=0.24\linewidth,height=0.13\textheight]{Padding-FIgs/ttft-slo-attainment-llama-70b.pdf} }}
    \hfill
    \subfloat[Tail TBT\vspace{-0.01in}\label{fig:tbt-llama-70b}]{{\includegraphics[width=0.24\linewidth,height=0.13\textheight]{Padding-FIgs/tbt-llama-70b.pdf} }}
    \hfill
    \DEL{\subfloat[TBT.\vspace{-0.01in}\label{fig:exp-tbt-175}]{{\includegraphics[width=0.32\linewidth,height=0.13\textheight]{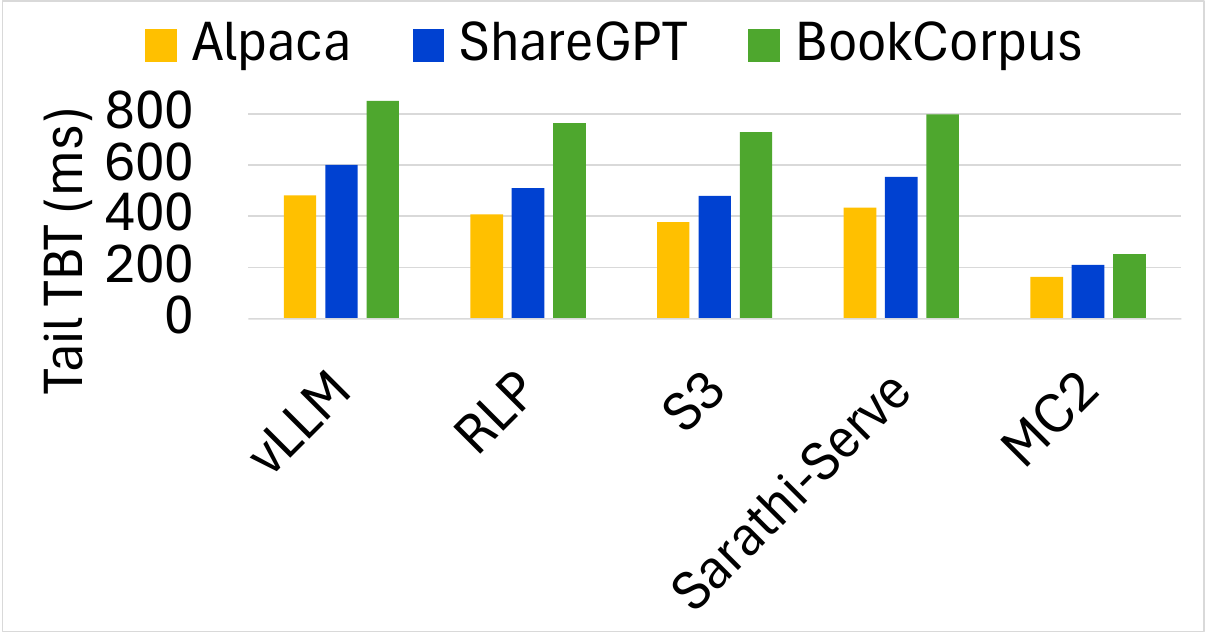} }}
    \hfill}
    \subfloat[TBT SLO attainment\vspace{-0.01in}\label{fig:tbt-slo-70b}]{{\includegraphics[width=0.24\linewidth,height=0.13\textheight]{Padding-FIgs/TBT-SLO-attainment-Llama-70b.pdf} }}
    \hfill
    \DEL{\subfloat[Overprovisioning rate.\vspace{-0.01in}\label{fig:overprovision-llama-70b}]{{\includegraphics[width=0.32\linewidth,height=0.13\textheight]{Padding-FIgs/overprovisioning-rate-llama-8b.pdf} }}
    \hfill
    \subfloat[Underprovisioning rate.\vspace{-0.01in}\label{fig:under-llama-70}]{{\includegraphics[width=0.32\linewidth,height=0.13\textheight]{Padding-FIgs/underprovisioning-rate-llama-8b.pdf} }}
    \hfill}
    \subfloat[Alpaca\vspace{-0.01in}\label{fig:alpaca-llama-70b}]{{\includegraphics[width=0.32\linewidth,height=0.13\textheight]{Padding-FIgs/padding-alpaca-llama-70b.pdf} }}
    \hfill
    \subfloat[ShareGPT\vspace{-0.01in}\label{fig:sha-llama-70b}]{{\includegraphics[width=0.32\linewidth,height=0.13\textheight]{Padding-FIgs/padding-sharegpt-llama-70b.pdf} }}
    \hfill
    \subfloat[Bookcorpus\vspace{-0.01in}\label{fig:book-llama-70b}]{{\includegraphics[width=0.32\linewidth,height=0.13\textheight]{Padding-FIgs/padding-bookcorpus-llama-70b.pdf} }}
    \hfill
    \DEL{\subfloat[TTFT for different arrival rates for OPT-175B.\vspace{-0.01in}\label{fig:exp-arr-175}]{{\includegraphics[width=0.32\linewidth,height=0.13\textheight]{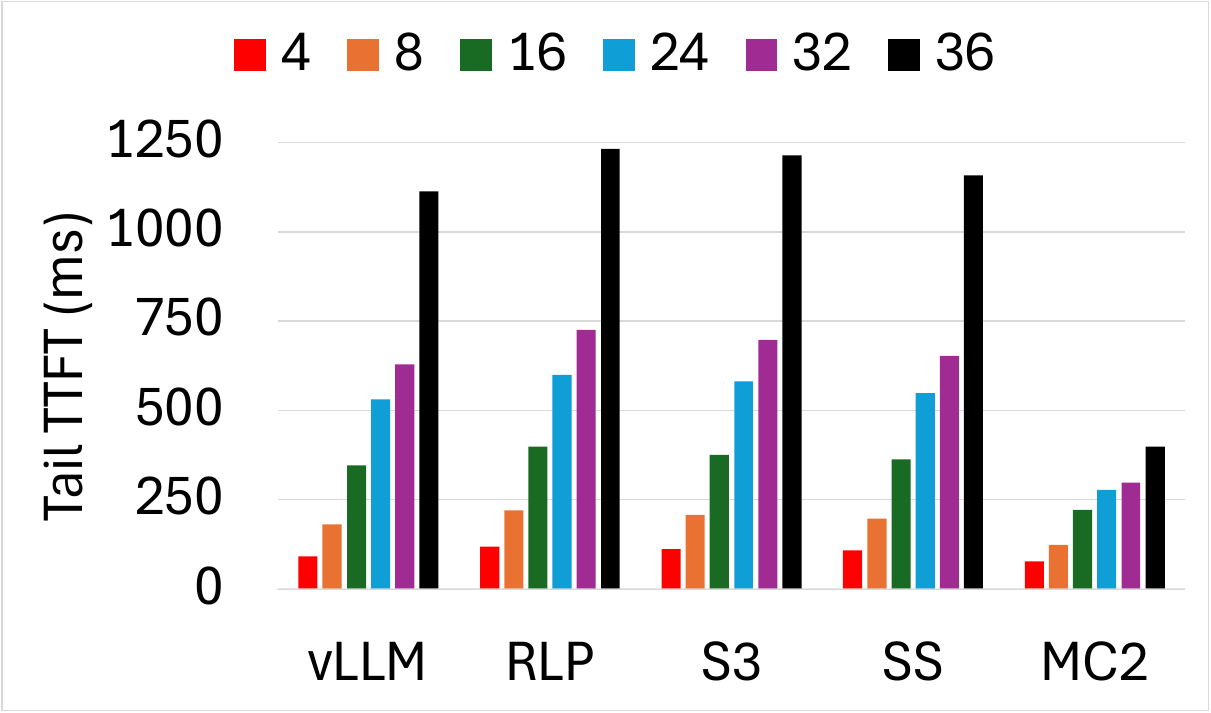} }}
    \hfill
    \subfloat[TBT for different arrival rates for OPT-175B.\vspace{-0.01in}\label{fig:exp-arr-tbt-175}]{{\includegraphics[width=0.32\linewidth,height=0.13\textheight]{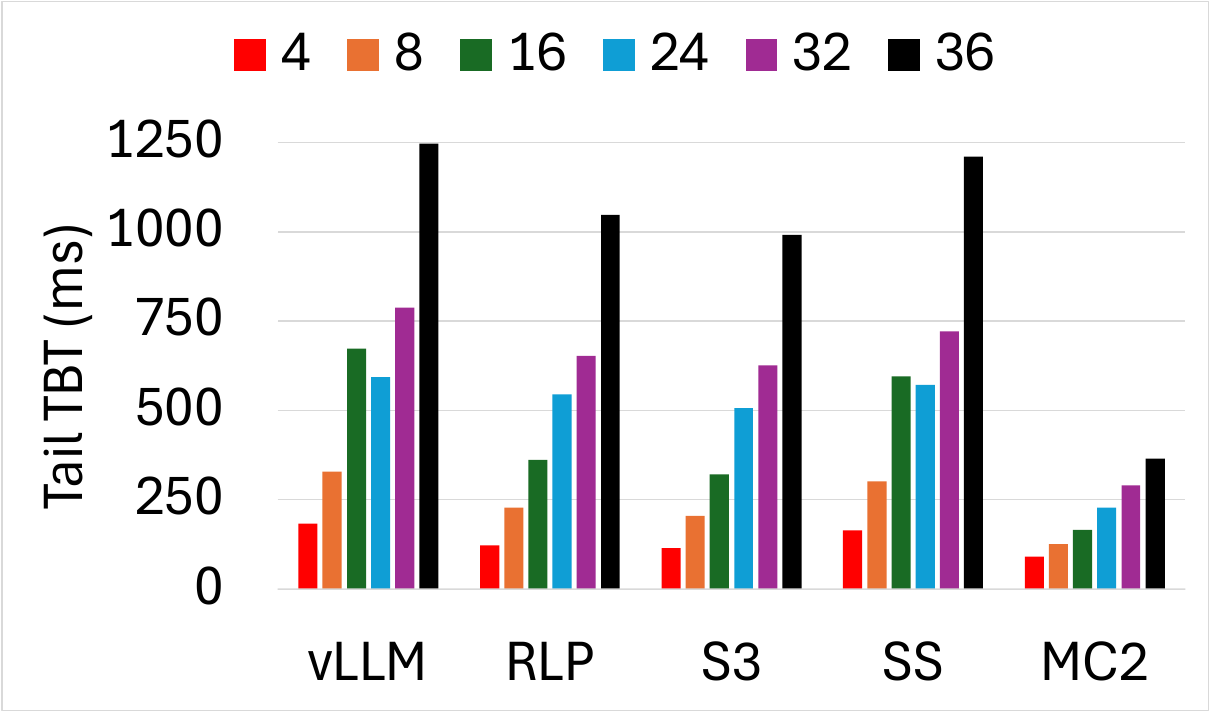} }}
    \hfill}
   \caption{\small{End-to-end latency performance for Llama-3-70B.\vspace{-0.0in}}}%
    \label{fig:overall-llama-70b}
\end{figure*}}

\DEL{\noindent{\textbf{Metrics.}}
We primarily measure the normalized latency of the system, which is the mean of every request's end-to-end latency divided by its output length~\cite{280922,vllm}. We also measure the severity of SLO violation (SVS), which defines the difference between the allocated KVC and the request's KVC demand. Finally, we report the scheduling overhead of all the methods. As part of the ablation study, we report the JCT and timeb between tokens (TBT) alongside the throughput (requests/seconds). }


\subsection{Overall Performance Comparison}

\noindent{\textbf{TTFT and TBT.}} Figures~\ref{fig:exp-tfft}, and ~\ref{fig:exp-tfft-175}illustrate the \textit{tail TTFT} during the entire running time versus the mean request arrival rate for different models. Figures~\ref{fig:exp-ttft-slo}, and~\ref{fig:exp-ttft-slo-175},show the \textit{TTFT SLO attainment}, which indicates the fraction of requests that meet their TTFT SLOs. We observe that the tail TTFT keeps increasing and the TTFT SLO attainment keep decreasing as the arrival rate increases. While the compared methods exhibit abrupt changes, \sys demonstrates a steady progression. \sys reduces the tail TTFT of vLLM, RLP, $S^3$, and Sarathi by 64\%-2.34$\times$, 1.21$\times$-2.83$\times$, 1.11$\times$-2.73$\times$, and 97\%-2.56$\times$ across all arrival rates and models, respectively. Following a similar TTFT trend, \sys achieves 0.94-0.97 TTFT SLO attainment, with the highest improvement of 47\% over Sarathi, the best compared system, at 40 req/s.
This improvement stems from advanced methods that mitigate KVC bottlenecks to enable more requests per batch while reducing both the number and duration of preemptions, ultimately decreasing waiting time. The \emph{confidence-based padding} method reduces preemptions, while the \emph{embedding method} mitigate the KVC bottleneck. \emph{Request selection and KVC allocation} tries to limit the TTFT. \textit{Proactive KVC allocation} and \textit{global KVC reservation} further reduce the preemptions. 
Additionally, \sys's advanced preemption policy 
reduces the number and duration of preemptions. Further, \emph{preemption strategy selection} minimizes swapping or recomputation time, allowing waiting requests to be scheduled earlier and reducing TTFT. Overall, the improvement for large models is smaller compared to small models due to reduced KVC competition when using more GPUs.


\DEL{\sh{double check and show me the statistics for the 2 groups-done} because of less KVC competition due to using \sh{how many-done} \tsr{8 GPUs and 4 GPUs for OPT-175B and Llama-3-70B, while both OPT-13B and Llama-3-8B use one GPU.}.}

\DEL{Unless otherwise specified, we discuss the average results across the three datasets.
Across the different models,  
\sys reduces the tail TTFT of vLLM, RLP, $S^3$, and Sarathi by 58\%-1.19$\times$, 1.16-1.86$\times$, 1.10-1.73$\times$, and 84\%-1.47$\times$\sh{update the statistics}.}
\DEL{Overall, the improvement on the large models is diminished compared to the small models\sh{double check and show me the statistics for the 2 groups-done, mentioned above } because of less KVC competition due to using \sh{how many-done} \tsr{8 GPUs and 4 GPUs for OPT-175B and Llama-3-70B, while both OPT-13B and Llama-3-8B use one GPU.}. }

\DEL{This translates\sh{how did you "translates"?} to absolute reductions of approximately 250ms, 400ms, 382ms, and 290ms for vLLM, RLP, $S^3$, and Sarathi, respectively. \tsr{We measure these values by comparing the difference between the average tail TTFT of \sys with other methods.} 
Notably, RLP and $S^3$ exhibit comparable tail TTFT values, with 554 ms and 536 ms on average\sh{"exceeding" what?-done}, respectively, primarily due to their reliance on prediction based KV cache allocation strategies, which limits the number of requests that can be accommodated in a batch and increases the waiting time. vLLM and Sarathi shows lower TTFT than RLP and $S^3$, reducing tail TTFT to approximately 408 ms and 475 ms\sh{how did you get these values} on average due to the use of block based KVC allocation strategy, which enables a batch to accommodate more requests and decrease waiting time. }

\DEL{\sys demonstrates the lowest tail TTFT, achieving a latency of 208 ms\sh{where did you get this value}.} 

\DEL{\sys continuously shows better performance for the increasing arrival rates in terms of Tail TTFT Compared to Sarathi\sh{is it the best performance one? if yes, indicate it}, for the six arrival rates from 4 reqs/s to 36 reqs/s, \sys reduces the tail TTFT by 23\%, 35\%, 1.59$\times$, 2.01$\times$, and 2.12$\times$, and 2.93$\times$, respectively.} 

\DEL{The results demonstrate that \sys achieves the highest SLO attainments with 0.96 for both models. In comparison, vLLM, RLP, $S^3$, and Sarathi exhibit TTFT SLO attainments below 0.7, highlighting their limitations in meeting TTFT SLO requirements. 
\sys shows 40-47\%, 28-41\%, 25-39\% and 35-41\% better performance compared to vLLM, RLP, $S^3$ and Sarathi, respectively on average for all the models. 
The superior performance of \sys can be attributed to its advanced strategies explaiend above.  
These strategies ensure high-priority requests consistently meet their SLOs, even under variable workloads, thereby significantly improving system responsiveness.}

\DEL{\tsr{With the increasing arrival rate, \sys shows increasing higher TTFT SLO attainments compared to the other methods. For the arrival rate of 4 and 8, \sys shows approximately 1.8\% and 5\% higher TTFT SLO attainments compared to Sarathi. However, for arrival rate of 16, 24, 32, and 36, \sys shows 14\%, 25\$, 41\% and 46\% higher TTFT SLO attainments, respectively.}


\sh{In the end of each metric's discussion, explain when the arrival rate increases, how the improvement increases.-done}}

Figures~\ref{fig:exp-tbt},and~\ref{fig:exp-tbt-175} illustrate the \emph{tail TBT} during the entire experiment.  Figures~\ref{fig:exp-tbt-slo}, and~\ref{fig:exp-tbt-slo-175} show the \textit{TBT SLO attainment}, which indicates the fraction of requests whose all iterations meet their TBT SLOs. Similarly, as the arrival rate increases, while the compared methods exhibit abrupt changes, \sys demonstrates a steady progression.
\sys achieves the lowest tail TBT among all systems. Specifically, \sys reduces the tail TBT of vLLM, RLP, $S^3$, and Sarathi by 1.82-3.29$\times$, 1.42-2.37$\times$, 1.31-2.30$\times$, and 1.47-2.71$\times$ across all arrival rates and models. 
\sys achieves 0.94-0.98 TBT SLO attainment, outperforming Sarathi by 53\% at 40 req/s.
The superior performance of \sys can be attributed to its advanced strategies explained above. Mitigating KVC competition also reduces preemptions. The reduction of the number and duration of preemptions directly reduces TBT and avoid TBT SLO violations. Considering TBT SLO in selecting requests in batching and for preemptions further reduce TBT and avoid TBT SLO violations. 

\DEL{For OPT-175B, \sys reduces their tail TBT by approximately 1.93-2.37$\times$, 1.43-2.03$\times$, 1.29-1.88$\times$, and 1.63-2.16$\times$, respectively. \tsr{\sys reduces the tail TBT of vLLM, RLP, $S^3$, and Sarathi by approximately 2.35-2.51$\times$, 1.69-2.13$\times$, 1.47-1.95$\times$, and 1.83-2.1$\times$ for Llama-3-8B. For Llama-3-70B, \sys reduces their tail TBT by approximately 1.91-2.32$\times$, 1.48-2.07$\times$, 1.33-1.95$\times$, and 1.68-2.19$\times$, respectively.} RLP and $S^3$ exhibit tail TBT of 560ms and 527 ms, respectively, vLLM and Sarathi show higher tail TBT of approximately 643ms and 594ms, respectively due to preemptions. In contrast, \sys achieves a latency of 203ms on average, underscoring its effectiveness in addressing tail latency challenges.
Such improvement of \sys over other methods is observed due to reducing the preemption time as per the reasons mentioned above.}

\DEL{\tsr{\sys continuously shows better performance for the increasing arrival rates in terms of Tail TBT Compared to Sarathi, for arrival rate 4 reqs/s and 36 reqs/s, \sys reduces the tail TTFT by 49\% and 3.39$\times$, respectively. For other request rates of 8, 16,24, and 32 \sys reduces the tail TTFT by 63\%, 2.74$\times$, 2.85$\times$, and 3.03$\times$, respectively.}}

\DEL{The improvement achieved by \sys can be attributed to its confidence-aware token padding and proactive KV cache allocation strategies. These innovations enable \sys to minimize both overprovisioning and underprovisioning, resulting in better resource utilization and reduced tail latency. Furthermore, \sys leverages an advanced preemption mechanism to dynamically adjust resource allocations, ensuring that high-priority requests meet their Service Level Objectives (SLOs).}

\DEL{\noindent{\textbf{SLO attainment.} }Figures~\ref{fig:exp-ttft-slo}, and~\ref{fig:exp-ttft-slo-175} show the \textit{TTFT SLO attainment} the OPT-13B, OPT-175B, Llama-3-8B, and Llama-3-175B model, respectively. The TTFT SLO attainment indicates the fraction of requests that meet their TTFT SLOs.
The results demonstrate that \sys achieves the highest SLO attainments exceeding 0.95 for both models. In comparison, vLLM, RLP, $S^3$, and Sarathi exhibit TTFT SLO attainments below 0.7, highlighting their limitations in meeting TTFT SLO requirements. 
\sys shows 30-41\%, 42-49\%, 36-44\% and 26-32\% better performance compared to vLLM, RLP, $S^3$ and Sarathi, respectively on average for both the models. 
The superior performance of \sys can be attributed to its advanced strategies explaiend above. 
These strategies ensure high-priority requests consistently meet their SLOs, even under variable workloads, thereby significantly improving system responsiveness.}


\DEL{Figures~\ref{fig:exp-tbt-slo},~\ref{fig:exp-tbt-slo-175},~\ref{fig:tbt-slo-8b}, and~\ref{fig:tbt-slo-70b} show the \textit{TBT SLO attainment} for the OPT-13B, OPT-175B model, Llama-3-8B, and Llama-3-70B models, respectively. The TBT SLO attainment indicates the fraction of requests whose all iterations that meet their TBT SLOs.}
\DEL{\sys again achieves the highest TBT SLO attainments of 0.95 for both models. Compared to vLLM, RLP, $S^3$ and Sarathi, \sys increases 37-44\%, 26-38\%, 30-35\% and 29-36\% higher TBT SLO attainment, for both the models.} 

\DEL{\tsr{With the increasing arrival rate, \sys shows increasing higher TBT SLO attainments compared to the other methods. For the arrival rate of 4, 8, 16, 24, 32, and 36 \sys shows approximately 3\%, 7\%, 14\%, 26\%, 38\% and 47\% higher TBT SLO attainemnts compared to Sarathi, respectively.}}

\DEL{its proactive and confidence-aware resource allocation strategies, which dynamically adjust KV cache allocation and minimize contention. These strategies ensure high-priority requests consistently meet their SLOs, even under variable workloads, thereby significantly improving system reliability and responsiveness.}

\DEL{\noindent{\textbf{Overprovisioning and underprovisioning.}} Figures~\ref{fig:exp-ov},~\ref{fig:exp-ov-175},~\ref{fig:overprovision-llama-8b}, and~\ref{fig:overprovision-llama-70b} illustrate the average \textit{overprovisioning rate} for the OPT-13B, OPT-175B, Llama-3-8B, and Llama-8-70B models, respectively. The overprovisioning rate represents the overprovisioned KVC beyond a request's actual usage expressed as the percentage of the total KVC. 
The results show that \sys achieves the lowest overprovisioning rates among all methods. Specifically, \sys reduces overprovisioning to less than 3\%, significantly outperforming other methods: approximately 5\% for vLLM, 18\% for RLP, 15\% for $S^3$, and 5.17\% for Sarathi, averaged across both models.
\sh{again, Sarathi's and vLLM's overprovisioning won't exceed 32 tokens.-done, reduced the valus-done}

\tsr{Compared to Sarathi, \sys shows 2.54\%, 2.61\%, 2.72\%, 2.82\%, 3.02\%, and 3.34\% lower overprovisioning rate for the request arrival rate of 4, 8, 16, 24, 32, 36, respectively. The overprovisoning improvement compared to the Sarathi appears to be low because Sarathi's overprovisioning don't exceed 32 tokens.}


Figures~\ref{fig:exp-under},~\ref{fig:exp-under-175},~\ref{fig:under-llama-8b}, and~\ref{fig:under-llama-70b} show the average \textit{underprovisioning rate} for the OPT-13B, OPT-175B, Llama-3-8B, and Llama-3-175B models, respectively. The underprovisioning rate reflects the underprovisioned KVC below the request's actual usage, expressed as the percentage of total KVC at the end of the final generation. For vLLM, we consider how many tokens were additionally allocated during all the preemptions of a request. 
The results show that \sys achieves the lowest underprovisioning rates among all methods. Specifically, \sys reduces overprovisioning to less than 17\%, compared to approximately 35\% for vLLM, 30\% for RLP, 28\% for $S^3$ and 32\% for Sarathi on average for both models. 
This superior performance can be attributed to \sys's proactive KV cache allocation strategies as mentioned above. 

\tsr{Compared to Sarathi, \sys shows 54\%, 59\%, 67\%, 71\%, 77\%, and 81\% lower underprovisioning rate for the request arrival rate of 4, 8, 16, 24, 32, 36, respectively.}}

\noindent{\textbf{Normalized latency.}} \emph{Normalized latency} is a request's end-to-end latency divided by its output length~\cite{280922,vllm}. 
Figures~\ref{fig:alpaca-13}-\ref{fig:book-13-e}, and~\ref{fig:alpaca-175}-\ref{fig:book-175-e}  show the average normalized latency of the systems versus the mean arrival rate. A high-throughput serving system should retain low normalized latency against high request rates. We compare the request rates that the systems can sustain while maintaining similar latencies. \DEL{For ShareGPT, \sys can sustain 1.25-1.75$\times$ higher request rates compared to vLLM, 1.25-2.25$\times$ compared to RLP and $S^3$, 1.13-1.75$\times$ compared to Sarathi, respectively. For BookCorpus, \sys can sustain 1.5-1.8$\times$ higher request rates compared to vLLM and 1.88-2.33$\times$ compared to RLP and $S^3$, and 1.5-1.8$\times$ compared to Sarathi. For Alpaca, \sys can sustain 1.13-2.14$\times$ higher request rates than vLLM, 1.2-1.24$\times$ compared to RLP and $S^3$, and 1.1-1.2$\times$ compared to Sarathi.} 
\sys can sustain 1.29-1.89$\times$ higher arrival rates than vLLM, 1.44-1.94$\times$ than RLP and $S^3$, and 1.24-1.58$\times$ than Sarathi on average for the three datasets.
\sys's advantages on BookCorpus is more pronounced because it contains longer sequences, allowing fewer requests to be batched and generating more KVC competition. 

Similar to the tail TTFT, for other metrics, \sys shows lower improvement for larger models compared to smaller models and as arrival rates increase, \sys demonstrates greater improvements over other systems due to the same reasons.


\DEL{Figure~\ref{fig:exp-arr-13} and~\ref{fig:exp-arr-175} illustrates the TTFT performance of five KV cache allocation methods—vLLM, RLP, $S^3$, Sarathi, and \sys—under varying request arrival rates (4, 8, 16, 24, 32, and 36 requests per second) for both the models. The results show that TTFT increases significantly with higher request arrival rates across all methods. Baseline methods such as vLLM, RLP, and $S^3$ exhibit steep increases, with TTFT \tsr{more than} 1000 ms under the highest load of 36 requests per second. Sarathi achieves moderate improvements, with tail TTFT peaking at around 750 ms. In contrast, \sys demonstrates superior scalability, maintaining tail TTFT well below 400 ms even under the highest arrival rates, which is 2.6$\times$ lower than the Sarathi. This improvement highlights \sys's proactive resource allocation and load management strategies, which ensure timely responses even under heavy workloads.

Figure~\ref{fig:exp-arr-tbt-13} and~\ref{fig:exp-arr-tbt-175} compares the TBT for five methods—vLLM, RLP, $S^3$, Sarathi, and \sys—across varying request arrival rates (4, 8, 16, 24, 32, and 36 requests per second) for both the models. 

The results indicate that TBT rises steeply with increasing request rates for vLLM, RLP, $S^3$, and Sarathi with values exceeding 1000 ms at 36 requests per second.  \sys outperforms all other methods, maintaining tail TBT below 400 ms across all arrival rates, which is 2.75$\times$ lower than the Sarathi. This consistent performance demonstrates \sys's ability to effectively manage resource contention and dynamically allocate KV cache to reduce latency for high-priority requests.}



\DEL{\begin{figure}
    \centering
    \includegraphics[width=0.75\columnwidth,height=0.13\textheight]{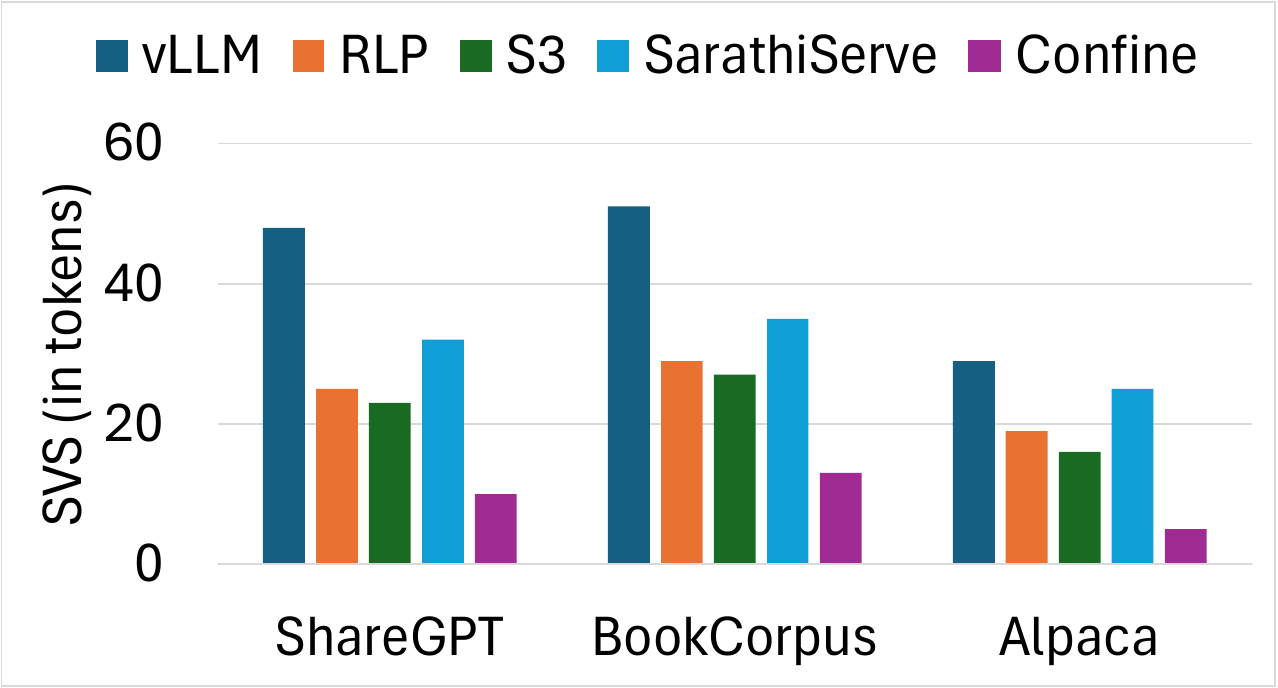}
    \caption{SLO violation severity.}
    \label{fig:svs}
\end{figure}

Figure~\ref{fig:svs} shows the SLO violation severity for the three datasets. Compared to vLLM, RLP,  $S^3$, and Sarathi, \sys shows 78\%, 61\%, 59\%, and 68\% lower SVS, respectively.

\sys outperforms other methods because it can efficiently use the KVC and avoid KVC overprovision and underprovision, as we observe from SVS. Moreover, its eviction method helps release occupied KVC earlier and expedite finding requests to utilize the KVC. In contrast, vLLM generates many KVC underprovisions, and hence in higher latency due to block allocation. RLP and $S^3$ also suffer from  KVC overprovision and underprovision, increasing latency. Sarathi also suffer from certain underprovision and overprovision as built top of vLLM.}

\subsection{Ablation Study}

We tested the following variants of \sys to evaluate the effectiveness of its components.

\squishlist
\item \emph{/CKA}: \sys without \underline{C}onfident-based \underline{K}VC \underline{A}llocation and it adds a fixed padding of 10\% of the predicted output length. 

\item \emph{/RSA:} \sys without \underline{R}equest \underline{S}election and KVC \underline{A}llocation and it uses FCFS for request selction and allocates KVC using RLP. 

\item \emph{/PA:} \sys without \underline{P}roactive \underline{K}VC allocation.

\item \emph{/GR:} \sys without \underline{G}lobal \underline{R}eservation 

\item \emph{/PP:} \sys without \underline{P}reemption \underline{P}olicy and it follows FCFS. 

\item \emph{/PSS:}  \sys without
\underline{P}reemption \underline{S}trategy \underline{S}election and its uses the default recomputation in the vLLM code.
\squishend



\begin{figure*}[t]
\centering
     \subfloat[Tail TTFT\vspace{-0.01in}\label{fig:ablation-ttft}]{{\includegraphics[width=0.24\linewidth,height=0.13\textheight]{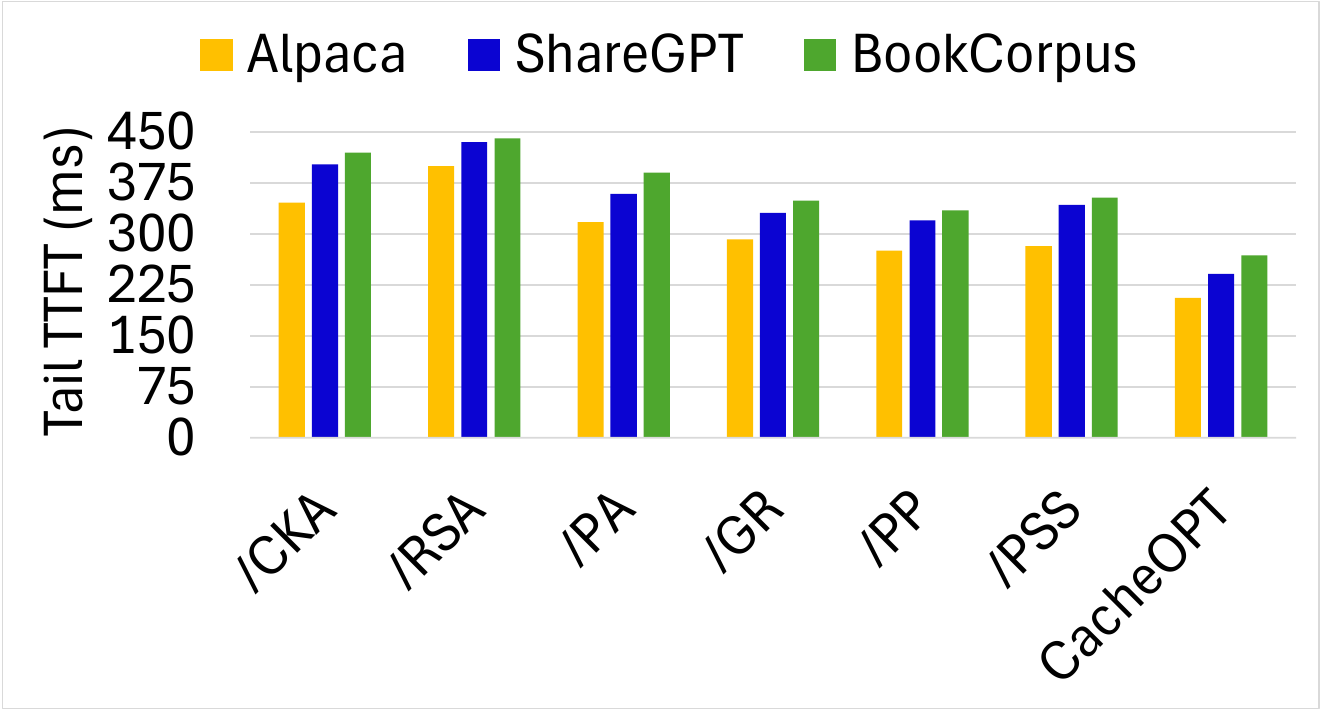} }}
    \hfill
    \subfloat[TTFT SLO attainment\vspace{-0.01in}\label{fig:ablation-slo}]{{\includegraphics[width=0.24\linewidth,height=0.13\textheight]{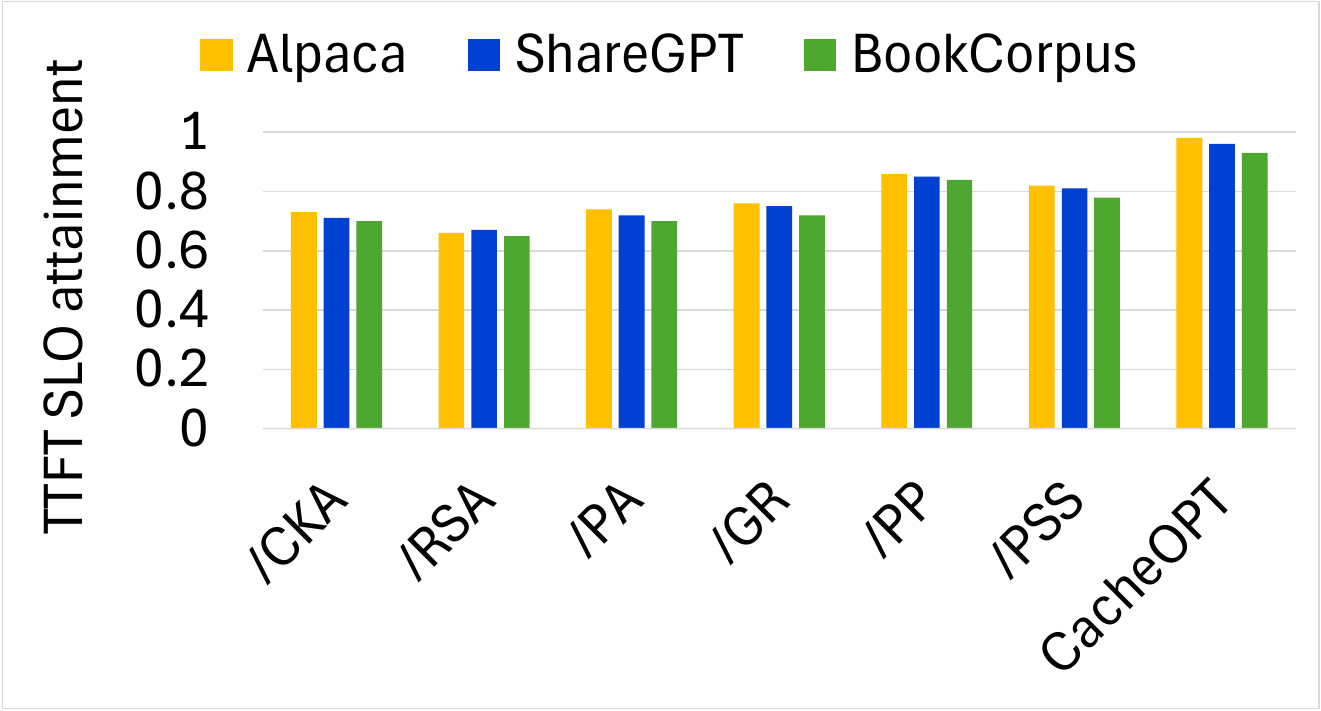} }}
    \hfill
    \subfloat[Tail TBT\vspace{-0.01in}\label{fig:ablation-tbt}]{{\includegraphics[width=0.24\linewidth,height=0.13\textheight]{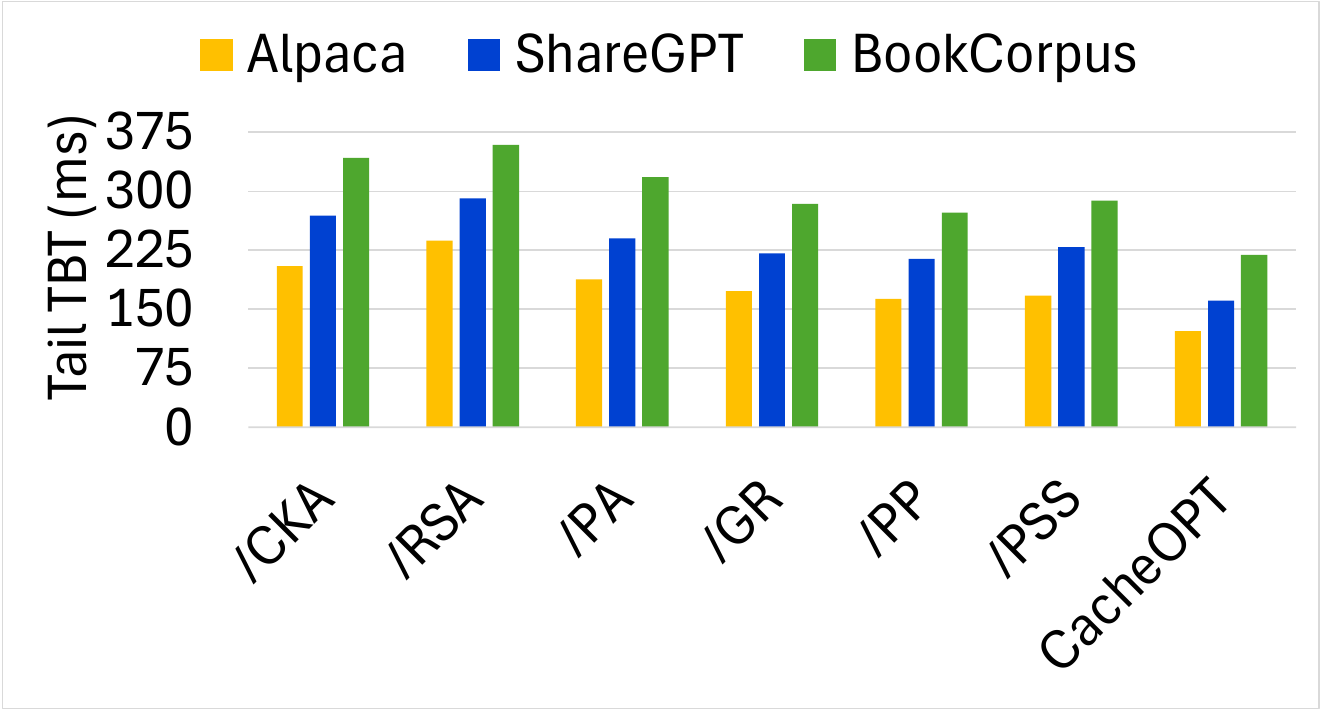} }}
    \hfill
\subfloat[TBT SLO attainment\vspace{-0.01in}\label{fig:tbt-ablation-slo}]{{\includegraphics[width=0.24\linewidth,height=0.13\textheight]{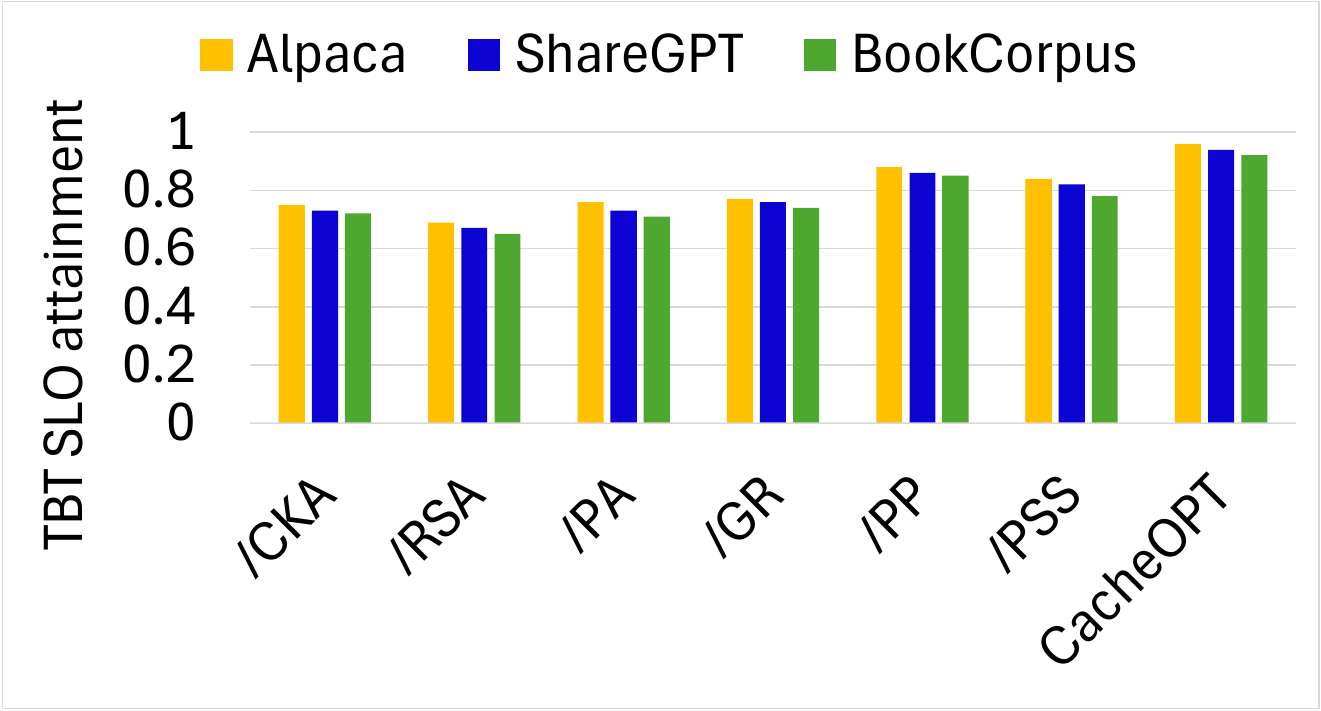} }}
    \hfill
\caption{\small{Ablation study for OPT-13B.\vspace{-0.0in}}}%
    \label{fig:ablation}
\end{figure*}
\begin{figure*}[t]
\centering
     \subfloat[Tail TTFT\vspace{-0.01in}\label{fig:ablation-ttft-175}]{{\includegraphics[width=0.24\linewidth,height=0.13\textheight]{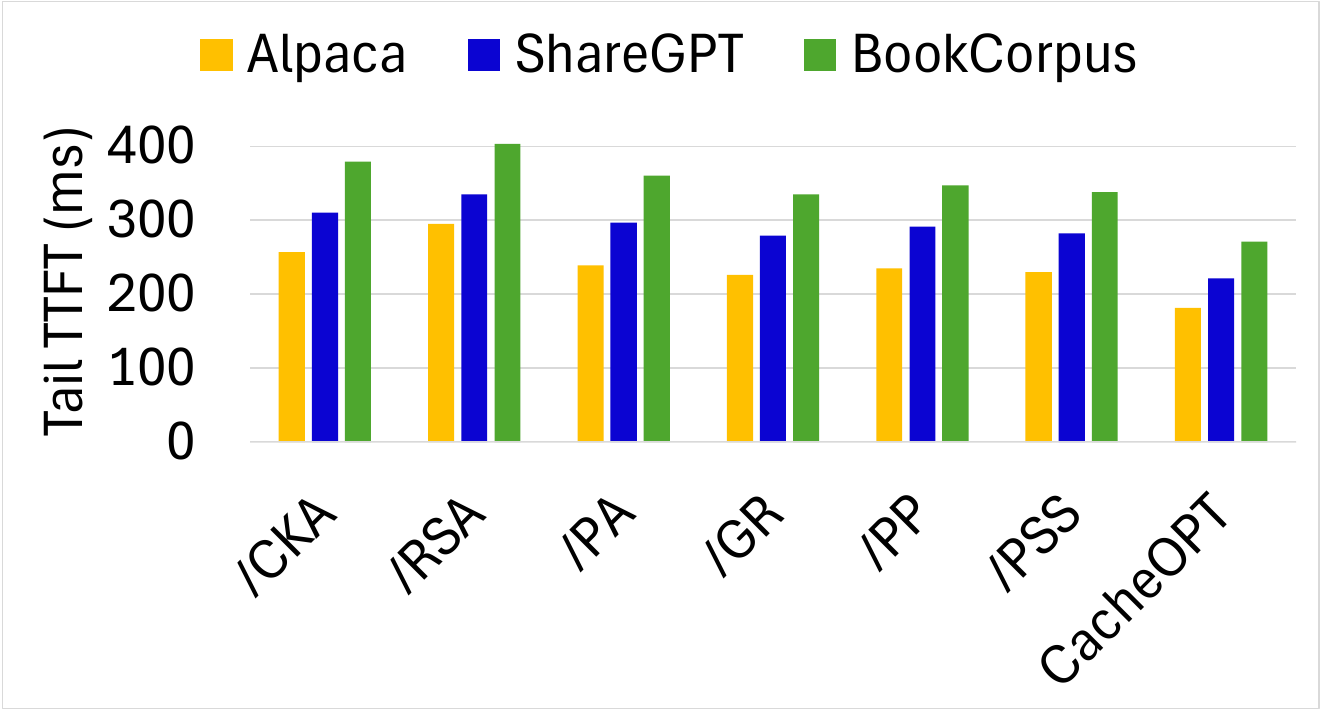} }}
    \hfill
    \subfloat[TTFT SLO attainment\vspace{-0.01in}\label{fig:ablation-slo-175}]{{\includegraphics[width=0.24\linewidth,height=0.13\textheight]{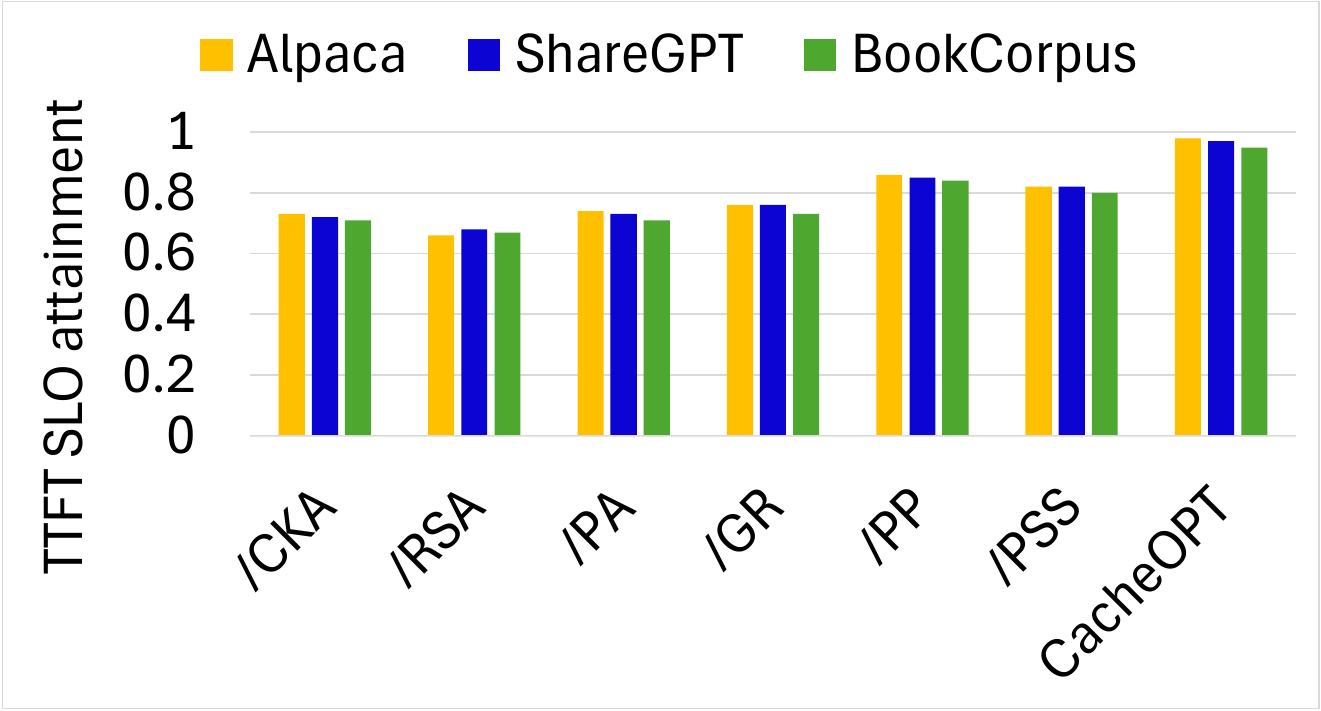} }}
    \hfill
    \subfloat[Tail TBT\vspace{-0.01in}\label{fig:ablation-tbt-175}]{{\includegraphics[width=0.24\linewidth,height=0.13\textheight]{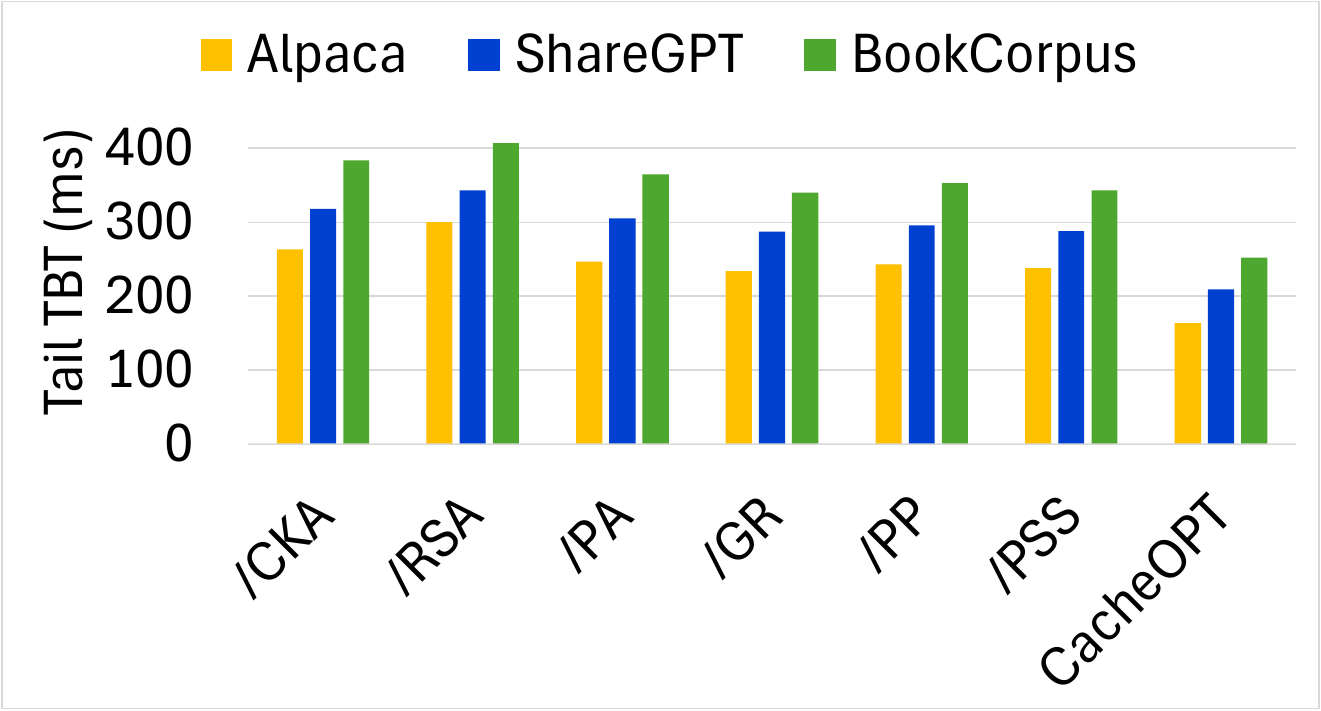} }}
    \hfill
\subfloat[TBT SLO attainment\vspace{-0.01in}\label{fig:tbt-ablation-slo-175}]{{\includegraphics[width=0.24\linewidth,height=0.13\textheight]{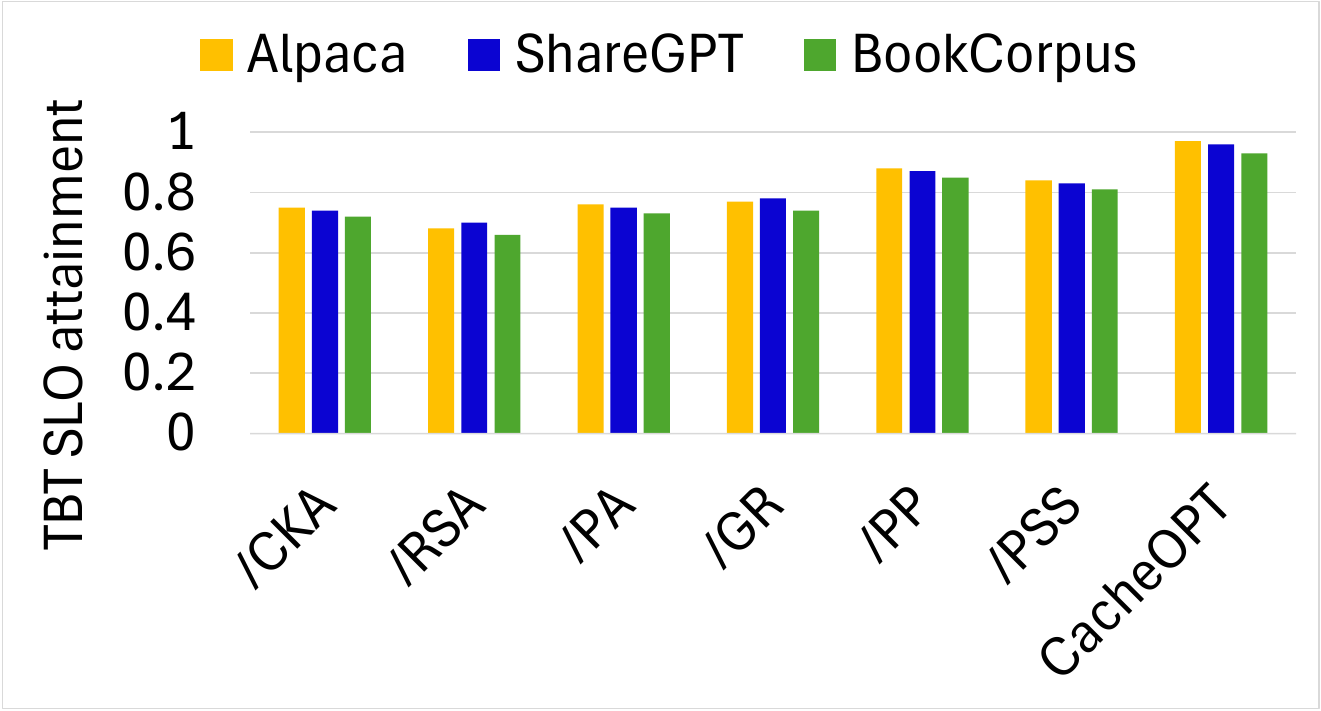} }}
    \hfill
\caption{\small{Ablation study for OPT-175B.\vspace{-0.0in}}}%
    \label{fig:ablation-175}
\end{figure*}

Figures~\ref{fig:ablation}, and~\ref{fig:ablation-175} show the different metrics of the variants in the four models. We now discuss the average results across the four models. \sys reduces the tail TTFT by 35\%, 44\%,	29\%, 24\%, 27\%, 28\% compared to /CKA, /RSA, /PA, /GR, /PP, and /PSS, respectively. For TTFT SLO attainment, these variants exhibit degradation of  27\%, 34\%, 26\%,	23\%,	14\%, and	17\%, respectively. For TBT, \sys demonstrates improvements of 40\%,	48\%,	35\%, 29\%,	25\%	and 27\%  compared to these variants, respectively. 
TBT SLO attainment for these variants shows degradations of  22\%, 28\%, 21\%, 20\%, 8\%, and 13\%, respectively. 
This demonstrates that removing or replacing critical design elements leads to performance degradation, and the critical role of all design components in reducing tail TTFT and TBT, and SLO violations. 

\DEL{Figures~\ref{fig:ablation-ttft},~\ref{fig:ablation-ttft-175},~\ref{fig:ablation-ttft-llama-8b}, and~\ref{fig:ablation-ttft-llama-70b}  show the tail TTFT of \sys variants for OPT-13B, OPT-175B, Llama-3-8B, and Llama-3-70B models, respectively. We observe that the full \sys achieves the lowest tail latencies, significantly outperforming all ablated variants. 

For TTFT, \sys reduces tail latency by approximately by 35\%,	44\%,	29\%, 24\%, 27\% compared to the /CKA, /RSA, /PA, /GR, /PP, and /PSS, variants, respectively. Figures~\ref{fig:ablation-slo}, \ref{fig:ablation-slo-175},~\ref{fig:ablation-ttft-slo-llama-8b}, and~\ref{fig:ablation-ttft-slo-llama-70b} focus on the SLO attainments for TTFT across the same set of configurations. The full \sys system consistently achieves the highest satisfaction ratios, nearing 1.0 for both metrics. In comparison for TTFT SLO, ablated variants such as /CKA, /RSA, /PA, /GR, /PP, and /PSS exhibit degradation of up to 27\%, 34\%, 26\%,	23\%,	14\%, and	17\%, respectively.

Figure~\ref{fig:ablation-tbt}, ~\ref{fig:ablation-tbt-175},~\ref{fig:ablation-tbt-llama-175b}, and~\ref{fig:ablation-tbt-llama-70b} show the tail TBT of \sys variants for OPT-13B, OPT-175B, Llama-3-8B, and Llama-3-70B models, respectively.
Similarly, for TBT, \sys demonstrates an improvement of 40\%,	48\%,	35\%, 29\%,	25\%	and 27\% compared to the /CKA, /RSA, /PA, /GR, /PP, and /PSS, variants, respectively. These results highlight the critical role of all design components in minimizing tail latencies. 
Figures~\ref{fig:tbt-ablation-slo},~\ref{fig:tbt-ablation-slo-175},~\ref{fig:tbt-ablation-slo-llama-8b} and~\ref{fig:tbt-ablation-slo-llama-70b} focus on the SLO attainments for TBT across the same set of configurations. The full \sys system consistently achieves the highest satisfaction ratios, nearing 1.0 for both metrics.  While comparison with TBT SLO, ablated variants such as /CKA, /RSA, /PA, /GR, /PP, and /PSS, variants exhibit degradation of up to 22\%, 28\%, 21\%, 20\%, 8\%, and 13\%, respectively. 
This demonstrates that removing or replacing critical design elements leads to reduced adherence to SLOs, further emphasizing the robustness and effectiveness of the full \sys system.

Overall, these ablation studies underline the necessity of each key component in \sys. Their combined effect ensures optimal performance, minimal tail latencies, and consistent adherence to SLOs under diverse workloads.
}

\DEL{Figure~\ref{fig:ablation} illustrates the performance of individual components of \sys in terms of average JCT, Time Between Tokens (TBT), and goodput. In the figure, ``$/ 10$'' means that the figure divides the results of the method by 10 to make the figure visible.

\sh{explain each method-done, red in compared method}
Compared to vLLM, \sys/C, \sys/E, \sys/P, and \sys achieve 16-36\%, 35-58\%, 43-64\%, and 64-91\% lower JCT, 14-32\%, 30-52\%, 41-62\%, and 60-72\% lower TBT, and 75-84\%, 91\%-1.13$\times$, 1.15-1.32$\times$, and 1.67-1.96$\times$ higher goodput, respectively.

\sys/C has the highest impact because, without the confidence-based batching and the length adjustment component, the overall latency increases the most. Then, \sys/E has less effect compared to JCT and TBT compared to \sys/C, as it only contributes to the preemption time. Finally, \sys/P has the most negligible impact because after making the preemption decision, the difference between swapping and the recomputation becomes similar for certain factors.}

\DEL{\begin{figure*}[t]
\centering
     \subfloat[Tail TTFT\vspace{-0.01in}\label{fig:ablation-ttft-llama-8b}]{{\includegraphics[width=0.24\linewidth,height=0.13\textheight]{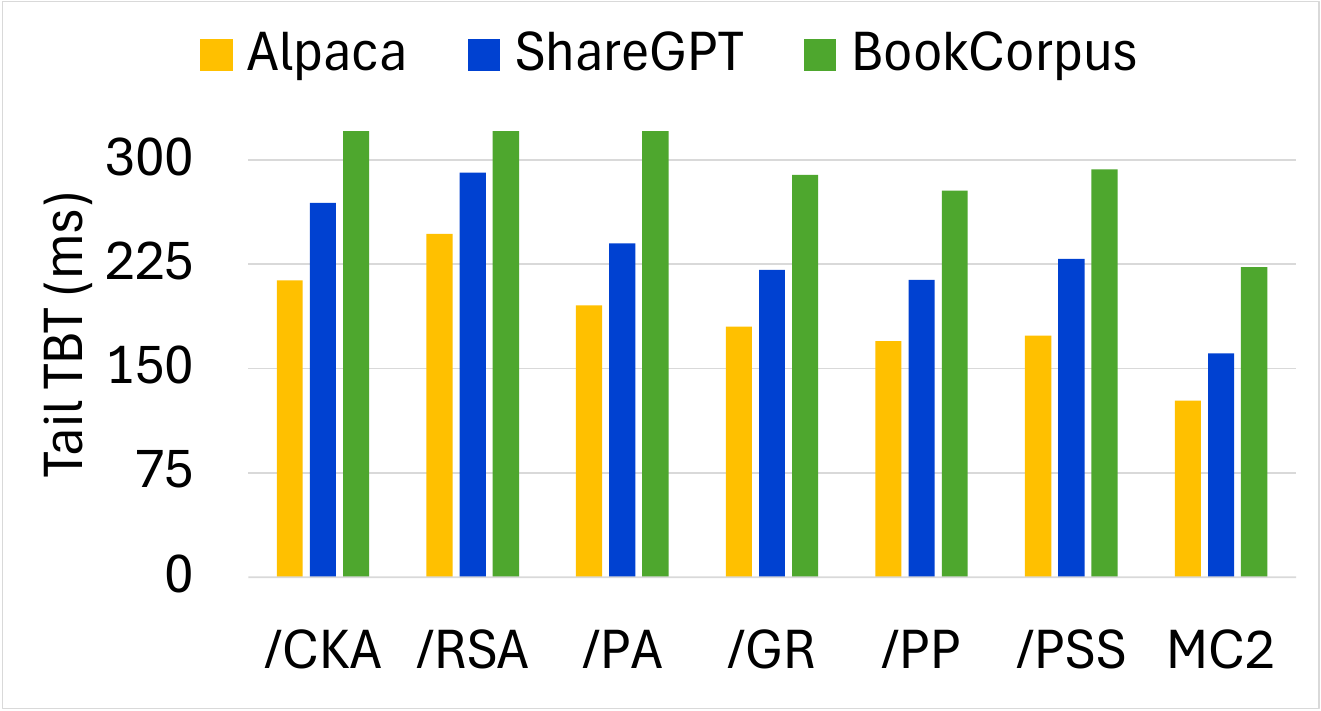} }}
    \hfill
    \subfloat[TTFT SLO attainment\vspace{-0.01in}\label{fig:ablation-ttft-slo-llama-8b}]{{\includegraphics[width=0.24\linewidth,height=0.13\textheight]{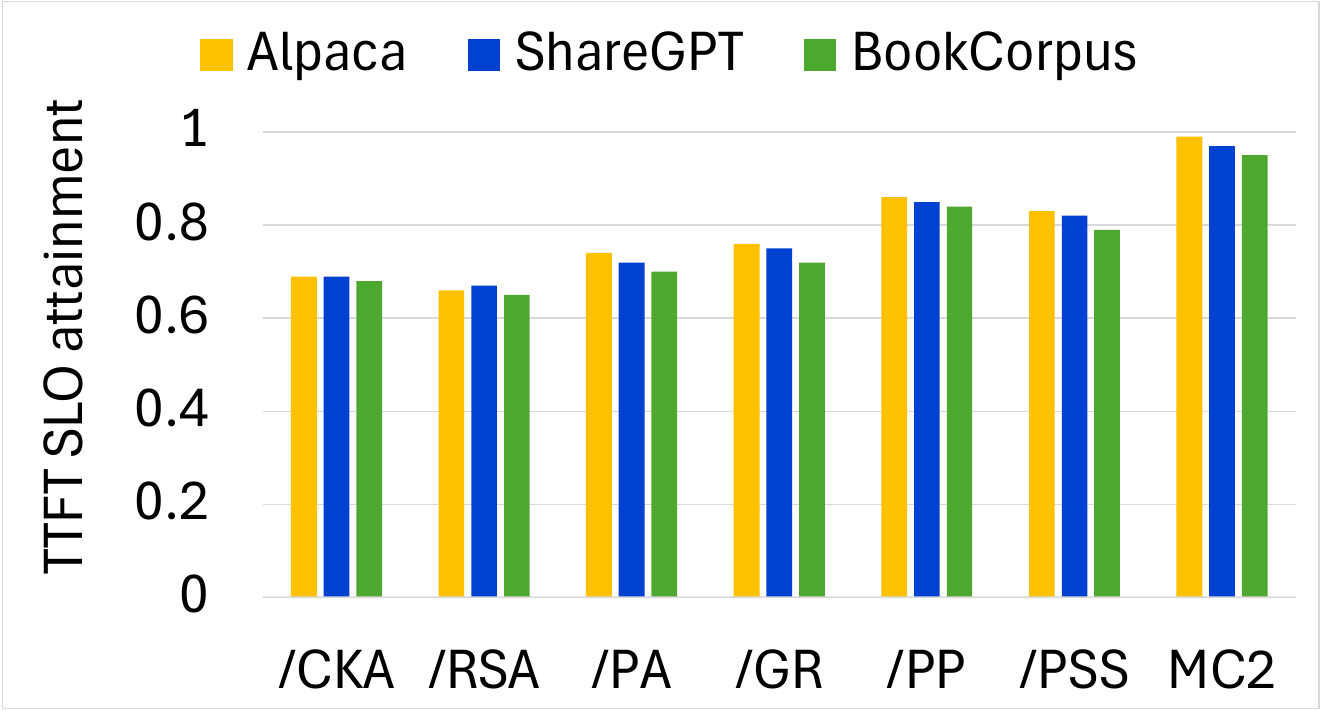} }}
    \hfill
    \subfloat[Tail TBT\vspace{-0.01in}\label{fig:ablation-tbt-llama-175b}]{{\includegraphics[width=0.24\linewidth,height=0.13\textheight]{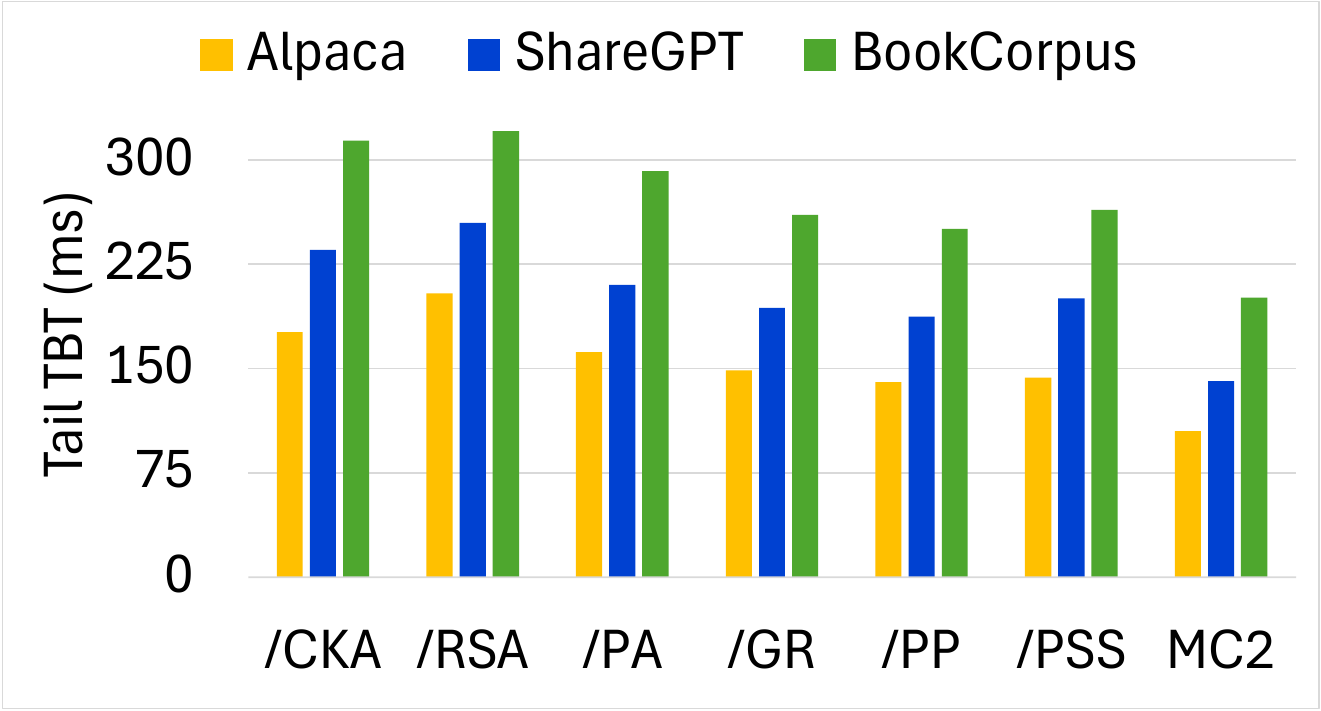} }}
    \hfill
\subfloat[TBT SLO attainment\vspace{-0.01in}\label{fig:tbt-ablation-slo-llama-8b}]{{\includegraphics[width=0.24\linewidth,height=0.13\textheight]{Padding-FIgs/TBT-slo-ablation-llama-8b.pdf} }}
    \hfill
\caption{\small{Ablation study for Llama-3-8B.\vspace{-0.0in}}}%
    \label{fig:ablation-llama-8b}
\end{figure*}}

\DEL{\begin{figure*}[t]
\centering
\subfloat[Tail TTFT\vspace{-0.01in}\label{fig:ablation-ttft-llama-70b}]{{\includegraphics[width=0.24\linewidth,height=0.13\textheight]{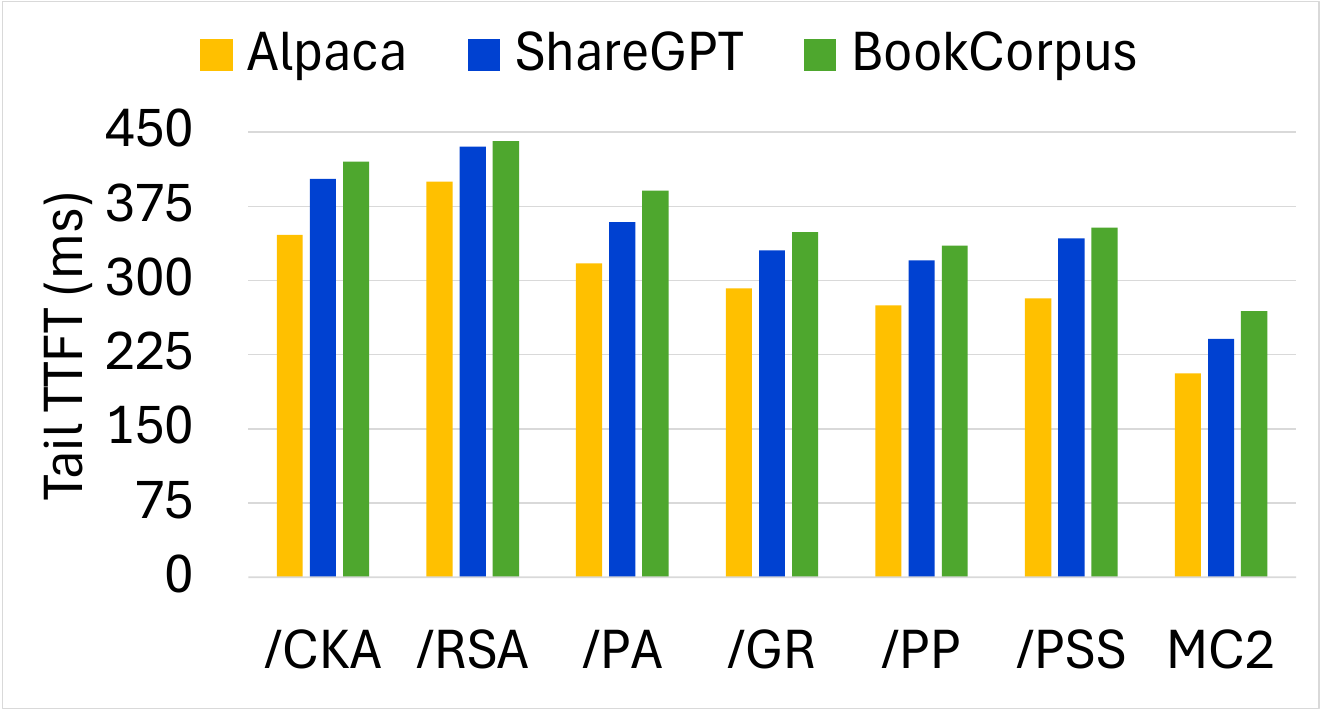} }}
    \hfill
    \subfloat[TTFT SLO attainment\vspace{-0.01in}\label{fig:ablation-ttft-slo-llama-70b}]{{\includegraphics[width=0.24\linewidth,height=0.060\textheight]{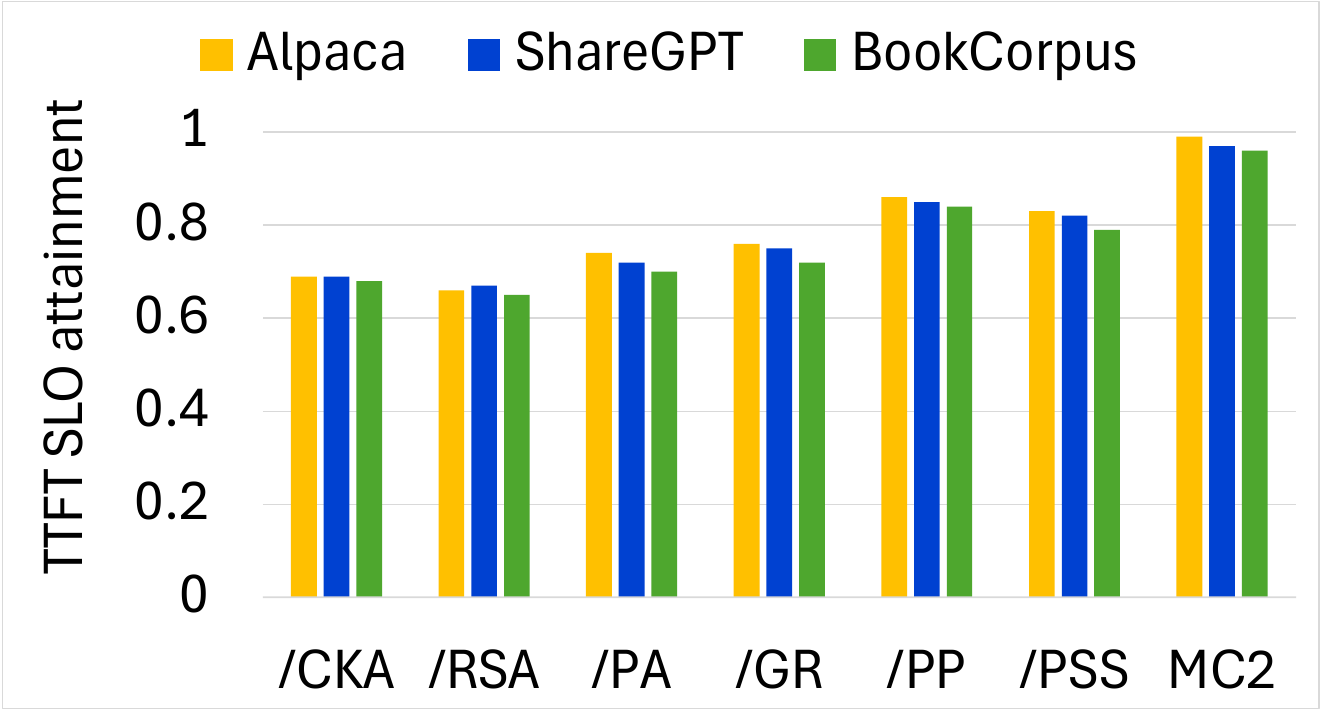} }}
    \hfill
    \subfloat[Tail TBT\vspace{-0.01in}\label{fig:ablation-tbt-llama-70b}]{{\includegraphics[width=0.24\linewidth,height=0.13\textheight]{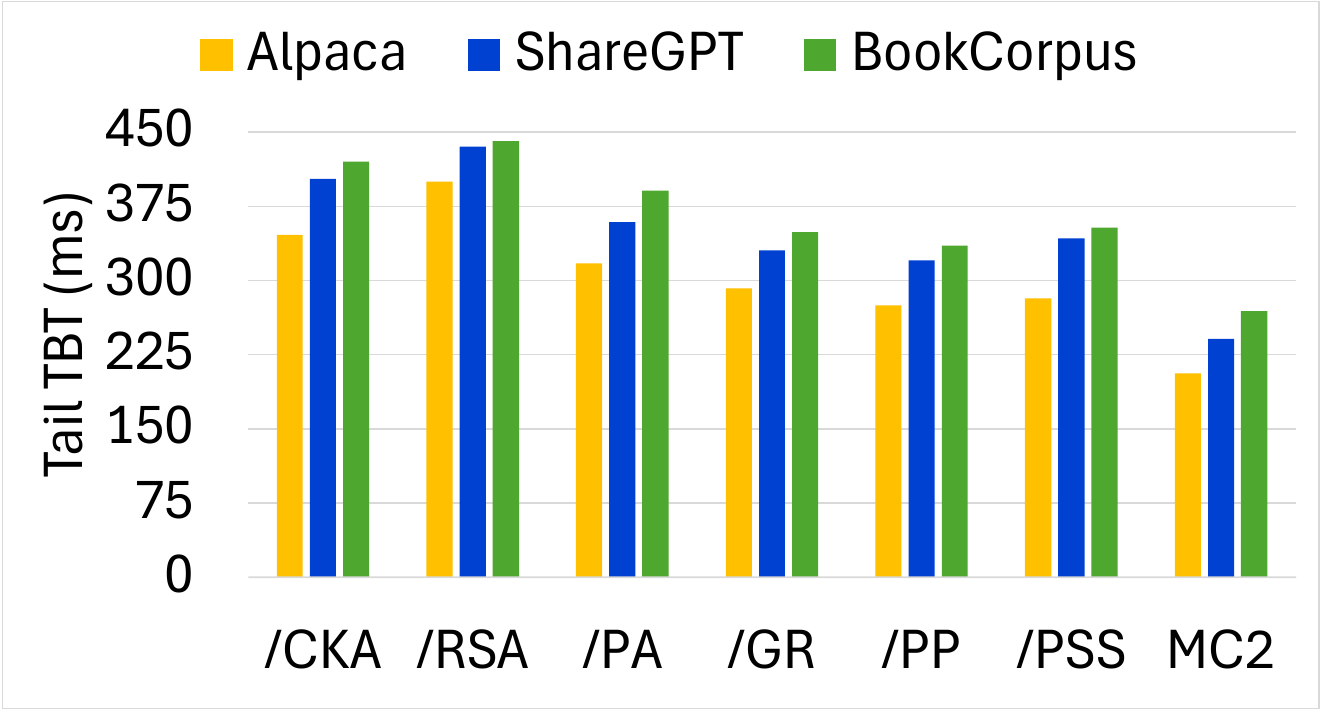} }}
    \hfill
\subfloat[TBT SLO attainment\vspace{-0.01in}\label{fig:tbt-ablation-slo-llama-70b}]{{\includegraphics[width=0.24\linewidth,height=0.13\textheight]{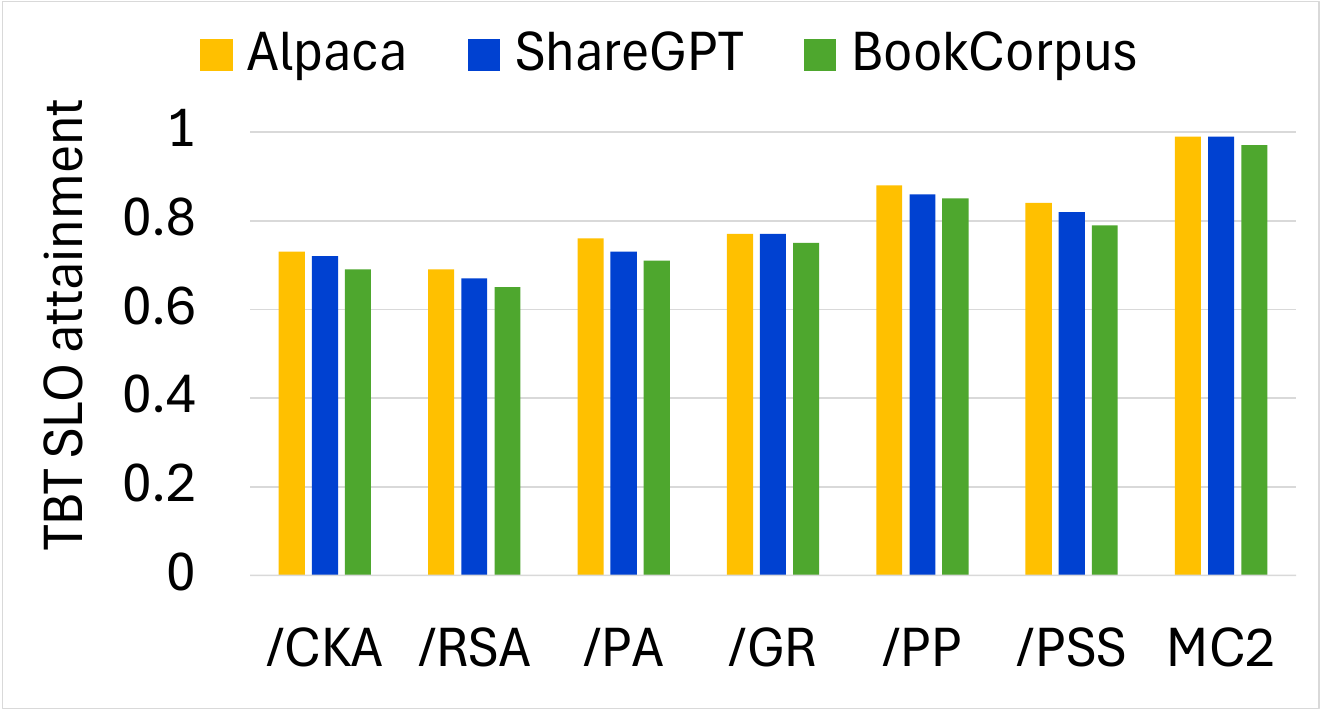} }}
    \hfill
\caption{\small{Ablation study for Llama-3-70B.\vspace{-0.0in}}}%
    \label{fig:ablation-llama-70b}
\end{figure*}}

\begin{figure}
    \centering
    \includegraphics[width=0.75\columnwidth,height=0.12\textheight]{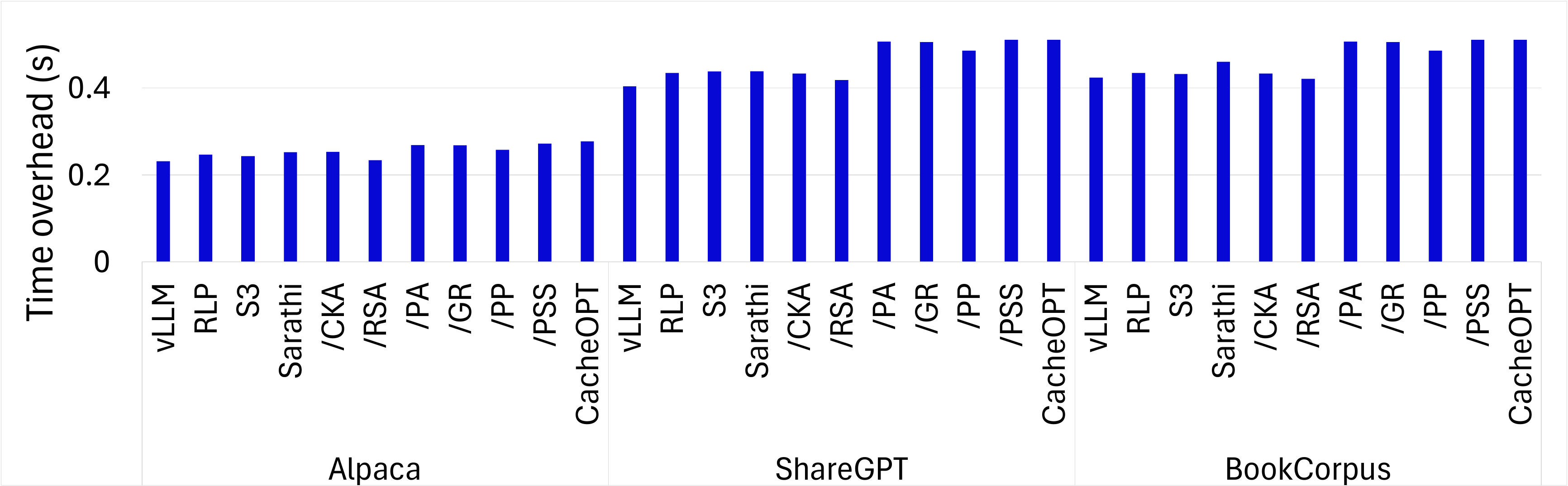}
    \caption{Scheduling time overhead.}
    \label{fig:overhead}
\end{figure}


\subsection{Time Overhead} 
Figure~\ref{fig:overhead} shows the time overhead for different systems and for different variants of \sys in each iteration.  \sys have 80\%, 77\%, 18\%, and 23\% higher time overhead than vLLM, Sarathi, RLP and $S^3$ respectively. 
These higher time overheads only constitute 0.003\% over the iteration. The overhead of the components of the \sys contributes to its time overhead. Specifically, compared to the \sys, /CKA, /RSA, /PA, /GR, /PP, /PPS show 17\%,	41\%,	3\% , 3.14\%,	7\%, 1.8\% less time overhead, indicating the time overhead of each component. \DEL{\tsr{\sys shows some additional overhead due to the shared functionalities among the components. These shared costs are included in the total time overhead and could not be isolated in the individual variant measurements.}}

\DEL{As we are using very basic strategies to replace the variants we can assume the approximate contribution of each component CKA, RSA, PA, GR, PP, and PPS  in time overhead contributes \sys's 80\% higher time overhead compared to vLLM. 
}
\DEL{/CKA, /RSA, /PA, /GR have 17\%, 2.81\%, and 2.86\%, and 2.92\% higher scheduling overhead than vLLM. /PP and /PSS have 2.83\% and 2.88\% higher overhead than vLLM.} 

\begin{figure*}[t]
\centering
    \subfloat[SLO scale\vspace{-0.0in}\label{fig:sensitivity-slo}]{{\includegraphics[width=0.24\linewidth,height=0.13\textheight]{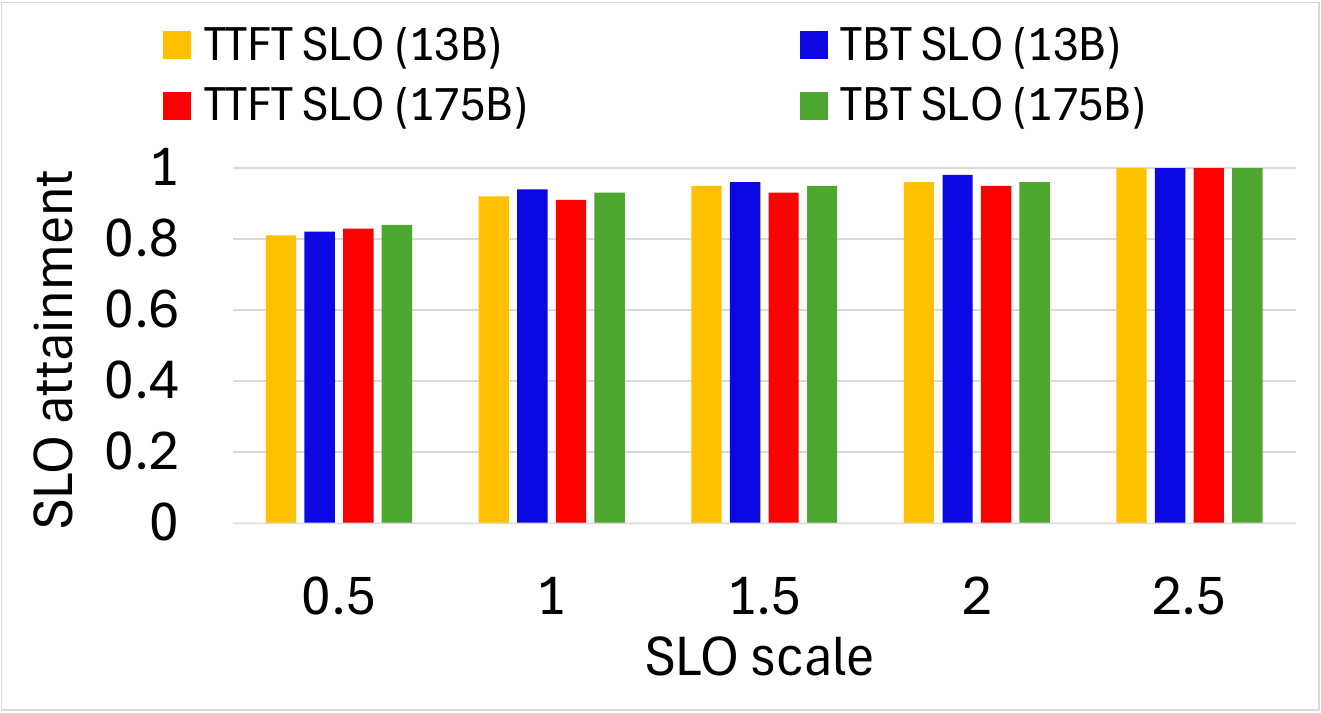} }}
    \hfill   
    \subfloat[Block size (B)\vspace{-0.0in}\label{fig:sensitivity-block-size}]{{\includegraphics[width=0.24\linewidth,height=0.13\textheight]{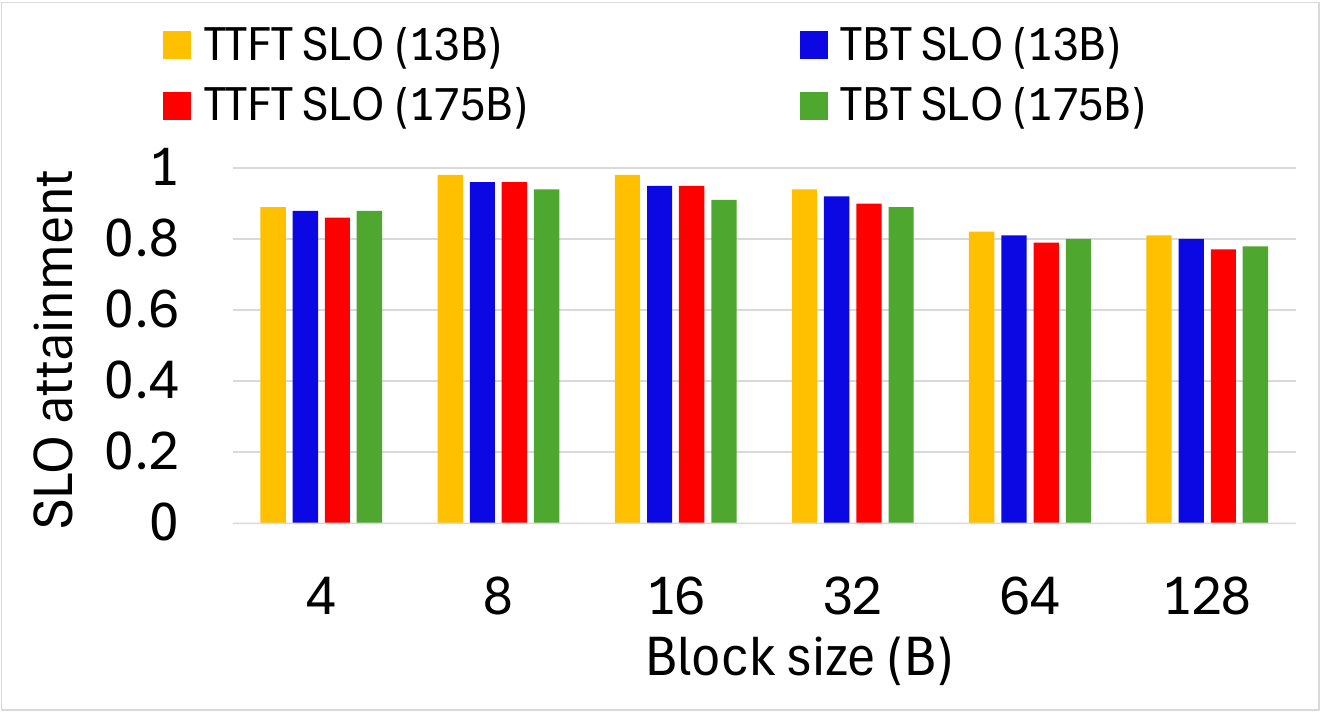} }}
    \hfill\subfloat[Iterations for preallocate (m)\vspace{-0.0in}\label{fig:sensitivity-m}]{{\includegraphics[width=0.24\linewidth,height=0.13\textheight]{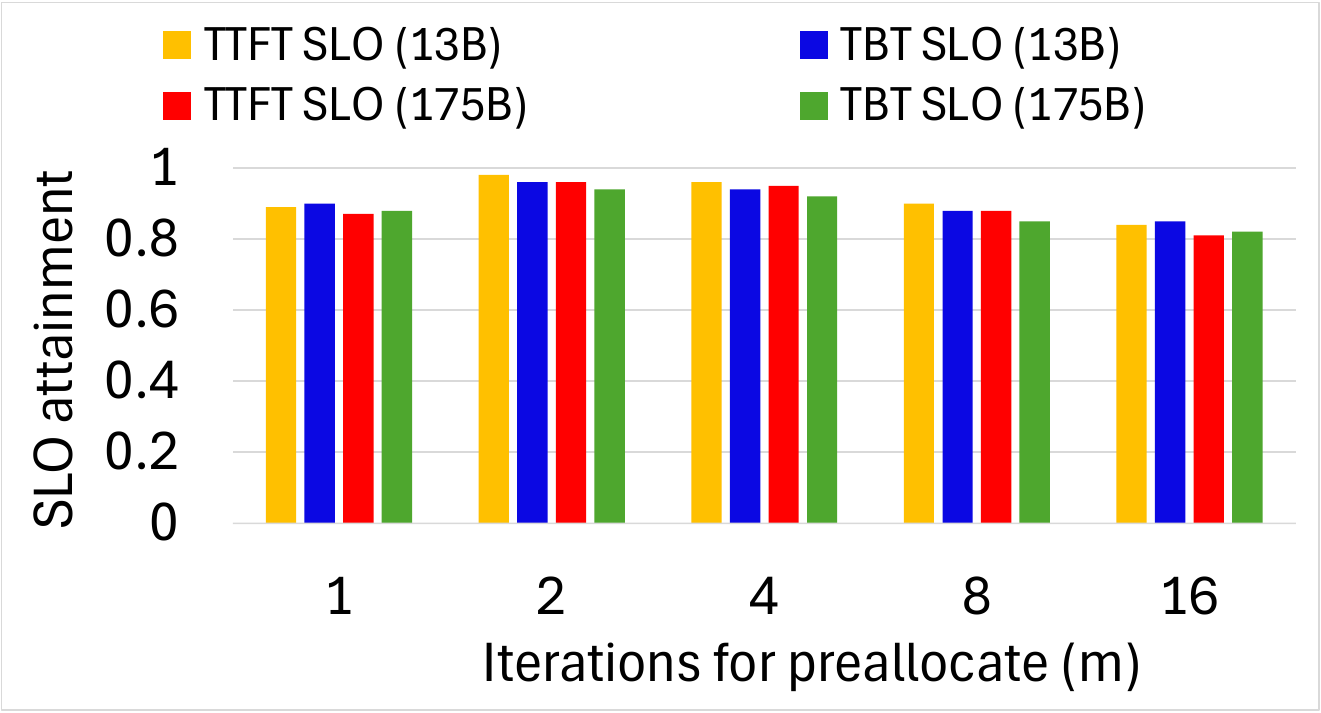} }}
    \hfill
    \subfloat[Number of reserved blocks\vspace{-0.0in}\label{fig:sensitivity-pad-block}]{{\includegraphics[width=0.24\linewidth,height=0.13\textheight]{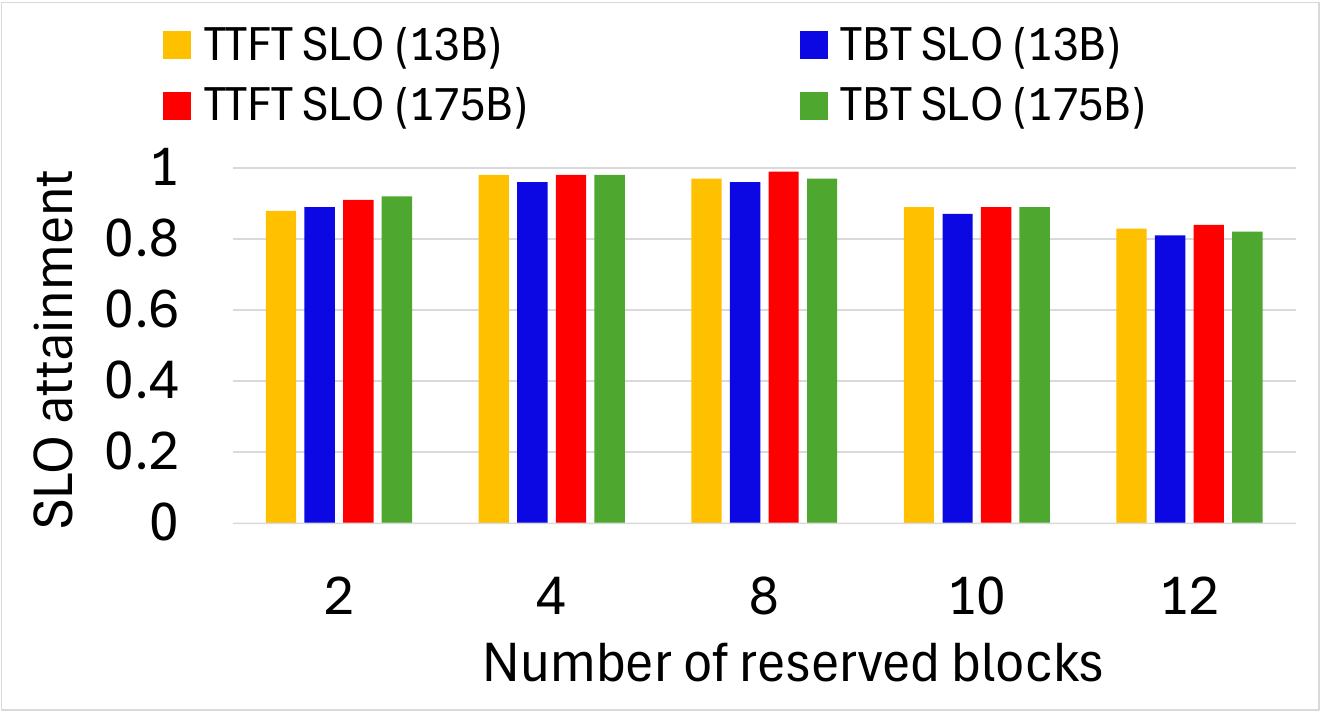} }}
    \hfill
\subfloat[$\alpha$\vspace{-0.0in}\label{fig:sensitivity-alpha}]{{\includegraphics[width=0.24\linewidth,height=0.13\textheight]{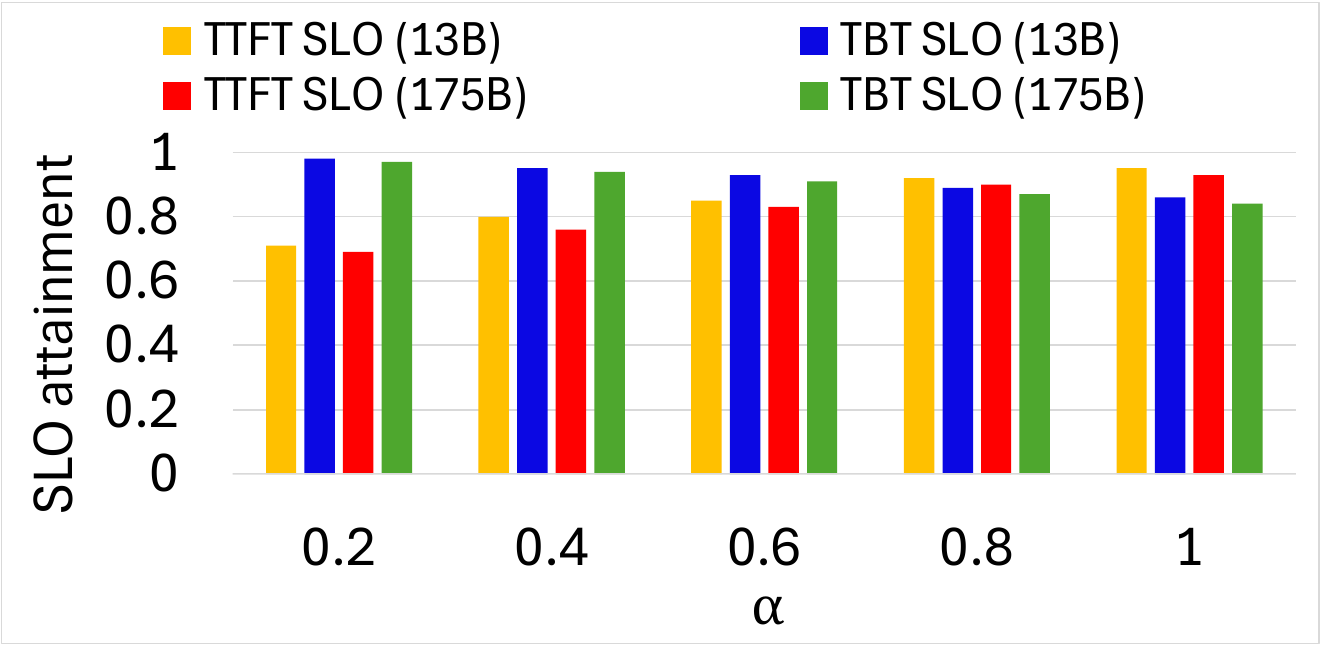} }}
    \hfill
\subfloat[$\beta$\vspace{-0.0in}\label{fig:sensitivity-beta}]{{\includegraphics[width=0.24\linewidth,height=0.13\textheight]{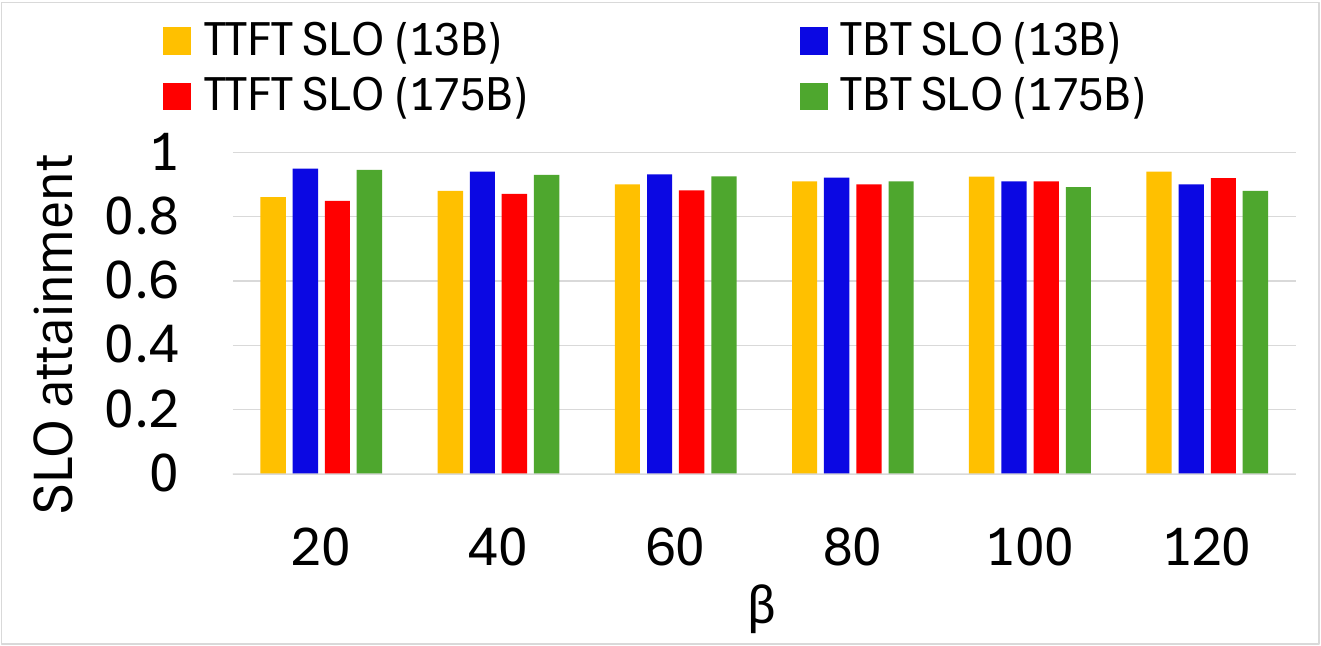} }}
    \hfill
    \subfloat[Confidence score. \vspace{-0.0in}\label{fig:sensitivity-conf}]{{\includegraphics[width=0.24\linewidth,height=0.13\textheight]{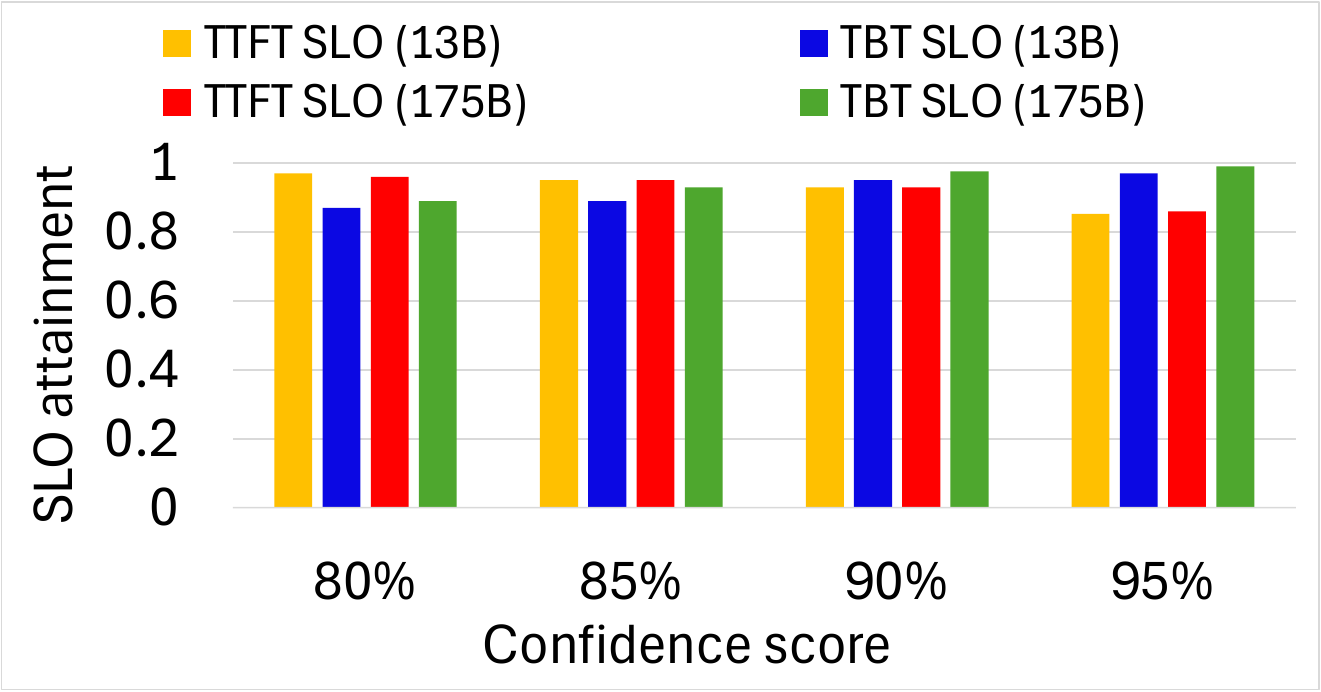} }}
    \hfill
    \subfloat[Response length predictor \vspace{-0.01in}\label{fig:performance-opt}]{{\includegraphics[width=0.24\linewidth,height=0.13\textheight]{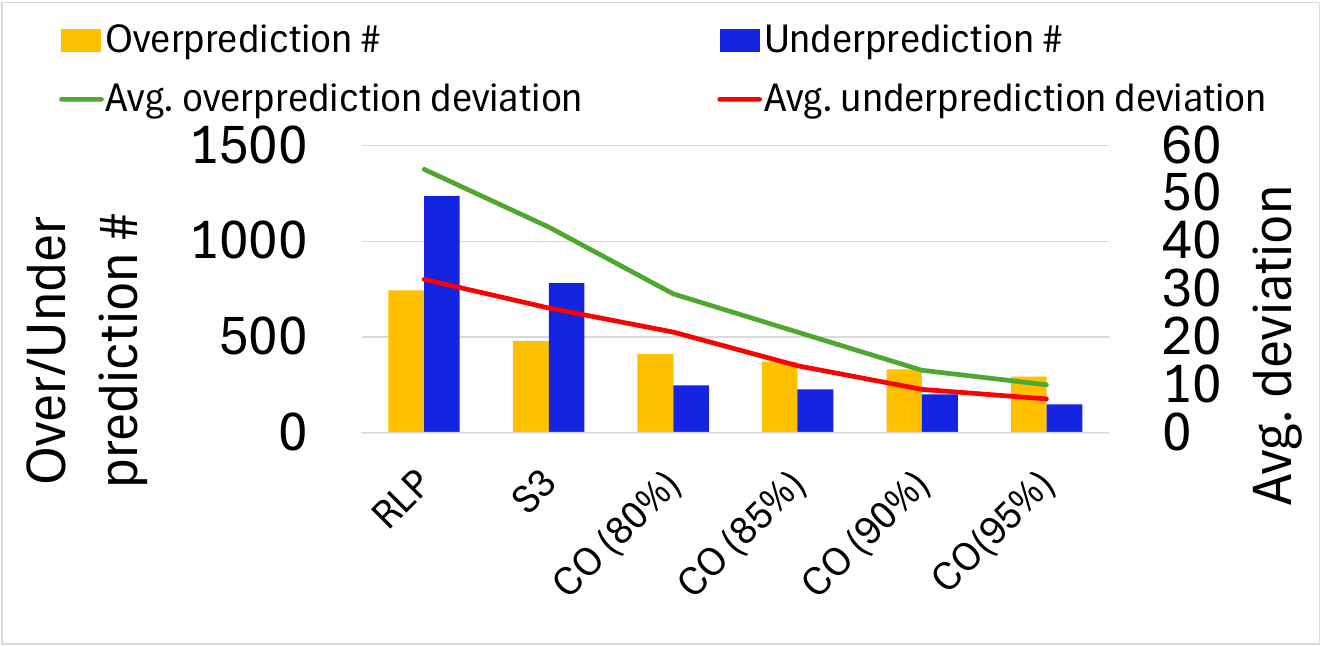} }}
    \hfill
    
    \DEL{\subfloat[$\alpha$\vspace{-0.0in}\label{fig:sensitivity-alpha}]{{\includegraphics[width=0.32\linewidth,height=0.13\textheight]{Padding-FIgs/sensitivity-alpha.pdf} }}
    \hfill
\subfloat[$\beta$\vspace{-0.0in}\label{fig:sensitivity-beta}]{{\includegraphics[width=0.32\linewidth,height=0.13\textheight]{Padding-FIgs/sensitiviti-beta-llama.pdf} }}
    \hfill}

    \vspace{-0.0in}
   \caption{Sensitivity testing. \vspace{-0.0in}}%
    \label{fig:sensitivity}
\end{figure*}

\DEL{\begin{figure*}[t]
\centering
    \subfloat[SLO scale\vspace{-0.0in}\label{fig:sensitivity-slo-llama}]{{\includegraphics[width=0.24\linewidth,height=0.13\textheight]{Padding-FIgs/slo-scale-llama-size.pdf} }}
    \hfill   
    \subfloat[Block size (B)\vspace{-0.0in}\label{fig:sensitivity-block-size-llama}]{{\includegraphics[width=0.24\linewidth,height=0.13\textheight]{Padding-FIgs/block-size-llama-size.pdf} }}
    \hfill\subfloat[Iterations for preallocate (m)\vspace{-0.0in}\label{fig:sensitivity-m-llama}]{{\includegraphics[width=0.24\linewidth,height=0.13\textheight]{Padding-FIgs/iterations-prealloc-llama-size.pdf} }}
    \hfill
    \subfloat[Number of reserved blocks\vspace{-0.0in}\label{fig:sensitivity-pad-block-llama}]{{\includegraphics[width=0.24\linewidth,height=0.13\textheight]{Padding-FIgs/reserved-block-llama-size.pdf} }}
    \hfill
    \DEL{\subfloat[Confidence score threshold\vspace{-0.0in}\label{fig:sensitivity-conf-llama}]{{\includegraphics[width=0.32\linewidth,height=0.13\textheight]{Padding-FIgs/confidence-llama-size.pdf} }}
    \hfill}
    \DEL{\subfloat[Confidence score controller (k).\vspace{-0.0in}\label{fig:sensitivity-conf-k-llama}]{{\includegraphics[width=0.32\linewidth,height=0.13\textheight]{Padding-FIgs/sensitivity-slo-llama-k.pdf} }}
    \hfill}
    
    \vspace{-0.0in}
   \caption{\small{Sensitivity testing for Llama models.\vspace{-0.0in}}}%
    \label{fig:sensitivity-llama}
\end{figure*}}

\DEL{\begin{figure}[t]
\centering
 \DEL{\subfloat[Used token budgets and allocated KVC tokens.\vspace{-0.01in}\label{fig:tk1}]{{\includegraphics[width=0.48\linewidth,height=0.13\textheight]{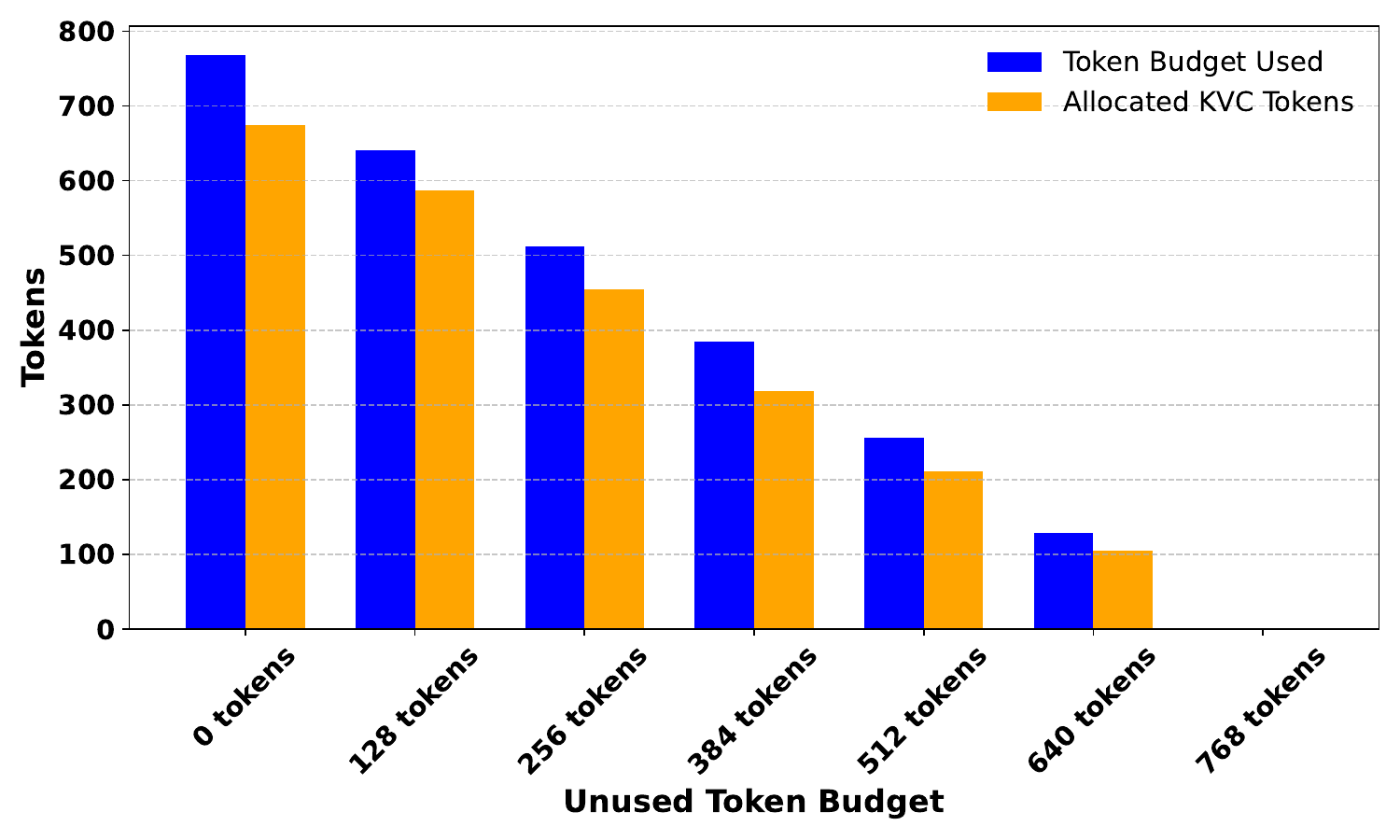} }}%
    \hfill}
    \subfloat[Confidence score \vspace{-0.0in}\label{fig:sensitivity-conf-llama}]{{\includegraphics[width=0.48\linewidth,height=0.13\textheight]{Padding-FIgs/confidence-score-up.pdf} }}
    \hfill
    \subfloat[Response length predictor.\vspace{-0.01in}\label{fig:performance-llama}]{{\includegraphics[width=0.48\linewidth,height=0.13\textheight]{Padding-FIgs/performance-llama-2.pdf} }}
    \hfill
    \DEL{\subfloat[vLLM (Y: average \%, two bars for each X point: 1) token budget used, Allocated KVC tokens, X points: 1)Not reach token budget, 2) Reach token budget, 3) all results) ---need to modify.\vspace{-0.01in}\label{fig:tk3}]{{\includegraphics[width=0.48\linewidth,height=0.13\textheight]{Padding-FIgs/vLLM_average_percentages_vs_categories.pdf} }}%
    \hfill
    \subfloat[RLP (Y: average \%, two bars for each X point: 1) token budget used, Allocated KVC tokens, X points: 1)Not reach token budget, 2) Reach token budget, 3) all results) ---need to modify.\vspace{-0.01in}\label{fig:tk-rlp}]{{\includegraphics[width=0.48\linewidth,height=0.13\textheight]{Padding-FIgs/RLP_average_percentages_vs_categories.pdf} }}%
    \hfill
    \subfloat[S3 (Y: average \%, two bars for each X point: 1) token budget used, Allocated KVC tokens, X points: 1)Not reach token budget, 2) Reach token budget, 3) all results) ---need to modify.\vspace{-0.01in}\label{fig:tk-s3}]{{\includegraphics[width=0.48\linewidth,height=0.13\textheight]{Padding-FIgs/S3_average_percentages_vs_categories.pdf} }}%
    \hfill}
    \DEL{\subfloat[CDF of iterations (Y: CDF of iterations, X: Unused token budget).\vspace{-0.01in}\label{fig:tk-cdf}]{{\includegraphics[width=0.48\linewidth,height=0.13\textheight]{{Padding-FIgs/cdf_iterations_vs_unused_token_budget.pdf}}}}
    \hfill}
   \DEL{\subfloat[CDF of requests vs waiting time.\vspace{-0.01in}\label{fig:tk4}]{{\includegraphics[width=0.48\linewidth,height=0.13\textheight]{{Padding-FIgs/cases.png}} }}%
    \hfill}
     \vspace{-0.1in}\caption{\small{Sensitivity of confidence score for Llama-3-8B.\vspace{-0.2in}}}
    \label{fig:sensitivity-score-llama}
\end{figure}}

\DEL{We measure the average latency to return to the running state for the different available memory (PCIe) bandwidth and available GPU(\%) and plot in Figure~\ref{fig:sensitivity} for the three traces. From the figure, we observe that with the increasing available memory (PCIe) bandwidth and available GPU(\%), the mentioned latency keep decreasing and it stabilizes after reaching a certain limit. The stabilization limit for available memory (PCIe) bandwidth is 50\%, while for available GPU (\%), it is 40\%.}


\subsection{Sensitivity Testing}


Figure~\ref{fig:sensitivity-slo}  shows the SLO attainment for the two OPT models, versus the SLO scale. The figure show a steady increase in SLO attainment as the scale increases. At the smallest scale of 0.5, \sys still can achieve around 0.85 TTFT and TBT SLO attainments. As the scale increases to 1.5, the SLO attainments surpass 0.95, and they reach nearly 1.0 at the scale of 2.5. The results verify the capability of \sys in providing high TTFT and TBT SLO attainments, and good user experience.

\DEL{OPT-13B and OPT-175B model and~\ref{fig:sensitivity-slo-llama} showsthe SLO attainment for both Llama-3-8B and Llama-3-70B model as the SLO scale changes. When X=0.5, we only use the 0.5 SLO scale for all requests, and the same for other X values. The figure show a steady increase in SLO attainment as the scale is increased from 0.5 to 2.5. At the smallest scale of 0.5, \sys still can achieve around 0.85 TTFT and TBT SLO attainments. As the scale increases to 1.5, the SLO attainments surpass 0.95, and they reach nearly 1.0 at the scale of 2.5. The results verify the capability of \sys in providing high TTFT and TBT SLO attainments, and high user experience.}



Figures~\ref{fig:sensitivity-block-size} shows the SLO attainments versus the block size (B). As the block size increases from 4 to 8, the SLO attainment increases, then as the block size increases, the SLO attainments gradually decrease. A smaller block size (e.g., 4) results in underprovisioning, which also leads to high TBT and hence high TTFT, though they do not increase significantly. This is because requests with insufficient KVC can still obtain additional KVC proactively or on demand. Conversely, a larger block size causes overprovisioning, reducing the number of requests per batch and increasing TTFT and KVC competition and hence TBT. Experimental results indicate that a block size of 8 performs best in our setup.


Figure~\ref{fig:sensitivity-m} shows the SLO attainments versus the number of iterations for preallocation ($m$). 
At $m=1$, SLO attainments are around 0.85, peaking at 1.0 when $m=2$, before gradually declining beyond $m=2$ due to reserved waste. Hence, $m=2$ is the optimal setting in our experimental setup.



Figure~\ref{fig:sensitivity-pad-block} shows the SLO attainments versus the number of reserved blocks. The SLO attainments increase as the number of reserved blocks increases from 2 to 4 and then to 8, but then decrease when it keeps decreasing. Insufficient reserved KVC cannot satisfy KVC demands from some requests but high reservation increases reserved waste. Therefore, 4 and 8 are suitable number of reserved blocks in our experiment settings.


Figure~\ref{fig:sensitivity-alpha} shows the impact of the variable $\alpha$ in Equation~\eqref{equ:c}, which controls the Hoeffding's inequality for \sys with $\beta$. For $\beta=100$, with the increase of $\alpha$, we observe that the TTFT SLO keeps increasing, but the TBT SLO keeps decreasing. This happens because increasing $\alpha$ increases $c_i$, which increases padding. Figure~\ref{fig:sensitivity-beta} shows the same plot for $\beta$, for $\alpha = 0.9$. For increasing $\beta$, we observe that TTFT SLO keeps decreasing while the TBT SLO keeps increasing, because increasing $\beta$ decreases $c_i$, which decreases padding.

Figure~\ref{fig:sensitivity-conf}  shows the impact of confidence score. 
The results demonstrate a trend of increasing TBT SLO attainments and decreasing TTFT SLO attainments as the confidence score increases. At lower confidence score (e.g., $80\%$), less padding is added, enabling more requests accommodated in a batch, which reduces waiting time but increases preemption time. 
At a confidence score of 0.9, both TTFT and TBT SLO attainments exceed 0.9, striking a balance between TTFT and TBT. 

\DEL{\tsr{Figure~\ref{fig:sensitivity-conf-k} and~\ref{fig:sensitivity-conf-k} show
\textit{SLO attainments} for TTFT and TBT for different confidence score controller (k) that decides the values of confidence in Hoeffding's inequality for \sys based on the arrival rate. For our tested arrival rates of [24,40], as we increase the values of $k$, the TTFT SLO keep decreasing, but the TBT SLO keep decreasing following the same trend as we observed for confidence score threshold. }}


Figure~\ref{fig:performance-opt} shows the impact of the confidence score on the number of overprovisions and underprovisions, along with their deviations. We also include those of RLP and $S^3$ as reference. 
From the figure, we observe compared to RLP and $S^3$ $\sys$'s response length predictor have 54\% and 29\% fewer overpridictions and 72\% and 57\% of lower underpredictions, respectively. We also see for overprediction, \sys's response length predictor show 72\% and 65\% less deviation and 57\% and 65\% less deviation for underprediction, compared to RLP and $S^3$, respectively. With the increase of the confidence score, more padding is added following Hoeffding's inequality. As a result, fewer overprediction  and  underprediction requests are observed as it increases both positive and negative padding. We also observe the same phenomenon on the average deviation.


 

Overall, these results demonstrate the adaptability of \sys across a wide range of configurations. By effectively adapting KV cache allocation with workload demands, \sys maintains high SLO attainments, ensuring reliable performance under diverse conditions.


\DEL{\begin{figure}[t]
\centering
\subfloat[OPT models\label{fig:cv-opt}]{\includegraphics[width=0.48\linewidth,height=0.13\textheight]{Padding-FIgs/cv-burstiness.pdf}}\hfill
\subfloat[Llama models \label{fig:cv-llama}]{\includegraphics[width=0.48\linewidth,height=0.13\textheight]{Padding-FIgs/cv-burstiness-llama.pdf}}
\hfill
\caption{\small Resilience to request burstiness.}
\label{fig:burstiness}
\end{figure}
\noindent{\textbf{Burstiness.}} Figure~\ref{fig:burstiness} shows the TTFT and TBT SLO attainments of \sys for the increasing coefficient of variance (CV) in the poisson distribution, which reflects burstiness. As CV increases, the SLO attainments decrease slowly. The results show that for CV up to 3, \sys sustains both TTFT and TBT SLO attainments greater than 0.9. This highlights \sys's ability to adhere to SLOs, even under conditions of high burstiness.}


{\section{Limitations and Discussions}
\label{sec:limitations}
\noindent{\textbf{Model Assumptions:} The effectiveness of Hoeffding's inequality relies on the assumptions of bounded and independent random variables. While request lengths are naturally bounded, the independence assumption might not hold perfectly in real-world LLM service scenarios due to potential correlations between requests or workload patterns.

\noindent{\textbf{Confidence Score Calibration:}} \sys heavily depends on the accuracy and calibration of the confidence scores associated with predicted request lengths. If the confidence scores are not well-calibrated (i.e., they don't accurately reflect the actual probabilities), the probabilistic guarantees offered by Hoeffding's inequality might not hold in practice.
}

\DEL{\noindent{\textbf{Computational Overhead:}} While the proposed approach avoids modifying the core optimization problem, calculating the length adjustments for each request using Hoeffding's inequality still adds some computational overhead. This overhead might become significant for huge batches or high request rates.}

\DEL{\noindent{\textbf{Beyond Adjustment:}} While length adjustment is a critical aspect of managing LLM service workloads, other techniques like dynamic batch sizing, priority scheduling, and adaptive resource allocation can also play a crucial role in optimizing performance and ensuring SLO compliance.}

\section{Related Work} 
\label{sec:related-works}

\Orca~\cite{280922} uses an iteration-level scheduling strategy combined with maximum resource allocation, which results in under-utilization of GPU resources. To address this inefficiency, vLLM~\cite{vllm} introduces a block-based allocation strategy, and prediction-based KVC allocation methods~\cite{jin2023s,Zheng2023ResponseLP} predict output lengths and allocate the KVC equal to the predicted size. Sarathi-Serve~\cite{agrawal2024taming} incorporates chunked-prefills and stall-free scheduling to address long sequences. FastServe~\cite{Wu2023FastDI} utilizes preemptive scheduling to minimize JCT. Llumnix~\cite{liu2024llumnix} reschedules requests across multiple model instances, effectively reducing tail latencies.  vAttention~\cite{Prabhu2024vAttentionDM} is a memory management system that uses virtual memory to store and manage the KV cache. Cheng \emph{et al.}~\cite{Cheng2024SliceLevelSF} developed a scheduling method that splits long-generation tasks into smaller parts, to make it easier to manage resources and predict serving times. ALISA~\cite{Zhao2024ALISAAL} combines a Sparse Window Attention algorithm to reduce the memory footprint of KVC with three-phase token-level dynamic scheduling to optimize the balance between caching and recomputation. Several approaches focus on resource optimization to improve LLM performance. ExeGPT~\cite{oh2024exegpt} optimizes resource usage and adjusts execution settings like batch size and parallelism. INFERMAX~\cite{Kim2024TheEO} analyzes different scheduling strategies, focusing on balancing costs and resource usage. Sheng~\emph{et al.}~\cite{Fairness-OSDI-2024} focused on achieving fairness in scheduling. Lee \emph{et al.}~\cite{298683} proposed prefetching only the essential KV cache entries for computing the subsequent attention layer. DistServe~\cite{298687} decouples the prefill and decode phases, running them on separate machines or
GPUs. Some other methods~\cite{298762,298764} aim to enhance the efficiency and performance of LLM applications either by optimizing application-level operations or by dynamic adapter management targeting LoRA. Unlike previous work, we study the impact of the allocated KVC amount on the tradeoff between satisfying the TTFT and TBT SLOs, and propose novel methods to maximize the attainment of both TTFT and TBT SLOs.

\DEL{Several approaches have been proposed to enhance the performance of LLMs through improved scheduling and memory management. \Orca~\cite{280922} adopts an iteration-level scheduling strategy combined with maximum resource allocation, which results in under-utilization of GPU resources. To address this inefficiency, vLLM~\cite{vllm} introduced a block-based allocation strategy that incrementally allocates memory to optimize utilization. Agarwal \emph{et al.}~\cite{agrawal2024taming} developed Sarathi, which incorporates advanced techniques such as chunked-prefills and stall-free scheduling to significantly enhance serving capacity and reduce latency across a variety of models and hardware setups.
Jin \emph{et al.}~\cite{jin2023s} introduced $S^3$, which predicts output length and allocates necessary memory to the request. Zheng \emph{et al.}~\cite{Zheng2023ResponseLP} proposed a system that predicts output lengths and batch queries with similar response lengths. Wu \emph{et al.}~\cite{Wu2023FastDI} introduced FastServe, utilizing preemptive scheduling to minimize JCT. Sun et al.~\cite{liu2024llumnix} introduced Llumnix, a system designed to reschedule requests across multiple model instances, effectively reducing tail latencies during inference.  Prabhu \emph{et al.}~\cite{Prabhu2024vAttentionDM} introduced vAttention, a memory management system that uses virtual memory to store and manage the KV cache for LLMs efficiently without modifying existing attention code. Cheng \emph{et al.}~\cite{Cheng2024SliceLevelSF} developed a scheduling method that splits long-generation tasks into smaller parts, to make it easier to manage resources and predict serving times.  Zhao \emph{et al.}~\cite{Zhao2024ALISAAL} introduced ALISA, a solution that combines a Sparse Window Attention algorithm to reduce the memory footprint of Key-Value (KV) caching with a three-phase token-level dynamic scheduling system. This approach optimizes the balance between caching and recomputation, enhancing performance on resource-constrained systems.

Several approaches focus on resource optimization to improve LLM performance. Oh \emph{et al.}~\cite{oh2024exegpt} proposed ExeGPT, a system that finds the best way to run LLMs by optimizing resource usage and adjusting execution settings like batch sizes and parallelism. Kim \emph{et al.}~\cite{Kim2024TheEO} created INFERMAX, a tool that analyzes different scheduling strategies for LLMs, focusing on balancing costs and resource usage during inference. Sheng\emph{et al.}~\cite{Fairness-OSDI-2024} focused on achieving fairness in scheduling.}
}



\section{Conclusion}
\label{sec:conclusion}
Our proposed \sys addresses the challenge to satisfy both TTFT and TBT SLOs. It incorporates four components: 1) confidence-based padding, 2) SLO-aware batching and KVC allocation, 3) preemption policy, and 4) preemption strategy selection.
Experimental results show that \sys
achieves up to a 3.29$\times$ and 2.83$\times$ lower tail TBT and TTFT,
and 47\% and 53\% higher TTFT and TBT SLO attainment
than the state-of-the-art methods. In the future, we will design methods to automatically determine the optimal parameters in \sys.


\bibliographystyle{unsrt}
\bibliography{myBib,myBib-2,nlp}

\end{document}